\documentclass[10pt,twocolumn,letterpaper]{article}

\usepackage{booktabs} %

\usepackage{cvpr}
\usepackage{times}
\usepackage{epsfig}
\usepackage{graphicx}
\usepackage{amsmath}
\usepackage{amssymb}
\usepackage{array}
\usepackage{comment}
\usepackage{subcaption}
\usepackage{capt-of}
\usepackage{booktabs}
\usepackage{tabularx}
\usepackage{soul}
\usepackage{siunitx}
\usepackage{microtype}
\usepackage{xspace}
\usepackage{caption}
\usepackage{amsmath,xparse,mleftright}
\usepackage{multirow}
\usepackage{mathtools}
\usepackage{pifont}
\usepackage{sidecap}

\usepackage{appendix}

\graphicspath{ {./images/} }

\usepackage[pagebackref=true,breaklinks=true,letterpaper=true,colorlinks,bookmarks=false]{hyperref}
\usepackage{cleveref}

\cvprfinalcopy %

\def\cvprPaperID{****} %

\newcommand{\w}{$\mathcal{W}$\xspace}
\newcommand{\wk}{$\mathcal{W}^k$\xspace}
\newcommand{\wstar}{$\mathcal{W}_{*}$\xspace}
\newcommand{\wkstar}{$\mathcal{W}^k_{*}$\xspace}
\newcommand{\cmark}{\ding{51}}

\newcommand{\norm}[1]{\left\lVert#1\right\rVert}

\newcommand*\ruleline[2]{\par\noindent\raisebox{.8ex}{\makebox[{#1}]{\hrulefill\hspace{1ex}\raisebox{-.8ex}{#2}\hspace{1ex}\hrulefill}}}

\begin{document}
\title{Designing an Encoder for StyleGAN Image Manipulation}

\vspace{-5cm}

\author{
\and
Omer Tov\\
Tel-Aviv University
\and
Yuval Alaluf\\
Tel-Aviv University
\and
Yotam Nitzan\\
Tel-Aviv University
\and \and
Or Patashnik\\
Tel-Aviv University
\and
Daniel Cohen-Or\\
Tel-Aviv University
}

\twocolumn[{%
\renewcommand\twocolumn[1][]{#1}%
\vspace{-2em}
\vspace{-0.1in}
\maketitle
\pagestyle{plain}
\vspace{-2.5em}
\begin{center}
    \centering
    \centering
\setlength{\tabcolsep}{1pt}
\begin{tabular}{c c c c c c c}
\vspace{-0.075cm}
\includegraphics[width=0.131\textwidth]{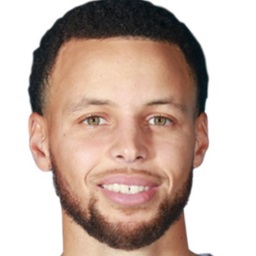} &
\includegraphics[width=0.131\textwidth]{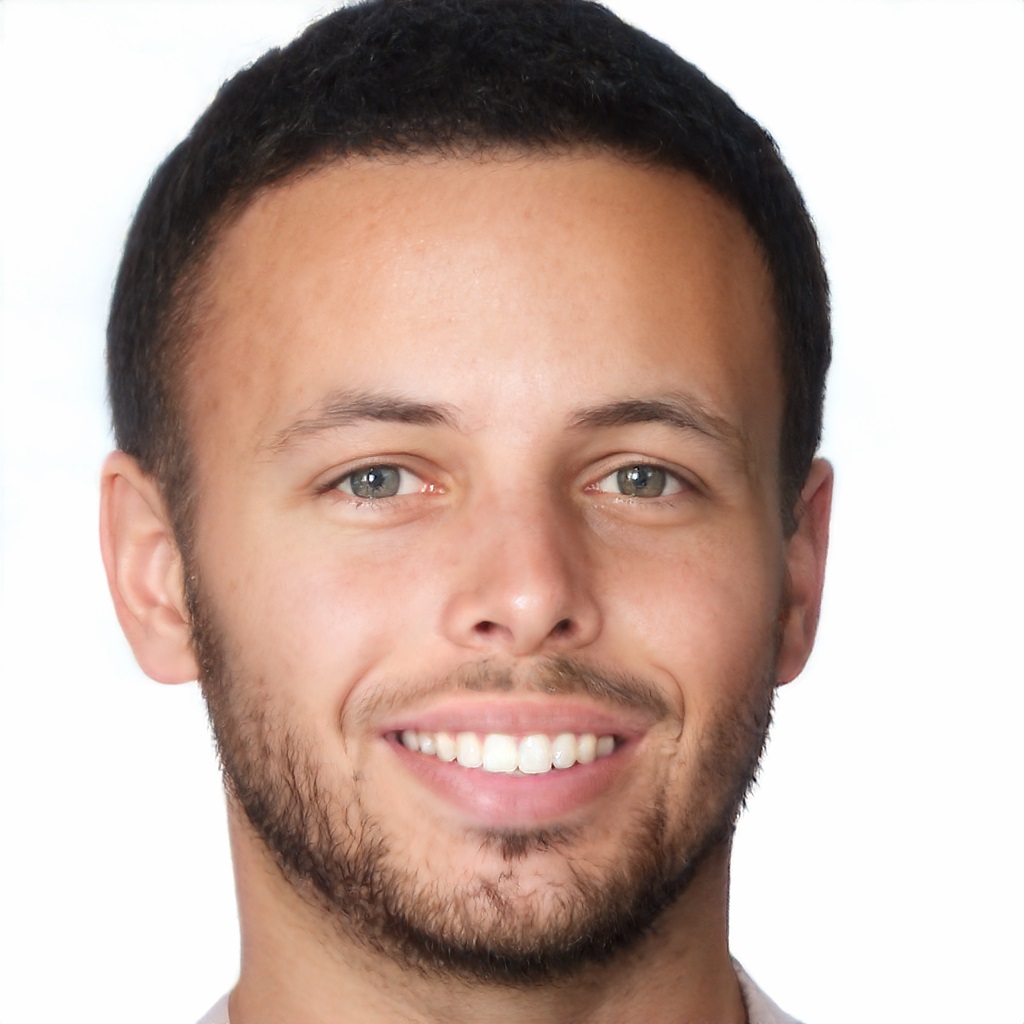} &
\includegraphics[width=0.131\textwidth]{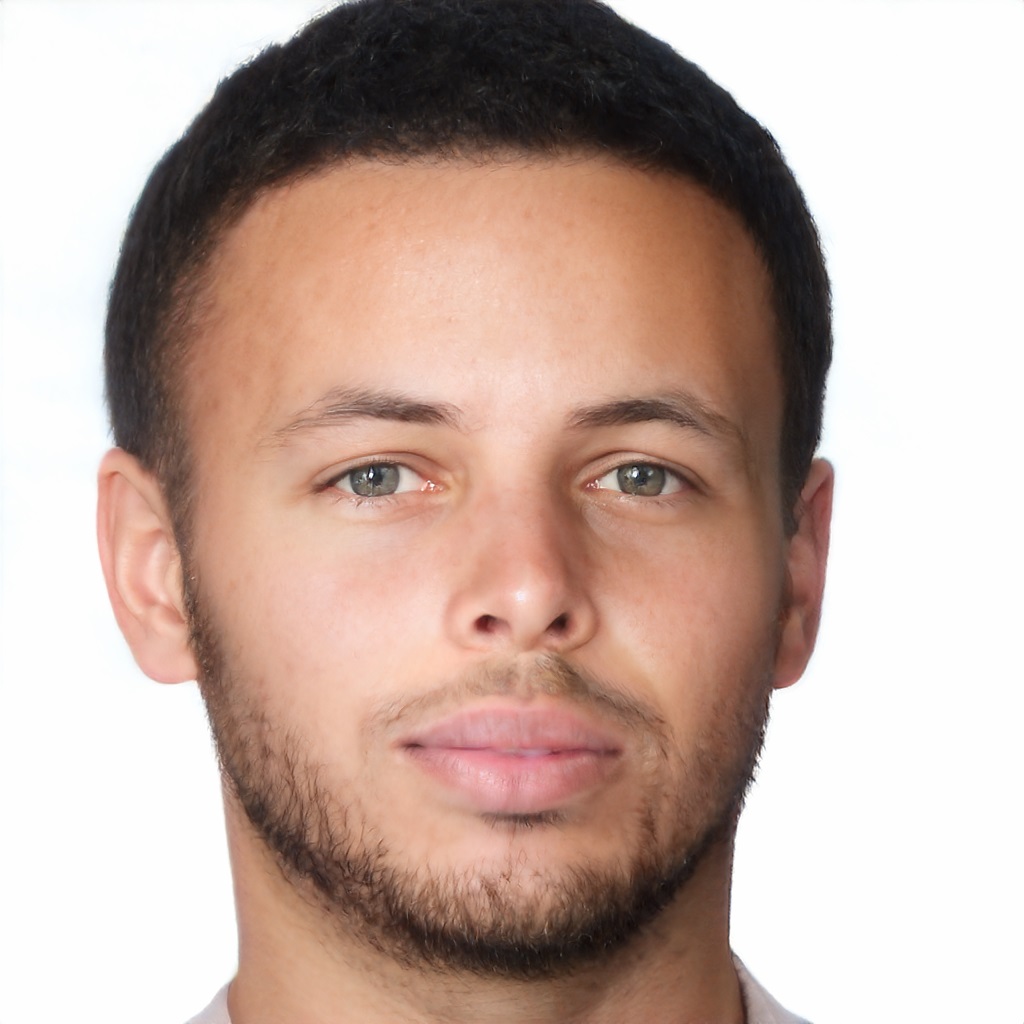}&
\includegraphics[width=0.131\textwidth]{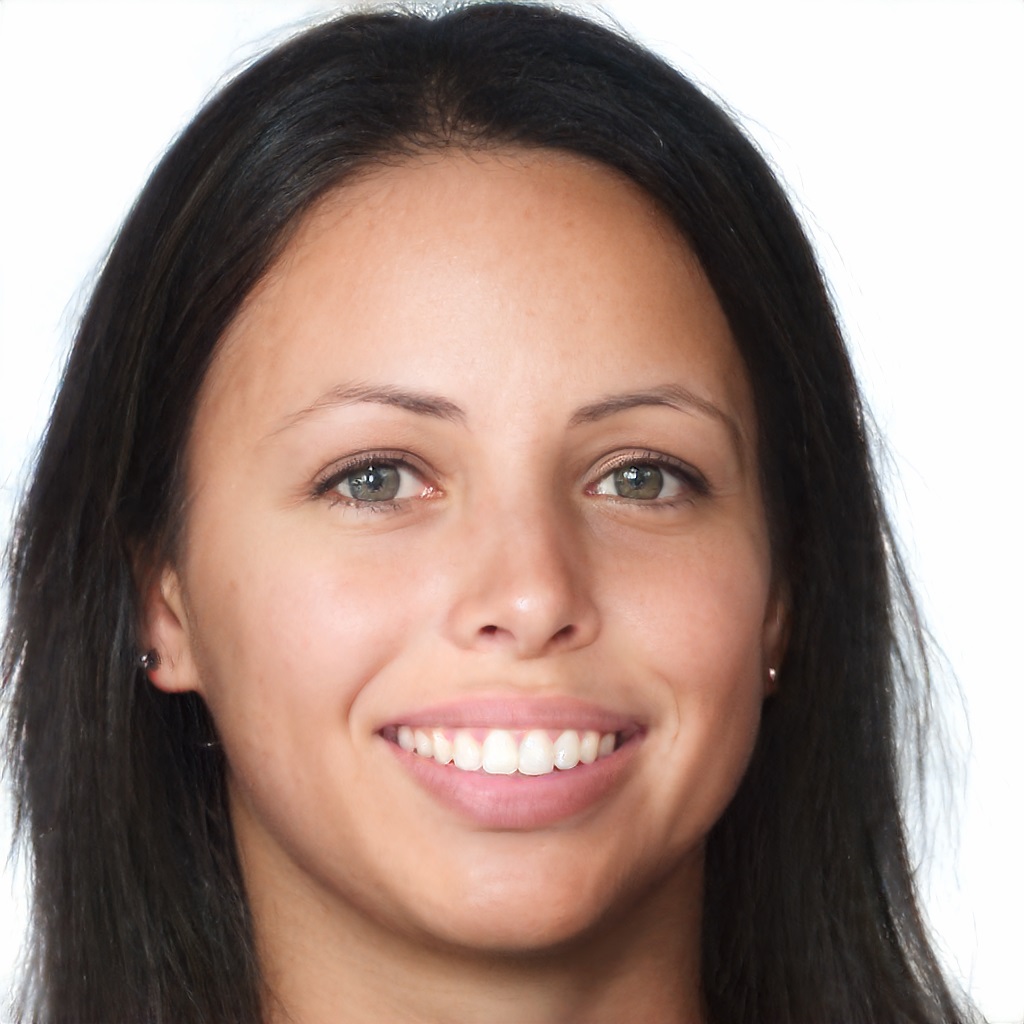}&
\includegraphics[width=0.131\textwidth]{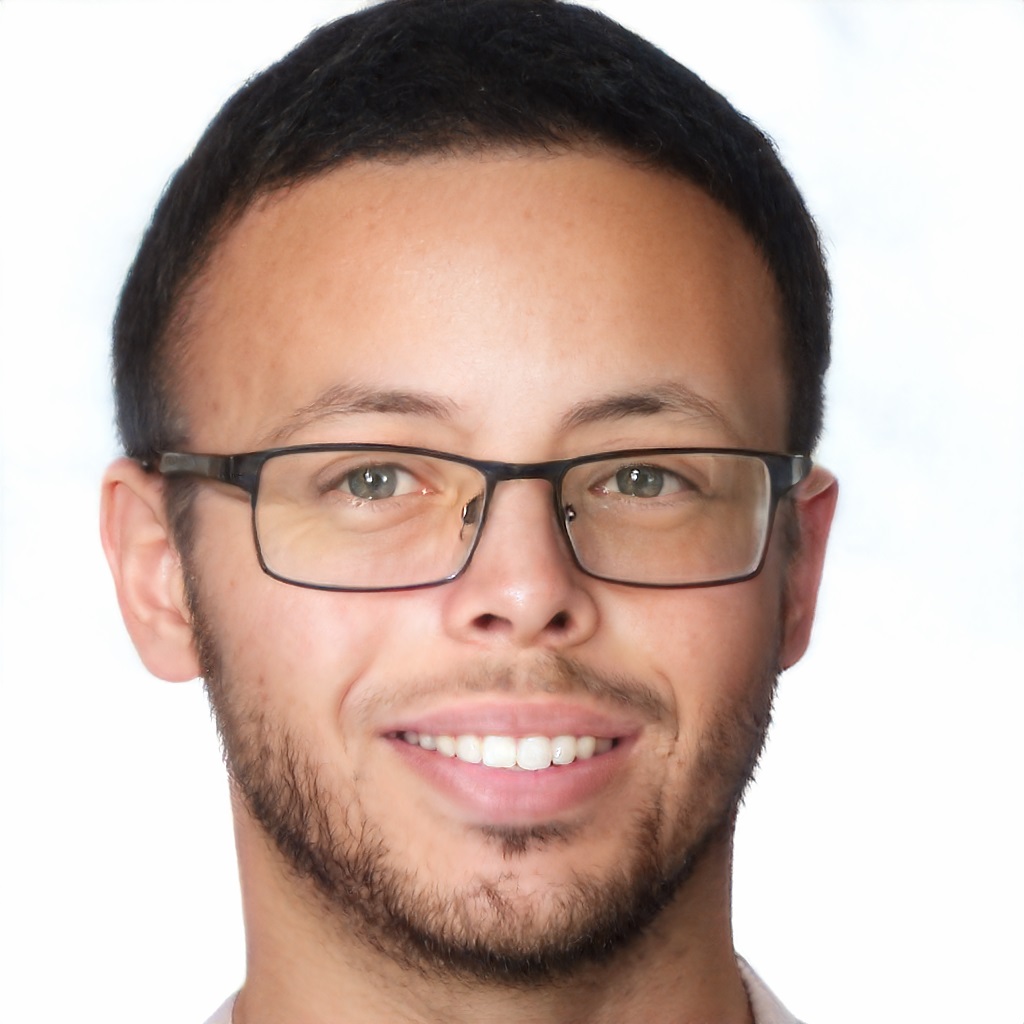} &
\includegraphics[width=0.131\textwidth]{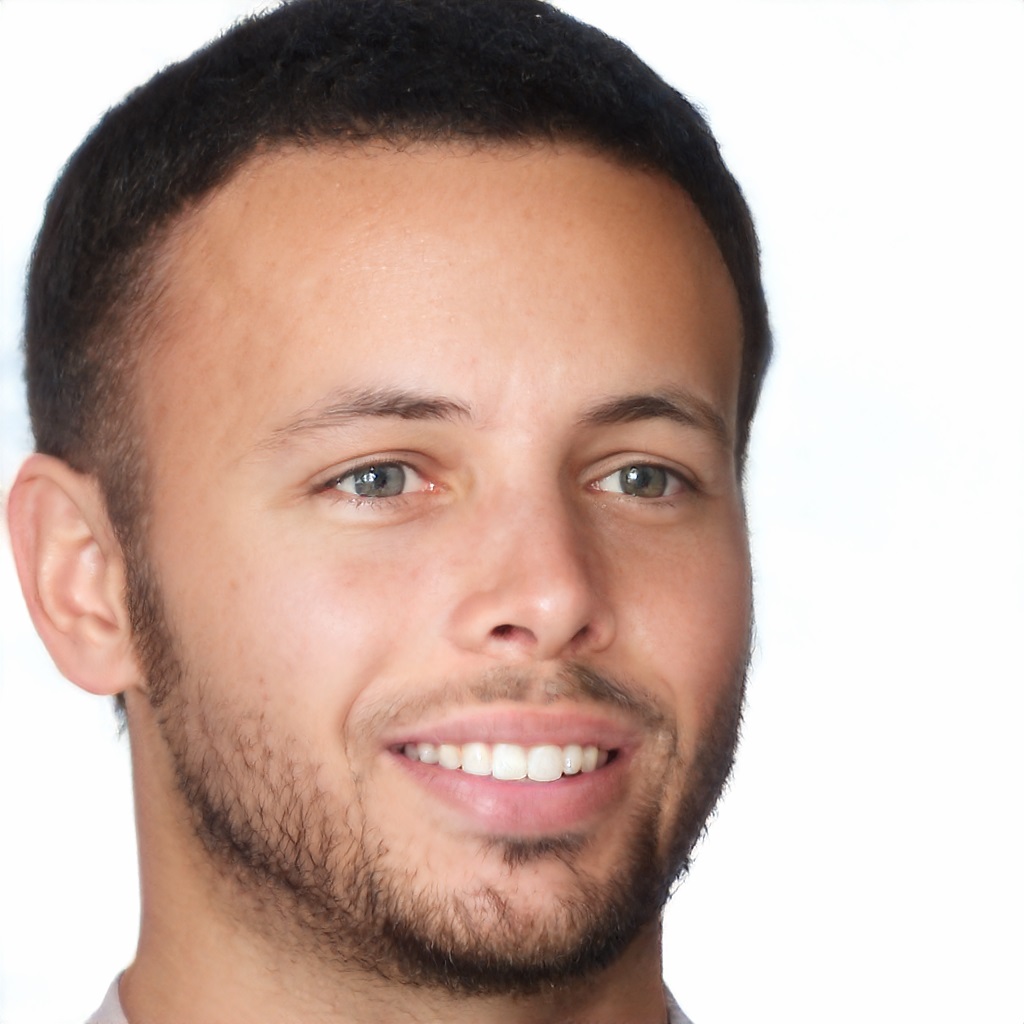} &
\includegraphics[width=0.131\textwidth]{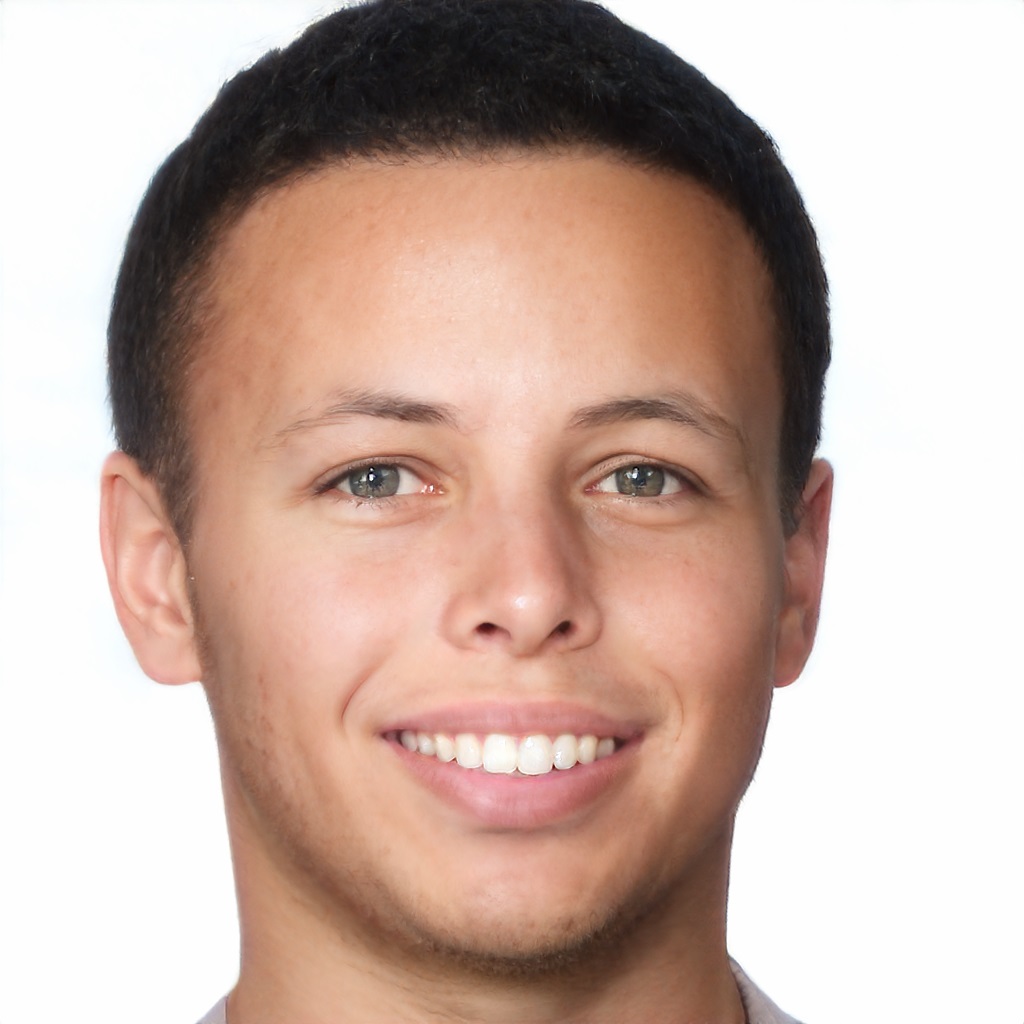}  \\
\vspace{-0.09cm}
\includegraphics[width=0.131\textwidth]{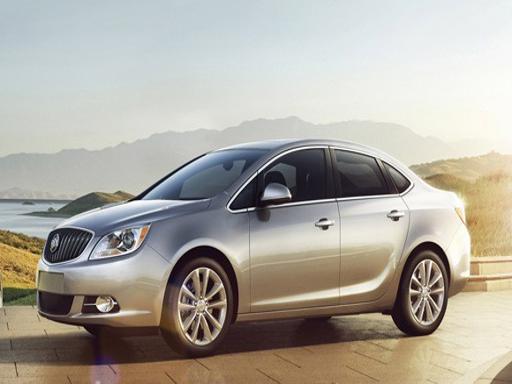} &
\includegraphics[width=0.131\textwidth]{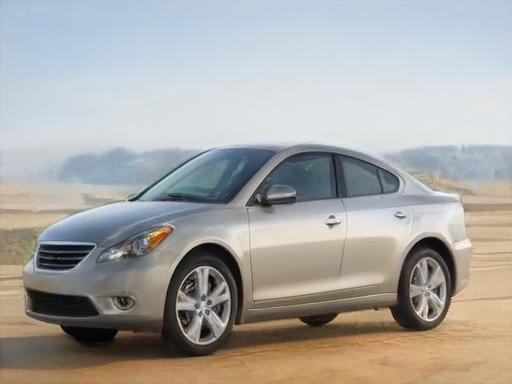}&
\includegraphics[width=0.131\textwidth]{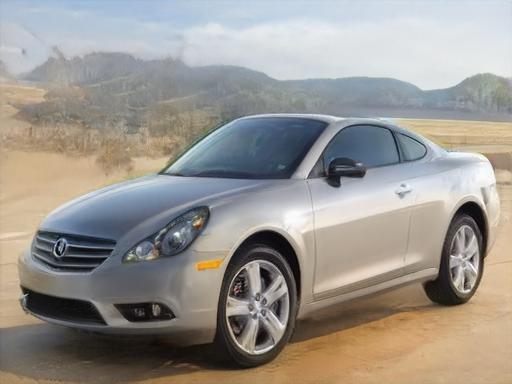} & 
\includegraphics[width=0.131\textwidth]{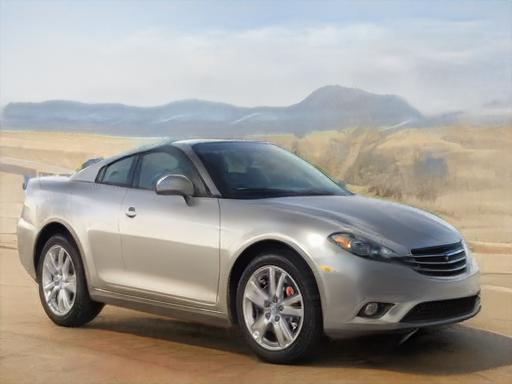} &
\includegraphics[width=0.131\textwidth]{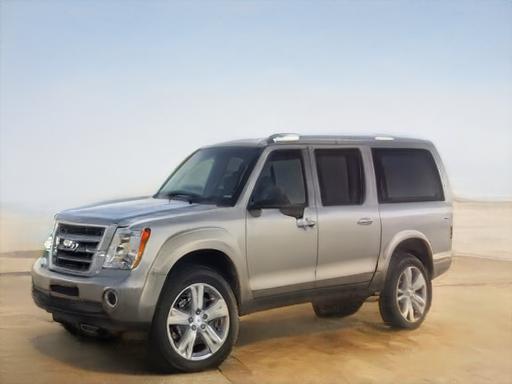} &
\includegraphics[width=0.131\textwidth]{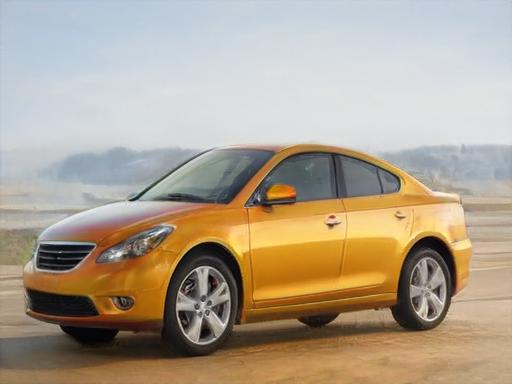} &
\includegraphics[width=0.131\textwidth]{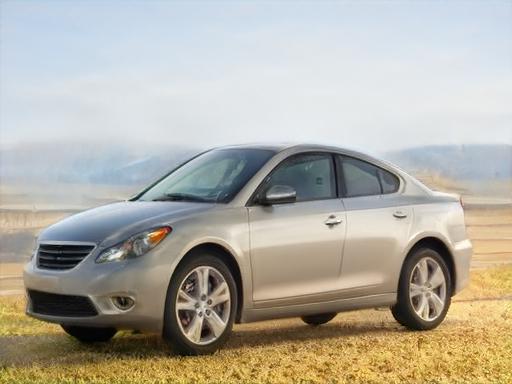} \\
\vspace{-0.09cm}
\includegraphics[width=0.131\textwidth]{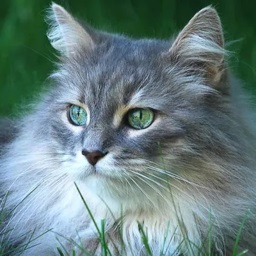}&
\includegraphics[width=0.131\textwidth]{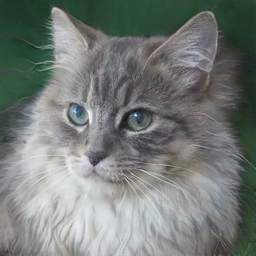} &
\includegraphics[width=0.131\textwidth]{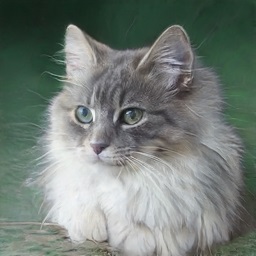}&
\includegraphics[width=0.131\textwidth]{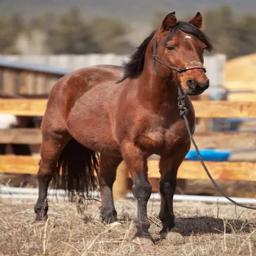} &
\includegraphics[width=0.131\textwidth]{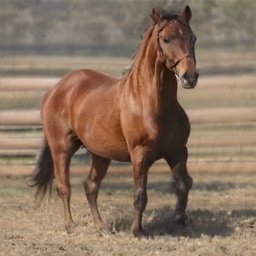} &
\includegraphics[width=0.131\textwidth]{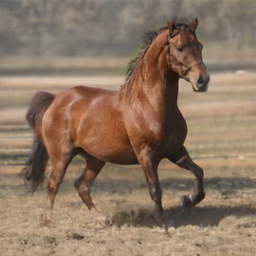} &  
\includegraphics[width=0.131\textwidth]{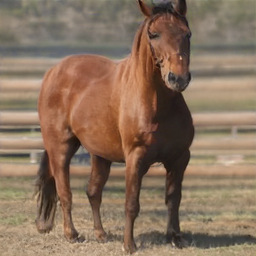} \\
\end{tabular}

    \vspace{-0.25cm}
    \captionof{figure}{Real image editing via StyleGAN inversion using our e4e method.
    For each domain we show from left to right: the original real image, the inverted image, and multiple manipulations performed using various editing techniques.}
    \label{fig:teaser}
\end{center}%
}]

\begin{abstract}
    \vspace{-0.395cm}
    Recently, there has been a surge of diverse methods for performing image editing by employing pre-trained unconditional generators. Applying these methods on real images, however, remains a challenge, as it necessarily requires the inversion of the images into their latent space. 
To successfully invert a real image, one needs to find a latent code that reconstructs the input image accurately, and more importantly, allows for its meaningful manipulation.
In this paper, we carefully study the latent space of StyleGAN, the state-of-the-art unconditional generator.
We identify and analyze the existence of a distortion-editability tradeoff and a distortion-perception tradeoff within the StyleGAN latent space.
We then suggest two principles for designing encoders in a manner that allows one to control the proximity of the inversions to regions that StyleGAN was originally trained on.
We present an encoder based on our two principles that is specifically designed for facilitating editing on real images by balancing these tradeoffs. 
By evaluating its performance qualitatively and quantitatively on numerous challenging domains, including cars and horses, we show that our inversion method, followed by common editing techniques, achieves superior real-image editing quality, with only a small reconstruction accuracy drop.

\end{abstract}
\section{Introduction}

In recent years, Generative Adversarial Networks (GANs)~\cite{Goodfellow2014GenerativeAN} have made significant progress in unconditional image synthesis.
In particular, StyleGAN~\cite{Karras2020ada,karras2019style, karras2020analyzing} has achieved unprecedented visual quality and fidelity in various domains.
Beyond its phenomenal realism, StyleGAN uses a learnt intermediate latent space, $\mathcal{W}$, which more faithfully reflects the distribution of the training data compared to the standard Gaussian latent space. Furthermore, numerous works have demonstrated that $\mathcal{W}$ has intriguing disentangled properties~\cite{collins2020editing, harkonen2020ganspace, shen2020interpreting, tewari2020stylerig, wu2020stylespace}, which allow one to perform extensive image manipulations by leveraging a pretrained StyleGAN. 

To apply such manipulations on real images, one must first \textit{invert} the given image into the latent space, i.e. retrieve a latent code, a so called style code, $w \in \mathcal{W}$, such that feeding the obtained style code as input to the pretrained StyleGAN returns the original image. 
Hence, a high-quality inversion scheme is vital for such editing techniques. 
High-quality inversion is characterized by two aspects. 
First, the generator should properly reconstruct the given image with the style code obtained from the inversion.
Second, it should be possible to best leverage the editing capabilities of the latent space to obtain meaningful and realistic edits of the given image. In short, we call the latter aspect \emph{editability}.
To define a proper reconstruction, we distinguish between two properties: (i) distortion and (ii) perceptual quality~\cite{blau2018perception}. As will be elaborately explained in the paper, \textit{distortion} measures per-image input-output similarity, whereas \textit{perceptual quality} measures how realistic the reconstructed image is. Note that ideally the distortion should be low and the perceptual quality high.

There is an intimate relation between distortion, perceptual quality, and editability.
The expressiveness of the $\mathcal{W}$ latent space has been shown to be limited~\cite{abdal2019image2stylegan, richardson2020encoding}, in that not every image can be accurately mapped into $\mathcal{W}$. To alleviate this limitation, Abdal \etal~\cite{abdal2019image2stylegan} demonstrate that any image can be inverted into an extension of $\mathcal{W}$, denoted $\mathcal{W}+$, where a style code consists of a number of style vectors.
The space $\mathcal{W}+$ has more degrees of freedom, and is thus significantly more expressive than $\mathcal{W}$.
Although this extension is expressive enough to represent real images, as we shall show, inverting images away from the original $\mathcal{W}$ space reaches regions of the latent space that are less editable and in which the perceptual quality is lower.
This tradeoff between distortion, perceptual quality, and editability is presented and extensively analyzed in the paper.

Recognizing that the main motivation for inverting an image is the downstream editing task, we focus on understanding the editability of inverted real images. 
Our key insight is that editability and perceptual quality are best achieved by inverting an image \textit{close} to $\mathcal{W}$. In the following, the term ``close'' will be characterized by two key properties. First, low \textit{variance} between the different style vectors. Second, each style vector should lie within the distribution $\mathcal{W}$. 

We design a new encoder, which is explicitly encouraged to invert images close to $\mathcal{W}$. 
This encoder serves two purposes: (i) it allows demonstrating that the distortion-editability and distortion-perception tradeoffs are controlled by the proximity of a latent code to \w, and (ii) it constitutes an effective encoder for editing real images. Thus, we name our encoder \textit{e4e} --- \textit{Encoder for Editing}.

Specifically, we design the encoder to map a given real image to a style code that consists of a series of style vectors with low variance, each close to the distribution of $\mathcal{W}$. Since the distribution of $\mathcal{W}$ cannot be explicitly modeled, we extend the adversarial training of the style code introduced in Nitzan \etal ~\cite{nitzan2020face}, and apply it to multiple codes to encourage the proper encoding of each into $\mathcal{W}$. 
To further assist in staying in editable regions, we additionally present a progressive training scheme where the variance between the style vectors is gradually increased during training.

We present quantitative and qualitative results that demonstrate the distortion-editability and distortion-perception tradeoffs, and the benefit of inverting ``close'' to $\mathcal{W}$. We evaluate our encoder, showing the generalization of our approach and its applicability for a variety of challenging domains which, unlike the facial domain, have no common structure and may contain numerous modes. In Figure \ref{fig:teaser}, we show inversions obtained by our encoder in multiple domains, followed by several manipulations performed using various editing methods~\cite{abdal2020styleflow, harkonen2020ganspace, shen2020interpreting, shen2020closedform}. As can be seen, with only a slight degradation in distortion, we are able to achieve plausibly edited images, while preserving the content and quality of the original images.

To summarize, we present four main contributions:
\begin{itemize}
    \item We analyze the complex latent space of StyleGAN and suggest a novel view of its structure.
    \item We present the innate tradeoffs among distortion, perception, and editability. 
    \item We characterize the tradeoffs and design two means for an encoder to control them.
    \item We present e4e, a novel encoder that is specifically designed to allow for the subsequent editing of inverted real images.
\end{itemize}

Source code and pretrained models can be found at our project page: \texttt{\url{https://github.com/omertov/encoder4editing}}.
\section{Background and Related Work}
\label{sec:rw}

\subsection{Latent Space Manipulation}
Recently, understanding and controlling the latent representation of pretrained GANs has attracted considerable attention. Notably, it has been shown that StyleGAN~\cite{Karras2020ada, karras2019style, karras2020analyzing}, with its novel style-based architecture, contains a semantically rich latent space that can be used to perform diverse image manipulations. Motivated by this, many works have presented diverse approaches for discovering and performing semantically meaningful manipulations on images. 
A commonly-used approach involves finding ``walking'' directions that control a specific attribute of interest. Early works~\cite{denton2019detecting, goetschalckx2019ganalyze, shen2020interpreting} use a fully-supervised approach and find latent directions for binary-labeled attributes such as young $\leftrightarrow$ old or smile $\leftrightarrow$ no-smile. Others have proposed self-supervised approaches that find latent space directions corresponding to a specific image transformation, such as zoom or rotation~\cite{jahanian2019steerability, plumerault2020controlling, spingarn2020gan}. Finally, several methods~\cite{harkonen2020ganspace, voynov2020unsupervised, wang2021a} find latent directions in an unsupervised manner and require manual annotations to determine the semantic meaning of each direction post hoc.

Extending the aforementioned works, there are other techniques that go beyond walking along linear directions.
Tewari \etal \cite{tewari2020stylerig} use a pretrained 3DMM to edit expression, pose and illumination of faces by borrowing them from other latent codes. 
Abdal \etal \cite{abdal2020styleflow} infer a modified latent code using an auxiliary pretrained face attributes classifier.
Shen \etal \cite{shen2020closedform} perform eigenvector decomposition of the weights of the generator's first layer to find edit directions. 
Collins \etal \cite{collins2020editing} perform local semantic editing by borrowing a subset of the latent code from other samples. 

Some recent works have also considered other latent spaces. Sendik \etal \cite{sendik2020unsupervised} suggest a modified style-based architecture that learns multiple input constants and achieves better disentanglement between data modes.
Last, ~\cite{liu2020style, wu2020stylespace} have considered the so called \textit{Style Space}, and are able to identify specific components that have a clear effect on spatial regions or on a single attribute. By modifying these components, they are able to achieve disentangled editing. 

\begin{table}
    \begin{center}
        \label{tab:latent_spaces}
        \begin{tabular}{|l | c | c|}
        \hline
        & \shortstack{Individual style codes \\ are limited to \w} & \shortstack{Same style code \\ in all layers} \\
        \hline
        \w & \cmark & \cmark \\
        \hline
        \wk & \cmark &  \\
        \hline
        \wstar &  & \cmark \\
        \hline
        \wkstar &  &  \\
        \hline
        \end{tabular}
    \end{center}
    \vspace{-0.45cm}
    \caption{A concise summary outlining the differences between various latent spaces. Note that, due to its ambiguity, $\mathcal{W}+$ is omitted from the table and is instead defined more explicitly using \wk and \wkstar.}
    \label{tab:latent_spaces}
\end{table}

\subsection{Latent Space Embedding}
To perform such manipulations on real images, one must first obtain the latent code from which the pretrained GAN can most accurately reconstruct the original input image. This task has been commonly referred to as \textit{GAN Inversion} \cite{richardson2020encoding, xia2021gan, zhu2020domain}.
Generally, inversion methods either (i) directly optimize the latent vector to minimize the error for the given image \cite{abdal2019image2stylegan,abdal2020image2stylegan++, creswell2018inverting, karras2020analyzing, lipton2017precise}, (ii) train an encoder to map the given image to the latent space \cite{guan2020collaborative, perarnau2016invertible, richardson2020encoding}, or (iii) use a hybrid approach combining both \cite{pbayliesstyleganencoder, zhu2020domain}. Typically, methods performing optimization are superior in achieving low distortion. However, they require a substantially longer time to invert an image and are less editable.

Given the high dimensionality of the latent space, finding meaningful directions is extremely challenging. Therefore, recent methods propose an end-to-end approach for performing latent manipulations using a well-trained generator. Specifically, Nitzan \etal~\cite{nitzan2020face} train an encoder to obtain a latent vector representing the identity of one image and the pose, expression, and illumination of another. Menon \etal~\cite{menon2020pulse} solve single-image super resolution by taking a low-resolution image and searching the latent space for a high-resolution version of the image using direct optimization. Finally, Richardson \etal~\cite{richardson2020encoding}
perform image-to-image translation by directly encoding input images into the latent codes representing the desired transformation. 

In contrast to previous works, our encoder is specifically designed to output latent codes that ensure further editing capabilities. As we shall see, there is an explicit trade-off between distortion and editability where one can obtain better editability by allowing a small degradation in distortion.
\begin{figure}
    \centering
    \includegraphics[width=0.95\linewidth]{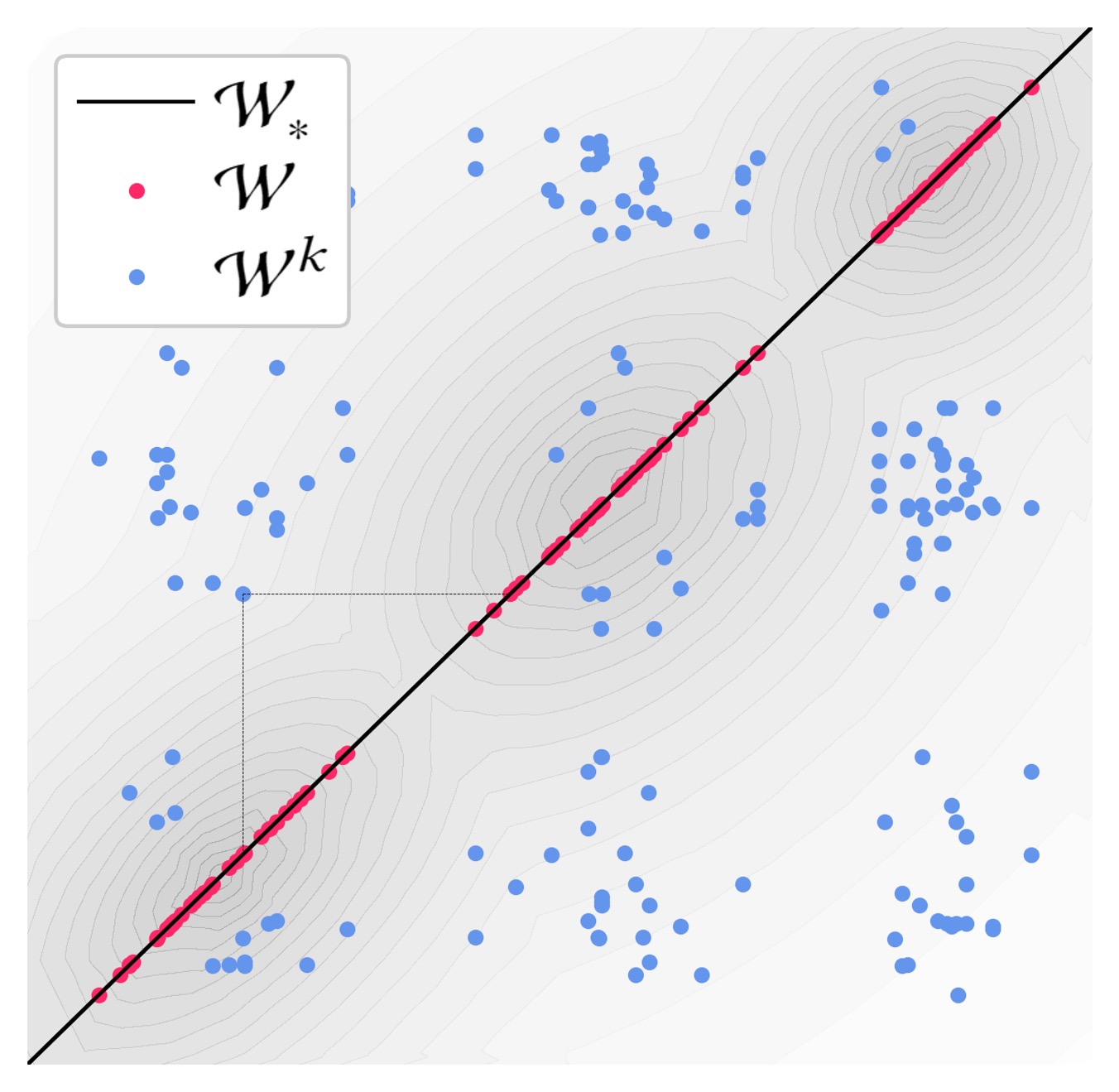}
    \caption{
    An illustration of our definitions for a 1-dimensional $\mathcal{W}$ and for $k=2$ (i.e., $\mathcal{W}^2$). 
    The diagonal line represents $\mathcal{W}_*$ since the two coordinates of each point on it are equal.
    The pink points are sampled from $\mathcal{W}$, which here is represented as a mixture of 1-dimensional Gaussians. 
    The entire gray region represents $\mathcal{W}^2_*$, corresponding to $\mathbb{R}^2$.
    The blue points are sampled from $\mathcal{W}^2$ since each coordinate of such a point is sampled from \w. 
    }
    \label{fig:w_illustration}
\end{figure}

\section{Terminology}
\label{sec:terminology}

StyleGAN~\cite{Karras2020ada, karras2019style, karras2020analyzing} consists of two key components: (i) a mapping function that maps a latent code $z \in \mathcal{Z} = \mathcal{N}(\mu,\sigma^2)$
into a \textit{style code} $w \in \mathcal{W} \subsetneq \mathbb{R}^{512}$, and (ii) a generator that takes in the style code, replicated several times (according to the desired resolution), and generates an image.
Note that while the distribution $\mathcal{Z}$ is known to be Gaussian, there is no known explicit model for the distribution of $\mathcal{W}$~\cite{wulff2020improving}. In the following, we will refer to this distribution as the \textit{range} of the mapping function. 

It has been shown~\cite{abdal2019image2stylegan} that not every real, in-domain image can be inverted into StyleGAN's latent space. 
To alleviate this limitation, one can increase the expressiveness of the StyleGAN generator by inputting $k$ \textit{different} style codes instead of a single vector. We denote this extended space by $\mathcal{W}^k \subsetneq \mathbb{R}^{k \times 512}$ where $k$ is the number of style inputs of the generator. For example, a generator capable of synthesizing images at a resolution of $1024\times1024$ operates in the extended $\mathcal{W}^{18}$ space corresponding to the $18$ different style inputs. 

Even more expressive power can be achieved by inputting style codes which are not necessarily from the true distribution of $\mathcal{W}$, i.e. outside the range of StyleGAN's mapping function. 
Observe that this extension can be applied by taking a single style code and replicating it, or by taking $k$ different style codes. We denote these extensions by $\mathcal{W}_{*}$ and $\mathcal{W}^k_{*}$, respectively. 
Note, that here, we depart from the commonly used $\mathcal{W}+$ notation due to its ambiguity with various works referring to it as both $\mathcal{W}^k$ and $\mathcal{W}^k_*$. 
However, note, that for simplicity and convenience, we refer to \w both as the $512$-dimensional distribution, and as a subset of \wk where all the $k$ style codes are equal and in \w.
In Table \ref{tab:latent_spaces} we present a summary describing the differences between the latent spaces. We provide an illustration of the differences between \w and \wk in Figure \ref{fig:w_illustration}.

\section{The GAN Inversion tradeoffs}
\label{sec:tradeoff}

\subsection{Preliminaries}
\label{subsec:preliminaries}
As discussed in Section \ref{sec:rw}, most of the research on StyleGAN's latent space revolves around two separate tasks: \textit{GAN inversion} and \textit{latent space manipulation}.
Previous works have considered the two tasks as follows. 
First, in the task of inversion, given an image $x$, we infer a latent code $w$, which is used to reconstruct $x$ as accurately as possible when forwarded through the generator $G$. 
In the task of latent space manipulation, for a given latent code $w$, we infer a new latent code, $w'$, such that the synthesized image $G(w')$ portrays a semantically meaningful edit of $G(w)$.

Inspecting the GAN inversion task, we stress that reconstruction alone has limited purpose since reconstructing an inverted image can at best return the original image. As previous works have mentioned ~\cite{abdal2019image2stylegan, abdal2020styleflow, zhu2020domain}, the central motivation for inversion is to allow for further latent editing operations. That is, a successful reconstruction should enable extensive and diverse manipulation of \emph{real} images with greater ease. This objective, however, is not trivial as some latent codes are more editable than others. 

Following the above observations, we suggest a broader perspective on how to evaluate GAN inversion methods. Specifically, they should be evaluated based on \textit{both} reconstruction and editability. 

The term reconstruction is ill-defined as an evaluation metric. Blau and Michaeli~\cite{blau2018perception} have distinguished between two separate properties of reconstruction --- \textit{distortion} and \textit{perceptual quality}. 
Formally, distortion is defined as $\mathbb{E}_{x \sim p_{X}}[\Delta(x, G(w)]$ where $p_X$ is the distribution of the real images, and $\Delta(x, G(w))$ is an image-space difference measure between images $x$ and $G(w)$.
Perceptual quality measures how realistic the reconstructed images are, with no relation to any reference image. 

Most previous works have considered only distortion \cite{abdal2019image2stylegan, nitzan2020face, richardson2020encoding} for evaluating the reconstruction. 
Distortion alone, however, does not capture the quality of the reconstruction. In fact, Blau and Michaeli ~\cite{blau2018perception} proved that not only is the perceptual quality different than distortion, there exists an explicit tradeoff between the two. 
Therefore, both the distortion and the perceptual quality of the reconstructed images must be evaluated to provide a complete evaluation of a reconstruction method or, in our case, a GAN Inversion method.

For editability, it is expected that given the inverted latent code, one can find many latent space directions corresponding to disentangled semantic edits in the image-space. 
Moreover, it is important to maintain a high \emph{perceptual quality} of the edited images.
In addition to the above, in Section~\ref{evaluation} we present a novel measure that is specifically designed to evaluate the success of an encoder for the editability of real images.

\begin{figure}
    \centering
    \includegraphics[width=0.95\linewidth]{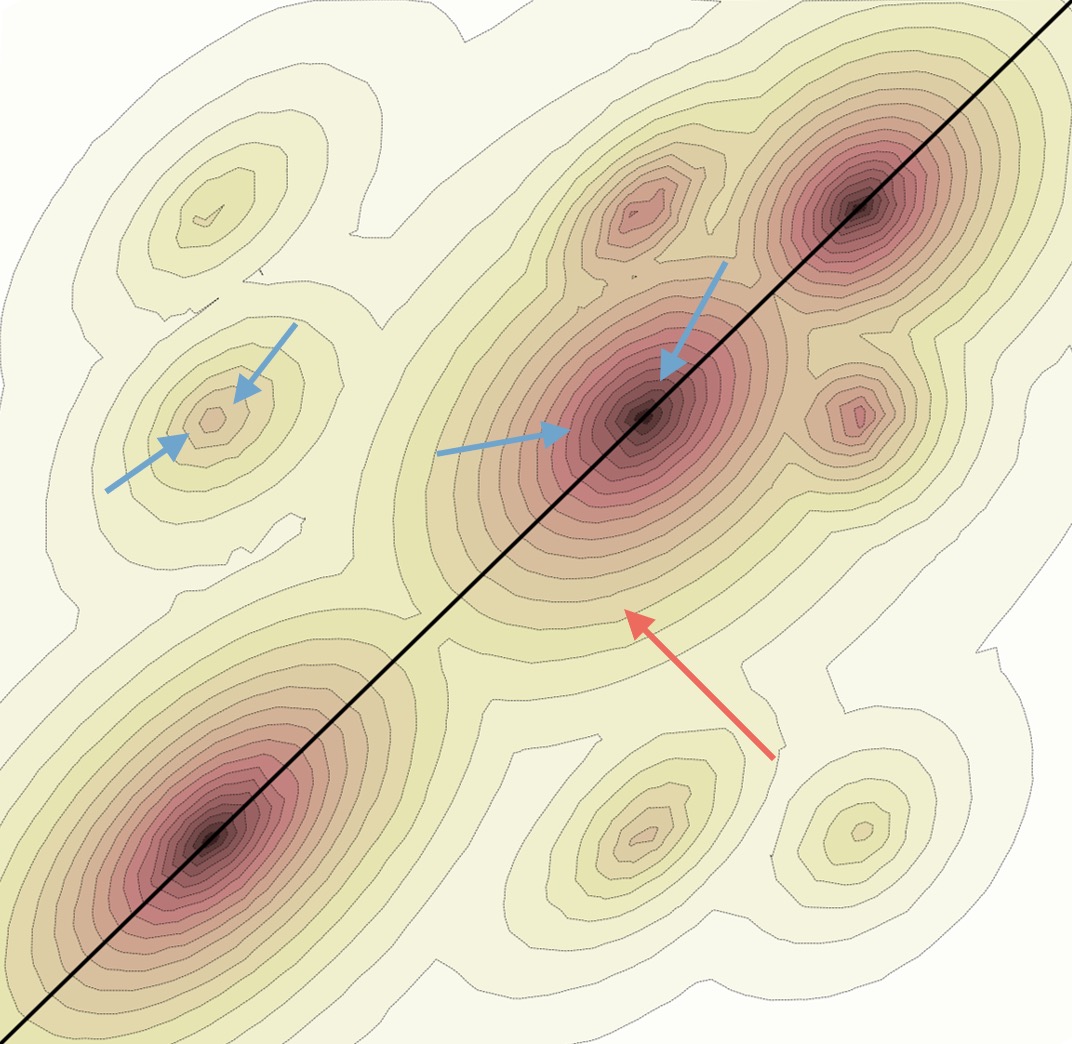}
    \caption{
    As in Figure~\ref{fig:w_illustration}, \w is 1-dimensional and $k=2$, that is $\mathcal{W}^{k}=\mathcal{W}^{2}$. The main diagonal represents \wstar, and warmer colors correspond to higher densities of \wk. As one traverses the space in parallel to the red arrow and approaches the diagonal, the latent codes become closer to \wstar since the variation between the coordinates of the latent code is decreased.  Likewise, traversing along a blue arrow results in latent codes that get closer to \wk since each coordinate, independently, gets closer to the distribution \w.
    }
    \label{fig:w_heatmap}
\end{figure}

\vspace{0.5cm}
\subsection{Distortion-Editability \& Distortion-Perception Tradeoffs}

We now turn to study the distortion, perceptual quality, and editability of different regions in StyleGAN's latent space. 
As discussed thoroughly in Section \ref{sec:terminology}, \wkstar differs from \w in two ways. 
First, \wkstar may contain different style codes at different style-modulation layers. Second, each style code, independently, is not bound to the true distribution of \w, but can instead take any value from $\mathbb{R}^{512}$. 

\begin{figure}
\setlength{\tabcolsep}{1pt}
    \begin{tabular}{c c c c c}
        \raisebox{0.04\linewidth}{\texttt{\w}} & 
        \includegraphics[width=0.22\linewidth]{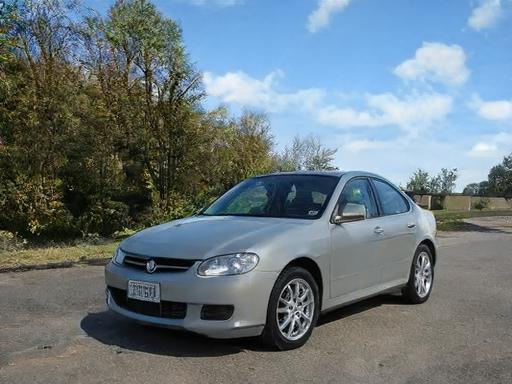} &
        \includegraphics[width=0.22\linewidth]{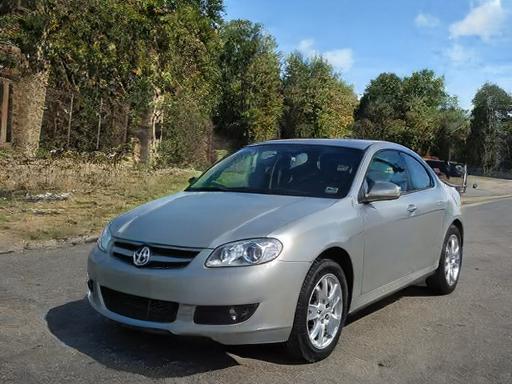} &
        \includegraphics[width=0.22\linewidth]{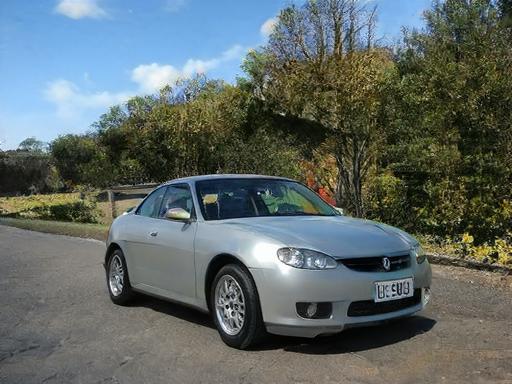} &
        \includegraphics[width=0.22\linewidth]{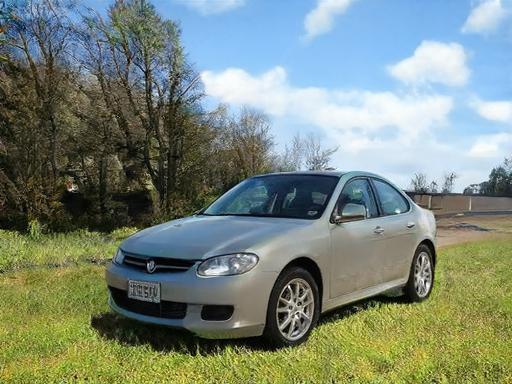} \\

        \raisebox{0.04\linewidth}{\texttt{\wkstar}} & 
        \includegraphics[width=0.22\linewidth]{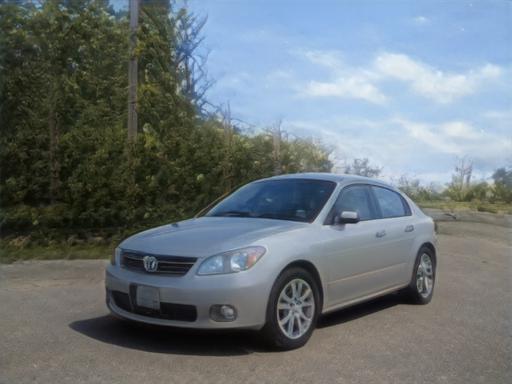} &
        \includegraphics[width=0.22\linewidth]{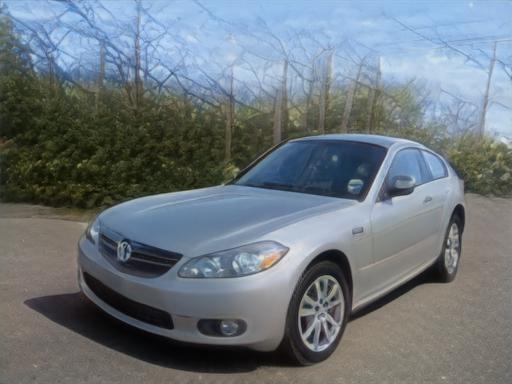} &
        \includegraphics[width=0.22\linewidth]{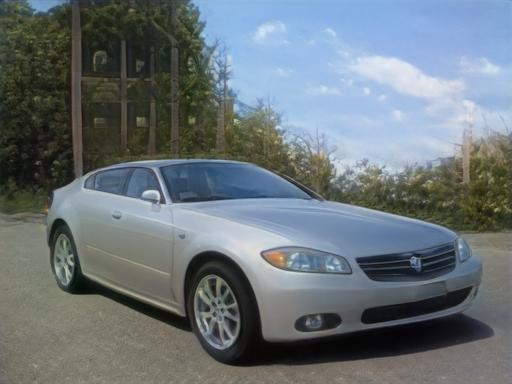} &
        \includegraphics[width=0.22\linewidth]{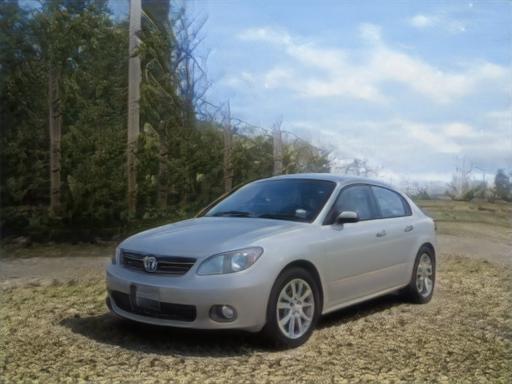} \\
        & Source & \multicolumn{3}{c}{\ruleline{0.67\linewidth}{Edits}}
    \end{tabular}
    \caption{
    The editability gap between \w and \wkstar. 
    The top row depicts several edits performed on an image generated from a latent code sampled from \w. 
    In the bottom row, the same image is inverted back into \wkstar and passed through the generator to obtain a visually-similar image.
    The same edits as those performed in the top row are then applied on the inverted \wkstar code.
    Note that although the two source images are similar in the image-space, the edits in \w are visually better than the corresponding ones in \wkstar, as can be observed by the warped shapes of the cars in the bottom row.
    }
    \label{fig:w_vs_wkstar}
\end{figure}

It is well known that \wkstar achieves lower, i.e. better, distortion than \w \cite{abdal2019image2stylegan, pbayliesstyleganencoder, richardson2020encoding, zhu2020domain}. 
Additionally, we find that \w is more editable, see Figure ~\ref{fig:w_vs_wkstar}. 
This provides the first evidence of the inherent tradeoff between distortion and editability. Observe that since StyleGAN is originally trained in the \w space, it is not surprising that \w is more well-behaved and has better perceptual quality compared to its \wkstar counterpart.
On the other hand, observe that due the significantly higher dimensionality of \wkstar and the architecture of StyleGAN, \wkstar has far greater expressive power.

\begin{figure}
\setlength{\tabcolsep}{1pt}
    \centering
    \begin{tabular}{c c c}
        \includegraphics[width=0.25\linewidth]{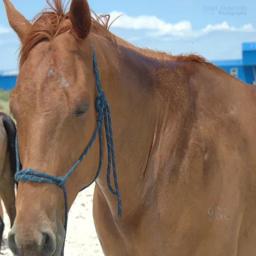} &
        \includegraphics[width=0.25\linewidth]{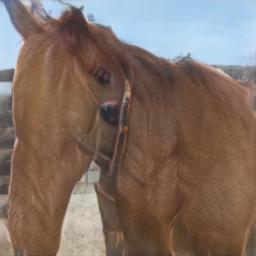} &
        \includegraphics[width=0.25\linewidth]{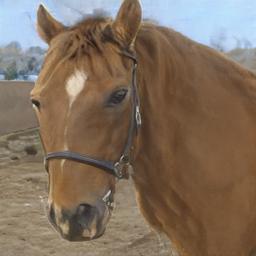} \\ 
        Source & \wkstar & \w
    \end{tabular}
    \caption{An example of the distortion-perception tradeoff. Inverting the source image using two different encoders yields significantly different results. While the middle image achieves low distortion (notice the overall similarity in shape and posture) and low perceptual quality (notice the warped head), the rightmost image achieves higher distortion but better perceptual quality.}
    \label{fig:perception-distortion-horses}
\end{figure}

We claim that the distortion-editability and the distortion-perception tradeoffs exist not only between \w and \wkstar, but also within the \wkstar space itself. Moreover, the tradeoffs are controlled by the proximity to \w. More specifically, as we approach \w, the distortion worsens while the editability and perceptual quality improve, see Figures~\ref{fig:w_vs_wkstar} and \ref{fig:perception-distortion-horses}. 
To validate this claim, it is necessary to develop a means for allowing one to control the proximity of an encoded image to \w.
In Section~\ref{sec:encoder}, we introduce two principles and mechanisms for controlling this proximity. 
Then, in Section \ref{sec:exp-tradeoff}, we perform extensive experimentation to support our claims.

Note that the concept of the distortion-editability tradeoff is not new. Previous works \cite{zhu2020improved}, and \cite{zhu2020domain} also observed the fact that different regions in the latent space have different editing properties and presented inversion methods that explicitly addressed this. However, here we identify the editable regions more explicitly by differentiating between the different extensions of \w.

\vspace{0.5cm}
\section{Designing an encoder}

\label{sec:encoder}
Building upon the observations from the previous section, we now present principles for designing an encoder and novel training scheme to explicitly address the proximity to \w.
Note that we make the important distinction between the two main inversion methodologies: (i) latent code optimization, and (ii) encoder-based methods. In this work, we concentrate on encoder-based methods for several key reasons. First, they are significantly faster as they infer a latent code with a single forward pass. This is in contrast to the costly, per-image optimization methodology. 
Second, due to the fact that CNNs are piece-wise smooth, the output of an encoder lies in a tight space which is more suitable for editing. Conversely, an optimization-based inversion may converge to an arbitrary point in the latent space. In the following, we consider a generic encoder which infers latent codes in the space of \wkstar.

We now present the two principles for controlling the proximity to \w, where each is defined using a dedicated training paradigm to encourage the encoder to map into regions in \wkstar that lie close to \w (see Figure~\ref{fig:w_heatmap}). We find these principles to be most effective when applied jointly.

\vspace{0.25cm}
\subsection{Minimize Variation}
\label{sec:var}
The first approach for getting closer to \w is to encourage the inferred \wkstar latent codes to lie closer to \wstar, i.e. minimize the variance between the different style codes, or equivalently, encourage the style codes to be identical. To this end, we propose a novel ``progressive'' training scheme.

Let $E(x)=(w_0, w_1, ..., w_{N-1})$ denote the output of the encoder, where $N$ is the number of style-modulation layers. Common encoders are trained directly into \wkstar, i.e., learn each $w_i$ \emph{separately} and simultaneously. Conversely, we infer a \emph{single} latent code, $w$, and a set of \emph{offsets} from $w$. More formally, we learn an output of the form $E(x)=(w, w + \Delta_1, ..., w + \Delta_{N - 1})$.

\begin{figure*}
    \centering
    \includegraphics[width=\linewidth]{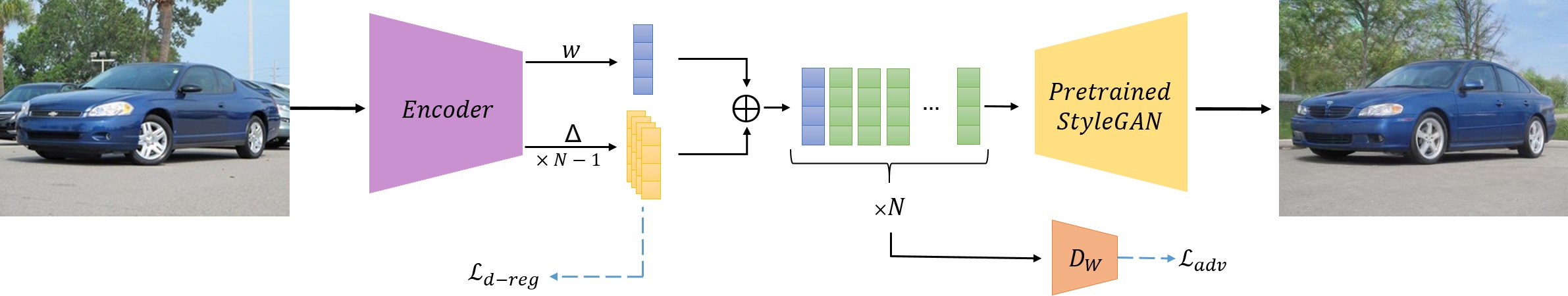}
    \caption{Our e4e network architecture. 
    The encoder receives an input image and outputs a single style code $w$ together with a set of offsets $\Delta_{1} .. \Delta_{N-1}$, where $N$ denotes the number of StyleGAN's style modulation layers.
    We obtain our final latent representation by replicating the $w$ vector $N$ times and adding each $\Delta_i$ to its corresponding entry.  
    During training, the $\mathcal{L}_{\text{d-reg}}$ regularization encourages small variance between different entries of the final representation, thereby remaining close to \wstar. $\mathcal{L}_{\text{adv}}$ guides each latent code towards the range of StyleGAN's mapping network, resulting in a final representation closer to \wk. As a result of applying both regularization terms, the encoder's final learned representation lies close to \w.}
    \label{fig:architecture}
\end{figure*}

Specifically, at the start of training we set $\forall i: \Delta_i=0$ and the encoder is trained to infer a single \wstar code. We then gradually allow the network to learn a \textit{different} $\Delta_i$ for each $i$ \textit{sequentially}. 
We find that doing so is useful in allowing the encoder to gradually expand from \wstar towards \wkstar. 
This scheme, together with the semantic meaning of specific input layers of StyleGAN, allows one to first learn a coarse reconstruction of the input image and then gradually add finer details while still remaining close to \wstar. 

We note that low frequency details greatly control the distortion quality. 
Thus, our progressive training scheme first focuses on improving the low frequency distortion by tuning the coarse-level offsets. Then, the encoder gradually complements these offsets with higher frequency details introduced by the finer-level offsets.

In terms of Figure~\ref{fig:w_heatmap}, this process is equivalent to starting from a style code that lies on the main diagonal, and then allowing small perturbations, thereby slightly diverging from it.

In order to explicitly enforce a proximity to \wstar, we add an $L_2$ delta-regularization loss:
\vspace{-0.2cm}
\begin{equation}
    \mathcal{L_{\text{d-reg}}}\left ( w \right ) = \sum_{i=1}^{N-1}|| \Delta_i ||_2.
\end{equation}
\vspace{-0.4cm}

\subsection{Minimize Deviation From \wk}~\label{sec:latent-disc}
The second approach for getting closer to \w is to encourage the \wkstar latent codes obtained by the encoder to lie closer to \wk. That is, encourage the individual style codes to lie within the actual distribution of \w. 
To do so, we adopt a latent discriminator~\cite{nitzan2020face} which is trained in an adversarial manner to discriminate between real samples from the \w space (generated by StyleGAN's mapping function) and the encoder's learned latent codes.

Observe that such a latent discriminator addresses the challenge of learning to infer latent codes that belong to a distribution which cannot be explicitly modeled. In doing so, the discriminator encourages the encoder to infer latent codes that lie within \w as opposed to \wstar. We use a single latent discriminator, denoted $D_\mathcal{W}$, that operates on each latent code entry separately. In every iteration, we calculate the GAN loss for every $E(x)_i$ and average over all $i$-s. 

We use the non-saturating GAN loss \cite{Goodfellow2014GenerativeAN} with $R_1$ regularization \cite{mescheder2018training},
\begin{equation}
\begin{gathered}
\mathcal{L}^{D}_{\text{adv}} =  -\underset{w \sim \mathcal{W}}{\mathbb{E}}[\log D_{\mathcal{W}}(w)]-\underset{x \sim p_{X}}{\mathbb{E}}[\log (1-D_{\mathcal{W}}(E(x)_i)]  + \\ 
\frac{\gamma}{2} \underset{w \sim  \mathcal{W}}{\mathbb{E}} \left[ \norm{\nabla_w D_{\mathcal{W}}(w)}_2^2 \right],
\end{gathered}
\end{equation}
\begin{equation}
    \mathcal{L}^{E}_{\text{adv}} = - \underset{x \sim p_{X}}{\mathbb{E}}[\log D_{\mathcal{W}}(E(x)_i)].    
\end{equation}

\section{e4e: Encoder for Editing}~\label{encoder_details}
We now turn to implement our encoder and novel training scheme. A high-level description of our encoder is illustrated in Figure \ref{fig:architecture}.
Our encoder builds upon the Pixel2Style2Pixel (pSp) encoder~\cite{richardson2020encoding}, but here, we design the encoder specifically for editing. Therefore, we name our new encoder \emph{e4e ---  ``Encoder for Editing''}.

Unlike the original pSp encoder which generates $N$ style-codes in parallel, here, we follow the principles described in Section~\ref{sec:var} and generate a single base style code, denoted by $w$, and a series of $N-1$ offset vectors (illustrated in yellow in Figure~\ref{fig:architecture}). The offsets are then summed up with the base style code $w$ to yield the final $N$ style codes which are then fed into a fixed, pretrained StyleGAN2 generator to obtain the reconstructed image. 

\subsection{Losses}~\label{losses}
To train our encoder, we employ losses that ensure low distortion, as commonly done when training an encoder, and losses that explicitly encourage the generated style codes to remain close to \w, thereby increasing the perceptual quality and editability of the generated images.

\paragraph{\textbf{Distortion}}
One of the key ideas presented in pSp is the identity loss, which is specifically designed to assist in the accurate inversion of real images in the facial domain. 
Motivated by recent strong self-supervised learning techniques, we generalize this identity loss and introduce a novel $\mathcal{L}_\text{sim}$ loss defined by,
\begin{equation}
    \mathcal{L}_{\text{sim}}\left (x \right ) = 1-\left \langle C(x),C(G(e4e(x)))) \right \rangle ,
\end{equation}
where $C$ is a ResNet-50~\cite{he2015deep} network trained with MOCOv2~\cite{chen2020improved} and $G$ is the pretrained StyleGAN2 generator. Here, $\mathcal{L}_\text{sim}$ explicitly encourages the encoder to minimize the cosine similarity between the feature embeddings of the reconstructed image and its source image. Observe that $\mathcal{L}_\text{sim}$ can be applied in any \textit{arbitrary} domain due to the general nature of the extracted features. Note that in the facial domain, we adopt the original identity loss used in pSp, and employ a pretrained ArcFace \cite{deng2019arcface} facial recognition network for extracting the feature embeddings.

In addition, we employ the commonly used $\mathcal{L}_2$ and $\mathcal{L}_{LPIPS}$~\cite{zhang2018unreasonable} losses to learn both pixel-wise and perceptual similarities.
That is, our distortion loss is defined as:
\begin{equation}
    \mathcal{L}_{\text{dist}}(x) =  \lambda_{l2}\mathcal{L}_{2}(x) + \lambda_{lpips}\mathcal{L}_{LPIPS}(x) + \lambda_{sim}\mathcal{L}_{\text{sim}}(x). \label{eq:rec_loss}
\end{equation}

\paragraph{\textbf{Perceptual quality and editability}}

To increase the perceptual quality and editability, we employ the two losses introduced in Section~\ref{sec:encoder}. First, we apply a delta-regularization loss to ensure proximity to \wstar when learning the offsets $\Delta_i$. Second, we use an adversarial loss using our latent discriminator, which encourages each learned style code to lie within the distribution \w. More formally, the editability loss is given by
\begin{equation}
    \mathcal{L}_{\text{edit}}(x) =   \lambda_{d-reg}\mathcal{L}_{\text{d-reg}}(x) + \lambda_{adv}\mathcal{L}_{\text{adv}}(x),
\end{equation}
where $\mathcal{L}_{\text{d-reg}}$ and $\mathcal{L}_{\text{adv}}$ are defined in Section~\ref{sec:var} and Section~\ref{sec:latent-disc}.

\paragraph{\textbf{Total loss}}
Our overall loss objective is defined as a weighted combination of the distortion and editability losses:
\begin{equation}
    \mathcal{L}(x) = \mathcal{L}_{\text{dist}}(x) + \lambda_{edit} \mathcal{L}_{\text{edit}}(x).
\end{equation}
\section{Evaluation}~\label{evaluation}
Evaluating our approach is particularly challenging. Our goal is to evaluate a tradeoff between distortion and two qualities --- perceptual quality and editability. Each alone is difficult to evaluate as they are perceptual in essence, and hard to objectively measure numerically. 
Perceptual quality is commonly quantitatively evaluated by measuring the discrepancy between the real and generated \emph{distributions} using algorithms, such as FID \cite{heusel2018gans}, SWD \cite{SWD} or IS \cite{salimans2016improved}. However, these methods do not always agree with human judgement, which is their true goal. Further, they are also affected by distortion, which makes them less suitable for evaluating the tradeoffs. As an example, let us consider a face reconstruction algorithm that perfectly reconstructs a given face, except that it adds to all men a realistic-looking beard. This is only a distortion problem, as the images all remain realistic to the human eye. However, the discrepancy between the distributions, which is measured only as a \emph{proxy of realism} will be affected.
Editability is even more difficult to evaluate because the quality of an edited image should be evaluated as a function of the magnitude of the edit change.
Additionally, qualitative measures are often susceptible to be biased.

To evaluate the results in a rigorous and fair manner, we provide numerous visual examples in the next section and in the supplementary materials to provide a large scale gallery for the reader's impression. It should be noted, that we take special care to avoid subjective bias in selecting the presented images. We do so by selecting the images according to their given order in their relevant test sets (avoiding the suspicion of cherry picking). We additionally conduct a user study to assess human opinion on the visual results. Last, we complete the evaluation by including all popular quantitative metrics. However, as we shall demonstrate below, such metrics often not only contradict the human opinion, but each other as well.

It should be emphasized that in the following examples, we present various editing results executed by different techniques. However, we do not evaluate their performance. Instead, we evaluate only the distortion-perception and distortion-editability tradeoffs which exist in all techniques.
Another important point to stress is that results presented by editing methods are often demonstrated on synthetic images generated by StyleGAN, with only a few editing examples on real images.
Here, however, since the focus of the paper is the inversion of real images, \textit{all} results are necessarily on real images. 

To properly evaluate our proposed method, one needs to evaluate the performance of distortion, perceptual quality, and editability.
Following the definitions introduced in Section \ref{subsec:preliminaries}, we now elaborate the evaluation protocols used.

\paragraph{\textbf{Distortion}}
We provide numerous qualitative results for the reader's impression.
To quantify these results, we apply the common metrics of $L_2$ and LPIPS~\cite{zhang2018unreasonable} on pairs of input and reconstructed images.

\paragraph{\textbf{Perceptual quality}} 
In addition to the large gallery shown, we provide the results of a user study to evaluate human subjective opinion.
We quantitatively evaluate the results by additionally measuring the FID \cite{heusel2018gans} and SWD~\cite{SWD} between the distributions of the real and reconstructed images.

\paragraph{\textbf{Editability}} We define editability as the ability to perform latent-space editing using any arbitrary technique while maintaining high-visual quality of the image obtained using the edited latent code. 
To this end, we follow our inversion method with several existing editing techniques: StyleFlow \cite{abdal2020styleflow}, InterFaceGAN \cite{shen2020interpreting}, GANSpace \cite{harkonen2020ganspace}, and SeFa \cite{shen2020closedform}. After performing inversion, we apply these techniques to edit the code in semantic manners such as pose, gender, and age for the human facial domain. We then generate images from the edited code and evaluate the perceptual quality of the generated images. 
We again provide numerous visual samples for the reader's impression and perform a user study. 
To further quantify these results, we also adopt the FID and SWD measures to compare the distributions of the original and the edited images. Note that FID and SWD measure perceptual quality for both reconstruction and editability. The difference lies with the distributions used to measure them.

\begin{figure}
    \centering
    \includegraphics[width=0.925\linewidth]{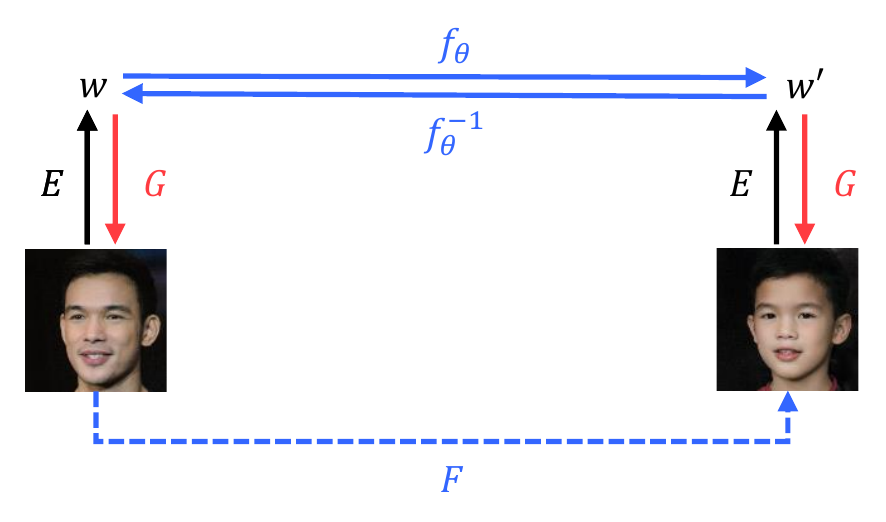}
    \vspace{-0.25cm}
    \caption{
    An Illustration of the Latent Editing Consistency (LEC) protocol. The color of the arrows correspond to their purpose, where black$\leftrightarrow$invert, blue$\leftrightarrow$edit and red$\leftrightarrow$generate. 
    We begin by inverting a real image, on the left, into the latent space of a pre-trained generator. 
    Then, the latent code is edited in some semantic manner by $f_\theta$.
    Using the obtained edited code, an image is synthesized using the generator.
    This resulting image, on the right, is then inverted back into the latent space. Finally, the obtained latent code is edited using the inverse of the original edit, $f^{-1}_\theta$, and an image is synthesized. 
    Ideally, the first and last images, as well as the latent codes, should be equal.}
    \label{fig:bnf_protocol}
\end{figure}

\paragraph{\textbf{Latent Editing Consistency}}
Here we present a new evaluation measure, which we call \textit{latent editing consistency (LEC)}, that combines two key components of GAN inversion methods meant for latent space editing. One captures the extent for which the inversion matches the true inverse of the generator, and the second captures how well-behaved the edits of the inversion outputs are. 
The protocol of this measure is illustrated in Figure \ref{fig:bnf_protocol}. We visually study the difference between the input and output images and quantitatively define the distance in the latent space by 
\begin{equation}
\label{eq:lec}
LEC(f_\theta) = \mathop{\mathbb{E}}_{x} \norm{E(x) - f_{\theta}^{-1}(E(G(f_{\theta}(E(x)))))}_2,
\end{equation}
where $E$ is an encoder and $f_{\theta}(w)$ is an invertible semantic latent editing function parameterized by $\theta$. For example, if the editing is performed by traversing along a linear direction $v$ in the latent space, then $f_{\theta}(w) = w + \alpha \cdot v$ and $\theta$ is composed of a scalar $\alpha$ and latent direction $v$.
Note that, in the optimal case, where $E$ is the perfect inverse of $G$, then $LEC=0$. In practice, however, the encoder is imperfect and thus $f_{\theta}(E(x))$ and $E(G(f_{\theta}(E(x))))$ are not exactly equal. By performing the inverse editing, LEC captures the extent for which the inherent inversion errors translate to errors in the subsequent editing. A well-behaved encoder suitable for latent editing should yield a small LEC difference in the latent space. 

\begin{figure*}
    \begin{subfigure}{0.29\textwidth}
        \setlength{\tabcolsep}{1pt}
        \centering
        \begin{tabular}{c c c c}
            \includegraphics[width=0.30\linewidth]{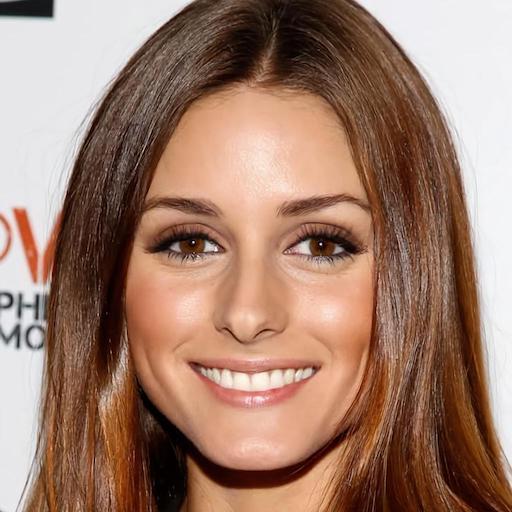} &
            \raisebox{0.15\linewidth}{\texttt{A}} & 
            \includegraphics[width=0.30\linewidth]{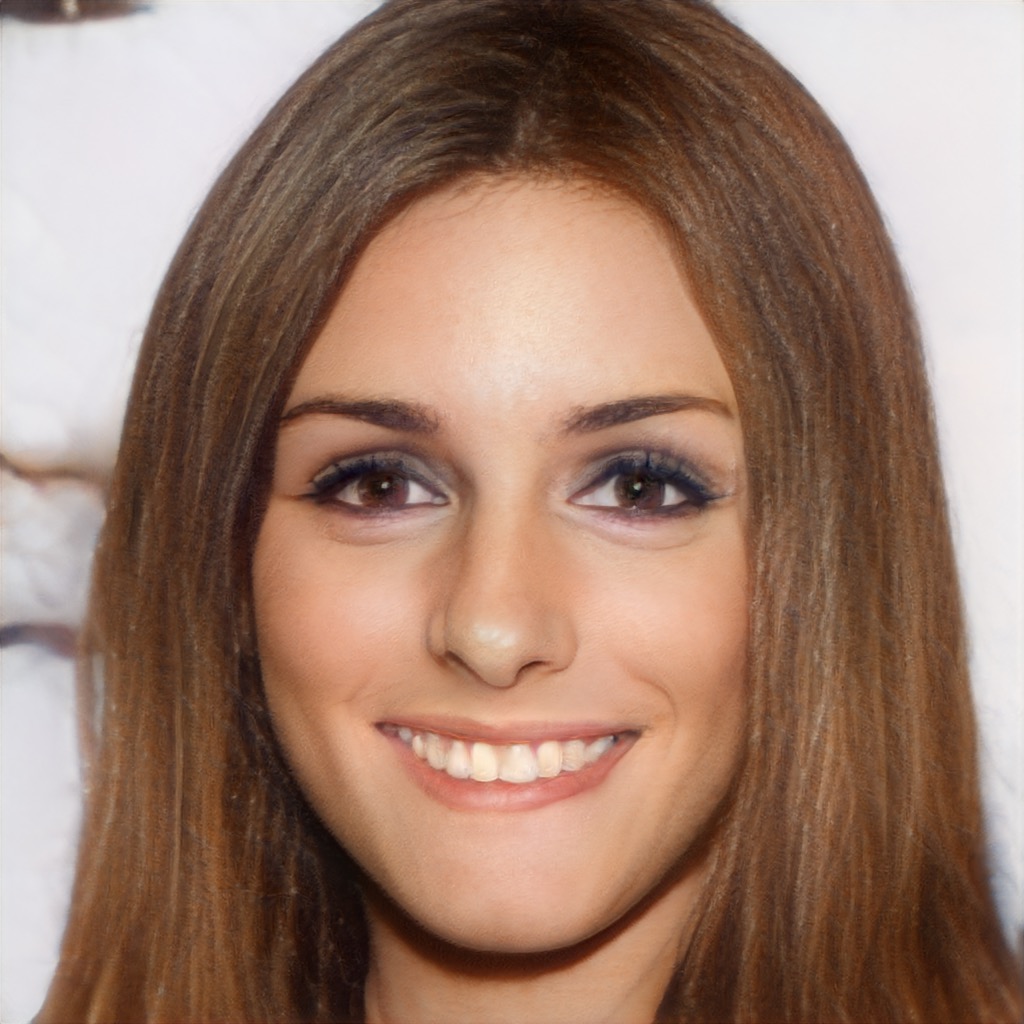} &
            \includegraphics[width=0.30\linewidth]{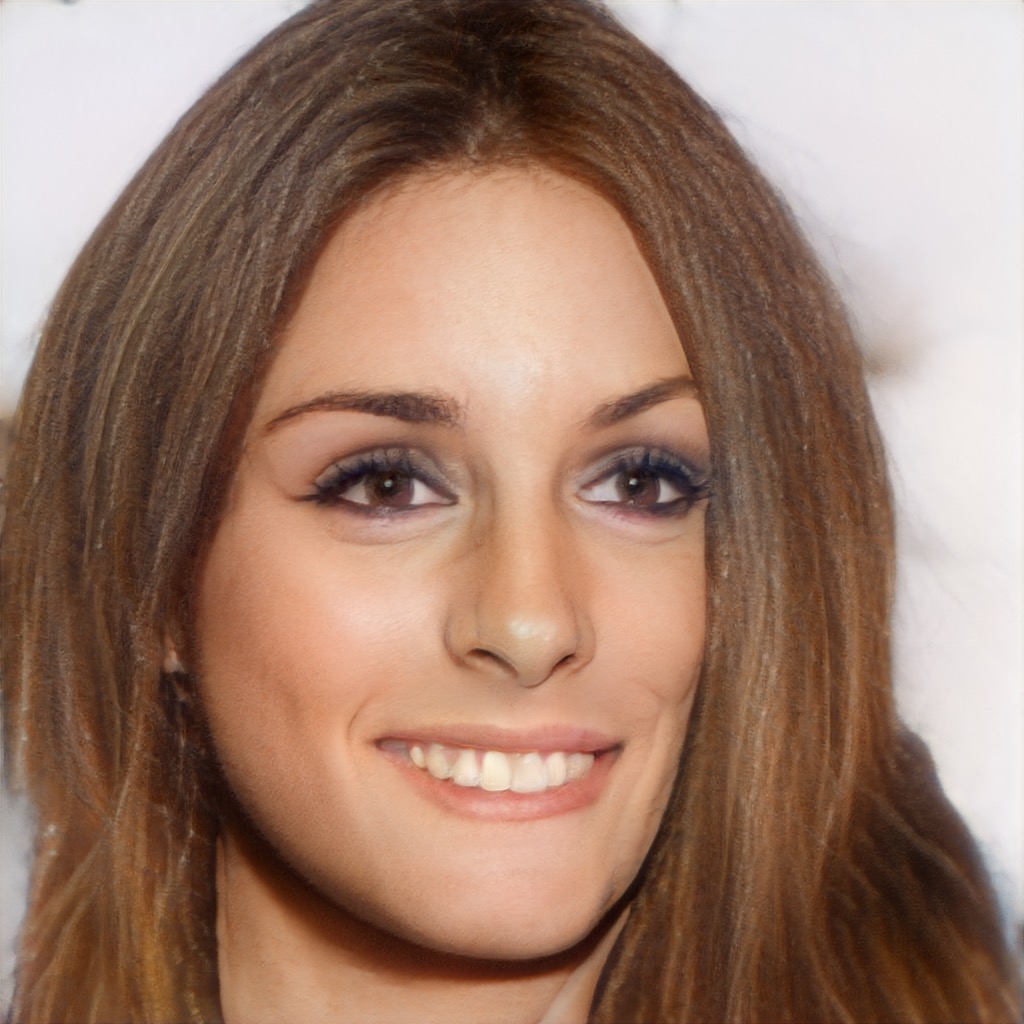} \\  
            \includegraphics[width=0.30\linewidth]{images/tradeoff/faces/1312.jpg} &
            \raisebox{0.15\linewidth}{\texttt{D}} & 
            \includegraphics[width=0.30\linewidth]{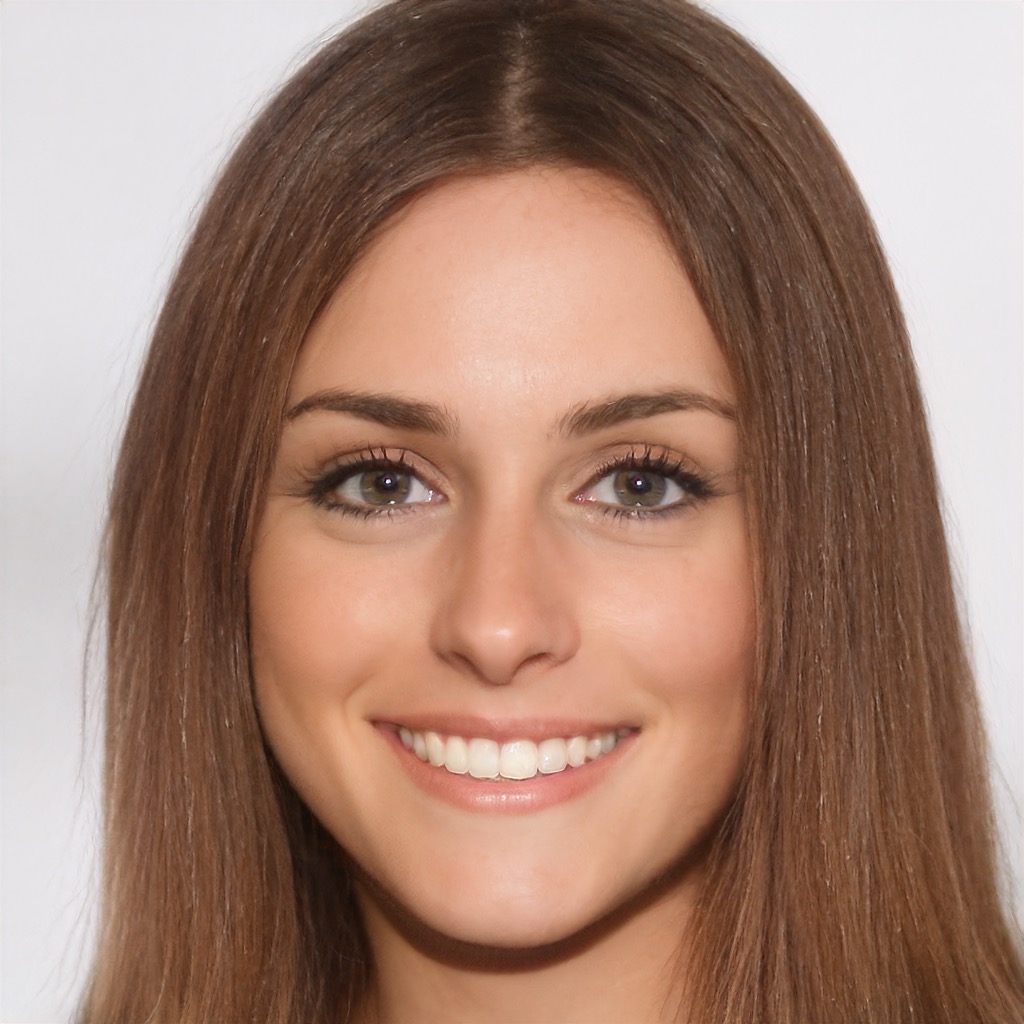} &
            \includegraphics[width=0.30\linewidth]{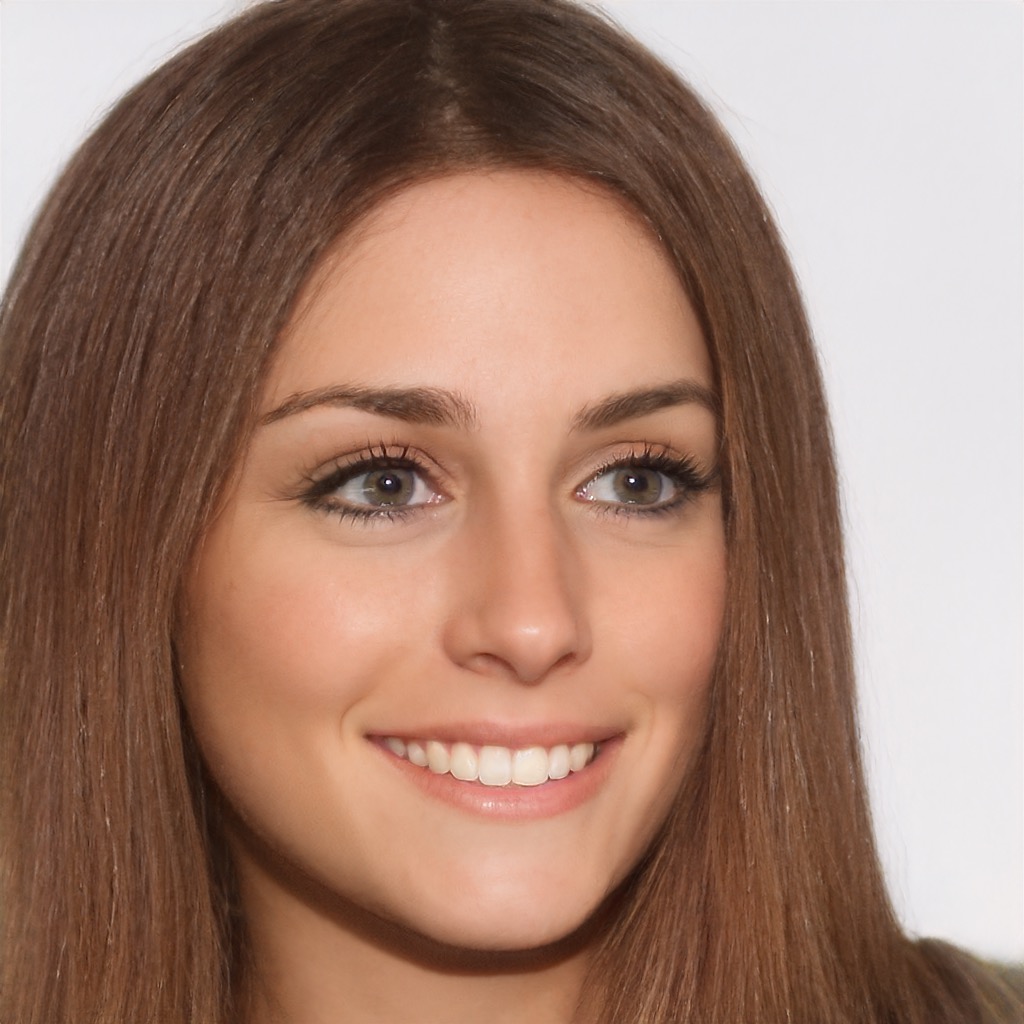} \\
            
            \includegraphics[width=0.30\linewidth]{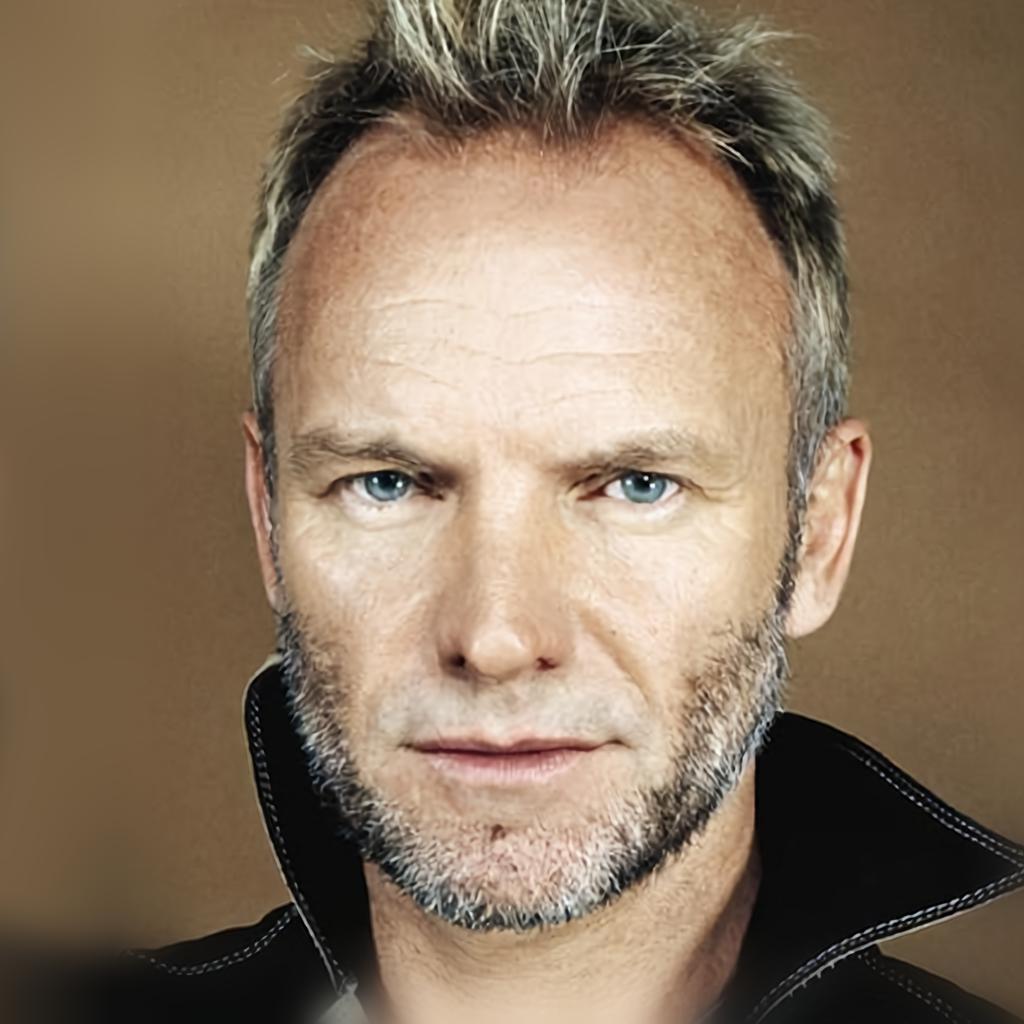} &
            \raisebox{0.15\linewidth}{\texttt{A}} & 
            \includegraphics[width=0.30\linewidth]{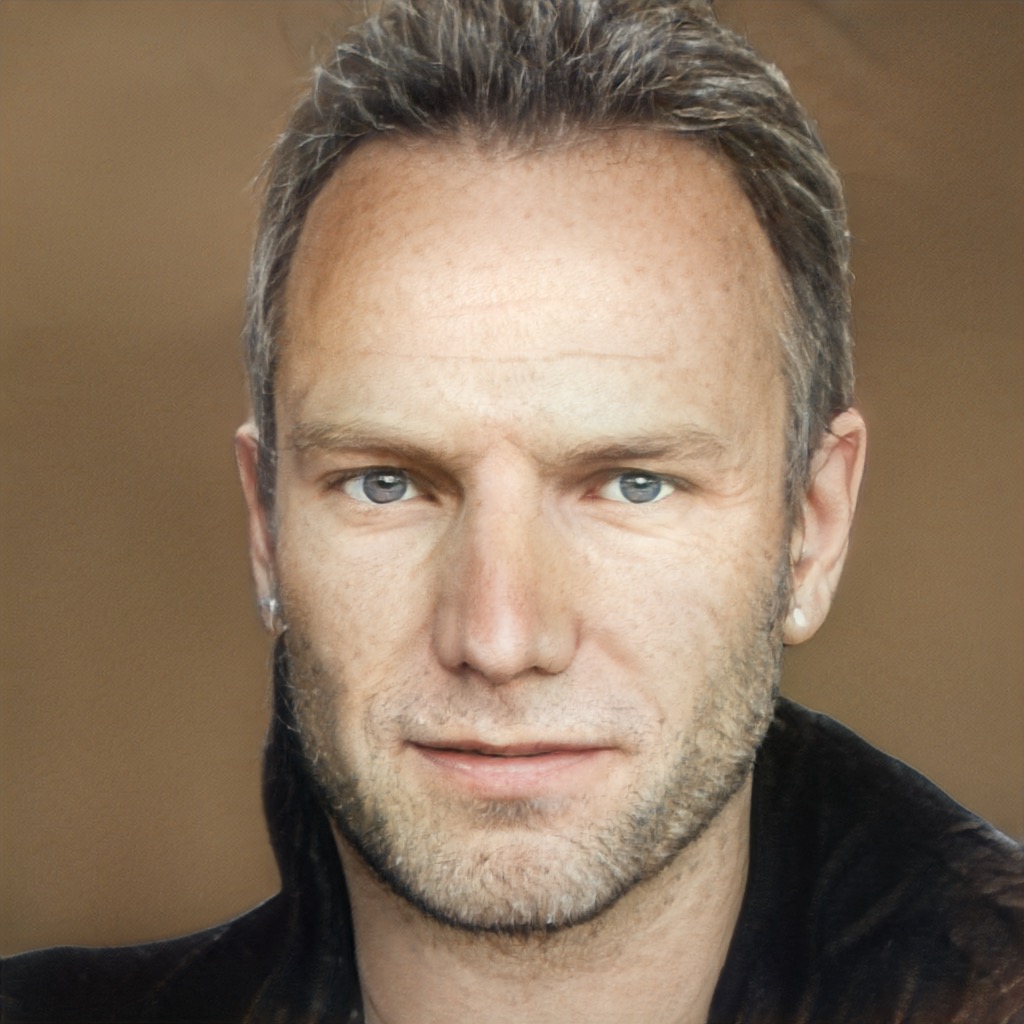} &
            \includegraphics[width=0.30\linewidth]{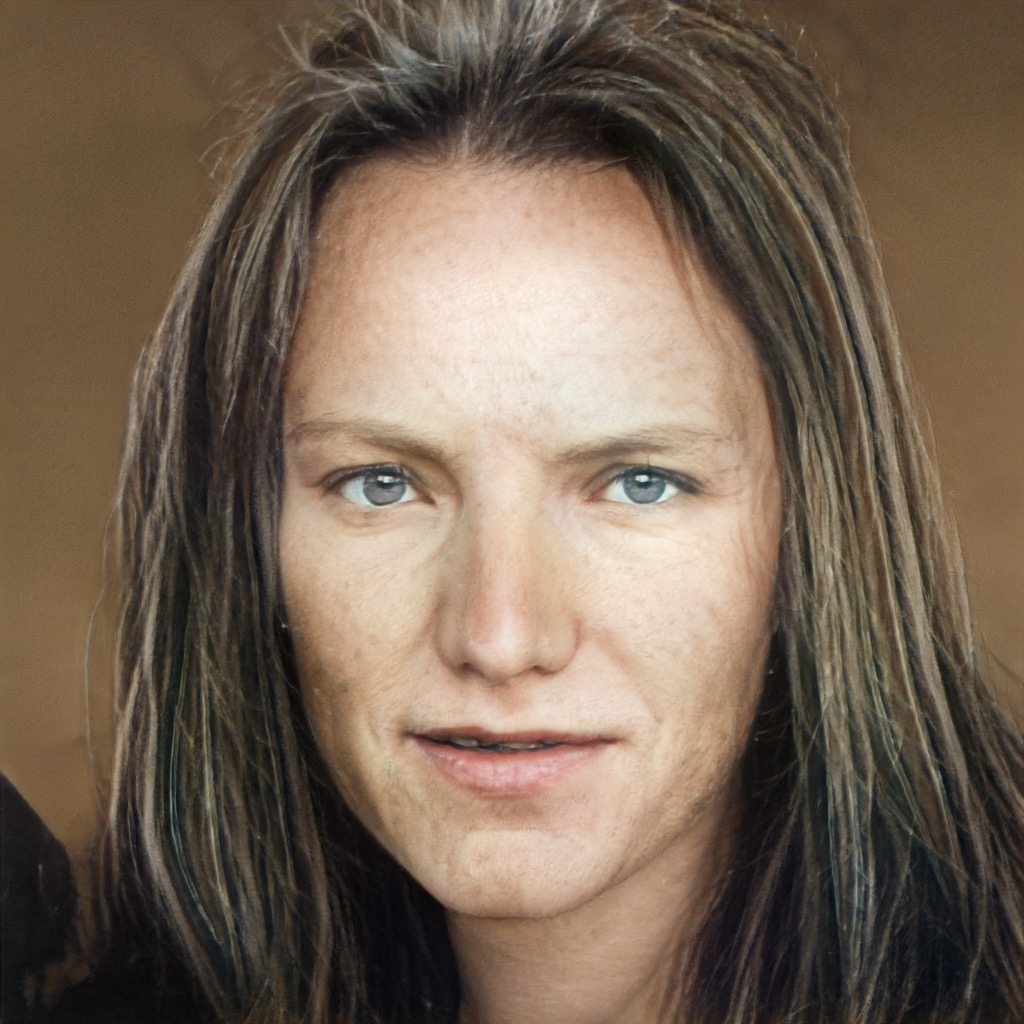} \\  
            \includegraphics[width=0.30\linewidth]{images/tradeoff/faces/1211.jpg} &
            \raisebox{0.15\linewidth}{\texttt{D}} & 
            \includegraphics[width=0.30\linewidth]{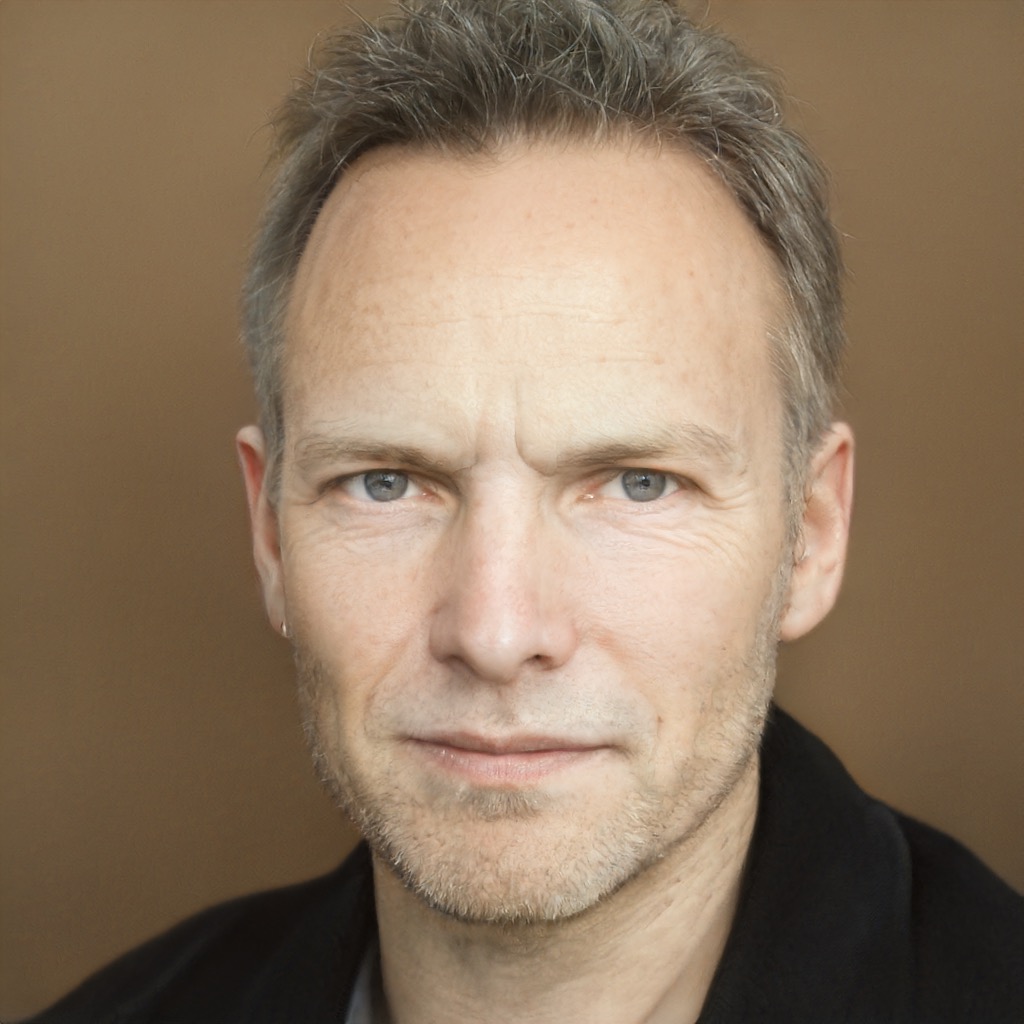} &
            \includegraphics[width=0.30\linewidth]{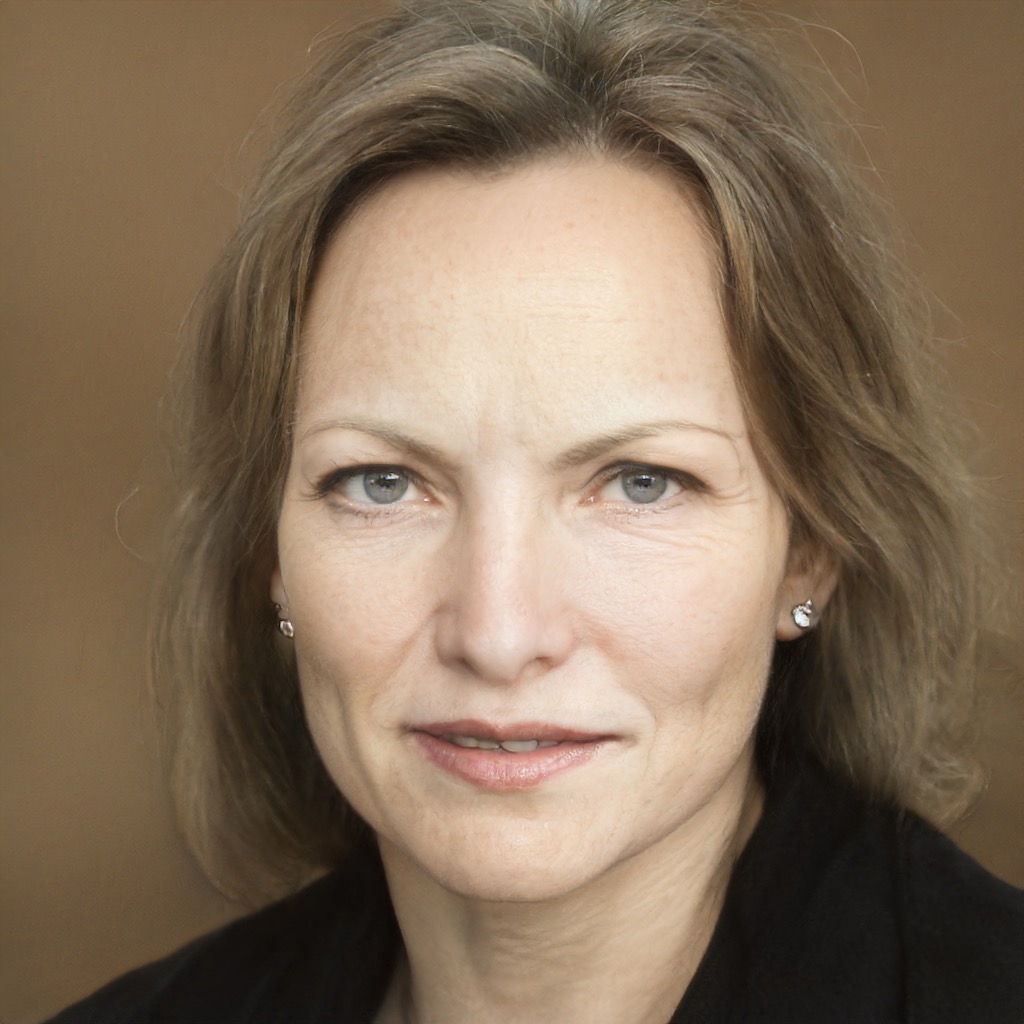} \\  
            
                        \includegraphics[width=0.30\linewidth]{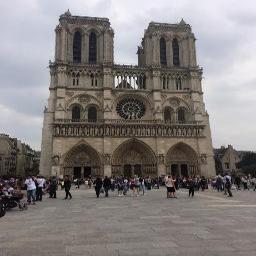} &
            \raisebox{0.15\linewidth}{\texttt{A}} & 
            \includegraphics[width=0.30\linewidth]{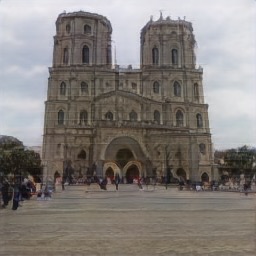} &
            \includegraphics[width=0.30\linewidth]{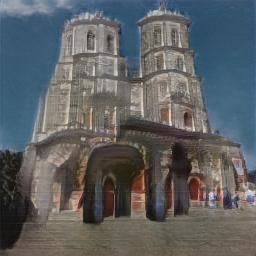} \\  
            \includegraphics[width=0.30\linewidth]{images/tradeoff/church/233.jpg} &
            \raisebox{0.15\linewidth}{\texttt{D}} & 
            \includegraphics[width=0.30\linewidth]{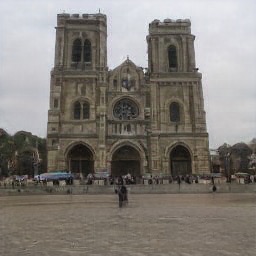} &
            \includegraphics[width=0.30\linewidth]{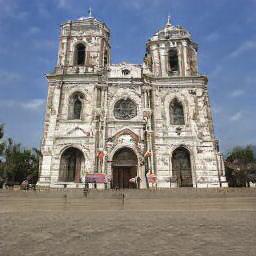} \\ 
            
            Source & & Inversion & Edit
        \end{tabular}
    \end{subfigure}%
    \begin{subfigure}{0.39\textwidth}
        \setlength{\tabcolsep}{1pt}
        \centering
        \begin{tabular}{c c c c}
                \includegraphics[height=0.2230769\linewidth]{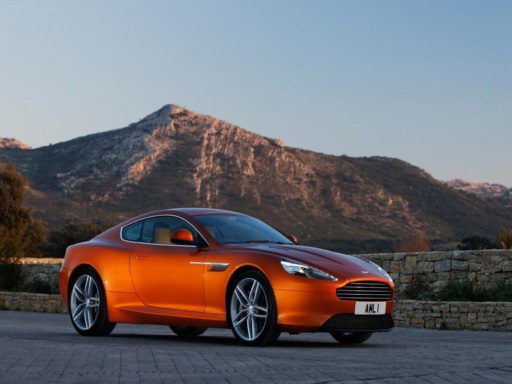} &
                \raisebox{0.11\linewidth}{\texttt{A}} & 
                \includegraphics[height=0.2230769\linewidth]{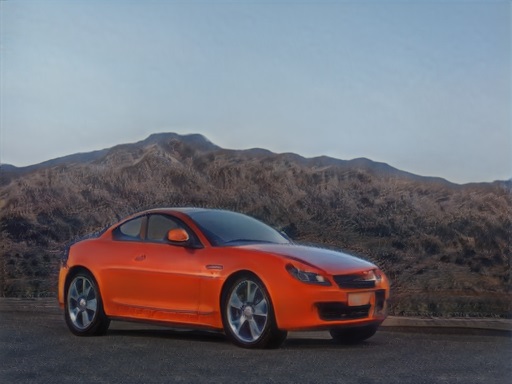} &
                \includegraphics[height=0.2230769\linewidth]{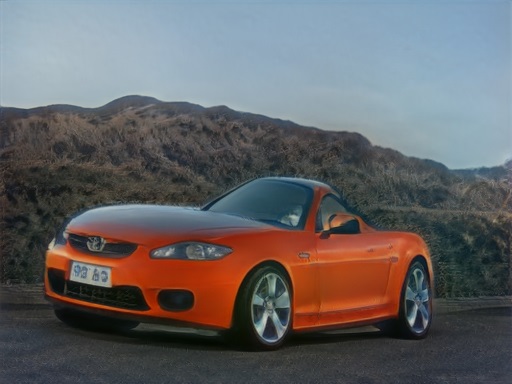} \\  
                \includegraphics[height=0.2230769\linewidth]{images/tradeoff/cars/01075_0_0.jpg} &
                \raisebox{0.11\linewidth}{\texttt{D}} & 
                \includegraphics[height=0.2230769\linewidth]{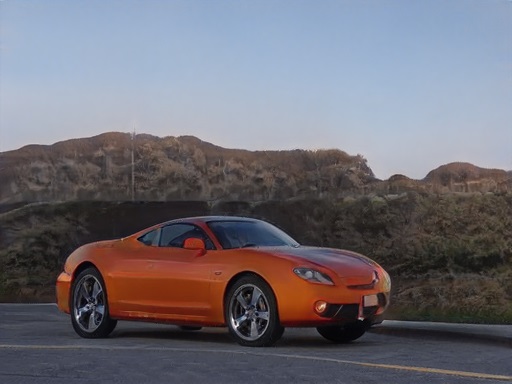} &
                \includegraphics[height=0.2230769\linewidth]{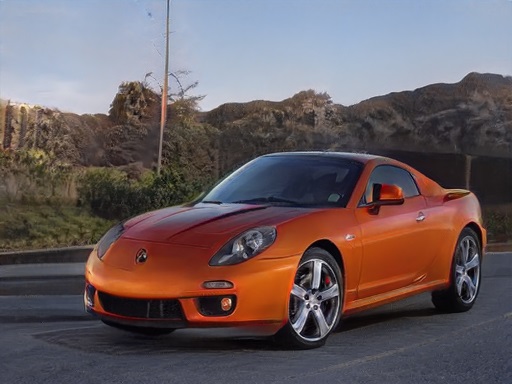} \\ 

                \includegraphics[height=0.2230769\linewidth]{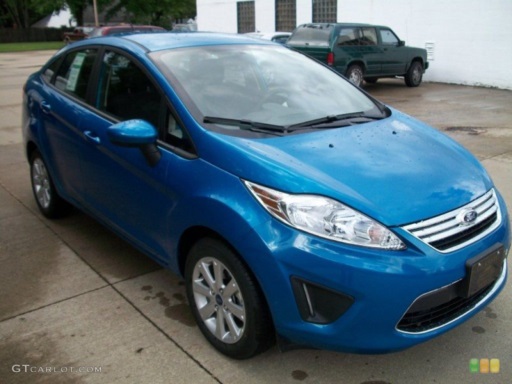} &
                \raisebox{0.11\linewidth}{\texttt{A}} & 
                \includegraphics[height=0.2230769\linewidth]{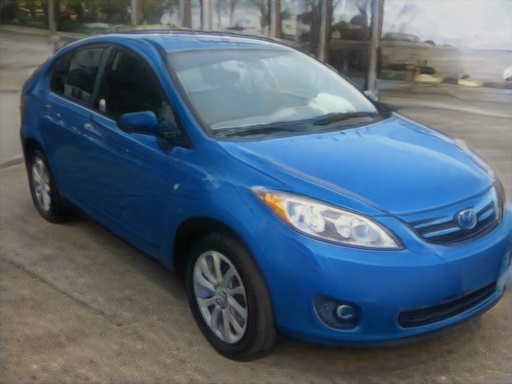} &
                \includegraphics[height=0.2230769\linewidth]{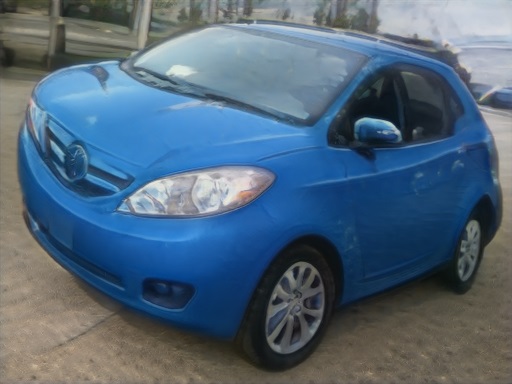} \\  
                \includegraphics[height=0.2230769\linewidth]{images/tradeoff/cars/00628_0_0.jpg} &
                \raisebox{0.11\linewidth}{\texttt{D}} & 
                \includegraphics[height=0.2230769\linewidth]{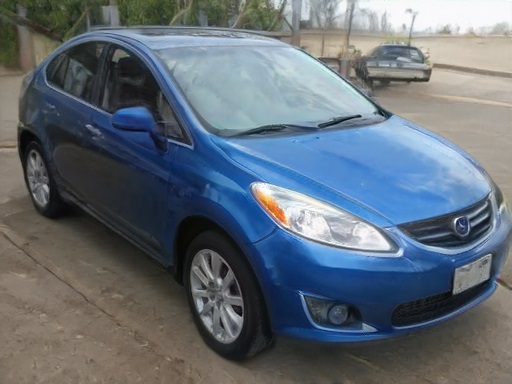} &
                \includegraphics[height=0.2230769\linewidth]{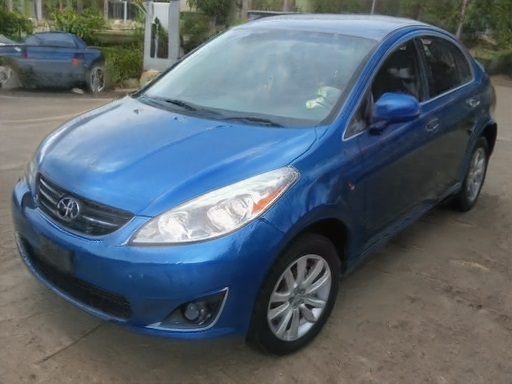} \\ 

                \includegraphics[height=0.2230769\linewidth]{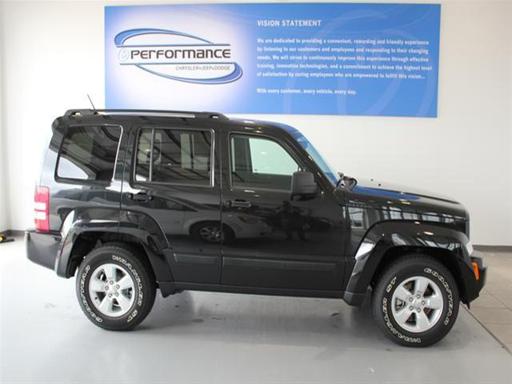} &
                \raisebox{0.11\linewidth}{\texttt{A}} & 
                \includegraphics[height=0.2230769\linewidth]{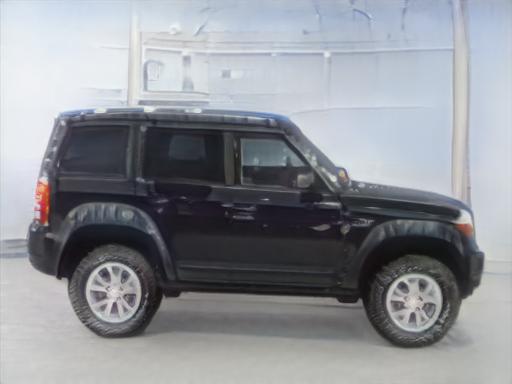} &
                \includegraphics[height=0.2230769\linewidth]{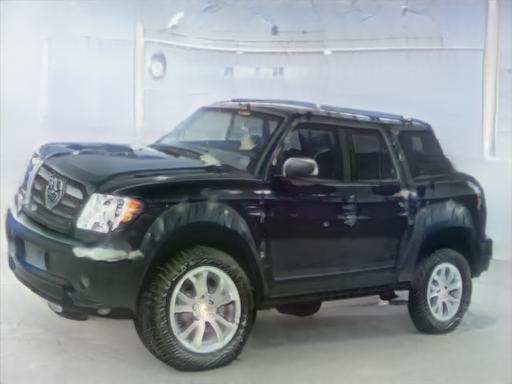} \\  
                \includegraphics[height=0.2230769\linewidth]{images/tradeoff/cars/00271_src.jpg} &
                \raisebox{0.11\linewidth}{\texttt{D}} & 
                \includegraphics[height=0.2230769\linewidth]{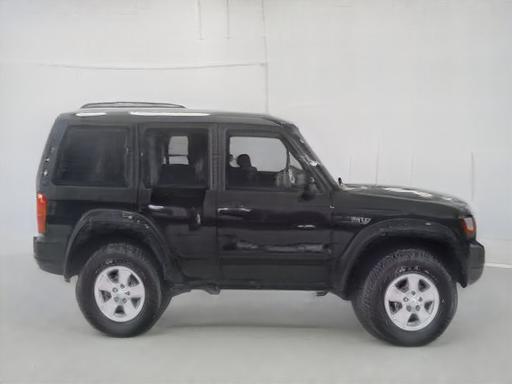} &
                \includegraphics[height=0.2230769\linewidth]{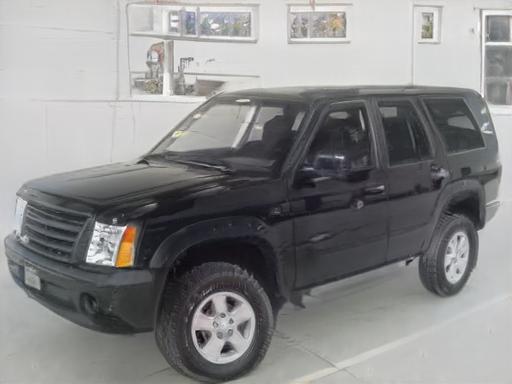} \\ 
                Source & & Inversion & Edit
            \end{tabular}
    \end{subfigure}
    \begin{subfigure}{0.29\textwidth}
        \setlength{\tabcolsep}{1pt}
        \centering
        \begin{tabular}{c c c c}
            
            \includegraphics[width=0.30\linewidth]{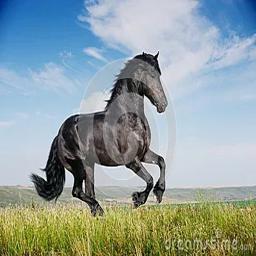} &
            \raisebox{0.15\linewidth}{\texttt{A}} & 
            \includegraphics[width=0.30\linewidth]{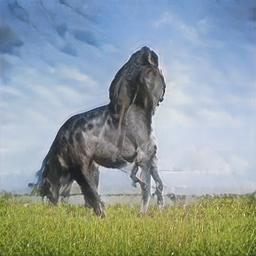} &
            \includegraphics[width=0.30\linewidth]{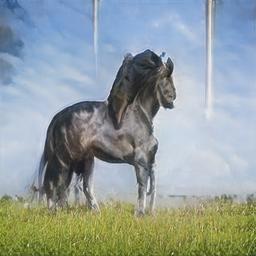} \\  
            \includegraphics[width=0.30\linewidth]{images/tradeoff/horse/01832.jpg} &
            \raisebox{0.15\linewidth}{\texttt{D}} & 
            \includegraphics[width=0.30\linewidth]{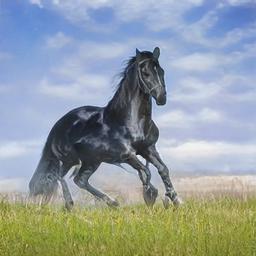} &
            \includegraphics[width=0.30\linewidth]{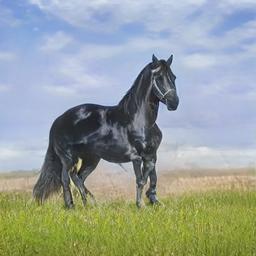} \\          
            
            \includegraphics[width=0.30\linewidth]{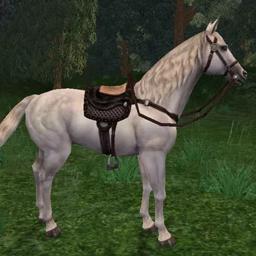} &
            \raisebox{0.15\linewidth}{\texttt{A}} & 
            \includegraphics[width=0.30\linewidth]{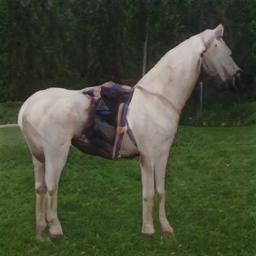} &
            \includegraphics[width=0.30\linewidth]{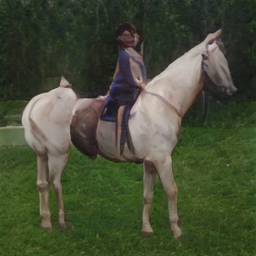} \\  
            \includegraphics[width=0.30\linewidth]{images/tradeoff/horse/01612.jpg} &
            \raisebox{0.15\linewidth}{\texttt{D}} & 
            \includegraphics[width=0.30\linewidth]{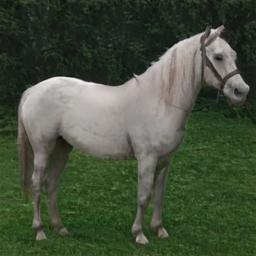} &
            \includegraphics[width=0.30\linewidth]{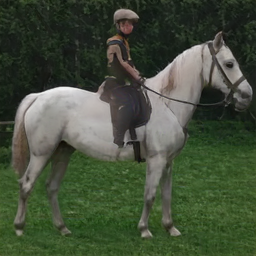} \\  
            
            \includegraphics[width=0.30\linewidth]{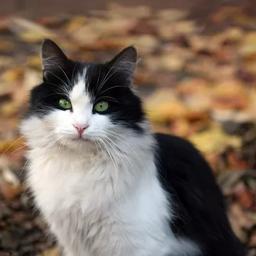} &
            \raisebox{0.15\linewidth}{\texttt{A}} & 
            \includegraphics[width=0.30\linewidth]{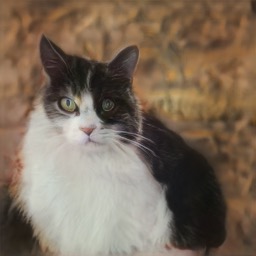} &
            \includegraphics[width=0.30\linewidth]{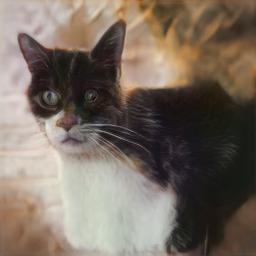} \\  
            \includegraphics[width=0.30\linewidth]{images/tradeoff/cat/2369.jpg} &
            \raisebox{0.15\linewidth}{\texttt{D}} & 
            \includegraphics[width=0.30\linewidth]{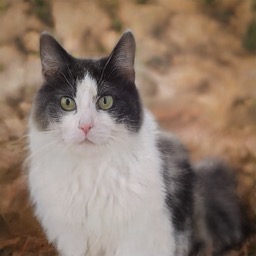} &
            \includegraphics[width=0.30\linewidth]{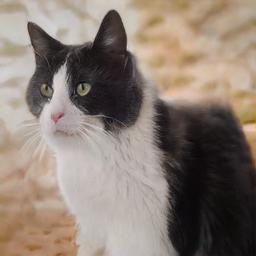} \\  

            Source & & Inversion & Edit
        \end{tabular}
    \end{subfigure}%
    \vspace{-0.15cm}
    \caption{
    We show triplets of a source image, its inversion, and an edit applied on the inverted image for multiple domains. 
    In the odd rows, the inversions are obtained by our baseline encoder (\texttt{A}). In the even rows, we use configuration \texttt{D}, which encodes images closer to \w. Observe the tradeoff between the distortion and perceptual quality of the inverted images. 
    For example, in the image of the white horse, observe the lower distortion of the inverted image using configuration \texttt{A} (e.g. the saddle is preserved). However, the perceptual quality is lower than that obtained by \texttt{D} (e.g. the horse's head is not realistic). 
    With respect to editability, notice how in the top left image of the female, the pose edit does not faithfully change the hair in \texttt{A}. Conversely, \texttt{D} obtains a realistic and visually-pleasing editing result at the cost of a subtle degradation in distortion.
    From top to bottom and left to right the edits are: head pose, gender, daylight, viewpoint (x3), horse pose, horse rider, cat pose.
    }
    \label{fig:tradeoff}
\end{figure*}

Note that edits with $f_{\theta}$ translate to visual results in the image-space after applying $G$. Therefore, there exists a corresponding function $F$ that performs the same edits in the image-space (see Figure \ref{fig:bnf_protocol}). If $F$ was tractable to compute, one could evaluate how far $E$ is from editing equivariance:
\begin{equation}
\label{eq:equivariance}
\mathop{\mathbb{E}}_{x}\norm{f_{\theta}(E(x)) - E(F(x))}_2.
\end{equation}
Since $F$ is usually intractable, however, LEC can be considered as a proxy for equivariance.
\vspace{0.4cm}
\section{Experiments}~\label{experiments}

\vspace{0.75cm}
\begin{SCtable}[50]
    \begin{tabular}{|l | c | c|}
        \hline
        & d-reg & $D_{\mathcal{W}}$\\
        \hline
        \texttt{A} &  &  \\
        \hline
        \texttt{B} & \cmark & \\
        \hline
        \texttt{C} & & \cmark  \\
        \hline
        \texttt{D} & \cmark & \cmark \\
        \hline
    \end{tabular}
    \vspace{0.5cm}
    \caption{Our encoder and training configurations. Recall, the delta-regularization (d-reg) refers to the variation minimization described in ~\ref{sec:var}.
    $D_w$ refers to the latent discriminator as explained in ~\ref{sec:latent-disc}.}
    \label{table:architecture_configs}
\end{SCtable}

\vspace{-1.25cm}
\subsection{Settings}
\paragraph{\textbf{Configurations}}
Throughout this section, we explore various configurations of our e4e method. We concisely summarize these configurations in Table~\ref{table:architecture_configs}. Note that the two editing losses listed in the table are used in conjunction with the distortion loss $\mathcal{L}_{\text{dist}}$ defined in Equation~\ref{eq:rec_loss}.

\paragraph{\textbf{Datasets}}
To illustrate the generalization of our approach, we perform extensive experimentation on a diverse set of challenging domains.
For the facial domain, we use the FFHQ~\cite{karras2019style} dataset for training and the CelebA-HQ~\cite{karras2017progressive} test dataset for evaluation. We use the Stanford Cars~\cite{KrauseStarkDengFei-Fei_3DRR2013} dataset for training and evaluation on the cars domain. We additionally provide results on the LSUN~\cite{yu2016lsun} Horse, Cats, and Churches datasets. The diversity of the LSUN domains is significantly greater than that of human faces and cars. We therefore find them to be more challenging.
Note that all results are obtained using official StyleGAN2 generators trained on images from that domain.

\subsection{Tradeoff} \label{sec:exp-tradeoff}
Here we show qualitative and quantitative results to support the existence of both the distortion-perception and distortion-editability tradeoffs. More specifically, we show that approaching \w tends to increase the distortion, perceptual quality, and editability.
We start by showing that inversions obtained by our complete method (denoted by configuration \texttt{D}) are much closer to \w than inversions obtained by configuration \texttt{A}, which reduces to pSp with the addition of the generic similarity loss defined in Section~\ref{losses}. Note that for faces, configuration \texttt{A} is equivalent to pSp. By comparing the distortion and editability of the two variants, we derive insights about the tradeoff of the latent space.

\setlength\tabcolsep{2.5pt}
\begin{table}[]
    \centering
    \begin{tabular}{|c c|c c| c c|c c|}
        \hline
         \multicolumn{2}{|c|}{}  & \multicolumn{2}{c|}{\textbf{Distortion}} & \multicolumn{2}{c|}{\textbf{Perception}} & \multicolumn{2}{c|}{\textbf{Editability}} \\
         \multicolumn{1}{|c}{\textbf{Domain}}& \textbf{Conf.} & $L_2$ & LPIPS & FID & SWD & FID & SWD  \\
         \hline
         \multirow{2}{*}{Faces} & \texttt{A} & \textbf{0.03} & \textbf{0.17} & \textbf{25.17} & 48.72 & \textbf{62.46} & 48.75 \\
         & \texttt{D} & 0.05 & 0.23 & 30.96 & \textbf{40.54} & 81.08 & \textbf{43.63} \\
         \hline
         \multirow{2}{*}{Cars} & \texttt{A} & 0.10 & 0.32 & \textbf{10.56} & \textbf{22.08} & \textbf{12.92} & \textbf{24.30} \\
         & \texttt{D} & 0.10 & 0.32 & 12.18 & 22.71 & 15.44 & 26.83 \\
        \hline
         \multirow{2}{*}{Horse} & \texttt{A} & \textbf{0.11} & \textbf{0.36} & \textbf{34.13} & 35.79 & 40.48 & \textbf{31.84} \\
         & \texttt{D} & 0.16 & 0.45 & 37.07 & \textbf{31.91} & \textbf{35.31} & 32.18 \\
        \hline
         \multirow{2}{*}{Cats} & \texttt{A} & \textbf{0.10} & \textbf{0.41} & \textbf{42.60} & \textbf{32.83} & \textbf{217.44} & \textbf{45.13} \\
         & \texttt{D} & 0.14 & 0.48 & 49.90 & 40.59 &  222.30 & 51.07 \\
        \hline
         \multirow{2}{*}{Church} & \texttt{A} &  \textbf{0.09} &  \textbf{0.32} &  \textbf{25.96} & 38.90 & 26.91 & 29.63 \\
         & \texttt{D} & 0.13 & 0.40 & 27.09 &  \textbf{31.18} & \textbf{21.87} & \textbf{23.32} \\
        \hline
    \end{tabular}
    \vspace{-0.2cm}
    \caption{Perceptual quality metrics computed on inversions and edited images obtained by configurations \texttt{A} and \texttt{D}. For all computed metrics, lower is better. Recall that latent codes obtained by \texttt{D} are encouraged to be close to \w, and therefore the distortion is higher. Note that FID and SWD contradict each other and do not reflect human judgments.}
    \label{tab:tradeoff-results}
\end{table}

\vspace{0.2cm}
\paragraph{\textbf{Which latent vectors are close to \w?}}
We begin by inspecting the variation of the latent codes produced by configurations \texttt{A} and \texttt{D} on the CelebA-HQ~\cite{karras2017progressive} test set. Specifically, given a latent code $w = (w_1, ..., w_k)\in$ \wkstar obtained from the encoder, we calculate $ || w - \overline{w}_{\mu}  ||_2$ where $w_{\mu} = \frac{1}{k}{\sum_{i=1}^k{w_i}}$ and $\overline{w}_{\mu} = (w_\mu, ..., w_\mu)$. We then take the average over all embeddings obtained from the test set. We obtain a variation of $324.76$ for codes generated by configuration \texttt{A} and $20.18$ for codes generated by configuration \texttt{D}, respectively. Therefore, latent codes obtained from \texttt{D} are closer to \w in the sense of having a lower variation.

The deviation from the distribution \w is more difficult to measure as there is no known explicit model for it. We therefore perform the following test. We sample a single $w \in $ \w by using StyleGAN's mapping function, and generate the corresponding image. We then encode the resulting image back into StyleGAN's latent space using both configurations \texttt{A} and \texttt{D}. We repeat this procedure on $5,000$ randomly sampled $w \in \mathcal{W}$ vectors, and measure the average Euclidean distance between the source $w$ and the resulting latent code $w' := E(G(w))$. That is, we compute $\mathbb{E}_{w \in \mathcal{W}} \left [|| w' - w ||_2 \right]$. 

For the facial domain, we obtain an average distance of $364.51$ for configuration \texttt{A} and $57.67$ for configuration \texttt{D}. For cars, we obtain $316.14$ for configuration \texttt{A} and $39.85$ for \texttt{D}. Note that while Euclidean distance does not necessary reflect the distance within the \w space, the distance obtained by encoding with configuration \texttt{A} is an order of magnitude greater than that of configuration \texttt{D}, suggesting that the encoding of configuration \texttt{D} is indeed closer to \w, even within the subspace. Therefore, latent codes obtained by configuration \texttt{D} are closer to \w in both the sense of their deviation from the distribution \w and in the sense of the variation between their different style codes.

\paragraph{\textbf{Showing the tradeoff}}
Having measured the proximity of each method to \w, we now turn to showing the existence of the distortion-editability and distortion-perception tradeoffs. 
Recall that our claim is that latents that are closer to \w have higher distortion, but better perceptual quality and editability. As we have shown, latents obtained by configuration \texttt{D} are closer to \w than those obtained by \texttt{A}. Therefore, to support our claim, we use their respective latent codes.

\setlength\tabcolsep{9pt}
\begin{table}[]
    \centering
    \begin{tabular}{|c c| c c|}
        \hline
         \textbf{Domain} &\textbf{Conf.} & \textbf{Perception} & \textbf{Editability} \\ 
         \hline
         \multirow{2}{*}{Faces} & \texttt{A}  & 43.67\% & 14.33\% \\
         & \texttt{D} &  \textbf{56.33\%} & \textbf{85.67\%} \\
         \hline
         \multirow{2}{*}{Cars} & \texttt{A}  & 25.67\% & 18.33\% \\
         & \texttt{D} & \textbf{74.33\%} & \textbf{81.67\%} \\
        \hline
         \multirow{2}{*}{Horse} & \texttt{A}  & 6.50\% & 13.50\% \\
         & \texttt{D}  & \textbf{93.50\%} & \textbf{86.50\%} \\
        \hline
    \end{tabular}
    \vspace{-0.2cm}
    \caption{
    User study results. Each cell displays the percentage of respondents that favored the corresponding configuration for each of the two evaluated aspects.
    Observe, that both the edited and reconstructed images produced by \texttt{D} were chosen as being more realistic across all evaluated domains.
    }
    \label{tab:user-study}
\end{table}

In Figure~\ref{fig:tradeoff}, we provide a visual comparison of the inversions obtained by configurations \texttt{A} and \texttt{D}. For each domain, we also perform an edit for each inverted image. For cars, we use directions obtained by GANSpace~\cite{harkonen2020ganspace}; for faces we use StyleFlow~\cite{abdal2020styleflow}; and for horses, cats, and churches we use SeFa \cite{shen2020closedform}. As can be observed, while the distortion of images inverted by configuration \texttt{A} is lower, the visual quality of both the reconstructed and edited images inverted by configuration \texttt{D} is much higher.

Next, we perform a quantitative evaluation by measuring metrics detailed in Section \ref{evaluation}. The results are provided in Table \ref{tab:tradeoff-results}.
We expect the distortion achieved by configuration \texttt{A} to be lower, as it encodes images further away from \w. This can indeed be seen in the results.
Note that for both the perception and editability, FID and SWD mostly \emph{contradict} each other. This provides additional evidence for the ``leakage" of distortion into the perceptual quality measurements. As shown in Figure~\ref{fig:tradeoff}, reconstructions obtained by \texttt{A} and the subsequent edits applied contain many artifacts which are not faithfully captured by the FID and SWD measures.

Finally, we conduct a qualitative human evaluation to evaluate the perceptual quality and editability of the two configurations. As editability is measured by the perceptual quality of the edited images, all questions were with regard to perceptual quality.
The user study consisted of two sections, each including several questions from the facial, cars, and horses domains. A total of 87 respondents participated in the study.
In each question, the respondents were given a side-by-side comparison of images generated by configurations \texttt{A} and \texttt{D}, and were asked to choose the more realistic image. In the first section, the images shown were reconstructions of real images sampled from the test set. In the second section, the shown images were of results obtained after performing editing on inverted latent codes. The results of the user study are shown in Table \ref{tab:user-study}. As can be seen, the perceptual quality of images obtained by configuration \texttt{D} is higher both on the reconstructed and edited images. This further shows that the common perceptual metrics (e.g. FID and SWD) do not accurately reflect human perceptual judgement.

\begin{figure*}

    \setlength{\tabcolsep}{1pt}
    \centering
        \centering
            \begin{tabular}{c c c c c c c c c c c}
            \raisebox{0.035\textwidth}{\texttt{A}} & 
            \includegraphics[height=0.08\textwidth]{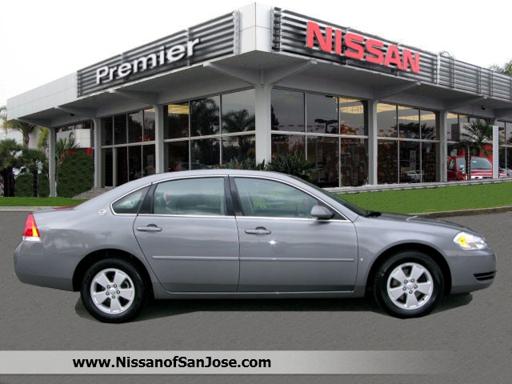} &
            \includegraphics[height=0.08\textwidth]{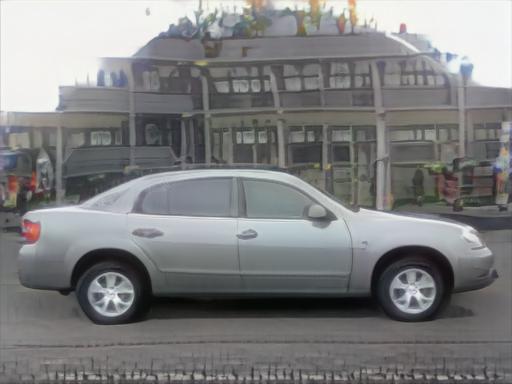} &
            \includegraphics[height=0.08\textwidth]{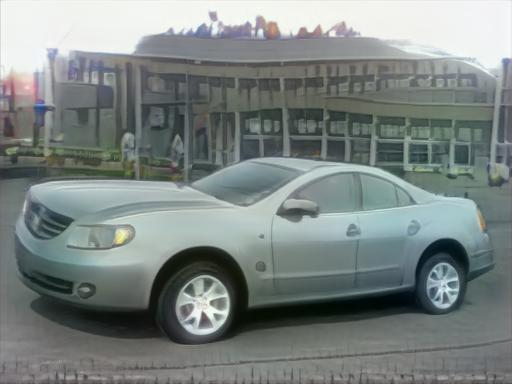} &
            \includegraphics[height=0.08\textwidth]{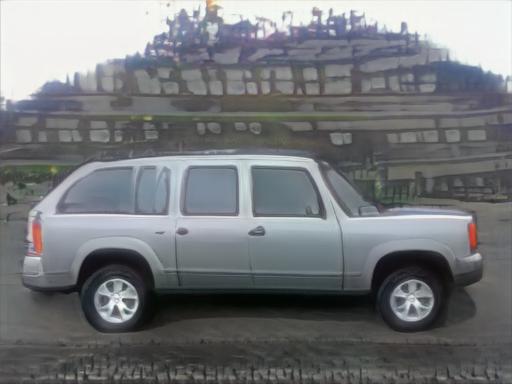} &
            \includegraphics[height=0.08\textwidth]{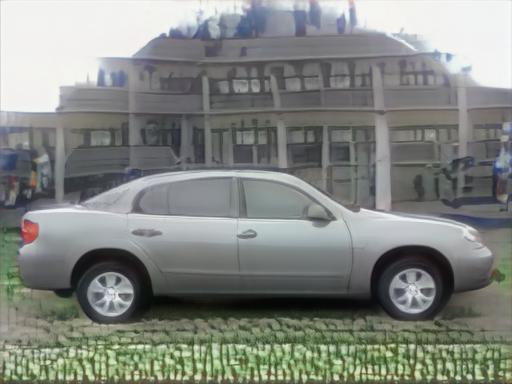} &
            \includegraphics[height=0.08\textwidth]{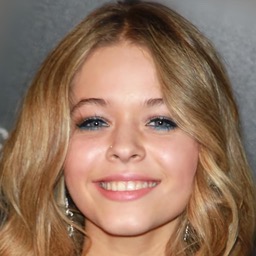} &
            \includegraphics[height=0.08\textwidth]{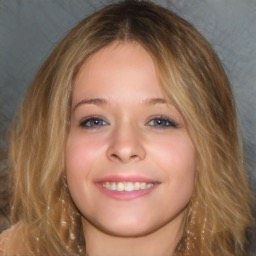} &
            \includegraphics[height=0.08\textwidth]{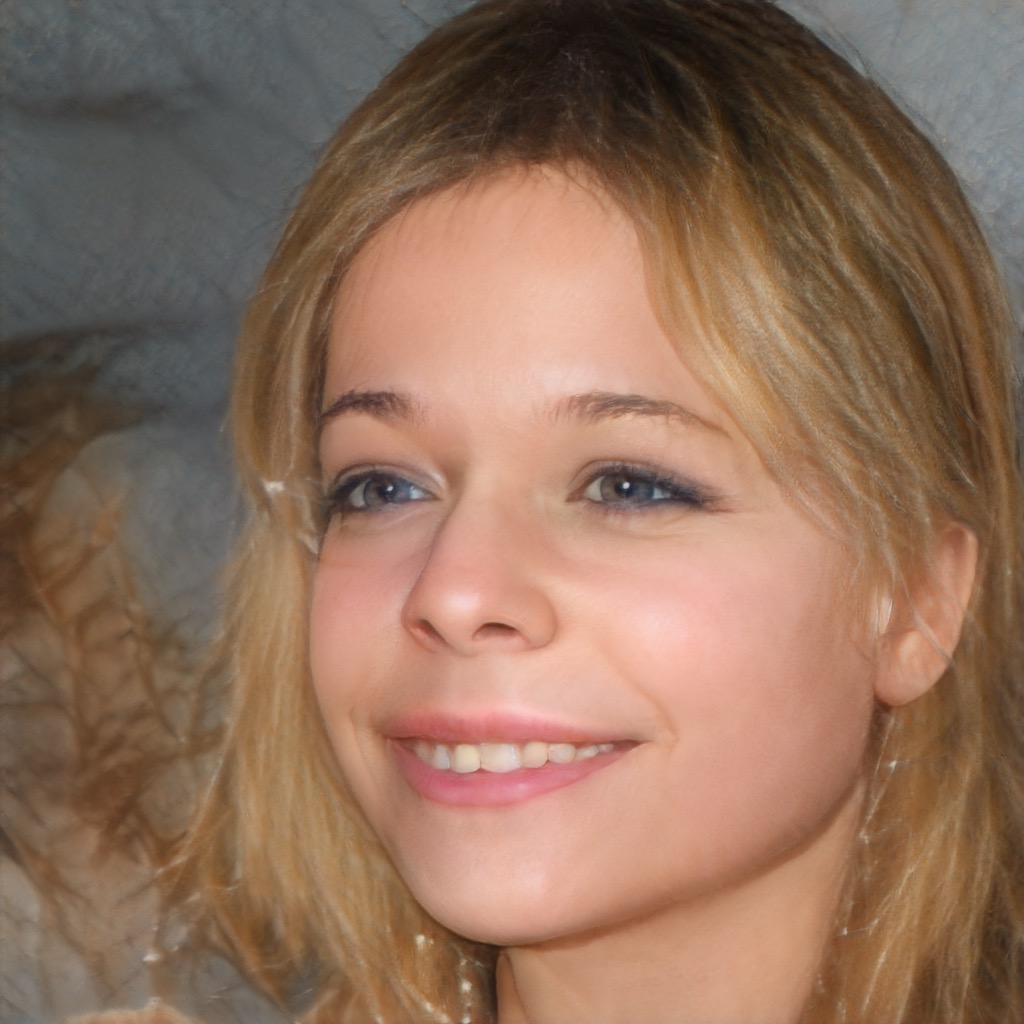} & 
            \includegraphics[height=0.08\textwidth]{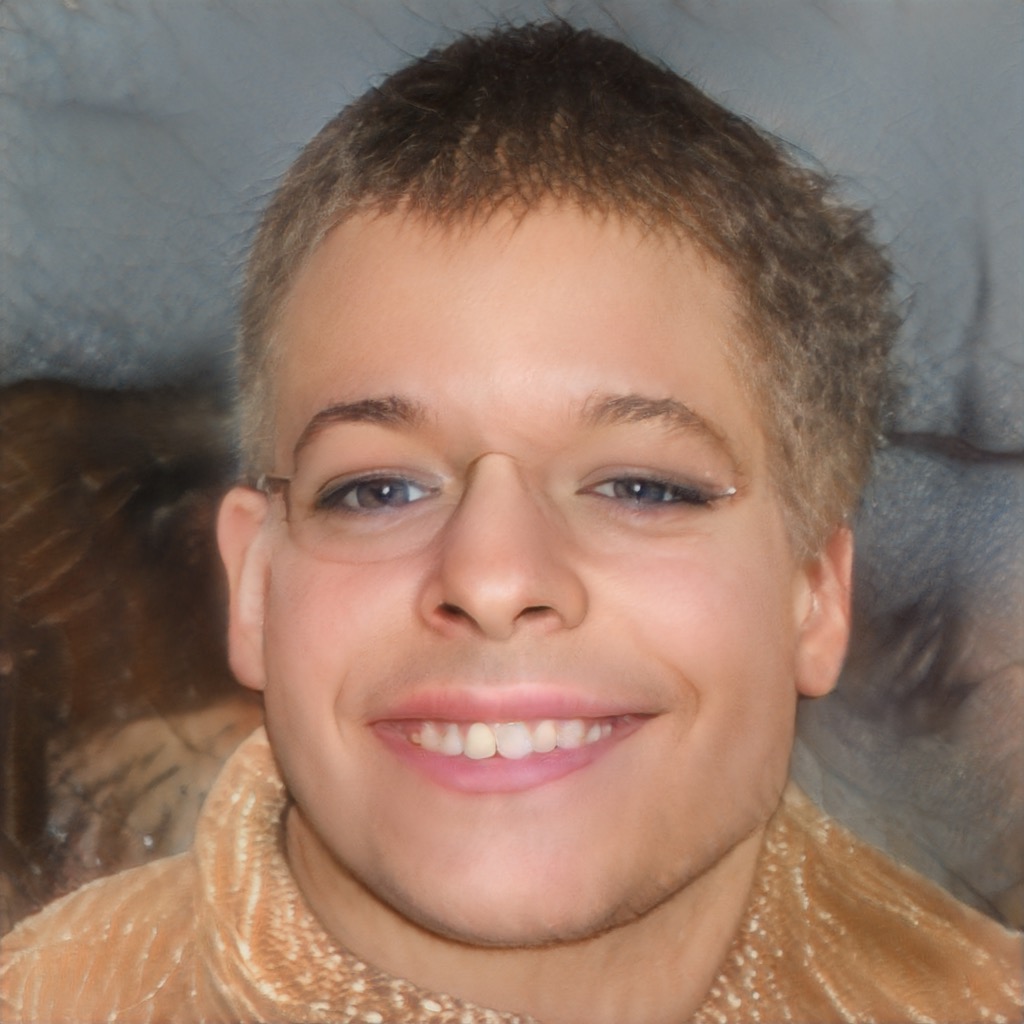} &
            \includegraphics[height=0.08\textwidth]{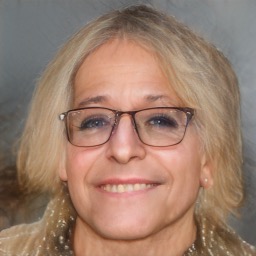}
            \\
            \raisebox{0.035\textwidth}{\texttt{B}} &
            \includegraphics[height=0.08\textwidth]{images/tradeoff_cars/source.jpg} &
            \includegraphics[height=0.08\textwidth]{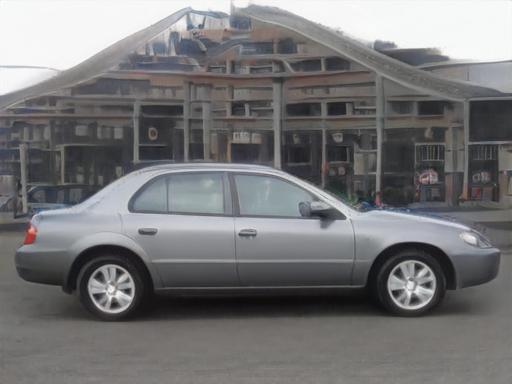} &
            \includegraphics[height=0.08\textwidth]{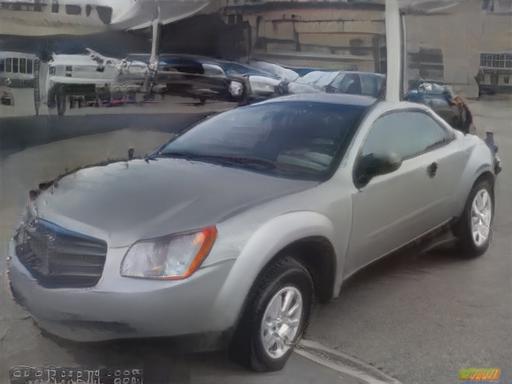} &
            \includegraphics[height=0.08\textwidth]{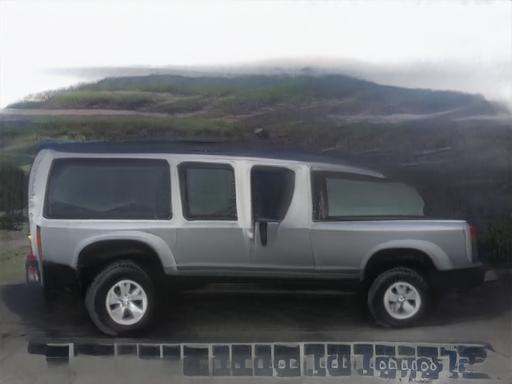} &
            \includegraphics[height=0.08\textwidth]{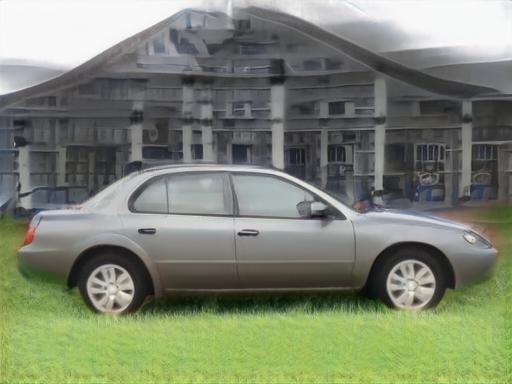} &
            \includegraphics[height=0.08\textwidth]{images/tradeoff_faces/source.jpg} &
            \includegraphics[height=0.08\textwidth]{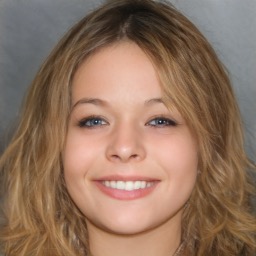} &
            \includegraphics[height=0.08\textwidth]{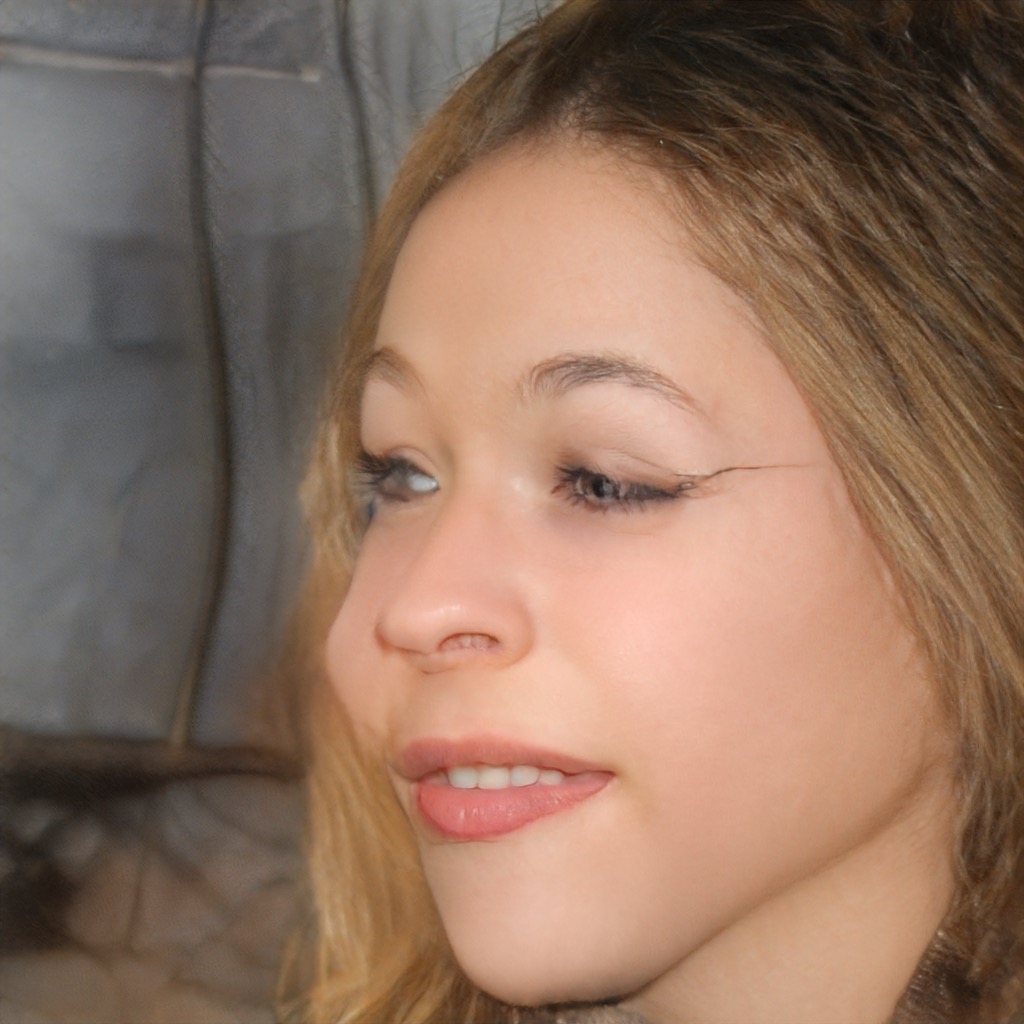} &
            \includegraphics[height=0.08\textwidth]{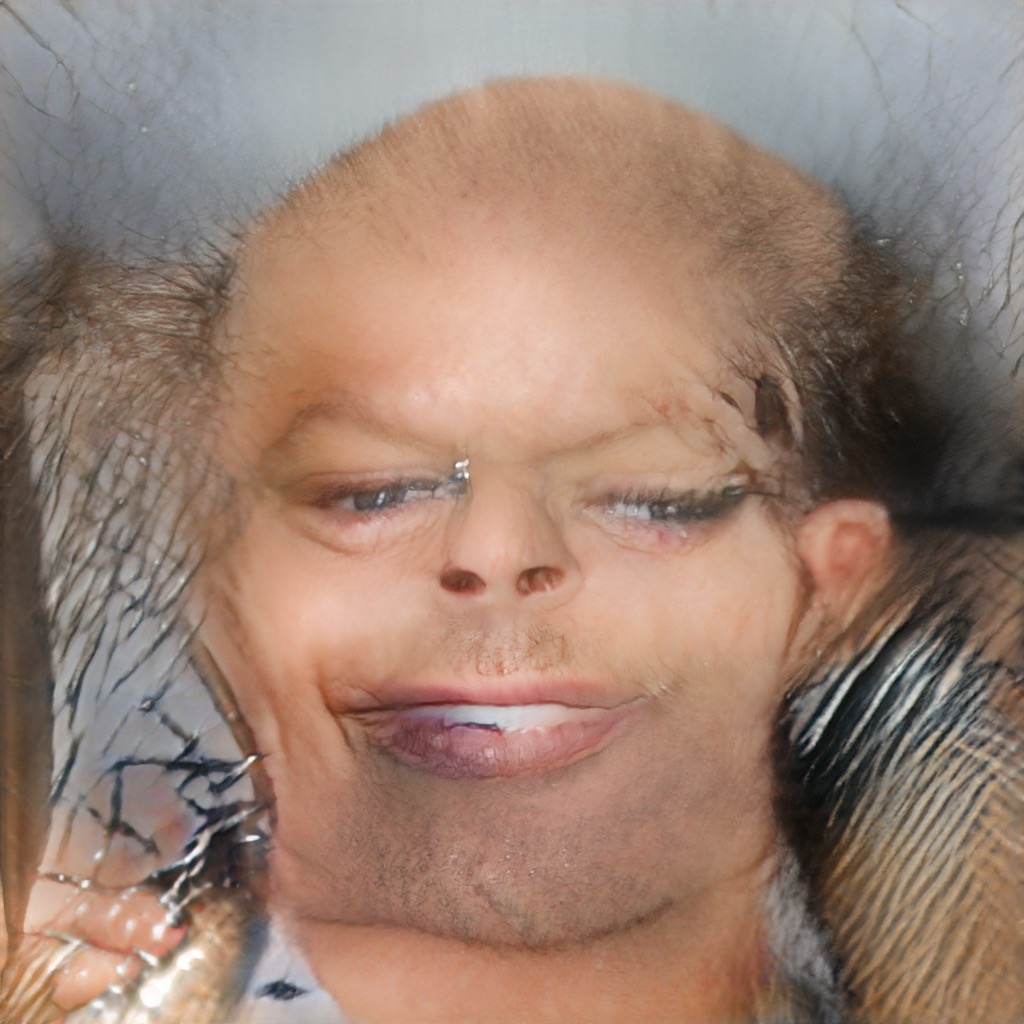} &
            \includegraphics[height=0.08\textwidth]{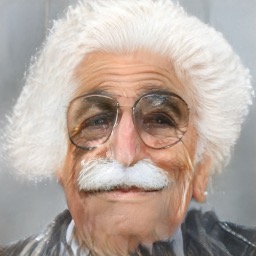}
            \\
            \raisebox{0.035\textwidth}{\texttt{C}} &
            \includegraphics[height=0.08\textwidth]{images/tradeoff_cars/source.jpg} &
            \includegraphics[height=0.08\textwidth]{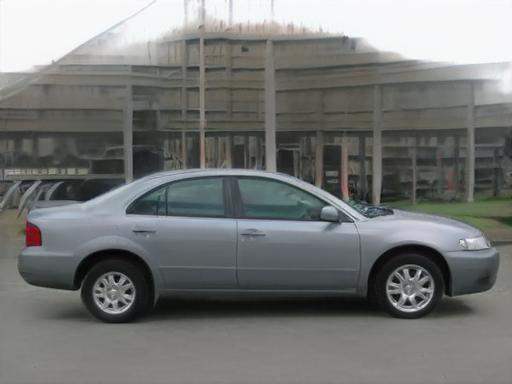} &
            \includegraphics[height=0.08\textwidth]{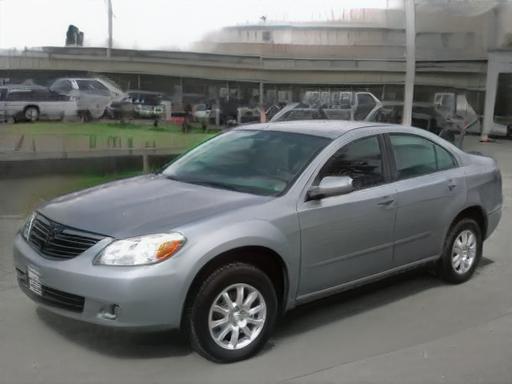} &
            \includegraphics[height=0.08\textwidth]{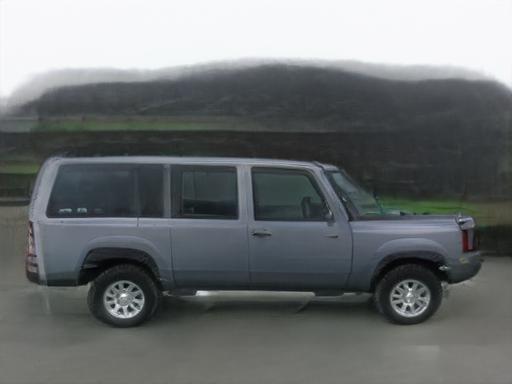} &
            \includegraphics[height=0.08\textwidth]{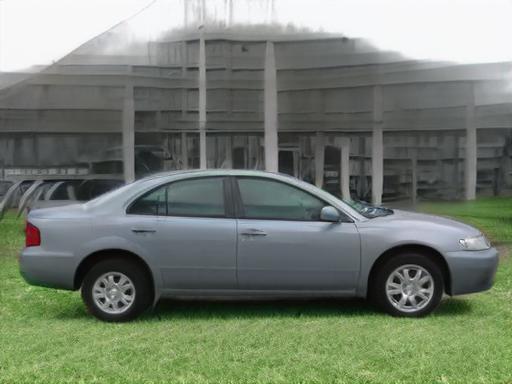} &
            \includegraphics[height=0.08\textwidth]{images/tradeoff_faces/source.jpg} &
            \includegraphics[height=0.08\textwidth]{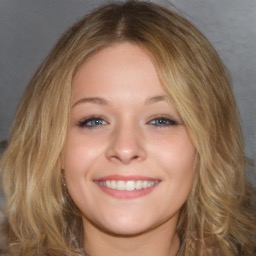} &
            \includegraphics[height=0.08\textwidth]{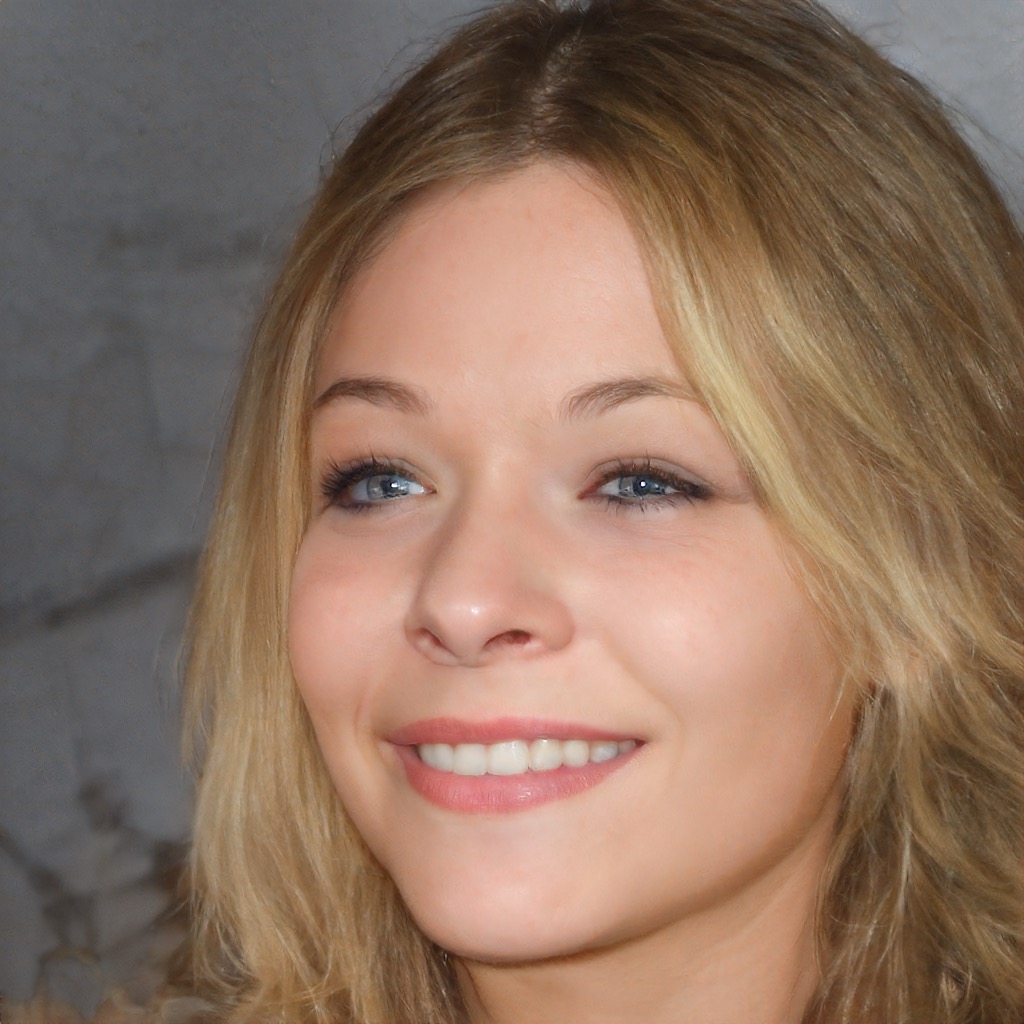} &
            \includegraphics[height=0.08\textwidth]{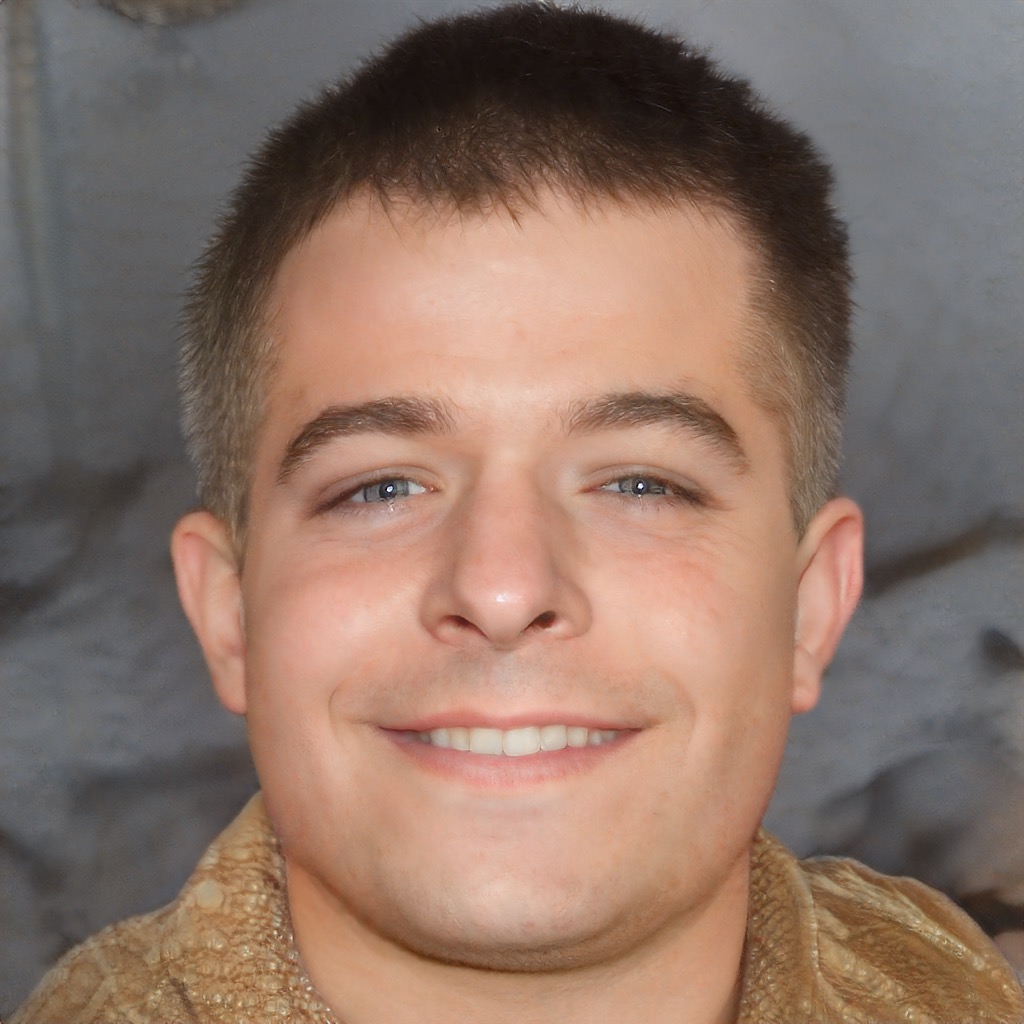} &
            \includegraphics[height=0.08\textwidth]{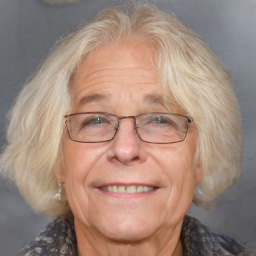}
            \\
            \raisebox{0.035\textwidth}{\texttt{D}} &
            \includegraphics[height=0.08\textwidth]{images/tradeoff_cars/source.jpg} &
            \includegraphics[height=0.08\textwidth]{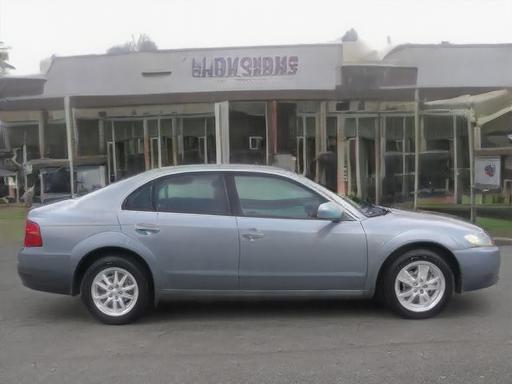} &
            \includegraphics[height=0.08\textwidth]{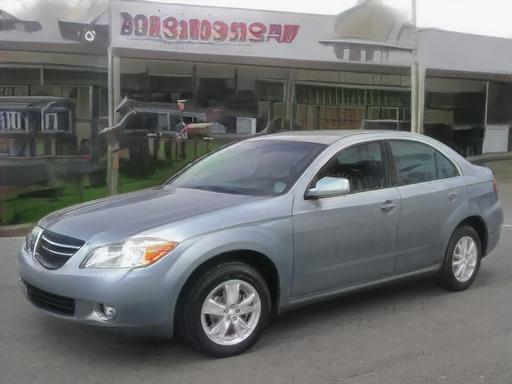} &
            \includegraphics[height=0.08\textwidth]{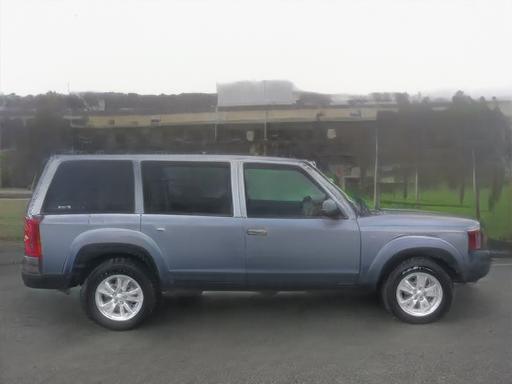} &
            \includegraphics[height=0.08\textwidth]{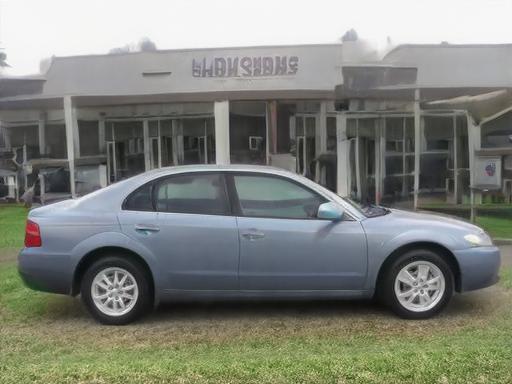} &
            \includegraphics[height=0.08\textwidth]{images/tradeoff_faces/source.jpg} &
            \includegraphics[height=0.08\textwidth]{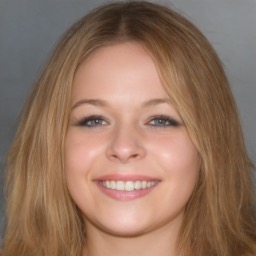} &
            \includegraphics[height=0.08\textwidth]{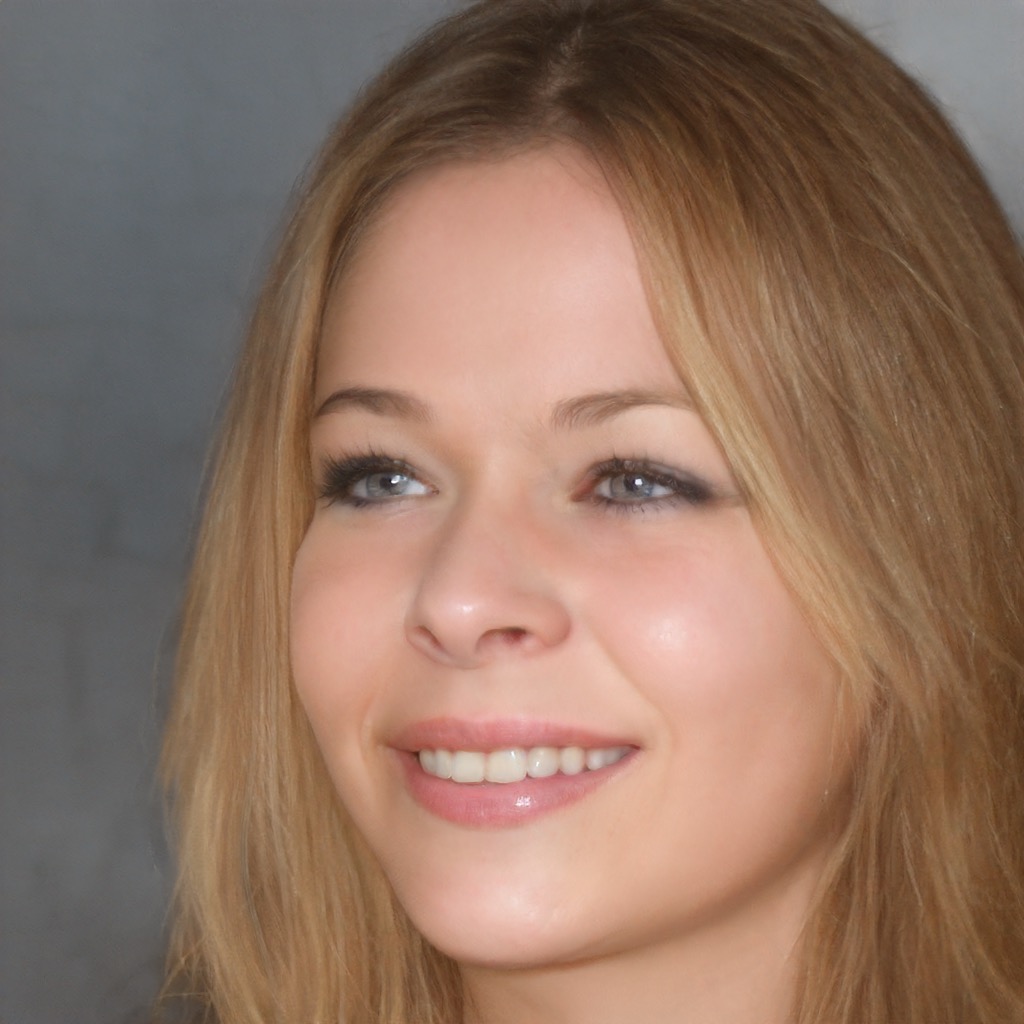} &
            \includegraphics[height=0.08\textwidth]{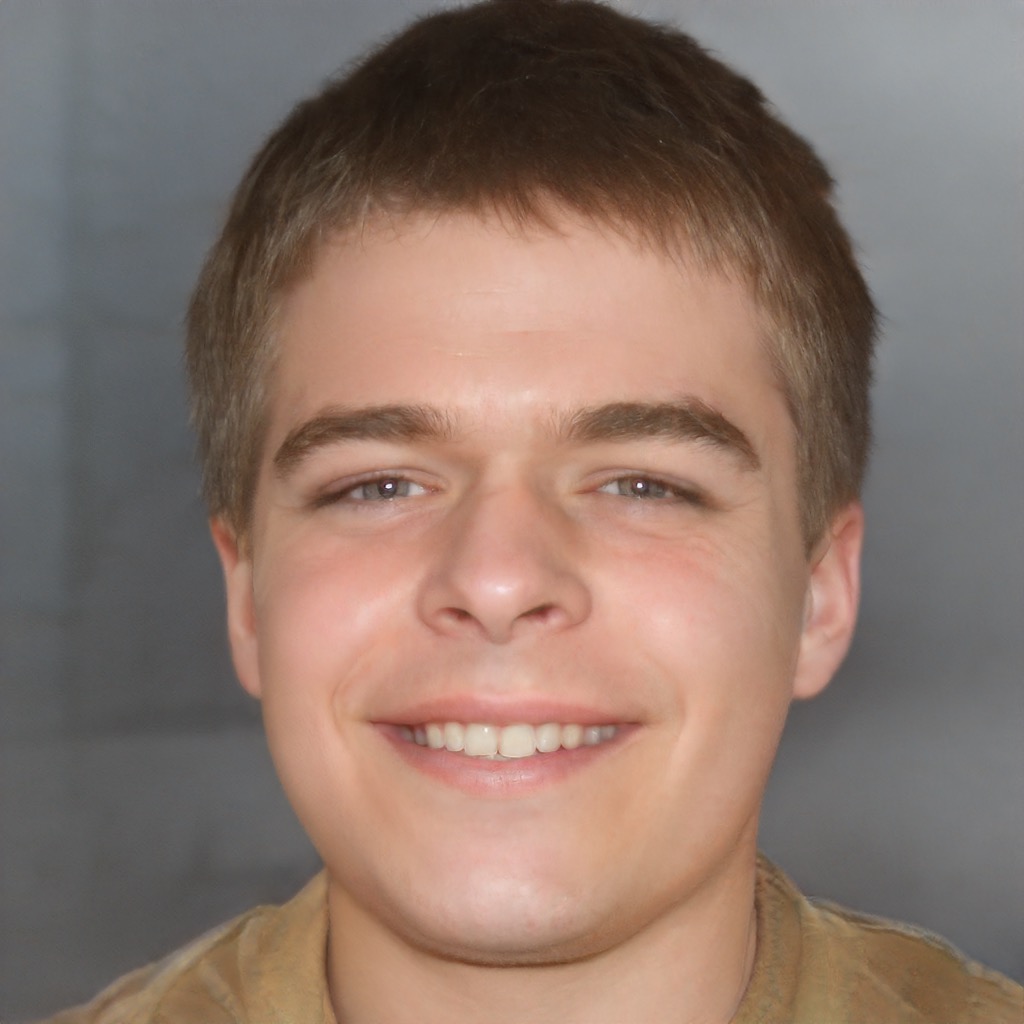} &
            \includegraphics[height=0.08\textwidth]{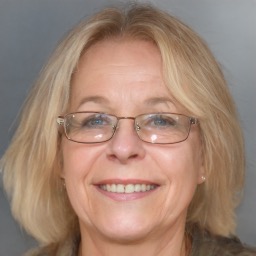}
            \\
             & Source & Inversion & Viewpoint & Cube & Grass & Source & Inversion & Pose & Gender & Age\\
            \end{tabular}
    \caption{
    Inversion followed by a series of edits for each of our configurations where configuration \texttt{D} is our complete e4e method. As can be seen, the distortion in the first row is the lowest while the perceptual quality of both the inverted and edited images in the last row is the highest.
    }
    \label{fig:abalation}
\end{figure*}

\begin{figure}
    \centering
    \includegraphics[width=0.15\linewidth]{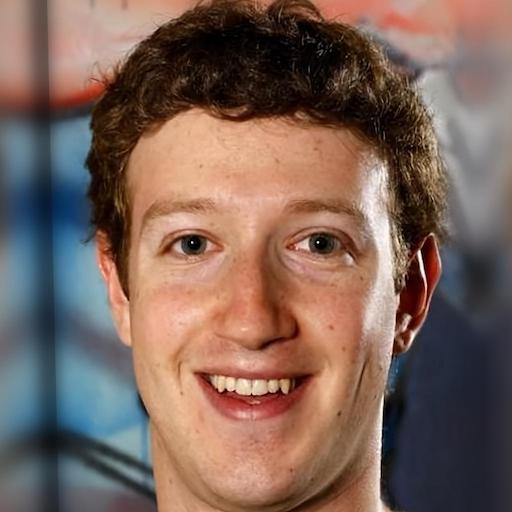}
    \includegraphics[width=0.75\linewidth]{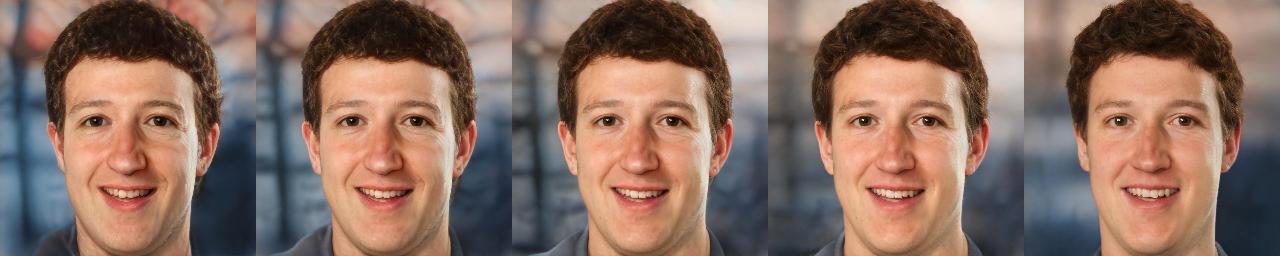} \\
    \includegraphics[width=0.15\linewidth]{images/tradeoff/zuckerberg/1721.jpg}
    \includegraphics[width=0.75\linewidth]{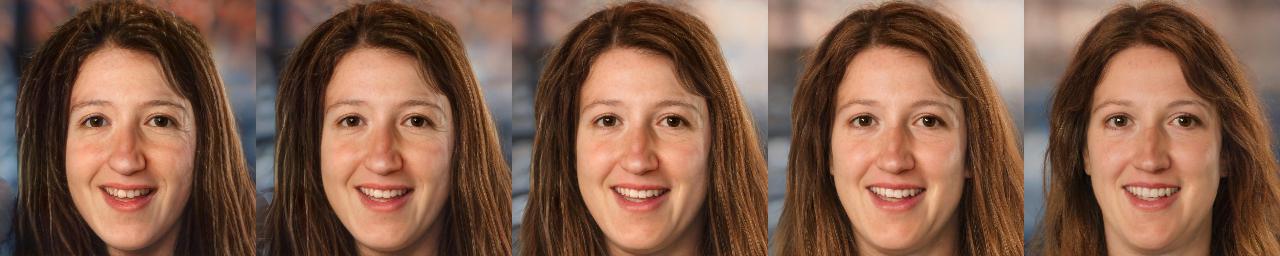}
    \vspace{-0.1cm}
    \setlength{\belowcaptionskip}{-5pt}
    \caption{The distortion-perception and distortion-editability tradeoffs. Zoom-in for details. The image on the left is the source image. In the top row, we show a series of images, where the leftmost image is the reconstruction obtained by pSp, and the rightmost is obtained by e4e. As we move to the right, the inversion approaches \w, the distortion becomes worse, and the perceptual quality becomes better. Then, for each of the inverted and interpolated images we perform a gender edit using StyleFlow. Notice that as the latent code used for editing approaches \w, the perceptual quality becomes significantly better. For example, observe the non-realistic hair in the leftmost edited image.
    }
    \label{fig:zuckerberg}
\end{figure}

\paragraph{\textbf{Controlling the tradeoff}}
The above analysis shows the existence of the distortion-editability and distortion-perception tradeoffs. We now show that it is a continuous tradeoff that can be controlled. We demonstrate this by considering inversions obtained by pSp and by our e4e method. We interpolate between their inverted latent codes and obtain additional latent codes with a monotonically increasing proximity to \w. Then, we manipulate the codes by applying the same edits using StyleFlow. In Figure \ref{fig:zuckerberg} we show an example of the above process.

\begin{figure}
\setlength{\tabcolsep}{1pt}
    \begin{tabular}{c c c c c c}
    
        \raisebox{0.035\textwidth}{\texttt{A}} &
        \includegraphics[width=0.18\linewidth]{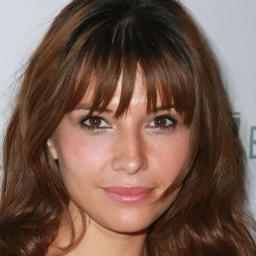} &
        \includegraphics[width=0.18\linewidth]{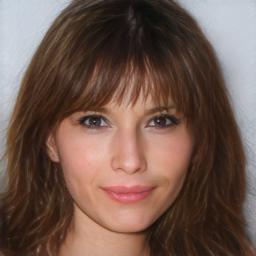} &
        \includegraphics[width=0.18\linewidth]{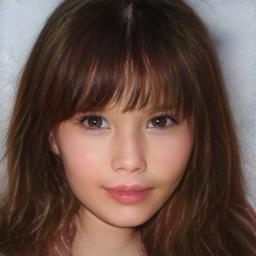} &
        \includegraphics[width=0.18\linewidth]{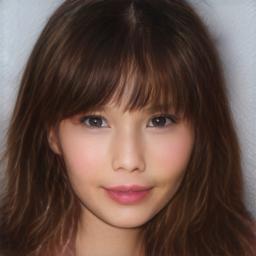} &
        \includegraphics[width=0.18\linewidth]{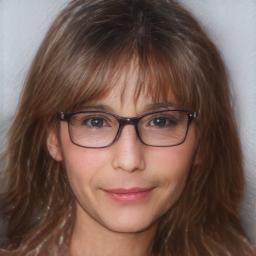} \\

        \raisebox{0.035\textwidth}{\texttt{D}} &
        \includegraphics[width=0.18\linewidth]{images/back_and_forth_edit/bnf_43_src.jpg} &
        \includegraphics[width=0.18\linewidth]{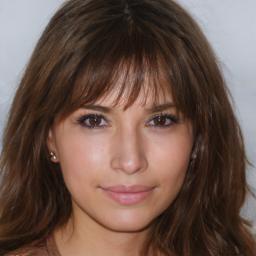} &
        \includegraphics[width=0.18\linewidth]{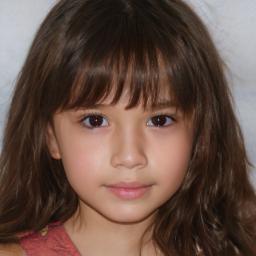} &
        \includegraphics[width=0.18\linewidth]{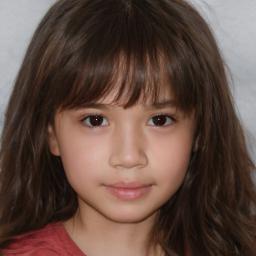} &
        \includegraphics[width=0.18\linewidth]{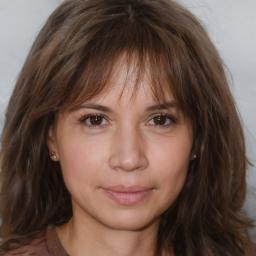} \\
        & Source & Inversion & Edit & Edit's & Inverse \\
        & & & & Inversion & Editing
    \end{tabular}
    \caption{Comparing the LEC measure of different configurations. The desired result is for the ``Inverse Editing'' image to be an exact reconstruction of the ``Inversion'' image. As can be seen, configuration \texttt{D} achieves superior results.}
    \label{fig:bnf_edit}
\end{figure}

\setlength\tabcolsep{2pt}
\begin{table}[]
    \vspace{0.25cm}
    \centering
    \begin{tabular}{|c c|c c c c c c|}
        \hline
        & & \multicolumn{6}{c|}{\textbf{Faces}} \\
        \hline
         \textbf{Conf.} & \textbf{$f_\theta$} & Young & Old & Smile & No Smile & Avg & \\
         \hline
         \texttt{A} & & 63.03 & 59.42 & 48.08 & 48.15 & 54.67 &  \\
         \texttt{D} & & \textbf{24.33} & \textbf{24.58} & \textbf{19.92} & \textbf{20.46} & \textbf{22.32} & \\
         \hline
         & & \multicolumn{6}{c|}{\textbf{Cars}} \\
         \hline
         \textbf{Conf.} & \textbf{$f_\theta$} & Pose I & Pose II & Cube & Color & Grass & Avg \\
         \hline
         \texttt{A} & & 186.95 & 181.50 & 133.93 & 83.93 & 89.78 & 135.22 \\
         \texttt{D} & & \textbf{56.28} & \textbf{56.37} & \textbf{70.46} & \textbf{28.19} & \textbf{33.06} & \textbf{48.87}\\
         \hline
    \end{tabular}
    \caption{The LEC measure computed for the facial and cars domains. For the facial domain, we perform editing using InterFaceGAN and for the cars domain we apply GANSpace. 
    The results are obtained by averaging over the entire test set.
    Note that lower is better.
    }
    \label{tab:bnf}
\end{table}

\subsection{Latent Editing Consistency}
We now turn to evaluating the e4e encoder using the newly proposed LEC method presented in Section~\ref{evaluation}. 
Qualitative and quantitative results are displayed in Figure \ref{fig:bnf_edit} and Table \ref{tab:bnf} with more results provided in the supplementary materials. The evaluation settings are described in Table~\ref{tab:bnf}. As can be observed, configuration \texttt{D} is superior to \texttt{A}. 

\vspace{1cm}
\subsection{Ablation Study}
In Section~\ref{sec:exp-tradeoff} we demonstrated that configuration \texttt{D} encodes real images into more editable regions comparable to configuration \texttt{A}.
Here, we perform an ablation study of the configurations described in Table~\ref{table:architecture_configs}. More specifically, we aim at understanding the importance of each of the principles described in Section~\ref{sec:encoder} for improving editability. In Figure~\ref{fig:abalation}, we provide an inversion and a series of edits for each configuration.

In configuration \texttt{A}, we see a low distortion and low perceptual quality in both the inverted and edited images. For example, observe the hair of the woman in the inverted image, and the shape of the car when editing the viewpoint.
In configuration \texttt{B}, which uses the delta-regularization, but no latent discriminator, we observe a slight improvement in image sharpness and overall realism of the inverted image compared to \texttt{A}. For example, observe the more realistic hair of the woman. On the other hand, the resulting latent code is highly un-editable (e.g., see the cube and age edits).
By using the latent discriminator, but no delta-regularization in configuration \texttt{C}, we observe a significant improvement in perceptual quality of both the inverted and edited images. For example, observe the realistic nature of the grass compared to the grass generated in \texttt{A}.
As one can see, by combining the delta-regularization with the latent discriminator in configuration \texttt{D}, we get the best perceptual quality in both the inverted and edited images. For example, observe the realistic head shape when editing the woman's gender compared to \texttt{C}.

\begin{table}
    \centering
    \begin{tabular}{|c c|c c | c c | c c|}
        \hline
         \multicolumn{2}{|c|}{}  & \multicolumn{2}{c|}{\textbf{Distortion}} &
         \multicolumn{2}{c|}{\textbf{Perception}} &
         \multicolumn{2}{c|}{\textbf{Editability}} \\
         \multicolumn{1}{|c}{\textbf{Domain}}& \textbf{Conf.} & $L_2$ & LPIPS & FID & SWD & FID & SWD  \\
         \hline
         \multirow{2}{*}{Faces} & \texttt{A} & \textbf{0.03} & \textbf{0.17} & \textbf{25.17} & 48.72 & \textbf{62.46} & 48.75 \\
         & \texttt{B} & 0.04 & 0.18 & 28.09 & 39.98 & 149.85 & 69.86 \\
         & \texttt{C} & 0.04 & 0.19 & 27.36 & \textbf{33.96} & 143.03 & \textbf{41.94} \\
         & \texttt{D} & 0.05 & 0.23 & 30.96 & 40.54 & 81.08 & 43.63 \\
         \hline
         \multirow{2}{*}{Cars} & \texttt{A} & \textbf{0.10} & 0.32 & \textbf{10.56} & \textbf{22.08} & \textbf{12.92} & \textbf{24.30} \\
         & \texttt{B} & 0.11 & 0.32 & 11.16 & 24.06 & 17.85 & 29.78 \\
         & \texttt{C} & 0.11 & 0.33 & 12.47 & 38.08 & 16.22 & 44.73 \\
         & \texttt{D} &\textbf{ 0.10} & 0.32 & 12.18 & 22.71 & 15.44 & 31.45 \\
        \hline
    \end{tabular}
    \caption{Quantitative comparison of the e4e configurations.}
    \label{tab:ablation}
\end{table}

\subsection{Comparison to Other Methods}
We now compare e4e with other state-of-the-art StyleGAN inversion methods. Specifically, we compare our e4e approach with both optimization-based and encoder-based inversion methodologies.

\begin{figure}
\setlength{\tabcolsep}{1pt}
    \begin{tabular}{c c c c c}
        \raisebox{0.26in}{\rotatebox[origin=t]{90}{Optimization}} & 
        \includegraphics[width=0.21\linewidth]{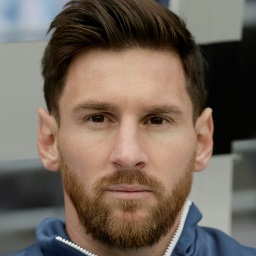} &
        \includegraphics[width=0.21\linewidth]{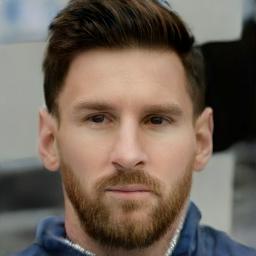} &
        \includegraphics[width=0.21\linewidth]{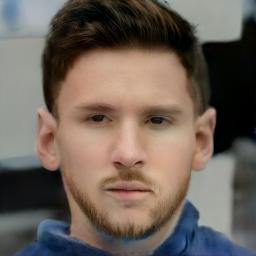} & 
        \includegraphics[width=0.21\linewidth]{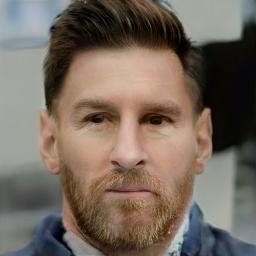}  \\
        \raisebox{0.26in}{\rotatebox[origin=t]{90}{e4e}} & 
        \includegraphics[width=0.21\linewidth]{images/optimization/messi/messi5_src.jpg} & 
        \includegraphics[width=0.21\linewidth]{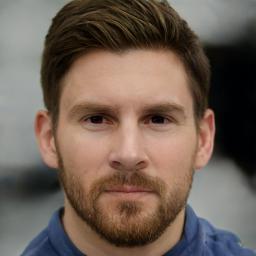} &
        \includegraphics[width=0.21\linewidth]{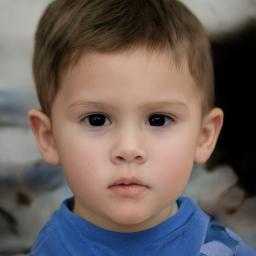} &
        \includegraphics[width=0.21\linewidth]{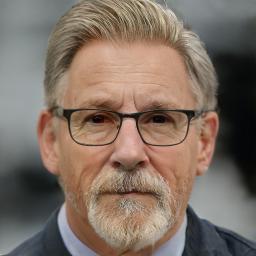} \\ 
        
        & Source & Inversion & Young & Old
    \end{tabular}
    \caption{
    Embedding the source image into \wkstar using StyleGAN2 optimization yields low distortion and a near perfect reconstruction. However, the obtained latent code behaves poorly when performing latent space editing (here age). On the other hand, latent codes obtained by our method are more suitable for editing, at the cost of a higher distortion.}
    \label{fig:optimization_editing}
\end{figure}

\begin{figure}
\setlength{\tabcolsep}{1pt}
    \begin{tabular}{c c c c}
        \includegraphics[width=0.21\linewidth]{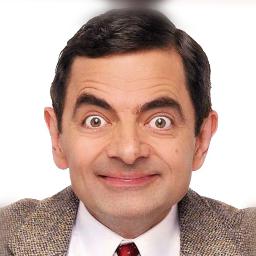} &
        \includegraphics[width=0.21\linewidth]{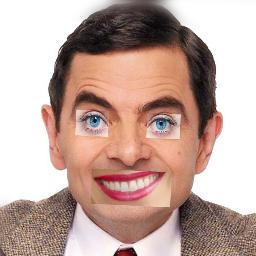} &
        \includegraphics[width=0.21\linewidth]{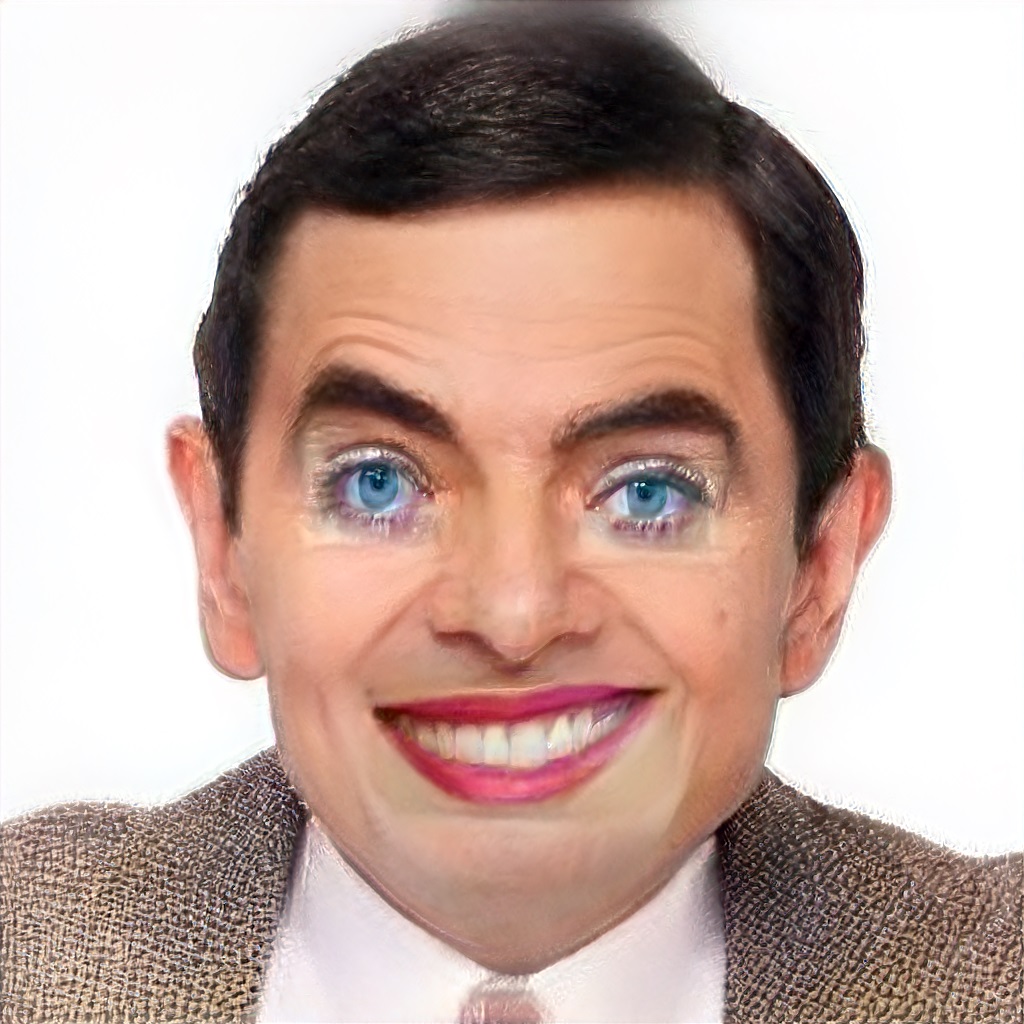} & 
        \includegraphics[width=0.21\linewidth]{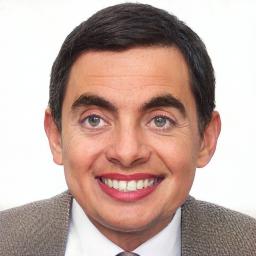} \\ 
        
        Source & Patch-Edited & Optimization & e4e
    \end{tabular}
    \caption{
    The applicability of our encoder-based approach for patch-editing. Embedding the patch-edited image into \wkstar using StyleGAN2 optimization yields a perceptually poor reconstruction, but of low distortion. Although it has a higher distortion, e4e's inversion is more realistic.}
    \label{fig:optimization_patch_editing}
\end{figure}

\paragraph{\textbf{Optimization-based methods}}
Although there has recently been significant progress in encoder-based inversion, most methods still resort to using a per-image optimization. There are several points that should be considered when using direct optimization: (i) it is computationally intensive; (ii) it tends to favor low distortion over editability (see Figure~\ref{fig:optimization_editing}); and (iii) it tends to over-fit on the input image, which is undesirable in many applications (see Figure~\ref{fig:optimization_patch_editing}).

\paragraph{\textbf{Encoder-based methods}}
We now compare our method with the current state-of-the-art encoder-based inversion methods: IDInvert~\cite{zhu2020domain} and pSp~\cite{richardson2020encoding}. We focus on showing that previous methods tend to prefer low distortion over high editability. In contrast, we show that e4e achieves a slightly higher distortion, but much better editability. We evaluate the three methods on the facial and cars domains. 

For the human facial domain, we use the official pre-trained inversion models for both IDInvert and pSp. For performing editing on the facial domain, we use editing directions from InterFaceGAN~\cite{shen2020interpreting}. For IDInvert we use their official editing directions. As InterFaceGAN is originally trained using StyleGAN1, we retrain their method on a StyleGAN2 generator and follow the same procedure as described in their paper to obtain the editing directions. These are then used for editing the latents obtained by pSp and e4e.

For the cars domain, we retrain both IDInvert and pSp with an input resolution of $512\times384$ using a StyleGAN2 generator. Note that while training IDInvert took over three weeks on two NVIDIA P40 GPUs, our e4e encoder was trained for only three days using a single P40 GPU. For performing editing on the inversions, we use editing directions obtained by GANSpace~\cite{harkonen2020ganspace}.

In Figure~\ref{fig:comparison}, we provide a qualitative comparison of the three methods. As can be seen, when editing the inverted images with e4e, the perceptual quality is significantly higher. For example, for IDInvert observe the grass in the first example (column $4$) and the unrealistic colors in the second example (column $3$). Also, observe the low quality faces generated when applying the age edit. 
For pSp, observe the car's warped shape when applying the pose edit (column $2$) and the unrealistic hair obtained on the edited images.

In Table~\ref{tab:comparison_to_other_methods} we provide a quantitative evaluation. For distortion we show both $L_2$ and LPIPS similarities. As can be seen, pSp achieves the lowest distortion of the three methods. For the completeness of the evaluation, we additionally show the FID and SWD metrics. However, as can be seen they contradict each other, again. 

\setlength\tabcolsep{2.0pt}
\begin{table}
    \centering
    \begin{tabular}{|c c|c c | c c | c c |}
        \hline
         \multicolumn{2}{|c|}{}  & \multicolumn{2}{c|}{\textbf{Distortion}} &
         \multicolumn{2}{c|}{\textbf{Perception}} &
         \multicolumn{2}{c|}{\textbf{Editability}} \\
         \multicolumn{1}{|c}{\textbf{Domain}}& \textbf{Method} & $L_2$ & LPIPS & FID & SWD & FID & SWD  \\
         \hline
        \multirow{2}{*}{Faces} & pSp & \textbf{0.03} & \textbf{0.17} & 29.57 & \textbf{21.35} & \textbf{73.70} & \textbf{32.30} \\
         & IDInvert & 0.06 & 0.22 & \textbf{20.10} & 31.31 & 129.00 & 36.97 \\
         & e4e & 0.05 & 0.23 & 35.47 & 31.24 & 96.93 & 36.02 \\
         \hline
         \multirow{2}{*}{Cars} & pSp & \textbf{0.10} & \textbf{0.30} & 17.63 & 25.95 & 21.00 & 24.46 \\
         & IDInvert & 0.13 & 0.31 & \textbf{8.26} & 24.79 & 18.18 & \textbf{18.94} \\
         & e4e & \textbf{0.10} & 0.32 & 12.18 & \textbf{22.71} & \textbf{15.44} & 26.83 \\
        \hline
    \end{tabular}
    \vspace{-0.2cm}
    \caption{Quantitative comparison to IDInvert and pSp. As can be seen, our inversion usually has greater distortion, which is expected. However, the perceptual metrics contradict each other, resulting in no clear winner. As discussed, perceptual quality is best evaluated using human judgement.}
    \label{tab:comparison_to_other_methods}
\end{table}

\begin{figure}
        \setlength{\tabcolsep}{1pt}
        \centering
        \begin{tabular}{c c c c c c}
            \includegraphics[width=0.18\linewidth]{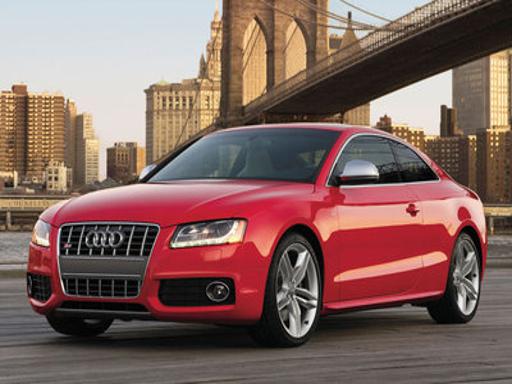} &
            \raisebox{0.15in}{\rotatebox[origin=t]{90}{IDInvert}} & 
            \includegraphics[width=0.18\linewidth]{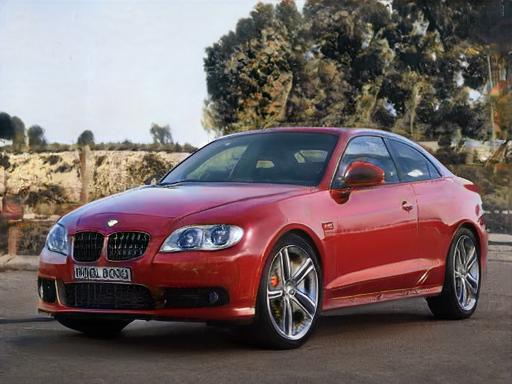} &
            \includegraphics[width=0.18\linewidth]{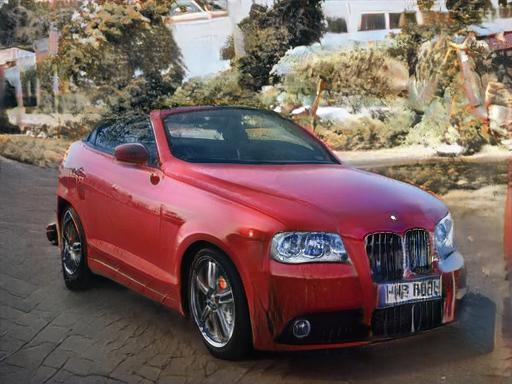} &
            \includegraphics[width=0.18\linewidth]{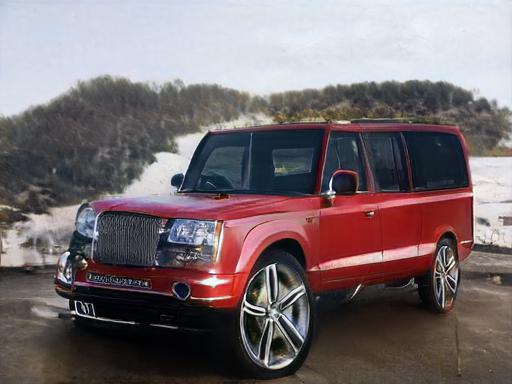} &
            \includegraphics[width=0.18\linewidth]{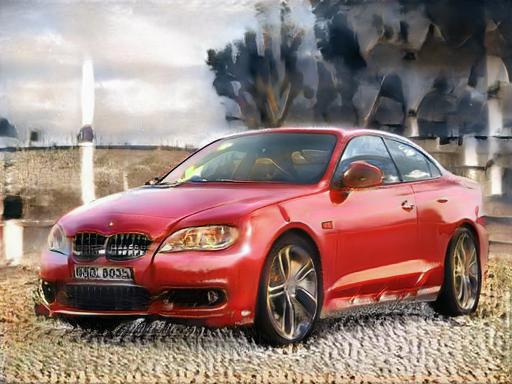} \\
            & \raisebox{0.15in}{\rotatebox[origin=t]{90}{pSp}} & 
            \includegraphics[width=0.18\linewidth]{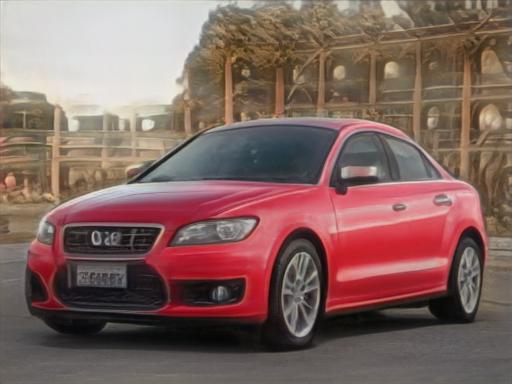} &
            \includegraphics[width=0.18\linewidth]{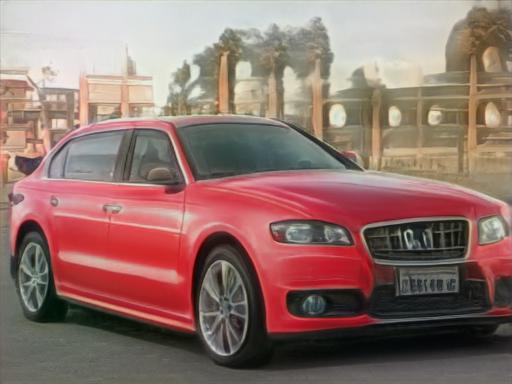} &
            \includegraphics[width=0.18\linewidth]{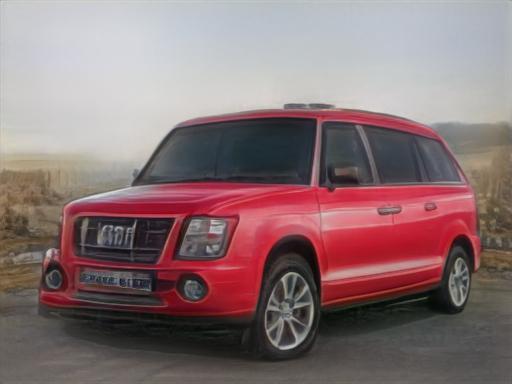} &
            \includegraphics[width=0.18\linewidth]{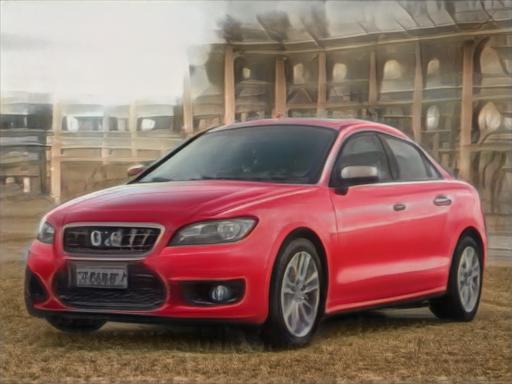} \\
            & \raisebox{0.15in}{\rotatebox[origin=t]{90}{Ours}} & 
            \includegraphics[width=0.18\linewidth]{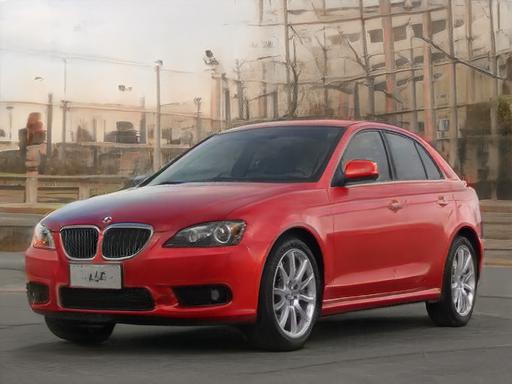} &
            \includegraphics[width=0.18\linewidth]{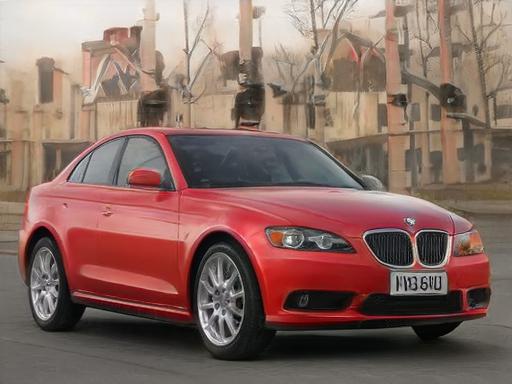} &
            \includegraphics[width=0.18\linewidth]{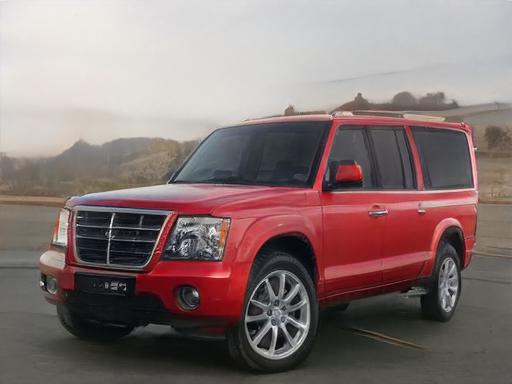} &
            \includegraphics[width=0.18\linewidth]{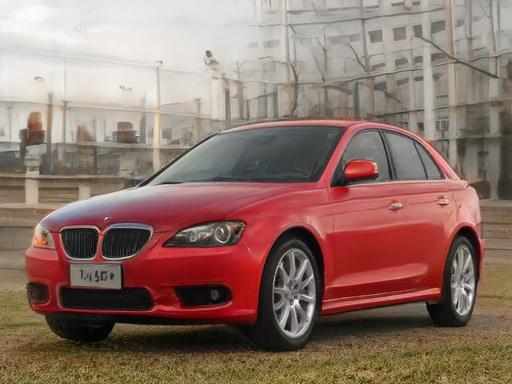} \\

            \includegraphics[width=0.18\linewidth]{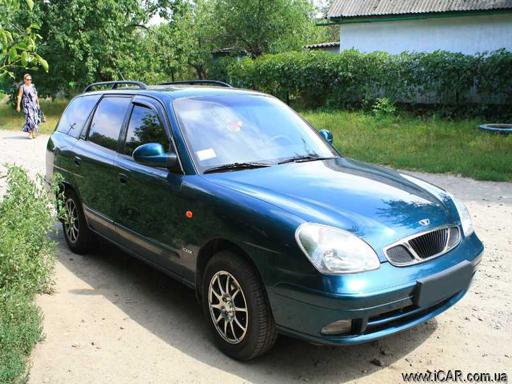} &
            \raisebox{0.15in}{\rotatebox[origin=t]{90}{IDInvert}} & 
            \includegraphics[width=0.18\linewidth]{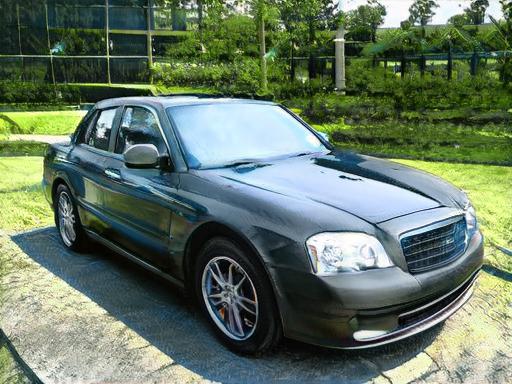} &
            \includegraphics[width=0.18\linewidth]{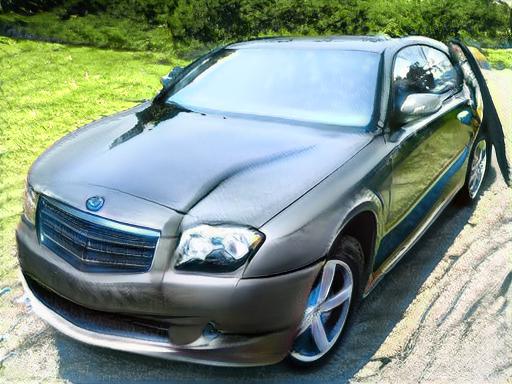} &
            \includegraphics[width=0.18\linewidth]{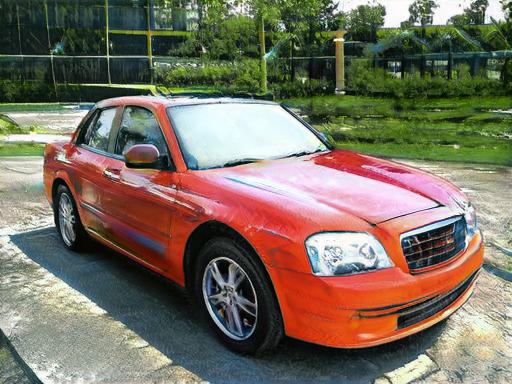} &
            \includegraphics[width=0.18\linewidth]{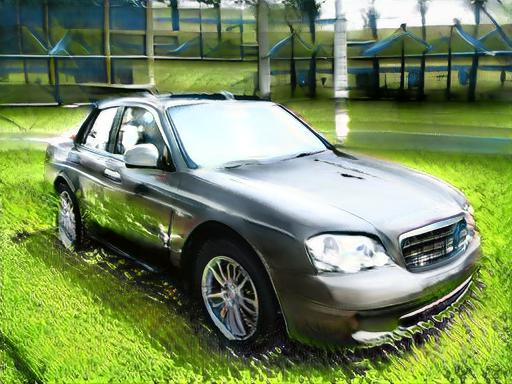} \\
            & \raisebox{0.15in}{\rotatebox[origin=t]{90}{pSp}} & 
            \includegraphics[width=0.18\linewidth]{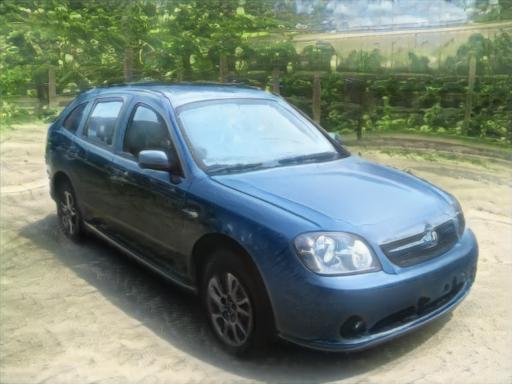} &
            \includegraphics[width=0.18\linewidth]{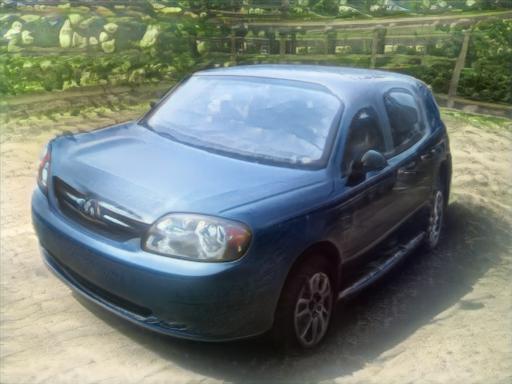} &
            \includegraphics[width=0.18\linewidth]{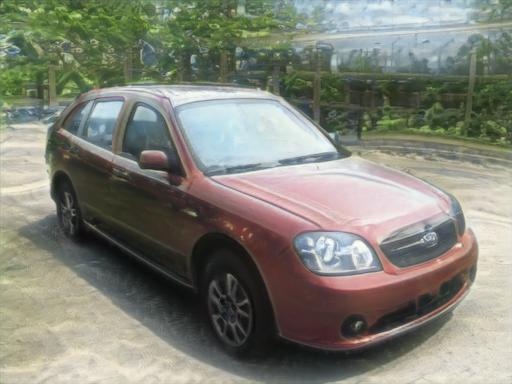} &
            \includegraphics[width=0.18\linewidth]{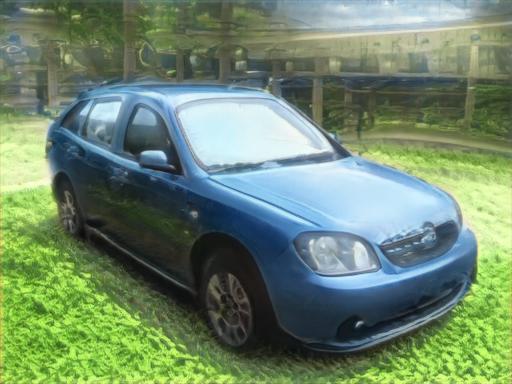} \\
            & \raisebox{0.15in}{\rotatebox[origin=t]{90}{Ours}} & 
            \includegraphics[width=0.18\linewidth]{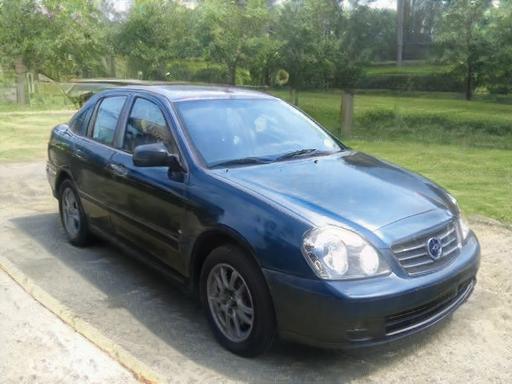} &
            \includegraphics[width=0.18\linewidth]{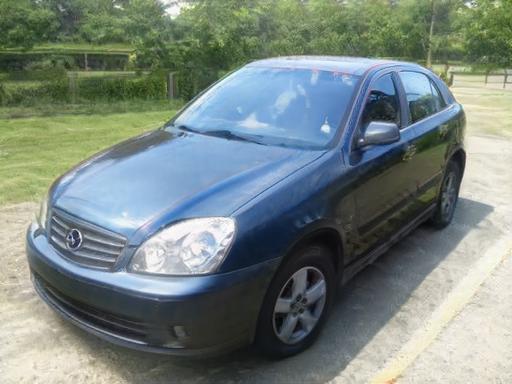} &
            \includegraphics[width=0.18\linewidth]{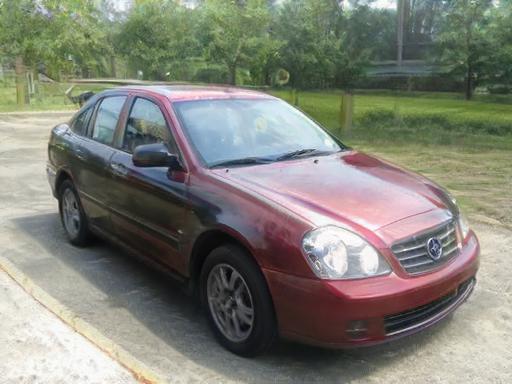} &
            \includegraphics[width=0.18\linewidth]{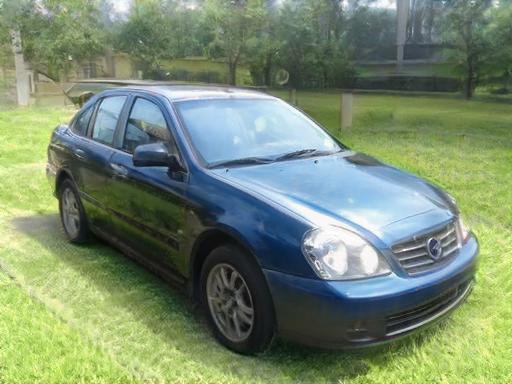} \\
            
            \includegraphics[width=0.18\linewidth]{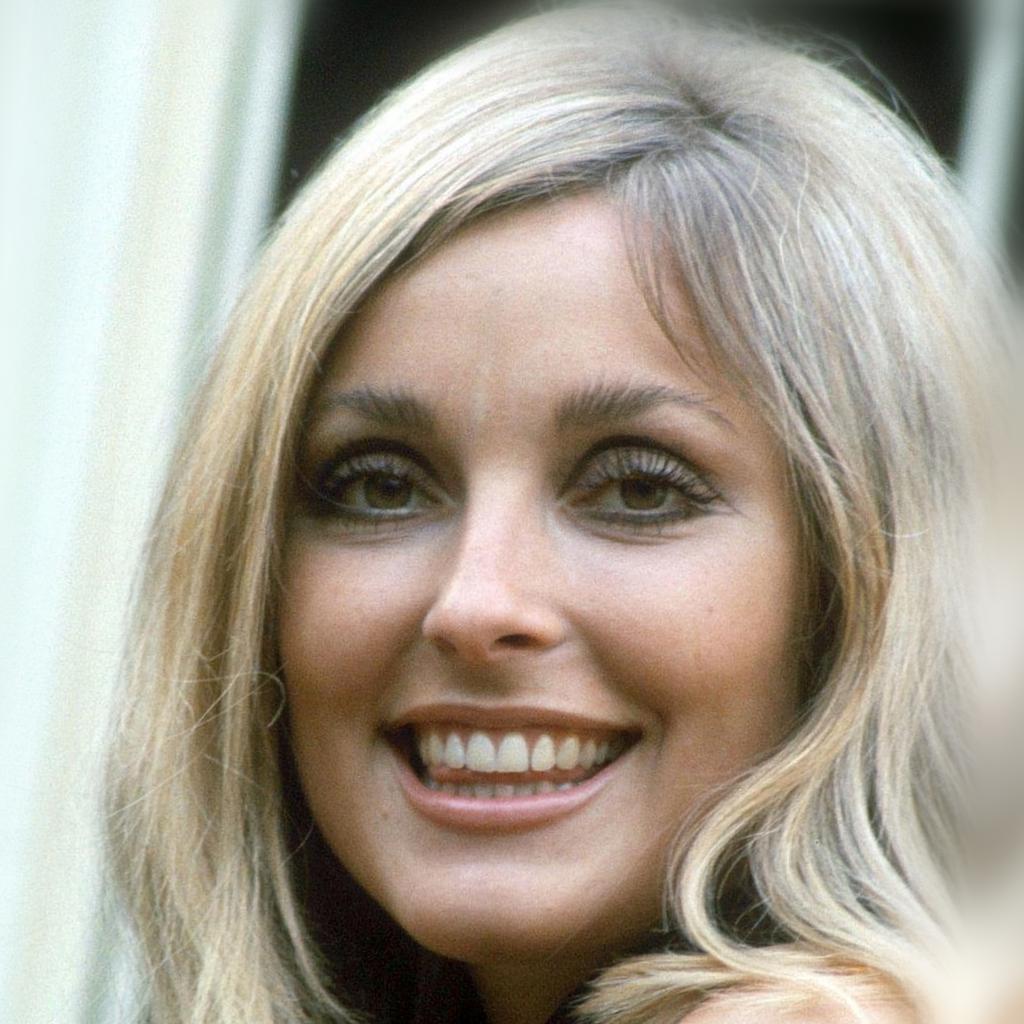} &
            \raisebox{0.22in}{\rotatebox[origin=t]{90}{IDInvert}} & 
            \includegraphics[width=0.18\linewidth]{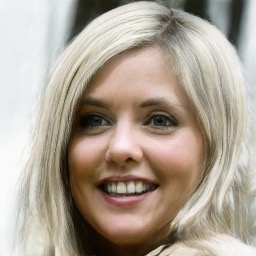} &
            \includegraphics[width=0.18\linewidth]{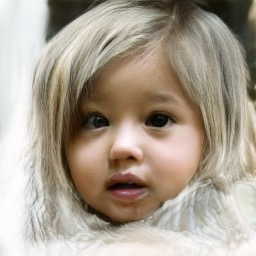} &
            \includegraphics[width=0.18\linewidth]{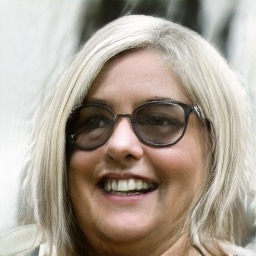} &
            \includegraphics[width=0.18\linewidth]{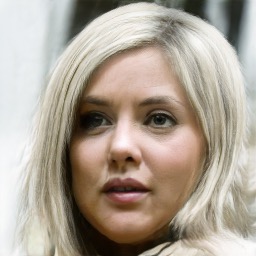} \\
            & \raisebox{0.22in}{\rotatebox[origin=t]{90}{pSp}} & 
            \includegraphics[width=0.18\linewidth]{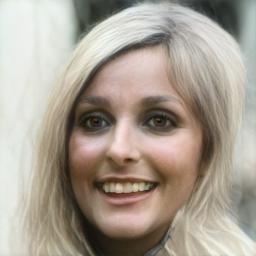} &
            \includegraphics[width=0.18\linewidth]{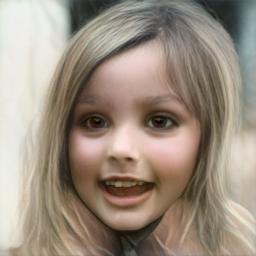} &
            \includegraphics[width=0.18\linewidth]{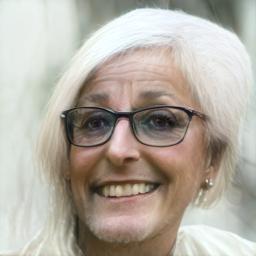} &
            \includegraphics[width=0.18\linewidth]{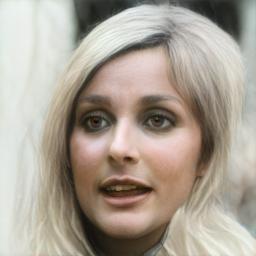}  \\ 
            & \raisebox{0.22in}{\rotatebox[origin=t]{90}{Ours}} & 
            \includegraphics[width=0.18\linewidth]{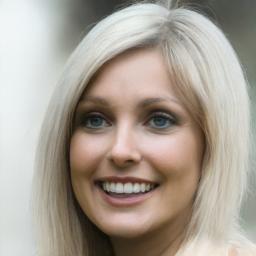} &
            \includegraphics[width=0.18\linewidth]{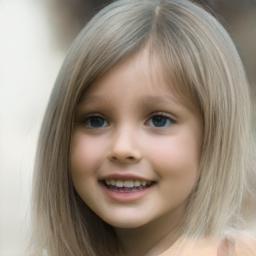} &
            \includegraphics[width=0.18\linewidth]{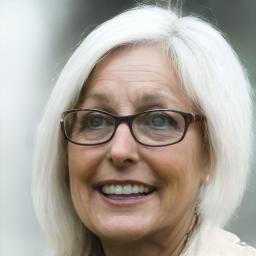} &
            \includegraphics[width=0.18\linewidth]{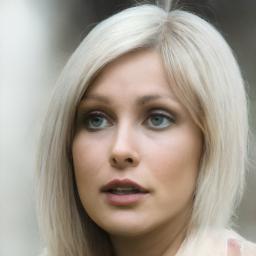} \\
        
            \includegraphics[width=0.18\linewidth]{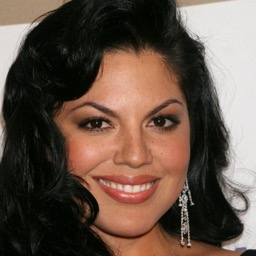} &
            \raisebox{0.22in}{\rotatebox[origin=t]{90}{IDInvert}} & 
            \includegraphics[width=0.18\linewidth]{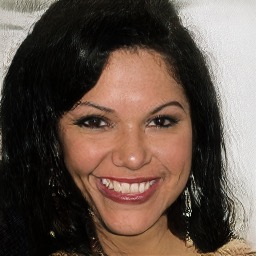} &
            \includegraphics[width=0.18\linewidth]{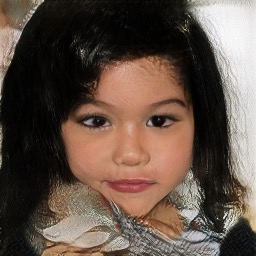} &
            \includegraphics[width=0.18\linewidth]{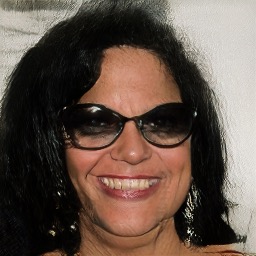} &
            \includegraphics[width=0.18\linewidth]{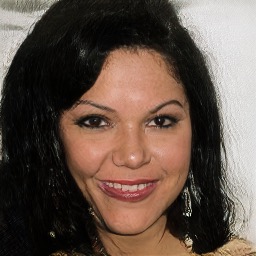} \\
            & \raisebox{0.22in}{\rotatebox[origin=t]{90}{pSp}} & 
            \includegraphics[width=0.18\linewidth]{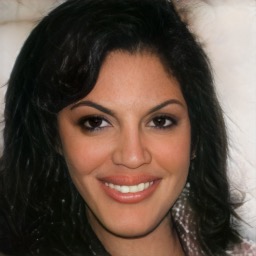} &
            \includegraphics[width=0.18\linewidth]{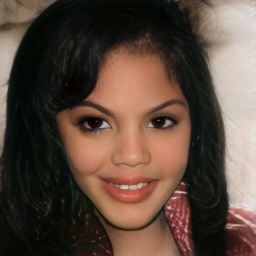} &
            \includegraphics[width=0.18\linewidth]{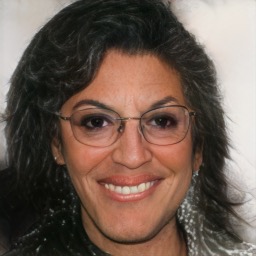} &
            \includegraphics[width=0.18\linewidth]{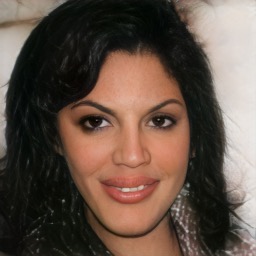} \\ 
            & \raisebox{0.22in}{\rotatebox[origin=t]{90}{Ours}} & 
            \includegraphics[width=0.18\linewidth]{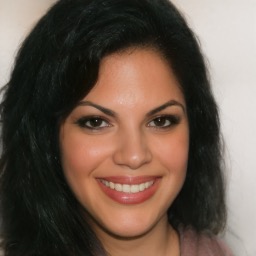} &
            \includegraphics[width=0.18\linewidth]{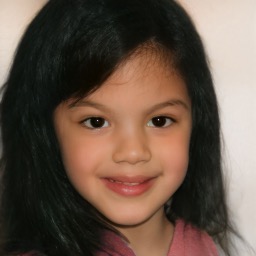} &
            \includegraphics[width=0.18\linewidth]{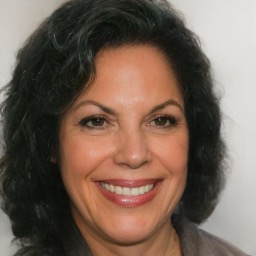} &
            \includegraphics[width=0.18\linewidth]{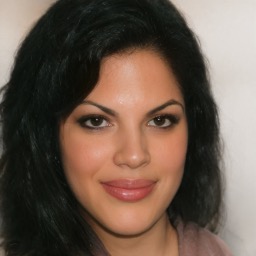} \\

            Source & & Inversion & \multicolumn{3}{c}{\ruleline{0.52\linewidth}{Edits}} \\
        \end{tabular}    
            
    \caption{Qualitative comparison to IDInvert and pSp. As can be seen, our inversion results have subtly greater distortion, but the subsequent editing is significantly more realistic and visually-appealing.}
    \label{fig:comparison}
\end{figure}
\section{Discussion and Conclusions}

The latent space of StyleGAN and its extensions are powerful, in the sense that they are semantically disentangled, and thus allow for various meaningful edits. However, they are complex spaces. Their quality is not fairly distributed across the whole space, where some regions are more well-behaved than others. Furthermore, the \wkstar space is huge --- much larger than the space of all natural images --- and thus, many images have more than a single latent representation. In our work, we make the first steps in analyzing the structure of this complex space, aiming to characterize well-behaved regions to guide the inversion of images into these regions.

\clearpage
\newpage
Our main contribution is twofold: (i) we propose means for encouraging the encoding of a real image to be mapped into well-behaved regions of \wkstar, and (ii) we design an encoder and demonstrate its performance, in light of the tradeoff between distortion and editability. We also discuss the difficulties in evaluating reconstruction and editability and propose evaluation protocols built upon commonly used measures. In a sense, our presented method complements image manipulation methods by facilitating higher editing quality on \textit{real} images.

Generally speaking, our encoder encourages mapping close to \w which works well since the space around \w is still surprisingly highly expressive. Moreover, this principle can be adopted for problems beyond image inversion. For example, it can be applied to map latent vectors that represent more than one image, or say a combination of two, like the disentangled representation of identity and pose \cite{nitzan2020face}, or a blend of two images, to a proper latent code of the target image which is likely to exist in proximity of \w. We are planning to explore this research direction. 

Our inversion scheme is generic, and we have demonstrated its performance on five challenging and diverse domains. Note, however, that some domains are harder than others. Human faces are well-structured, simplifying the training of the encoder. The domain of horses, for example, is much more complex as it is unstructured and it has many modes. Hence, training an encoder for such a domain is much more challenging. In the future, we would like to consider multi-modal generators, like that of Sendik \etal \cite{sendik2020unsupervised}, and develop an encoder into multi-modal latent spaces.

Finally, here we consider the inversion into a given latent space. In the future, it would be interesting and challenging to consider fine-tuning the generator and train both the encoder and decoder for their common target objective for specific downstream tasks. 

{\small
\bibliographystyle{ieee}
\bibliography{egbib}
}

\newpage
\appendix
\appendixpage
In this supplemental document we provide implementation details and further experiments. Zooming-in to better observe fine details is recommended in all figures.

\section{Implementation Details}~\label{implementation_details}
Here, we provide additional implementation details to complement the architecture, training scheme, and objective functions as described in the main text. 

\paragraph{\textbf{Losses}} 
We set the loss weights as follows: we set $\lambda_{l2} = 1$ and $\lambda_{lpips} = 0.8$. When training with the latent discriminator $D_w$, we use $\lambda_{adv}=0.1$. When applying a progressive delta-based training, we use $\lambda_{d-reg}=2e^{-4}$. For the cars, horses, and cats domains we set $\lambda_{sim}=0.5$ for our MOCO-based similarity loss. For the facial domain we set $\lambda_{sim}=0.1$ over a pre-trained ArcFace~\cite{deng2019arcface} facial recognition network. Finally, we set $\lambda_{edit}=1$.

\paragraph{\textbf{Progressive Training}} Only the first $w$ style vector is trained in the first $20,000$ training steps. After the first $20,000$ steps, we gradually add a delta for the next latent code entry every $2,000$ training steps.

\paragraph{\textbf{Discriminator}} For our latent discriminator, we use a $4$-layer MLP network using $0.2$ LeakyReLU activations. We train the discriminator using the Adam optimizer with a fixed learning rate of $2e^{-5}$. 

\section{Comparing Configurations \texttt{A} and \texttt{D}}
\begin{figure*}
    \setlength{\tabcolsep}{1pt}
    \centering
        \centering
            \begin{tabular}{c c c c c c c}
            & Source & Inversion & \multicolumn{4}{c}{\ruleline{0.52\linewidth}{Edits}} \\
            \raisebox{0.06\textwidth}{\texttt{A}} &\includegraphics[width=0.135\textwidth]{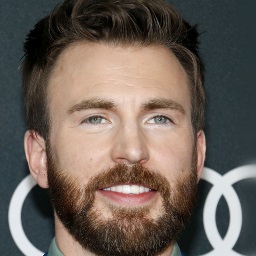} & 
            \includegraphics[width=0.135\textwidth]{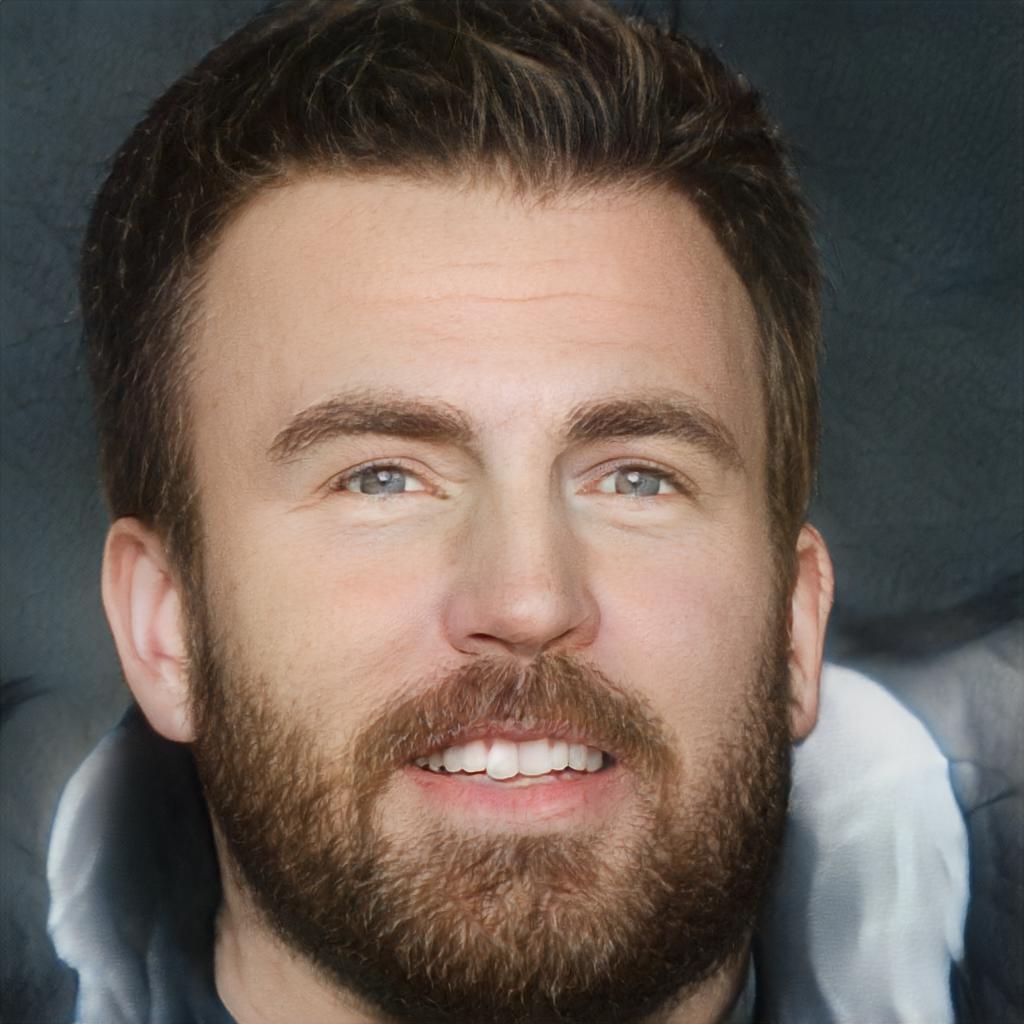} &
            \includegraphics[width=0.135\textwidth]{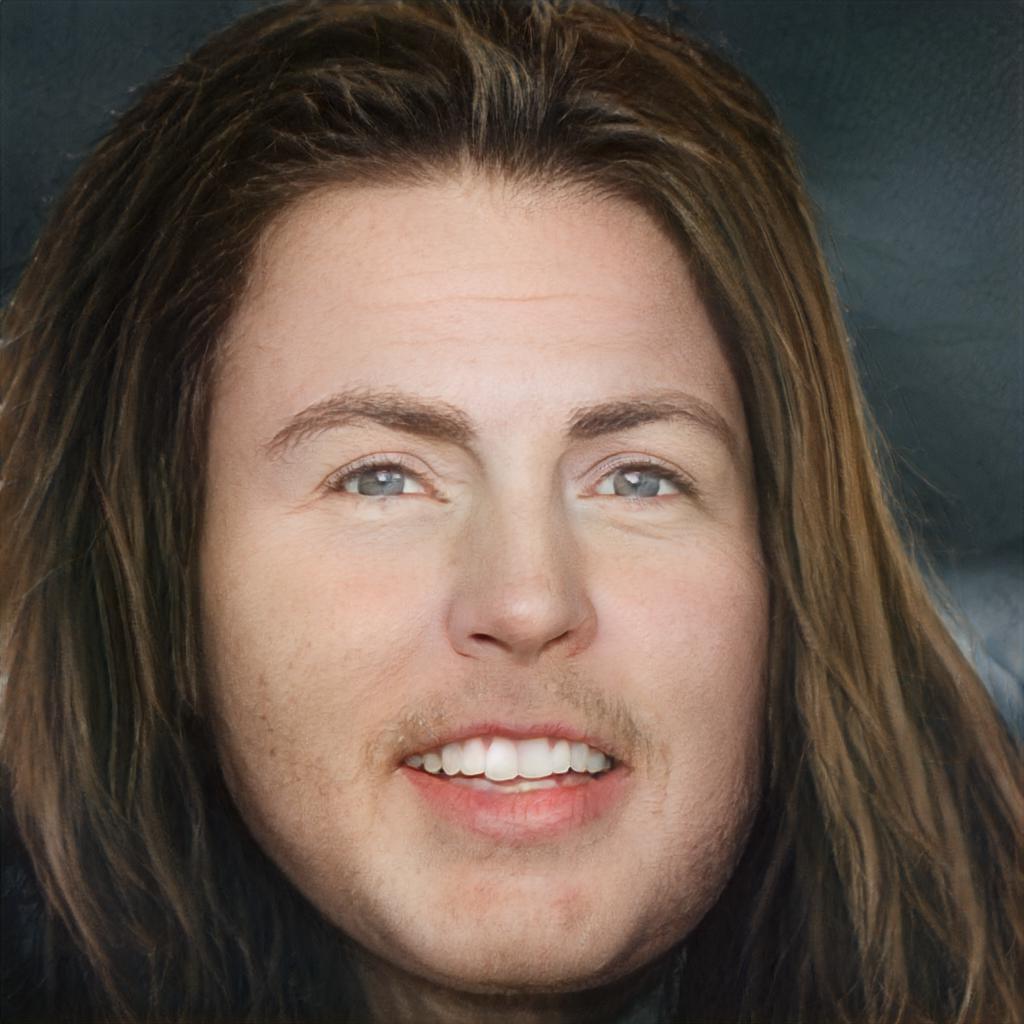} &
            \includegraphics[width=0.135\textwidth]{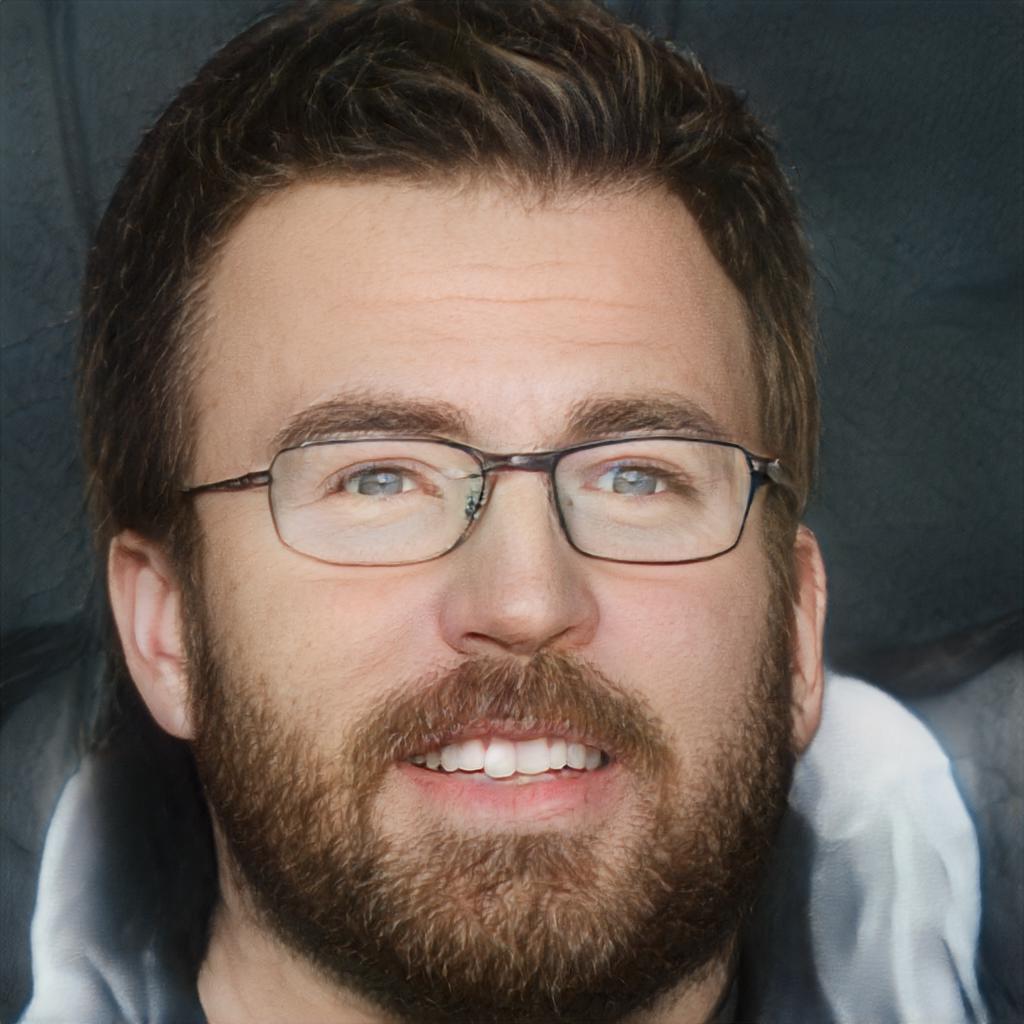} &
            \includegraphics[width=0.135\textwidth]{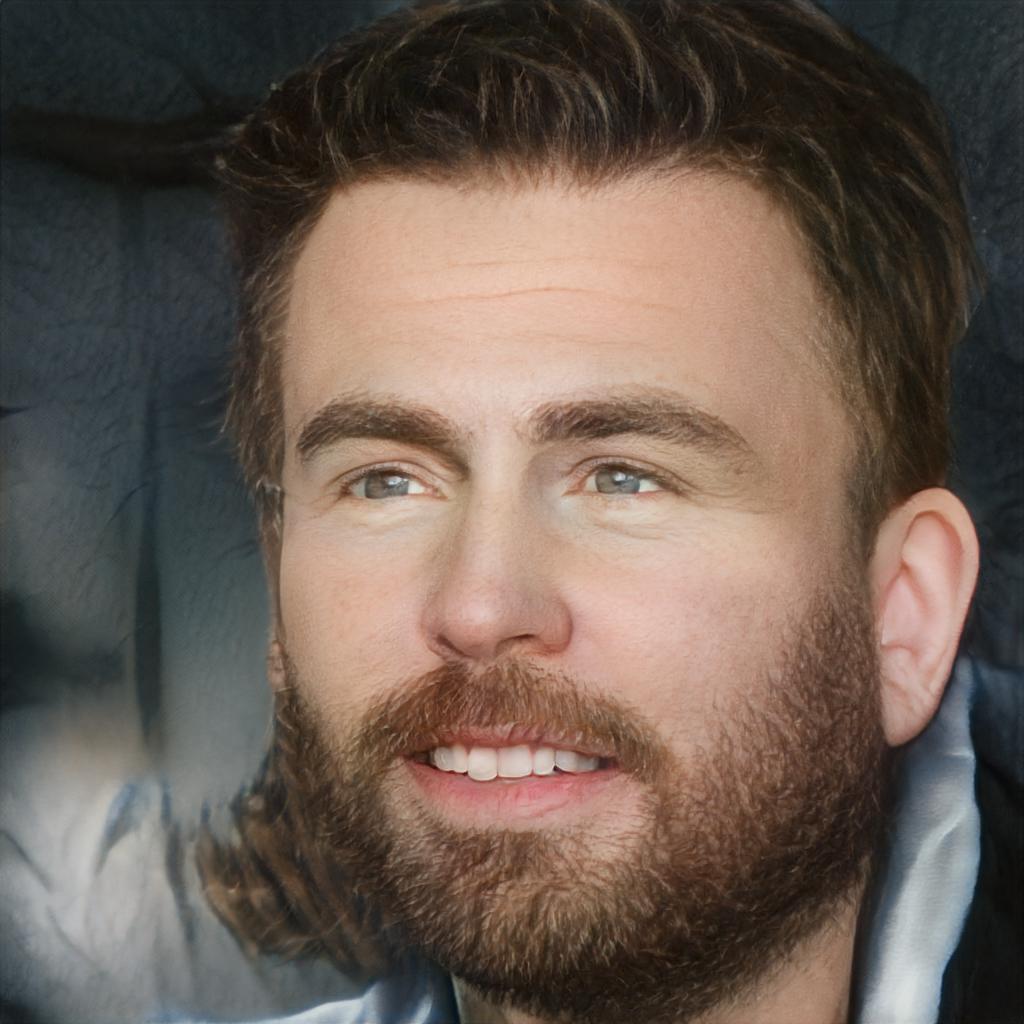} &
            \includegraphics[width=0.135\textwidth]{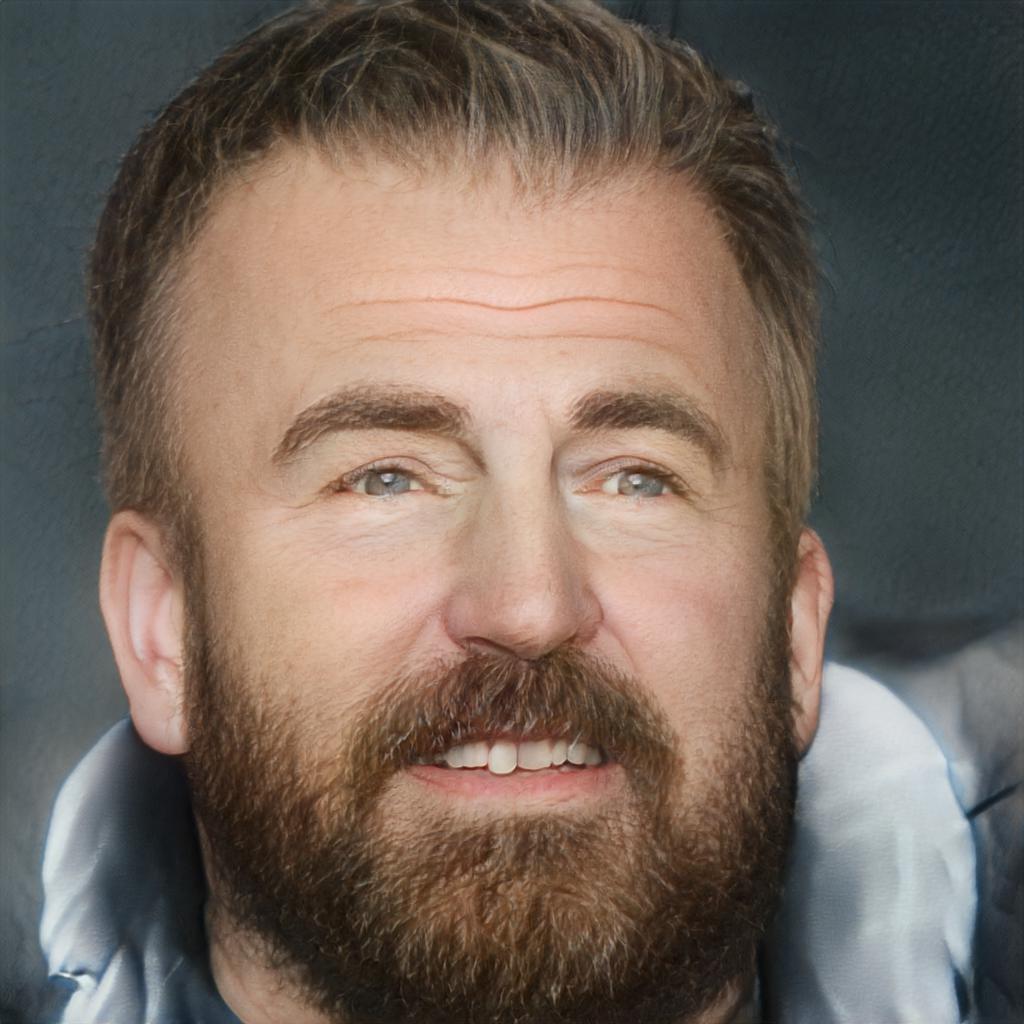}  \\
            \raisebox{0.06\textwidth}{\texttt{D}} &\includegraphics[width=0.135\textwidth]{images/appendix/styleflow_edit_celebs_ours/chris_evans_src.jpg} & 
            \includegraphics[width=0.135\textwidth]{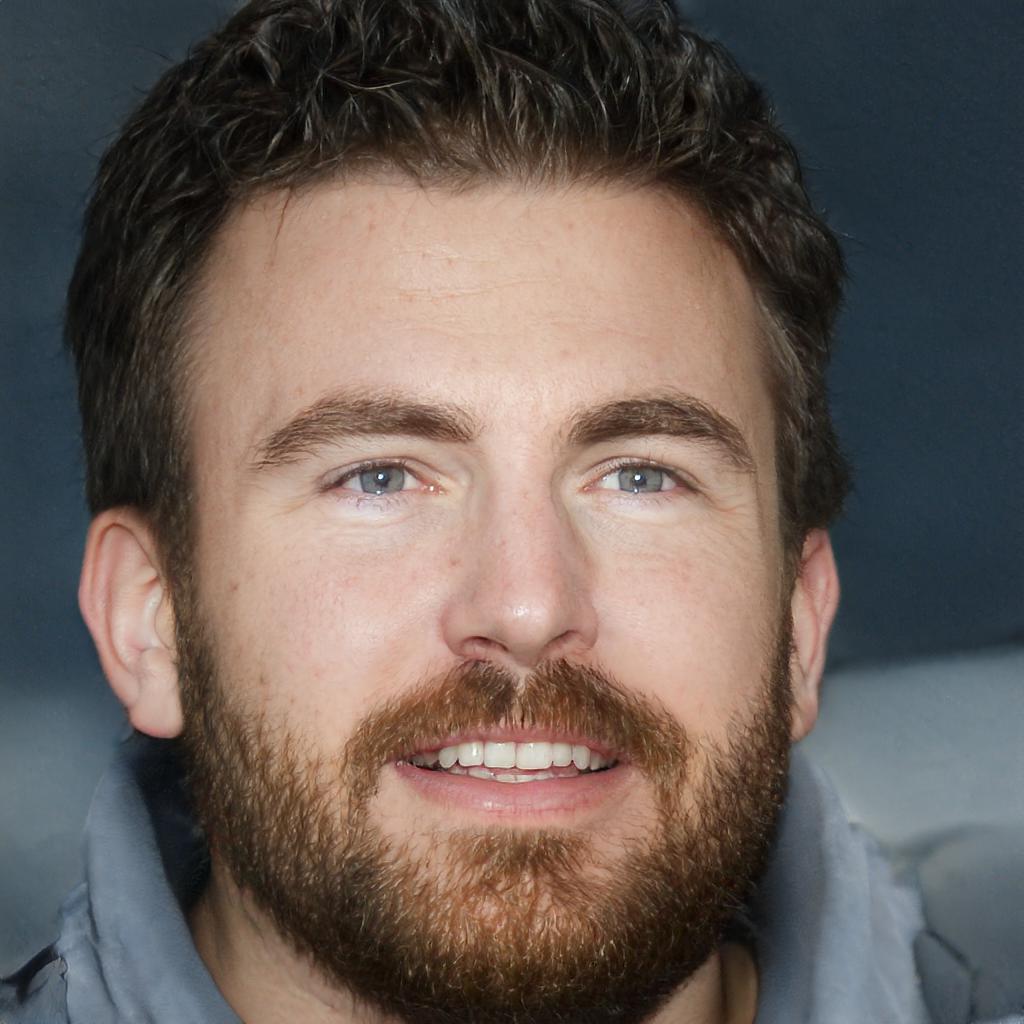} &
            \includegraphics[width=0.135\textwidth]{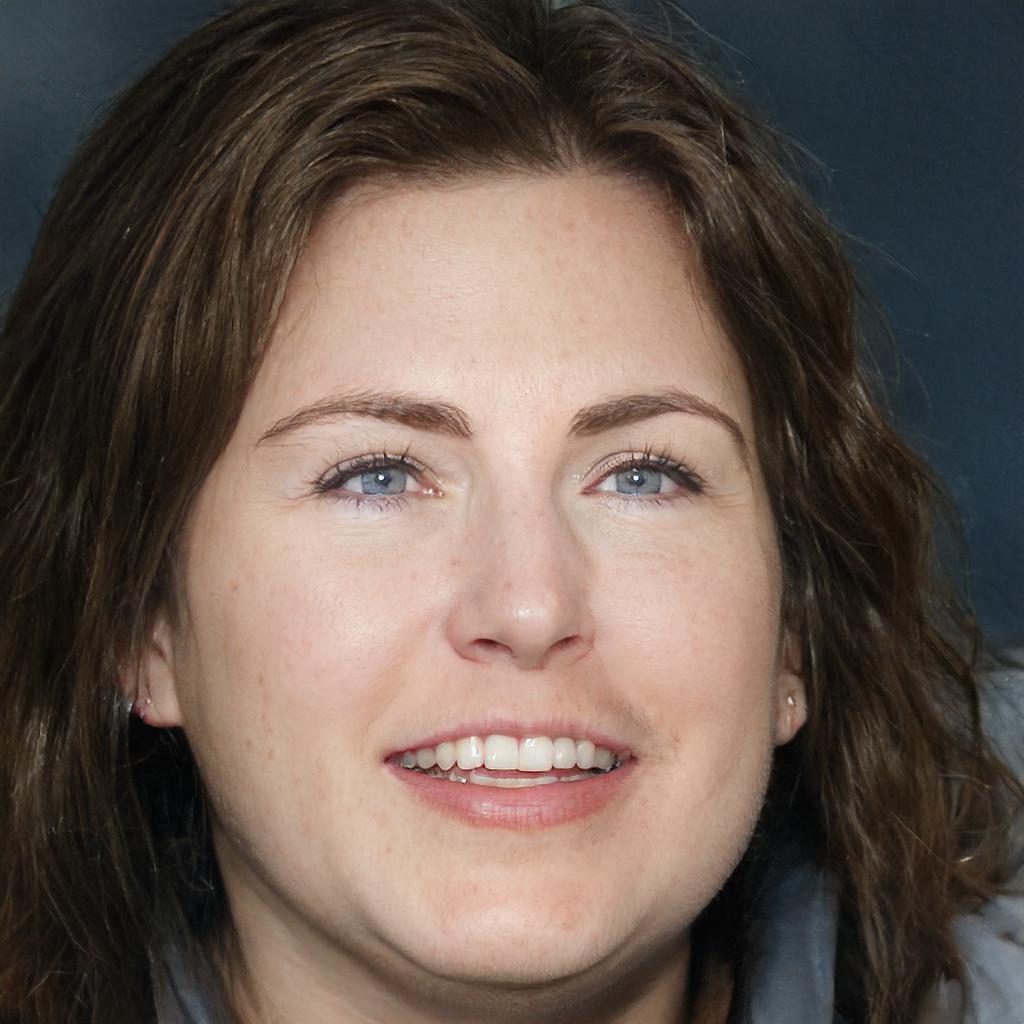} &
            \includegraphics[width=0.135\textwidth]{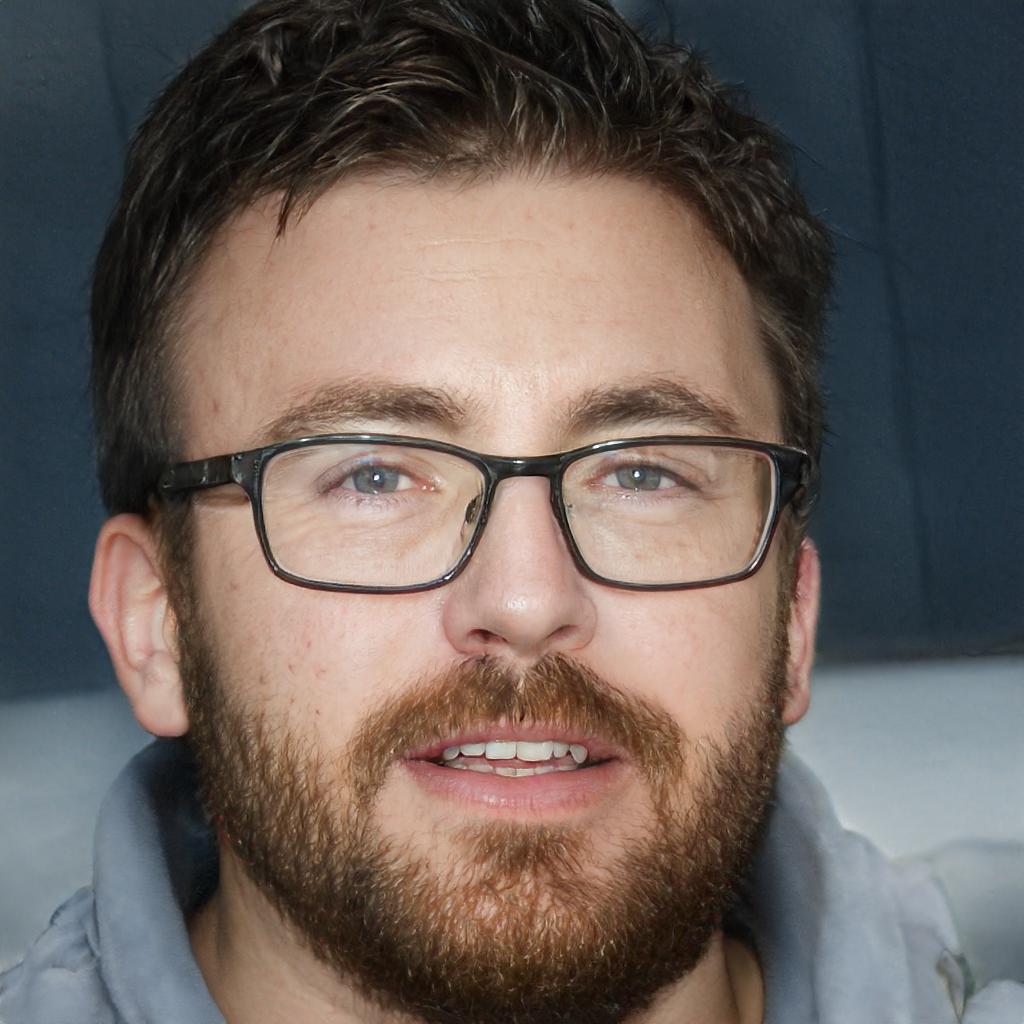} &
            \includegraphics[width=0.135\textwidth]{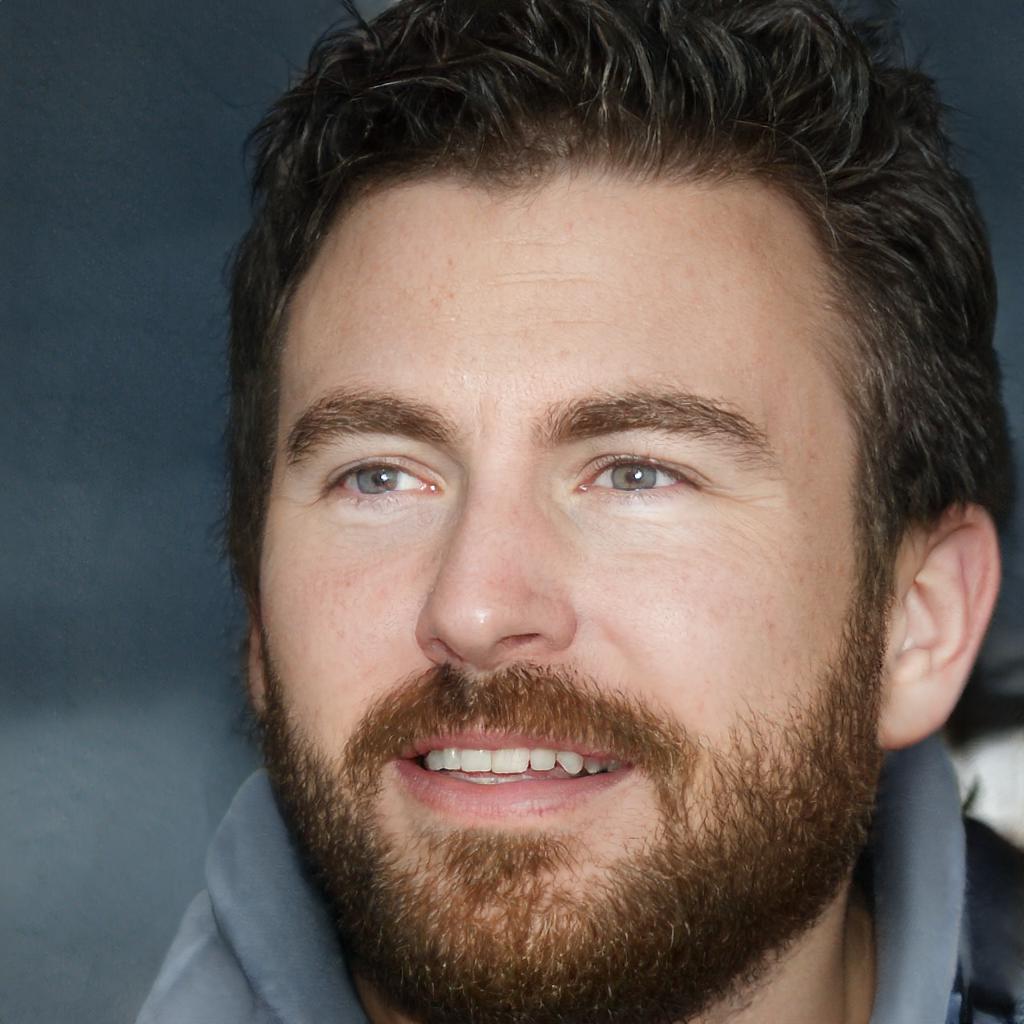} &
            \includegraphics[width=0.135\textwidth]{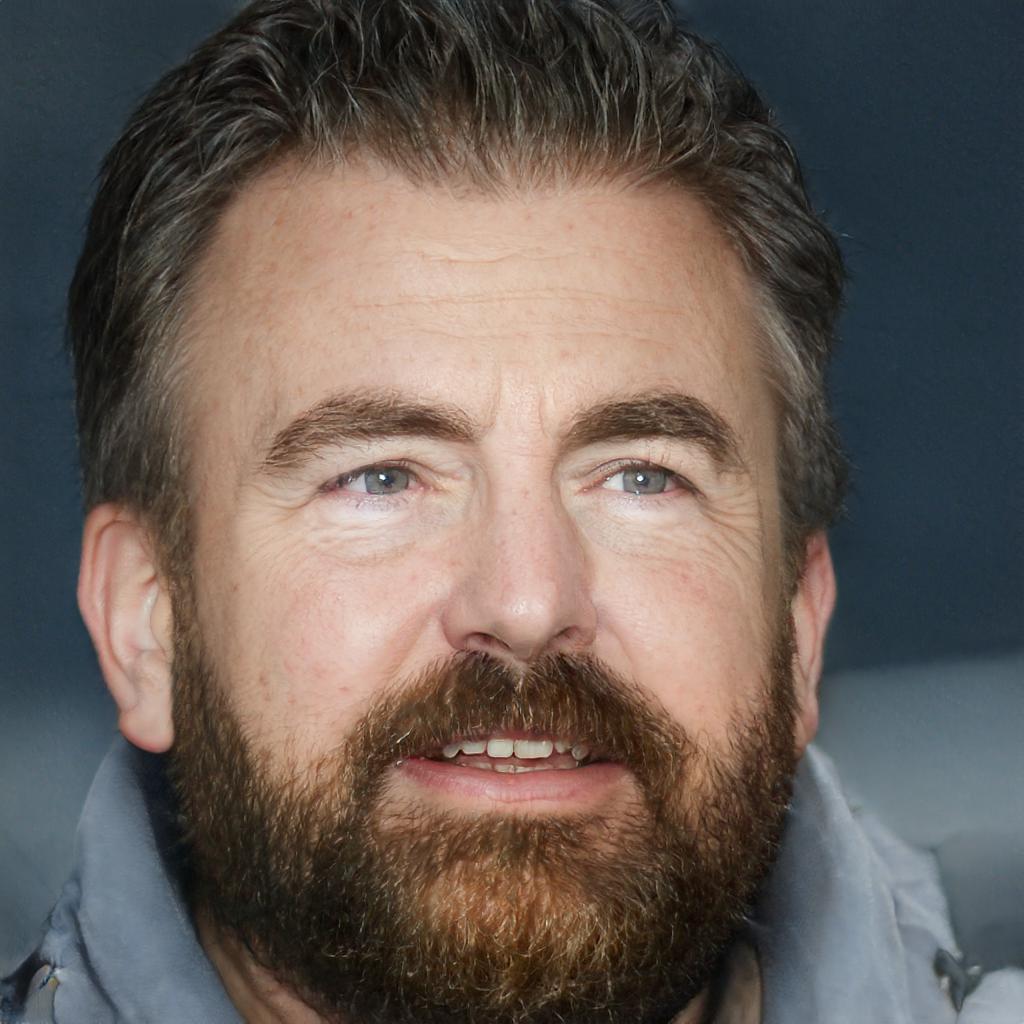} \\
            
            \raisebox{0.06\textwidth}{\texttt{A}} & \includegraphics[width=0.135\textwidth]{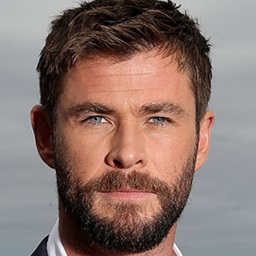} & 
            \includegraphics[width=0.135\textwidth]{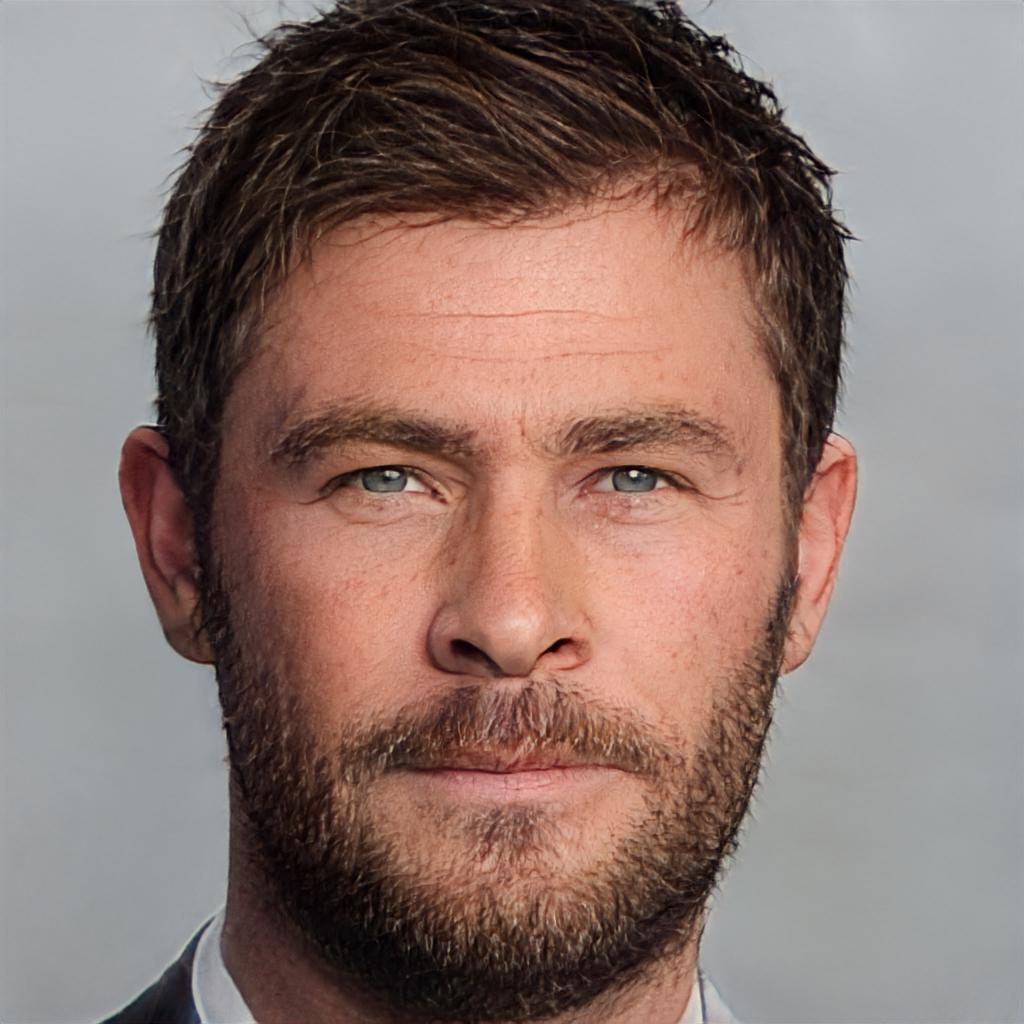} &
            \includegraphics[width=0.135\textwidth]{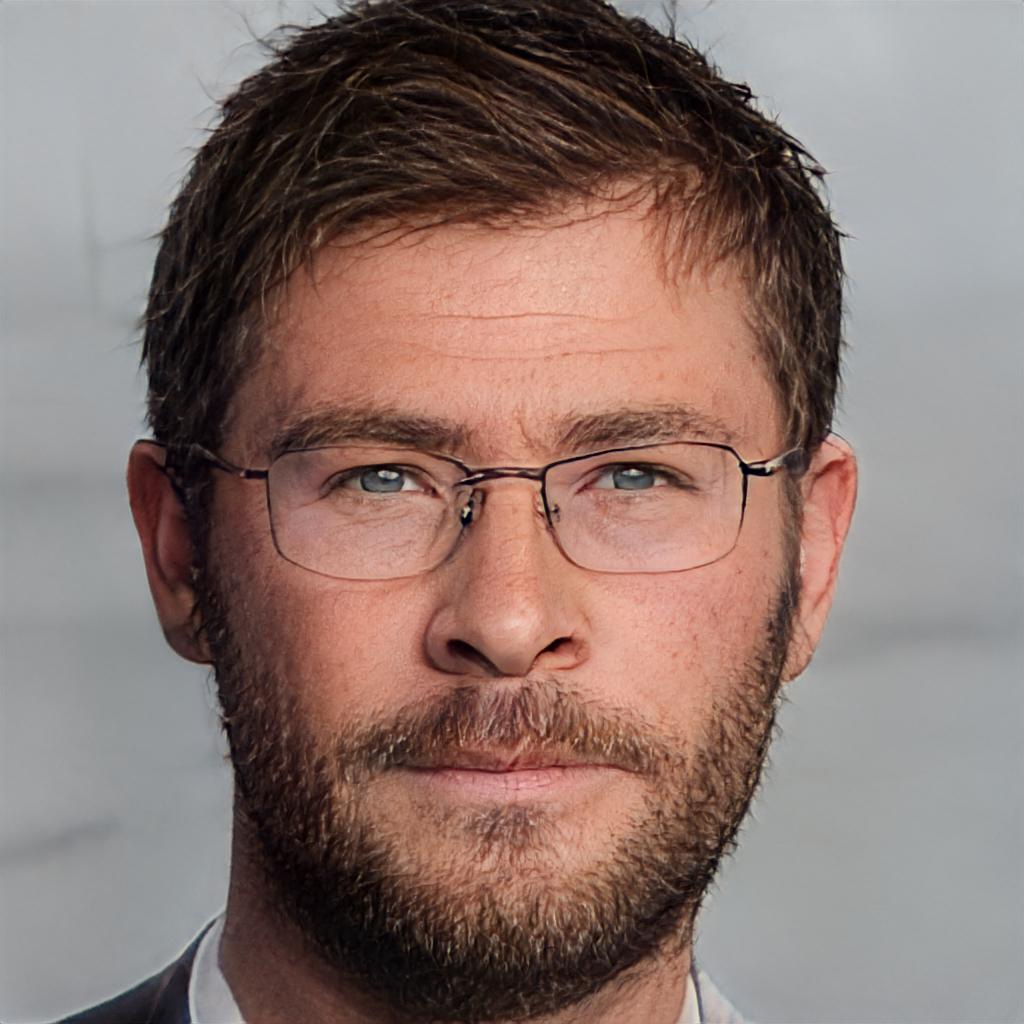} &
            \includegraphics[width=0.135\textwidth]{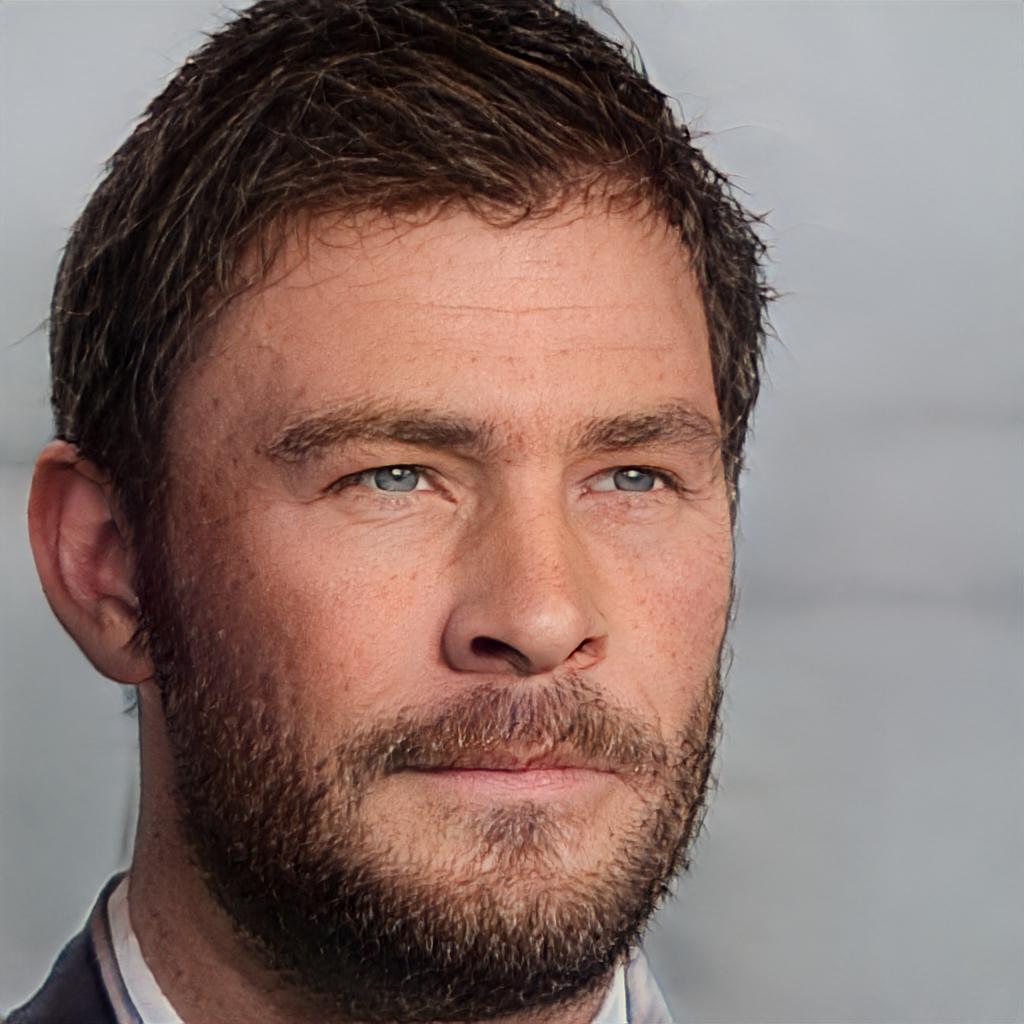} &
            \includegraphics[width=0.135\textwidth]{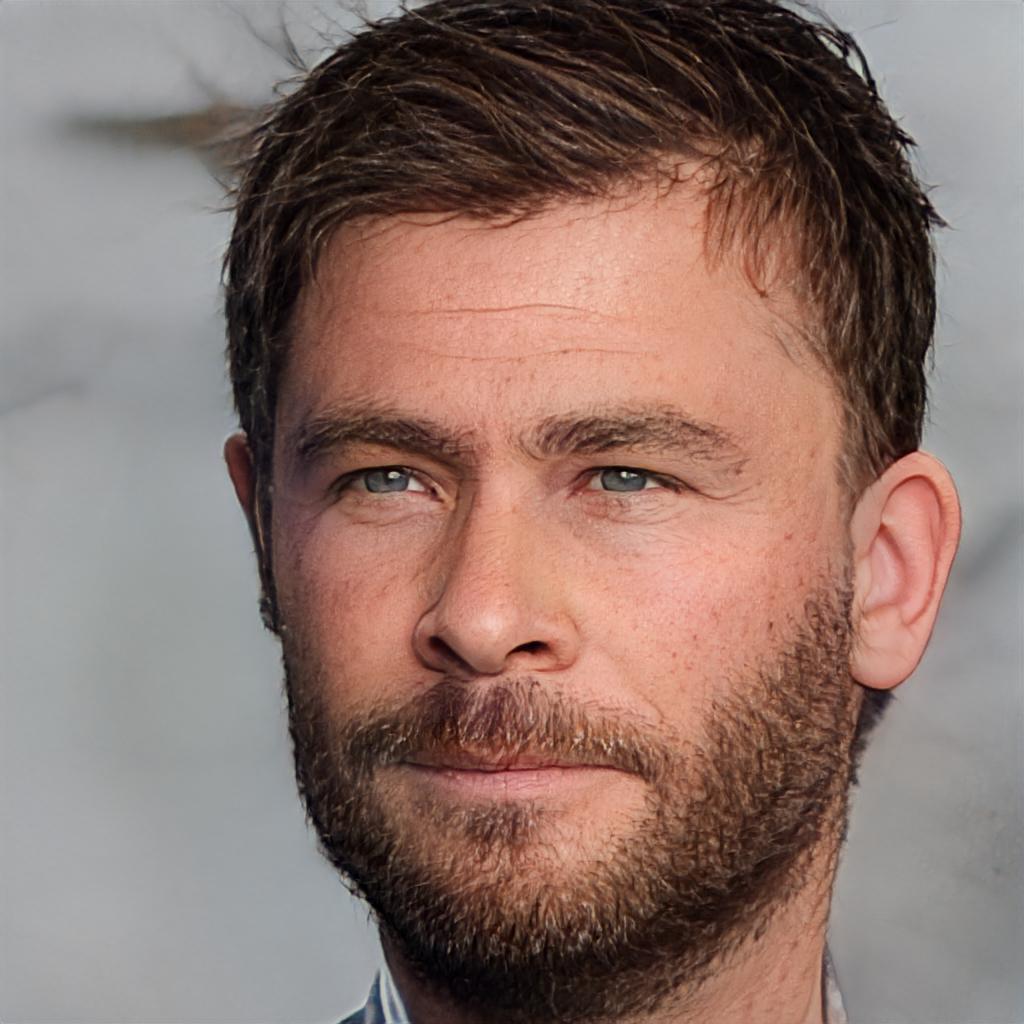} &
            \includegraphics[width=0.135\textwidth]{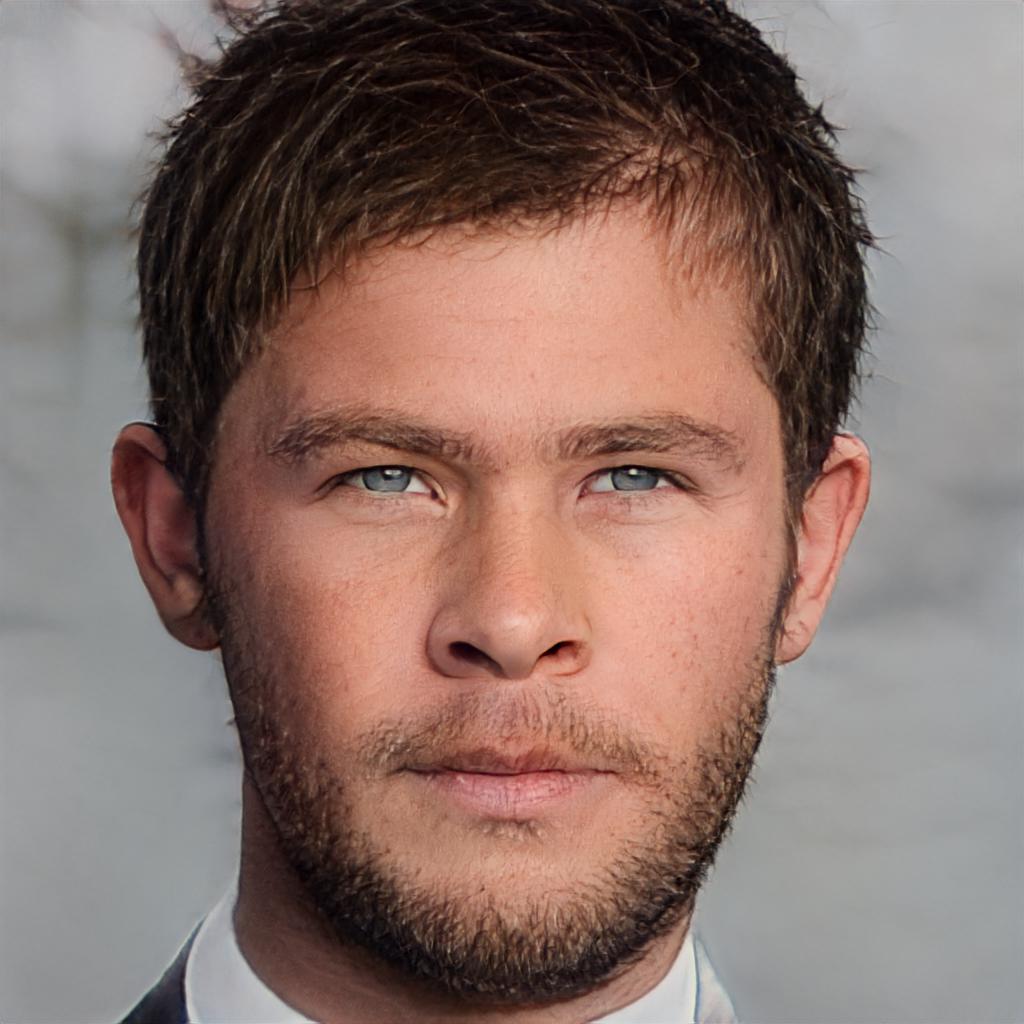} \\
            \raisebox{0.06\textwidth}{\texttt{D}} & \includegraphics[width=0.135\textwidth]{images/appendix/styleflow_edit_celebs_ours/chris_src.jpg} & 
            \includegraphics[width=0.135\textwidth]{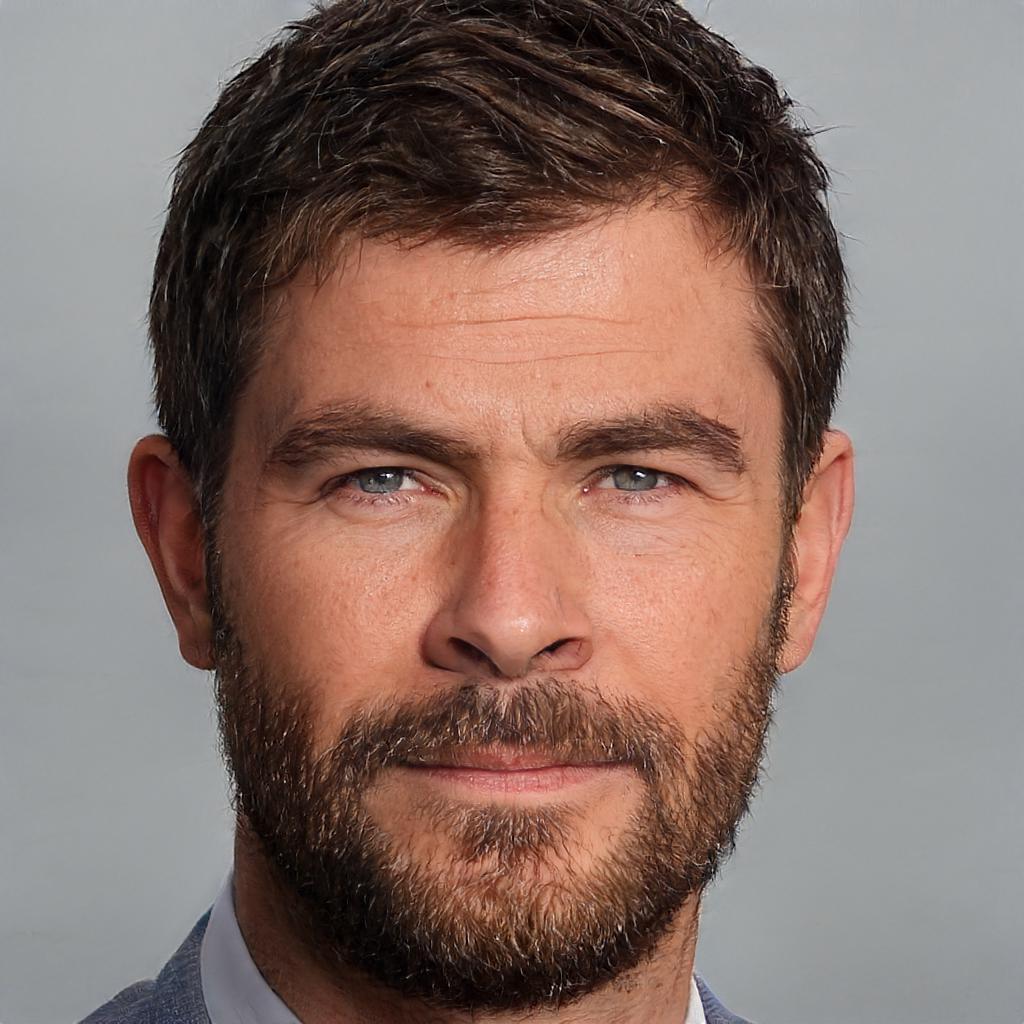} &
            \includegraphics[width=0.135\textwidth]{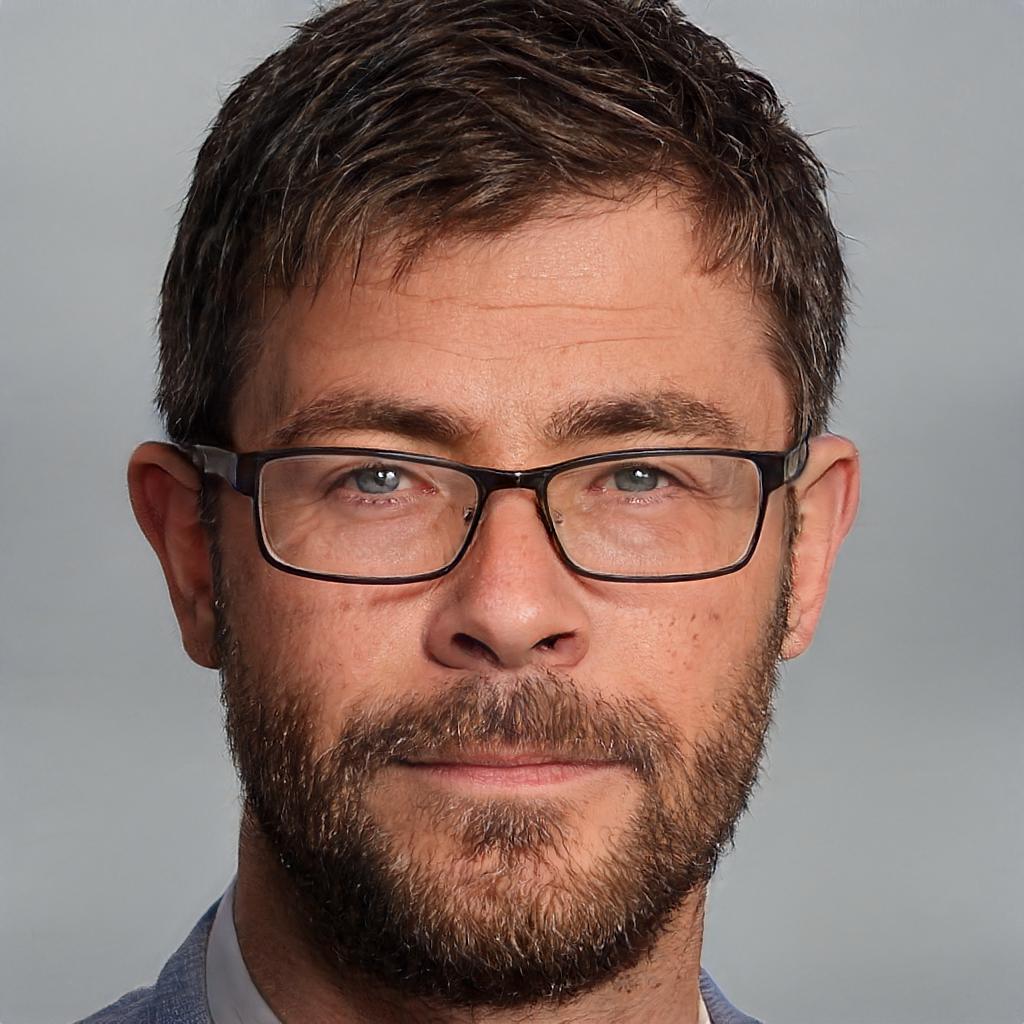} &
            \includegraphics[width=0.135\textwidth]{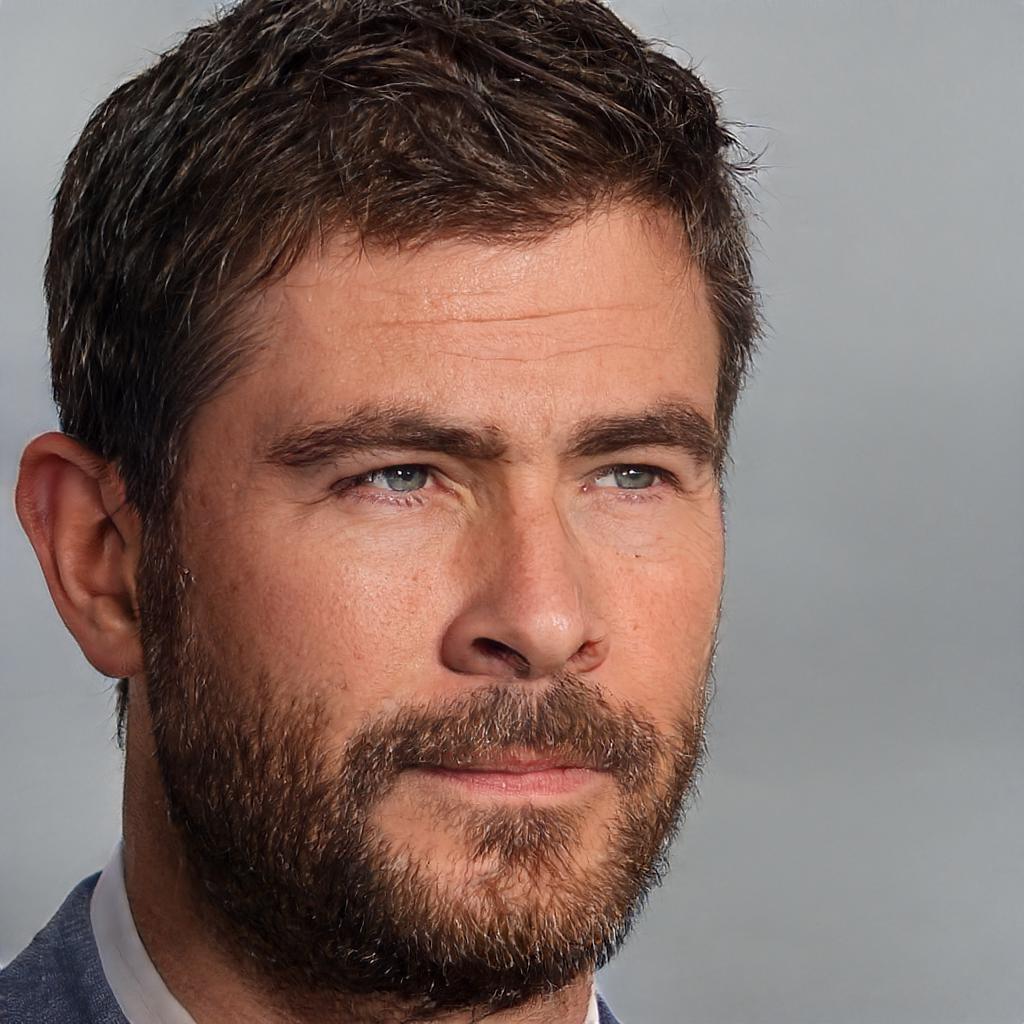} &
            \includegraphics[width=0.135\textwidth]{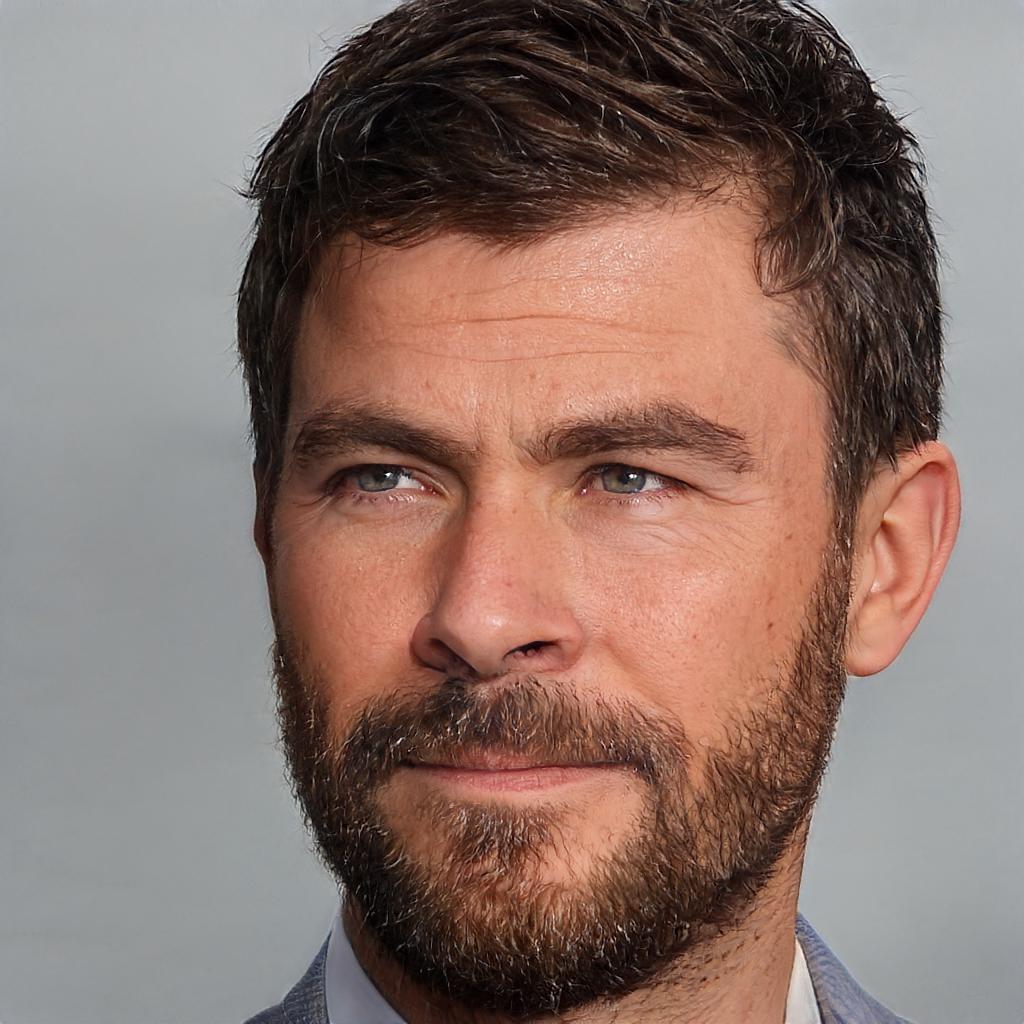} &
            \includegraphics[width=0.135\textwidth]{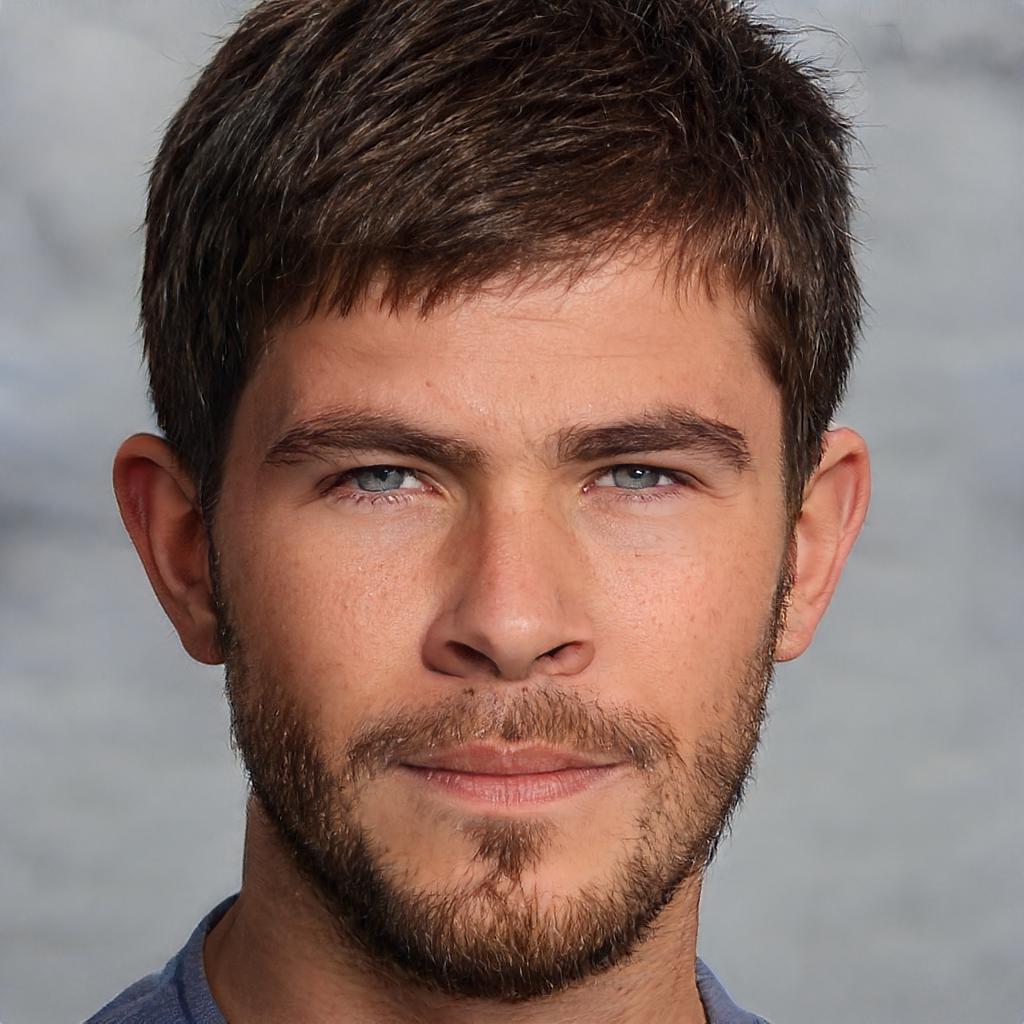} \\
            \raisebox{0.06\textwidth}{\texttt{A}} & \includegraphics[width=0.135\textwidth]{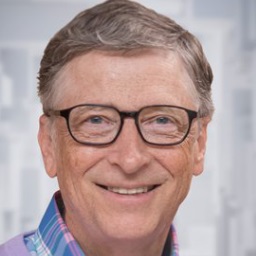} & 
            \includegraphics[width=0.135\textwidth]{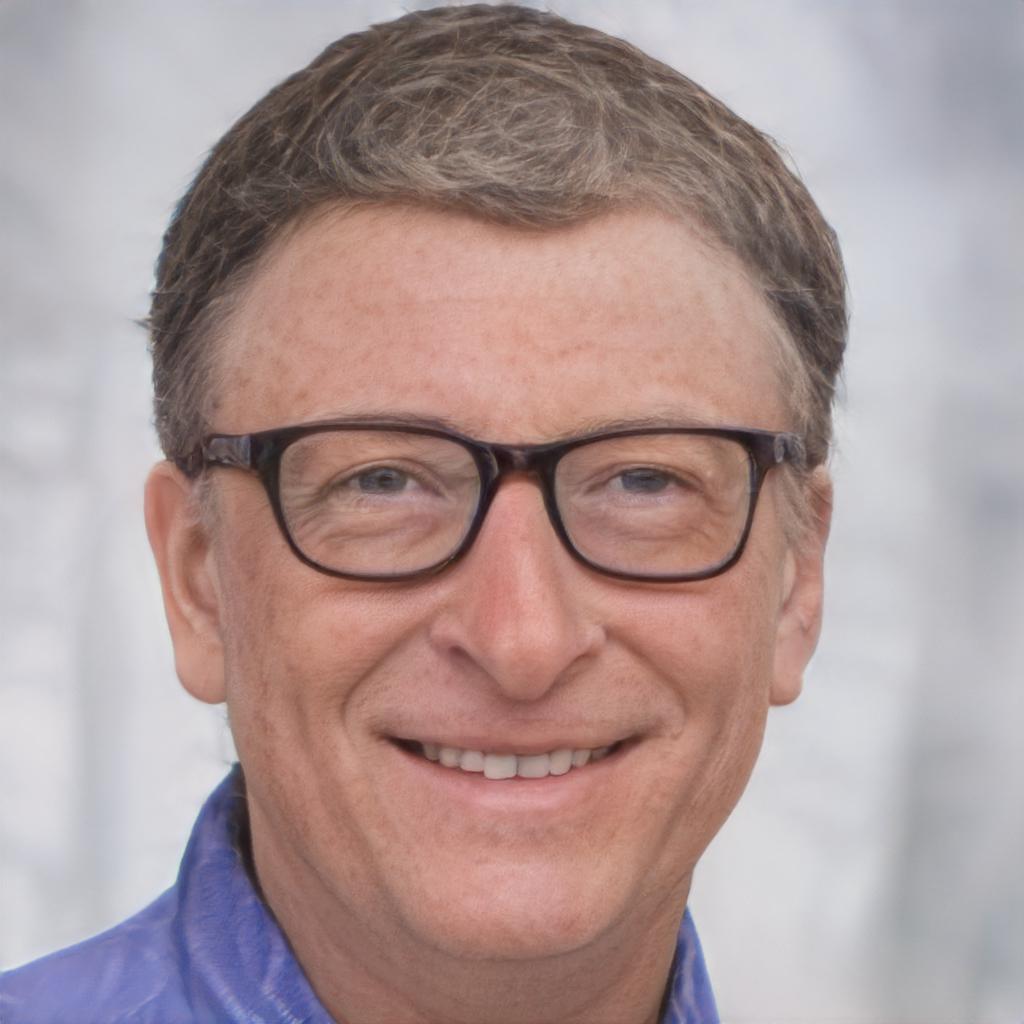} &
            \includegraphics[width=0.135\textwidth]{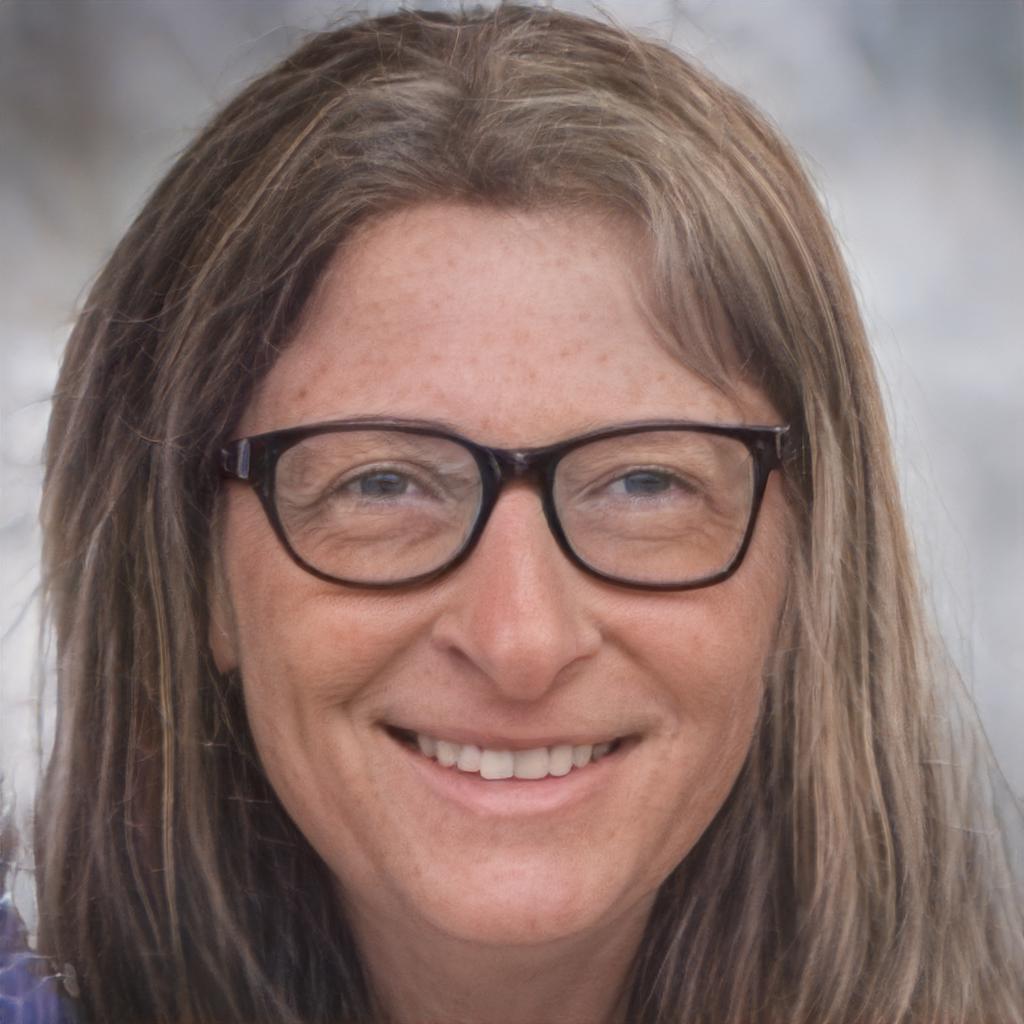} &
            \includegraphics[width=0.135\textwidth]{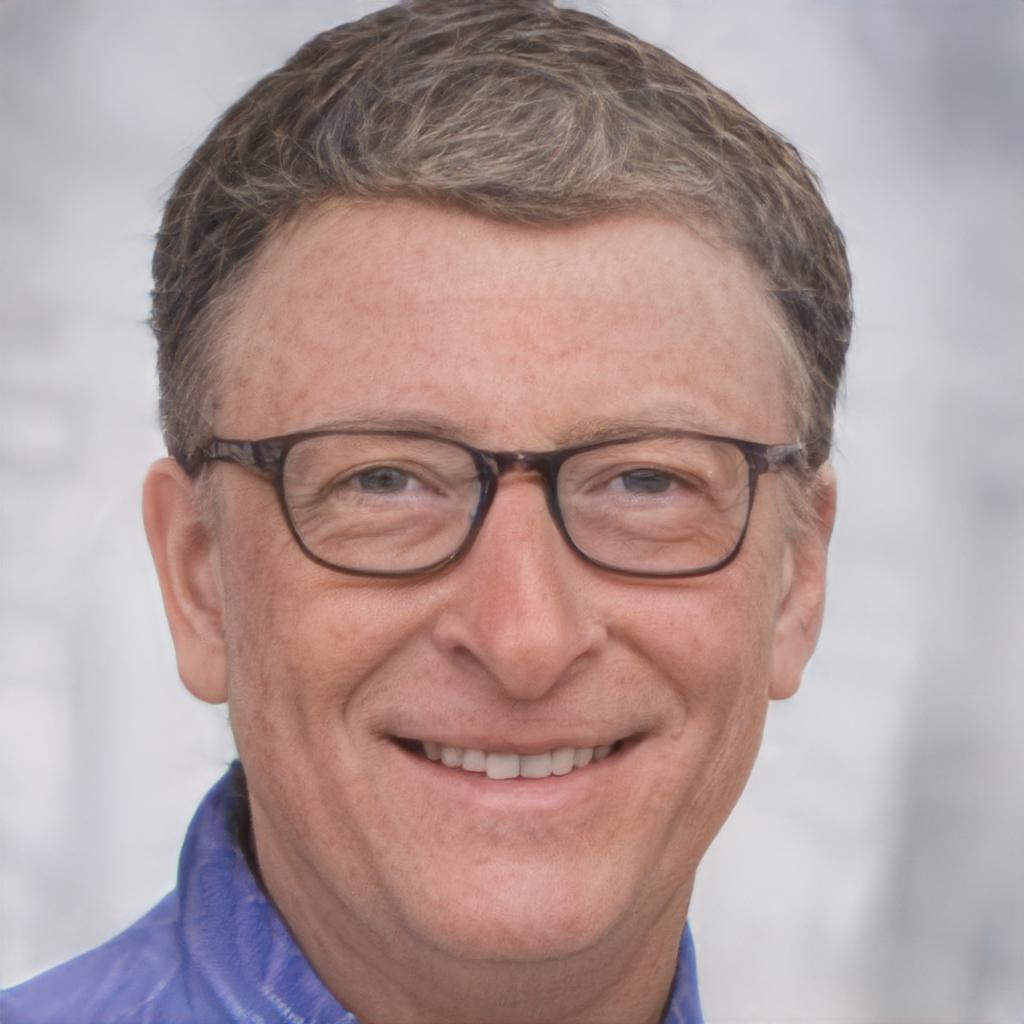} &
            \includegraphics[width=0.135\textwidth]{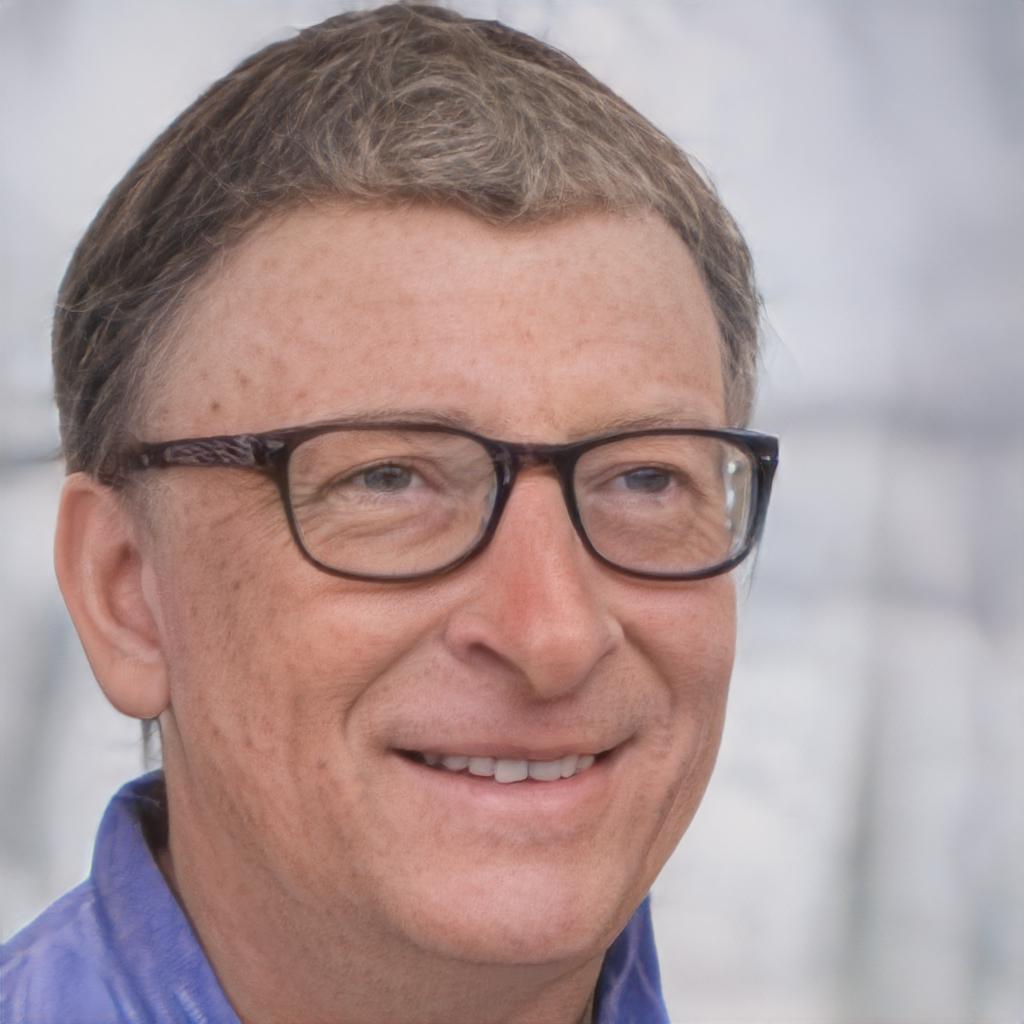} &
            \includegraphics[width=0.135\textwidth]{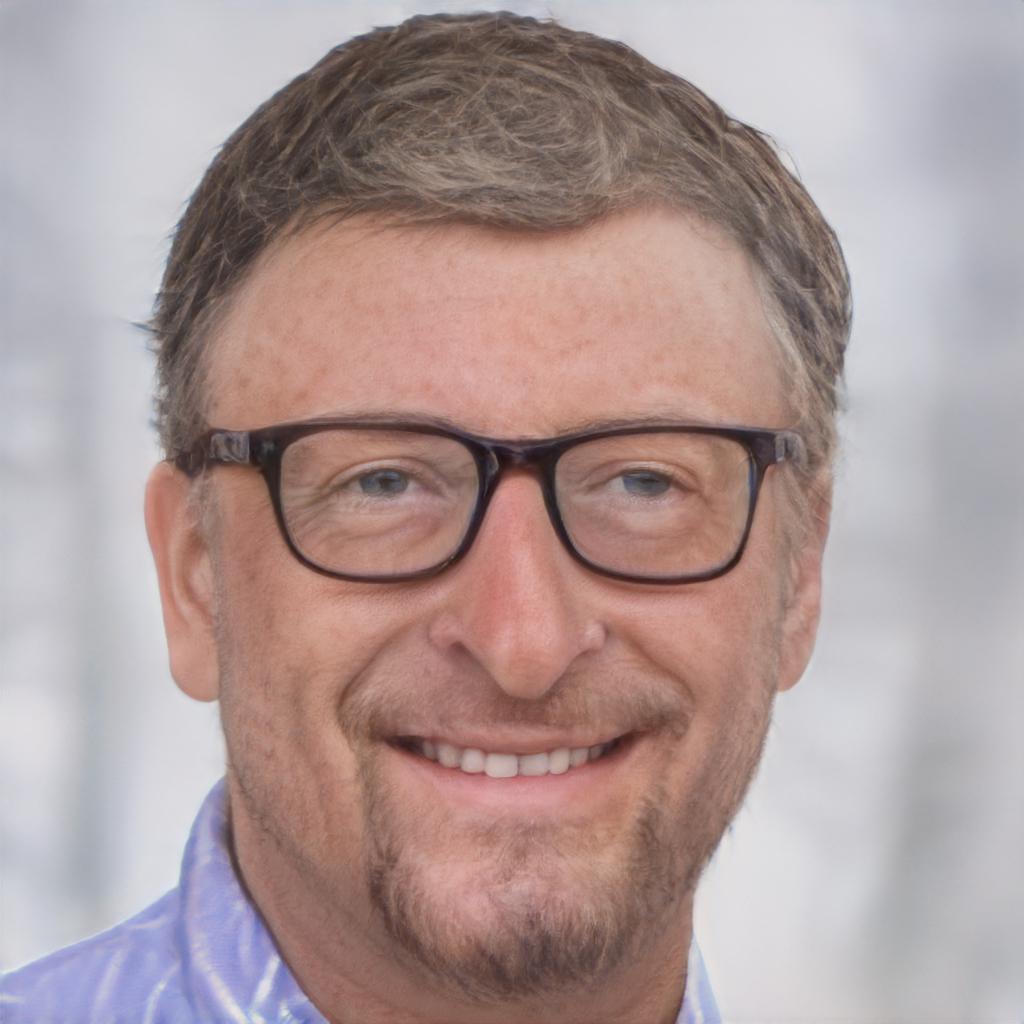}  \\
            \raisebox{0.06\textwidth}{\texttt{D}} & \includegraphics[width=0.135\textwidth]{images/appendix/styleflow_edit_celebs_ours/gates2_src.jpg} & 
            \includegraphics[width=0.135\textwidth]{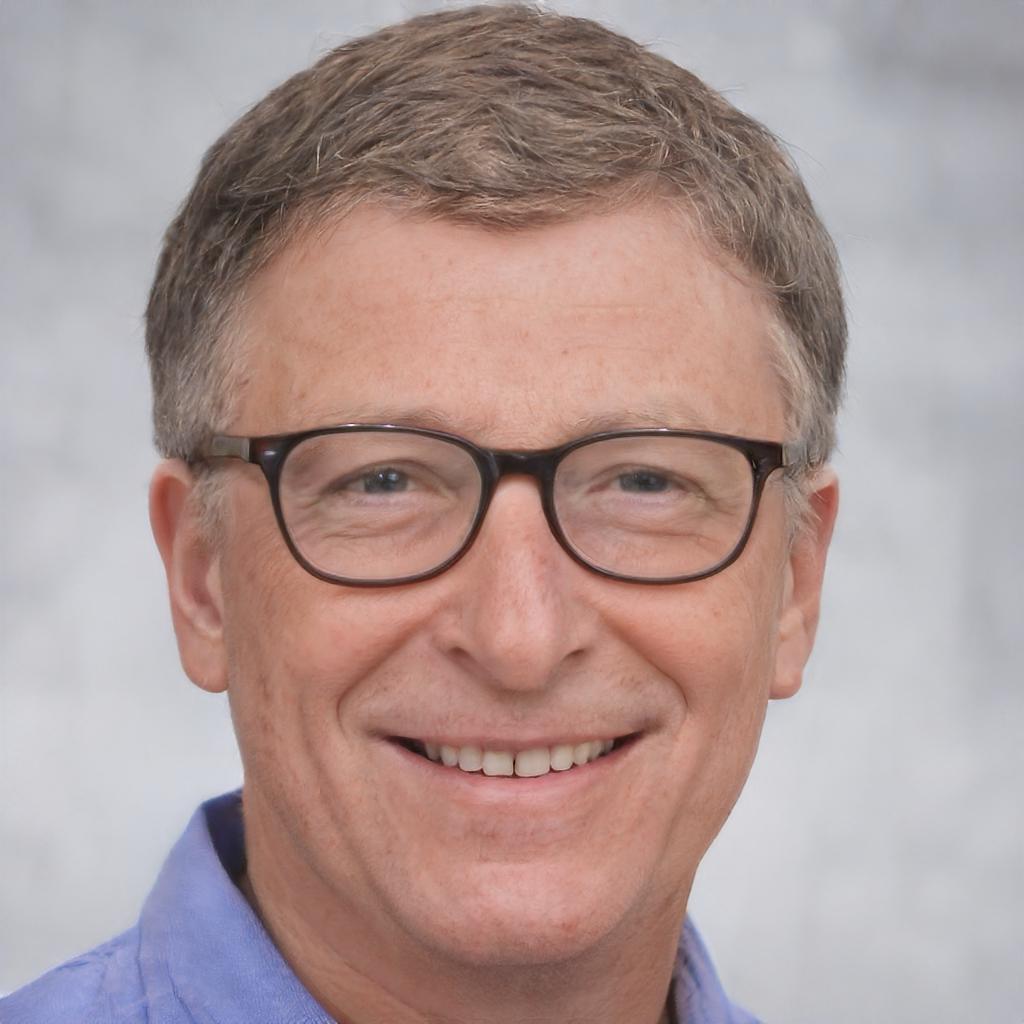} &
            \includegraphics[width=0.135\textwidth]{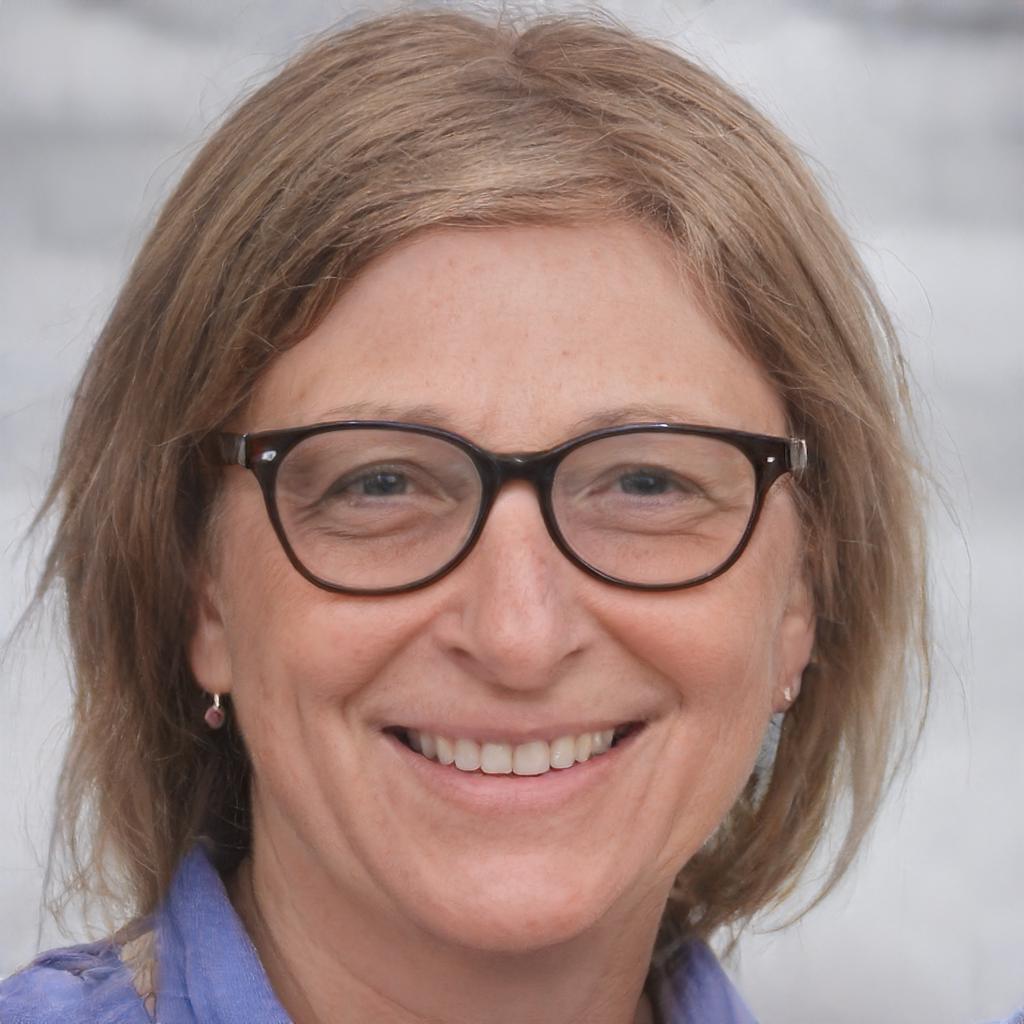} &
            \includegraphics[width=0.135\textwidth]{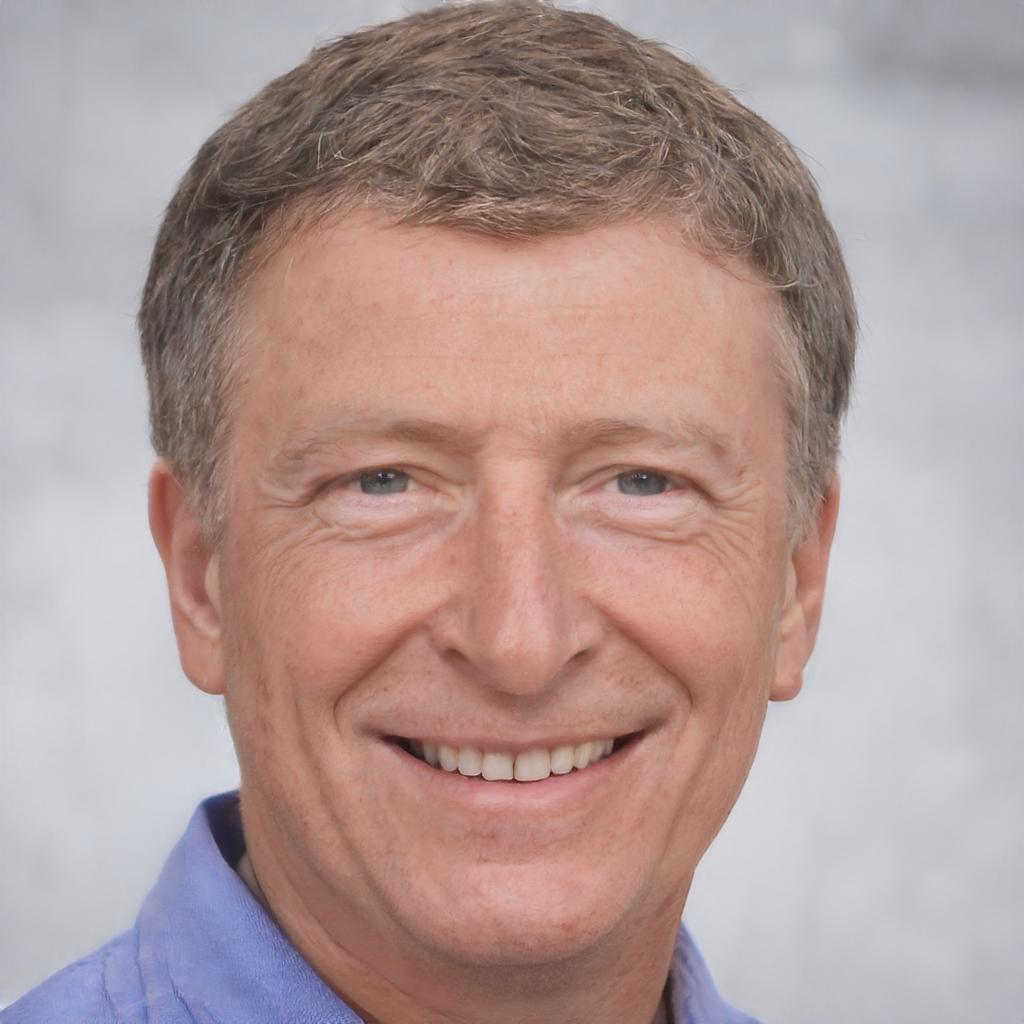} &
            \includegraphics[width=0.135\textwidth]{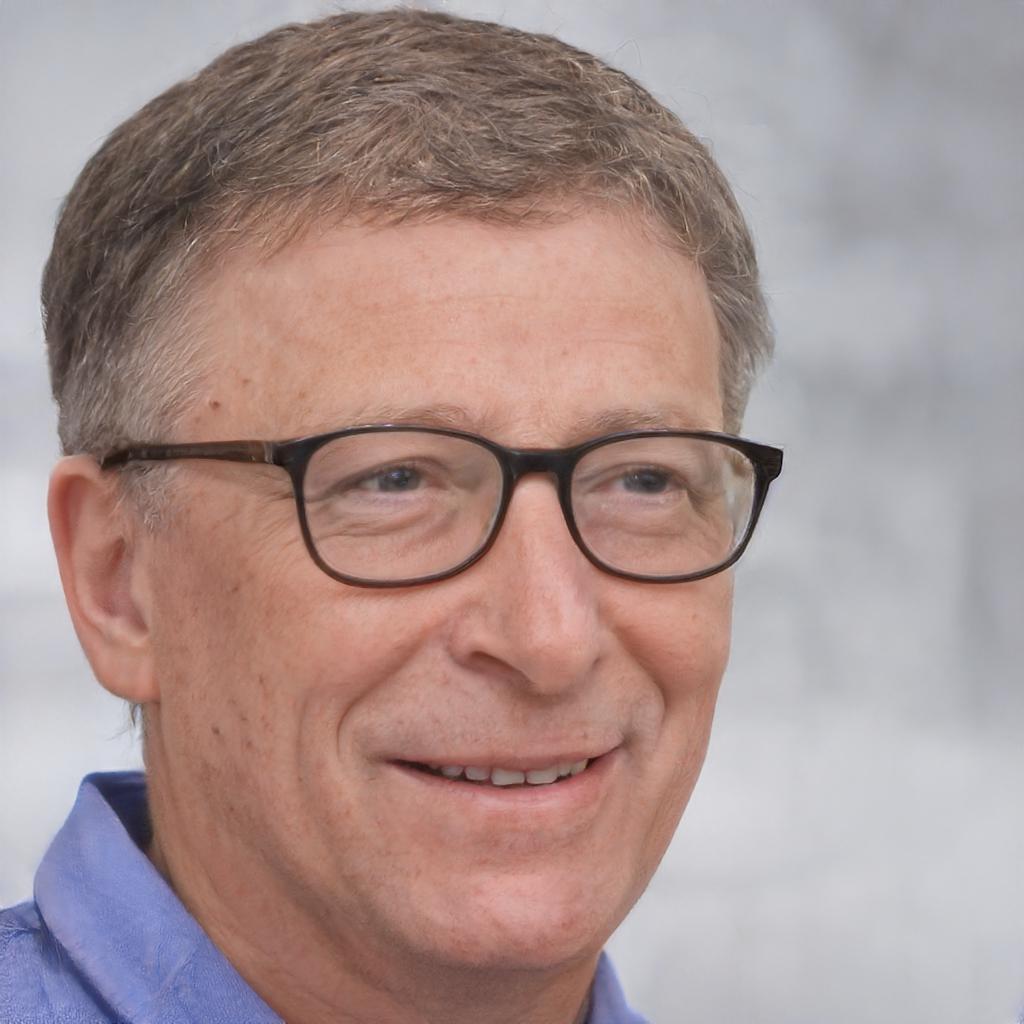} &
            \includegraphics[width=0.135\textwidth]{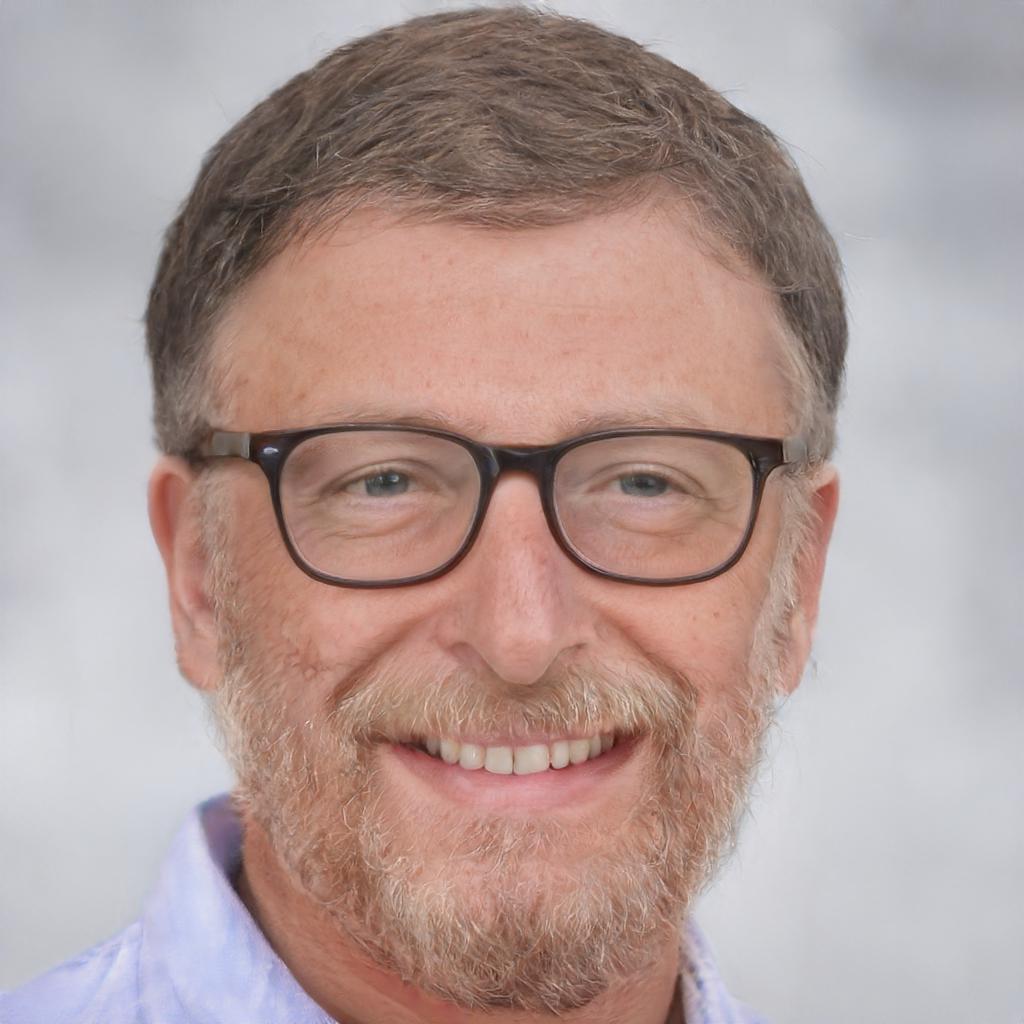} \\
            \raisebox{0.06\textwidth}{\texttt{A}} & \includegraphics[width=0.135\textwidth]{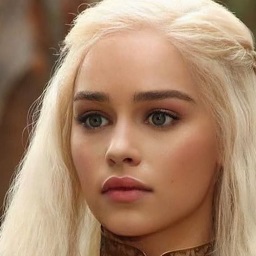} & 
            \includegraphics[width=0.135\textwidth]{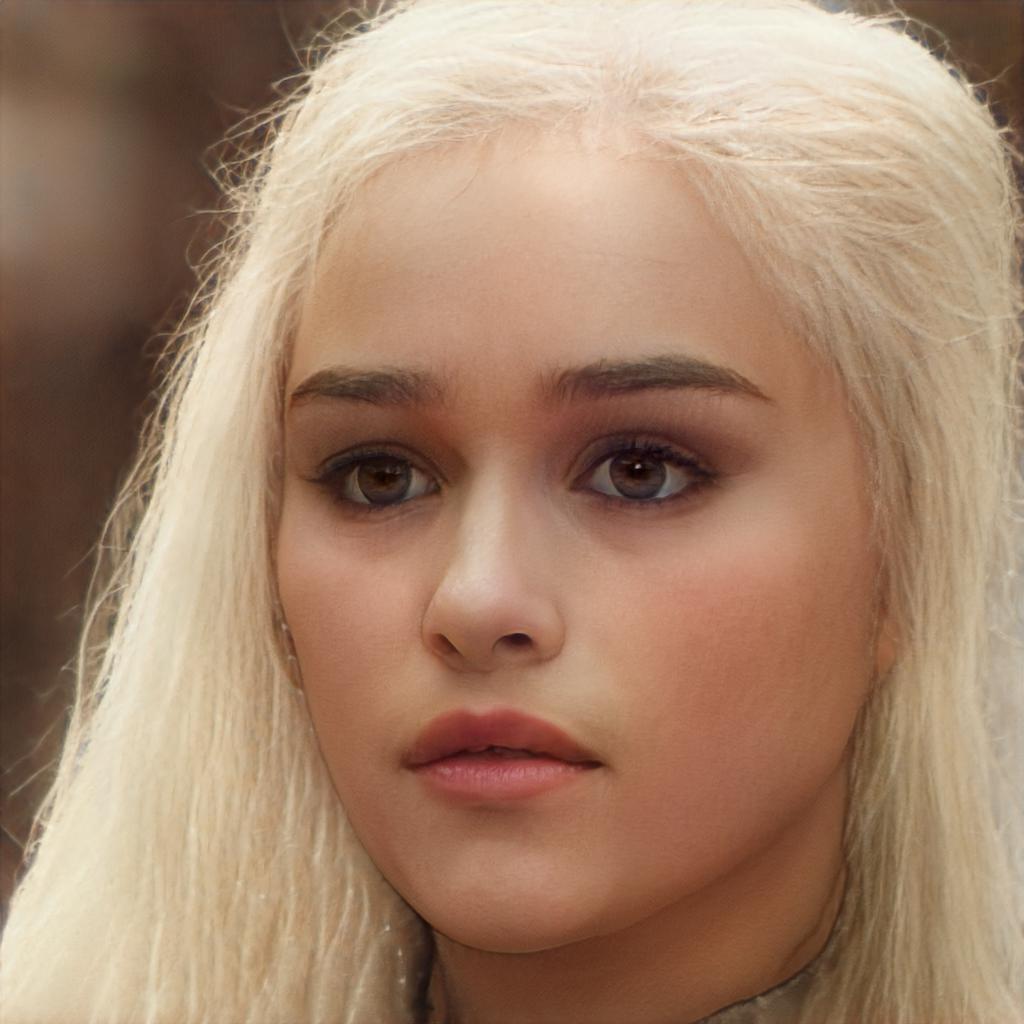} &
            \includegraphics[width=0.135\textwidth]{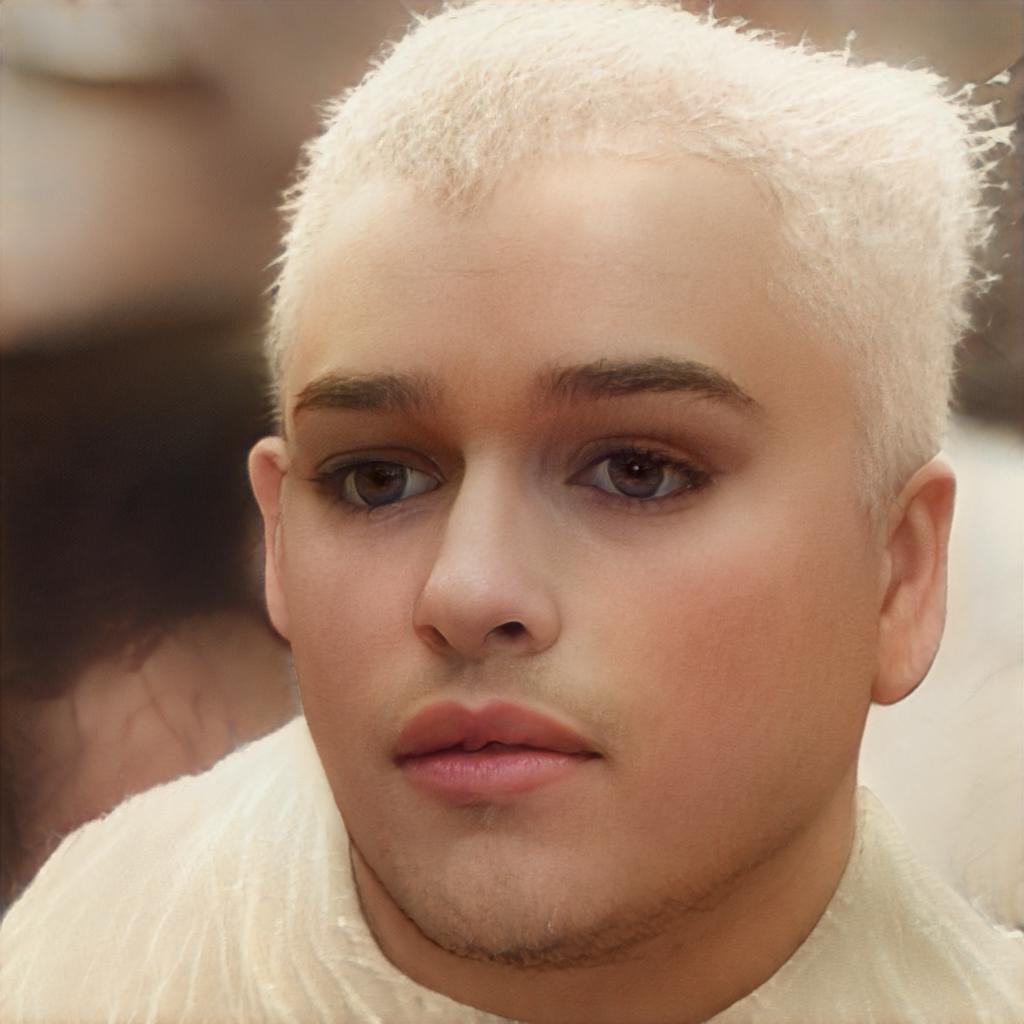} &
            \includegraphics[width=0.135\textwidth]{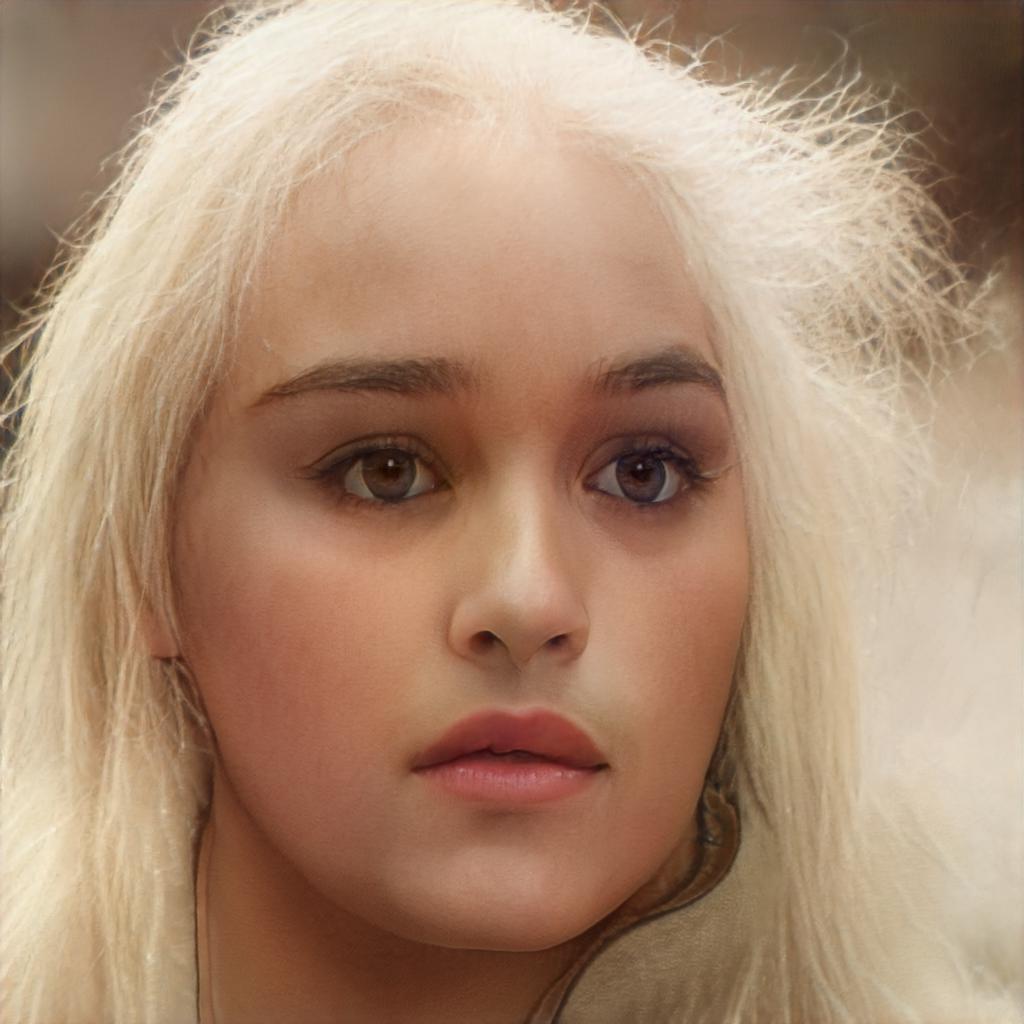} &
            \includegraphics[width=0.135\textwidth]{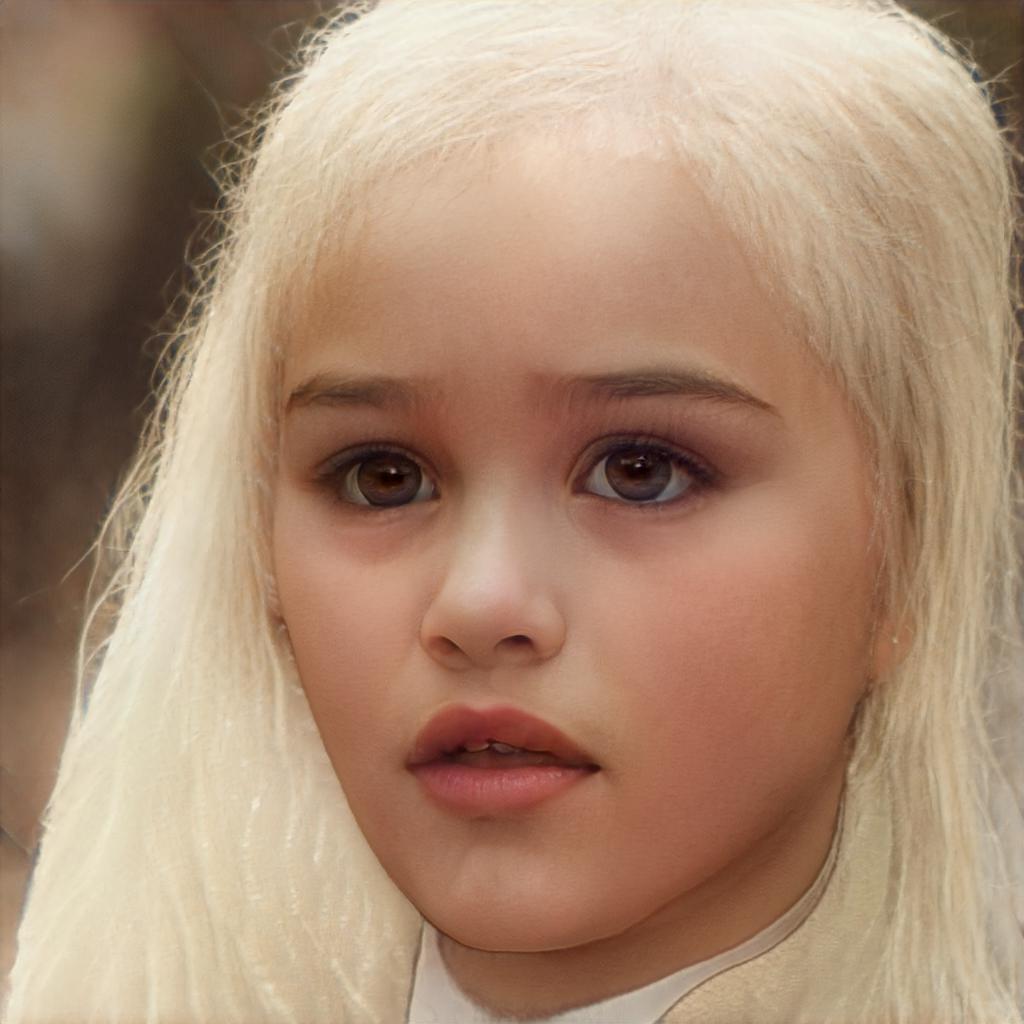} &
            \includegraphics[width=0.135\textwidth]{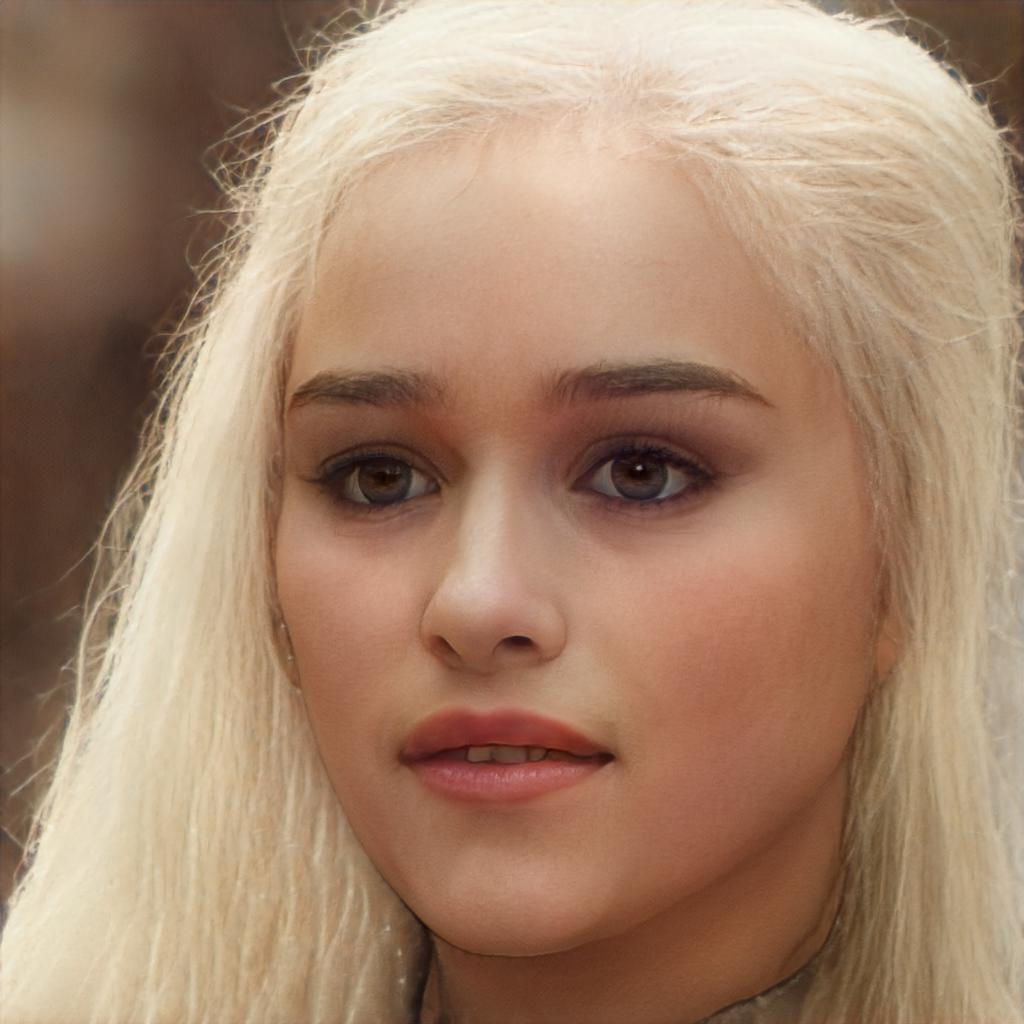}  \\
            \raisebox{0.06\textwidth}{\texttt{D}} & \includegraphics[width=0.135\textwidth]{images/appendix/styleflow_edit_celebs_ours/got2_src.jpg} &
            \includegraphics[width=0.135\textwidth]{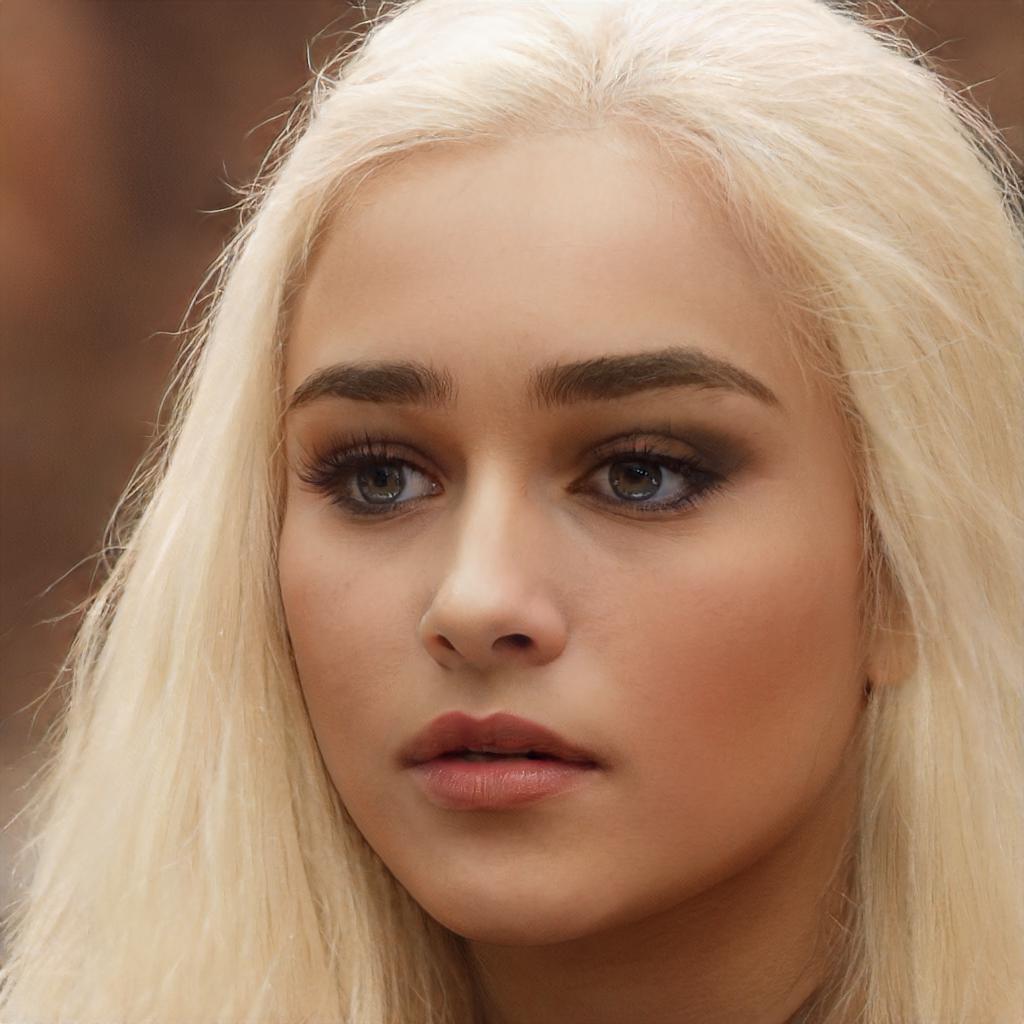} &
            \includegraphics[width=0.135\textwidth]{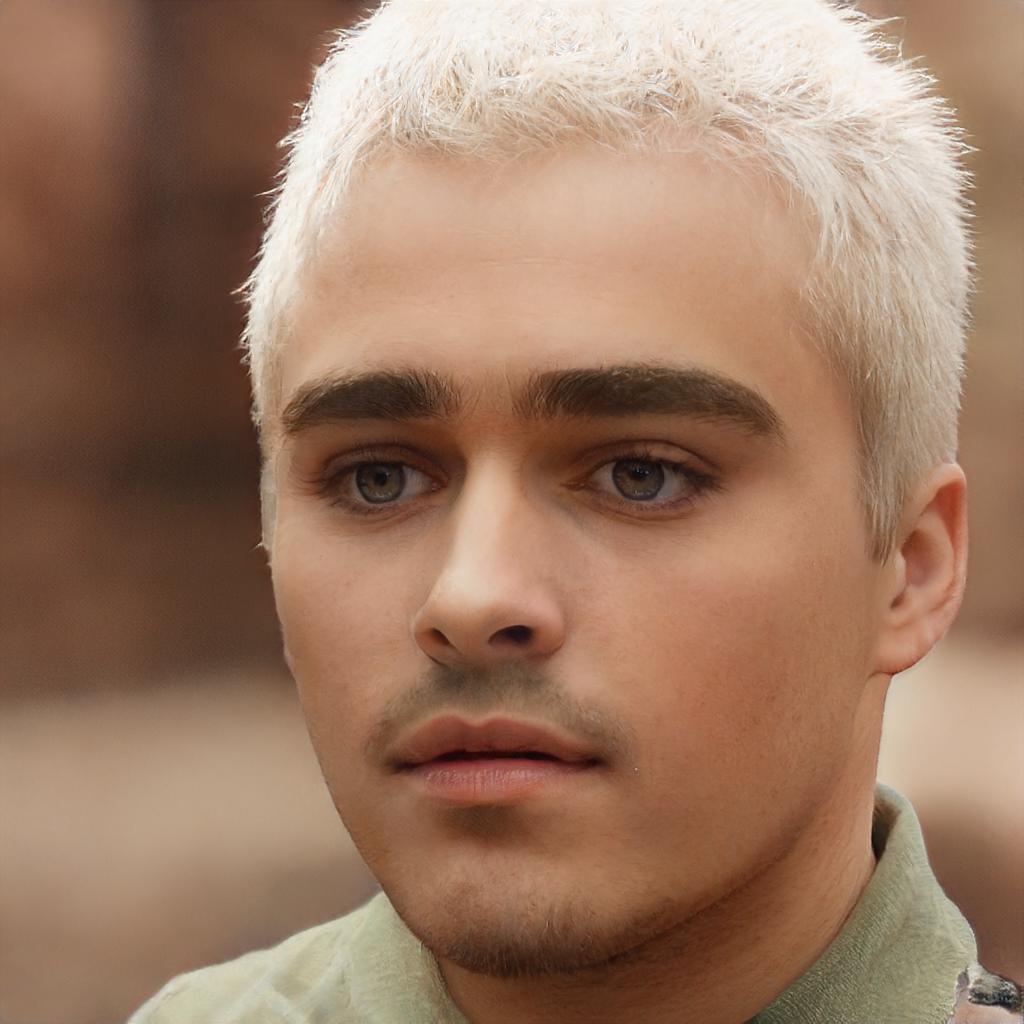} &
            \includegraphics[width=0.135\textwidth]{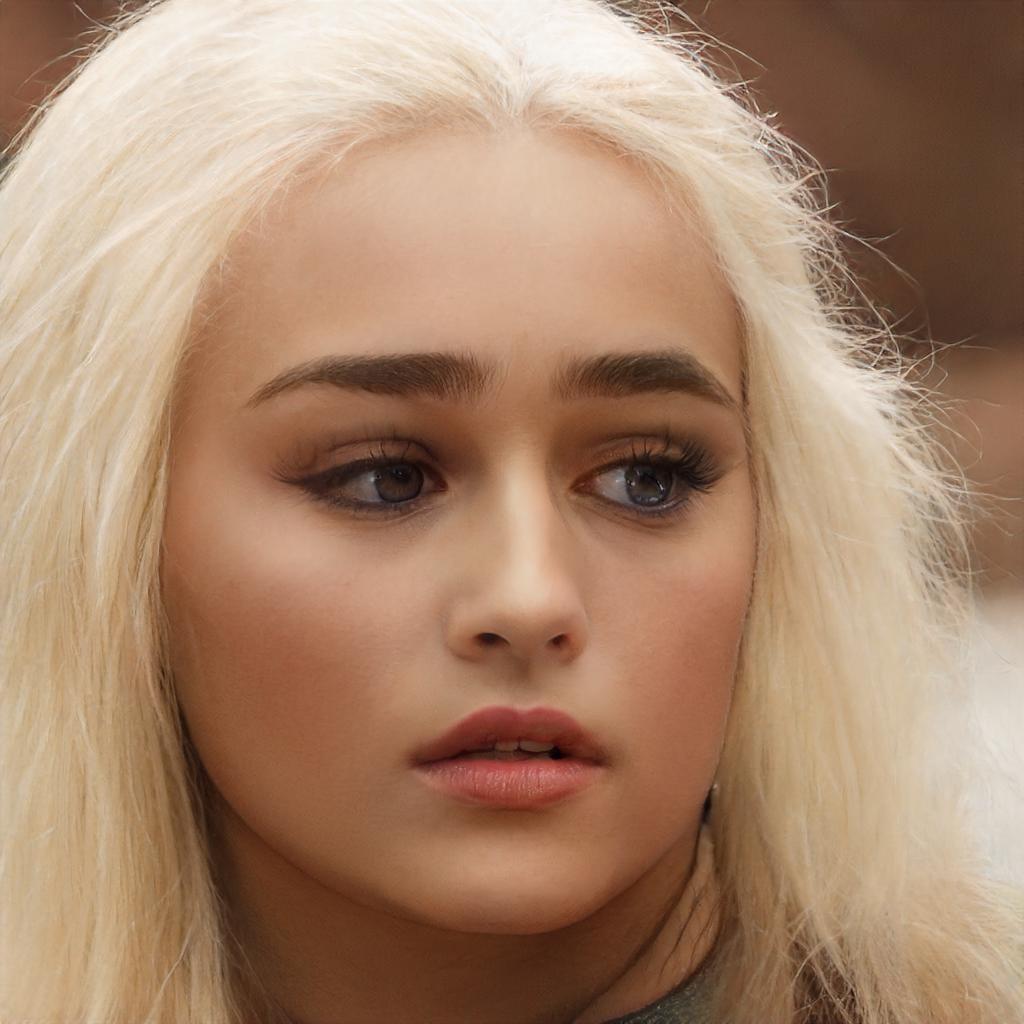} &
            \includegraphics[width=0.135\textwidth]{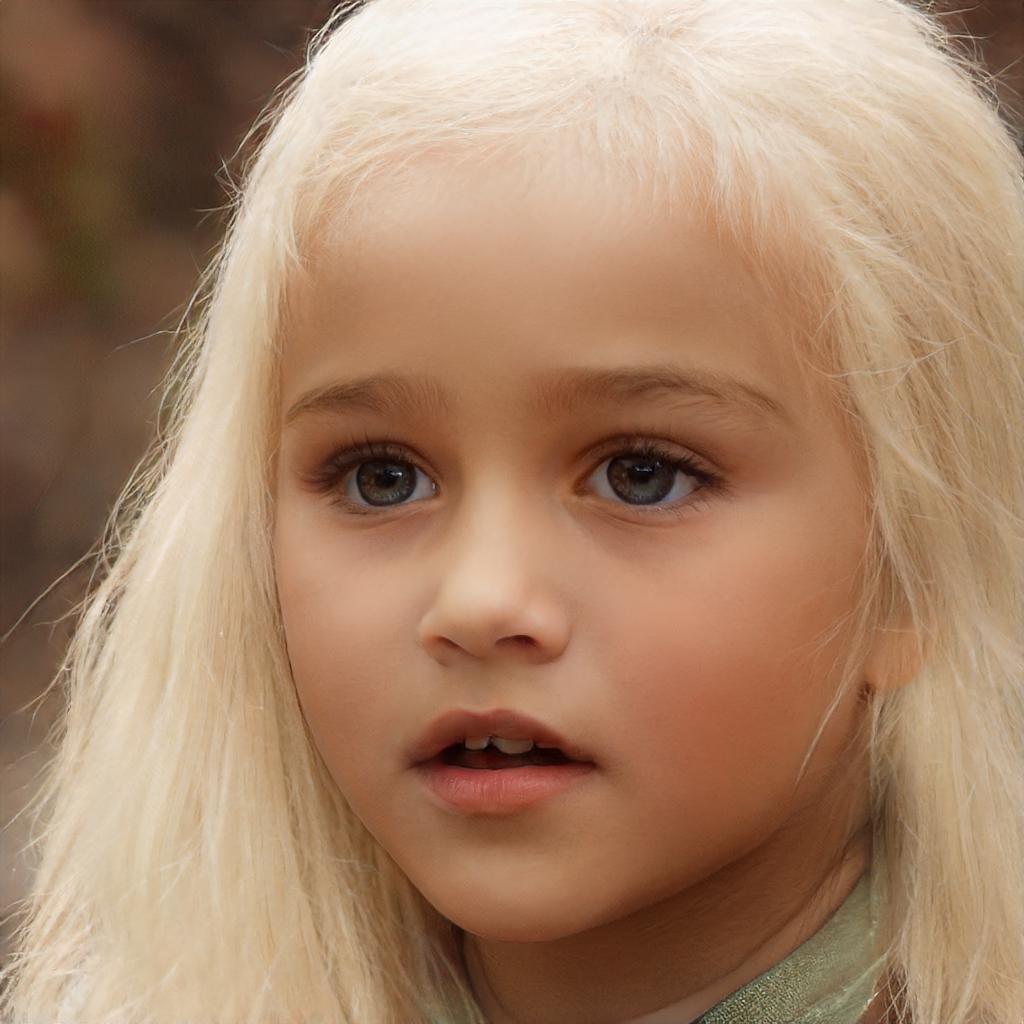} &
            \includegraphics[width=0.135\textwidth]{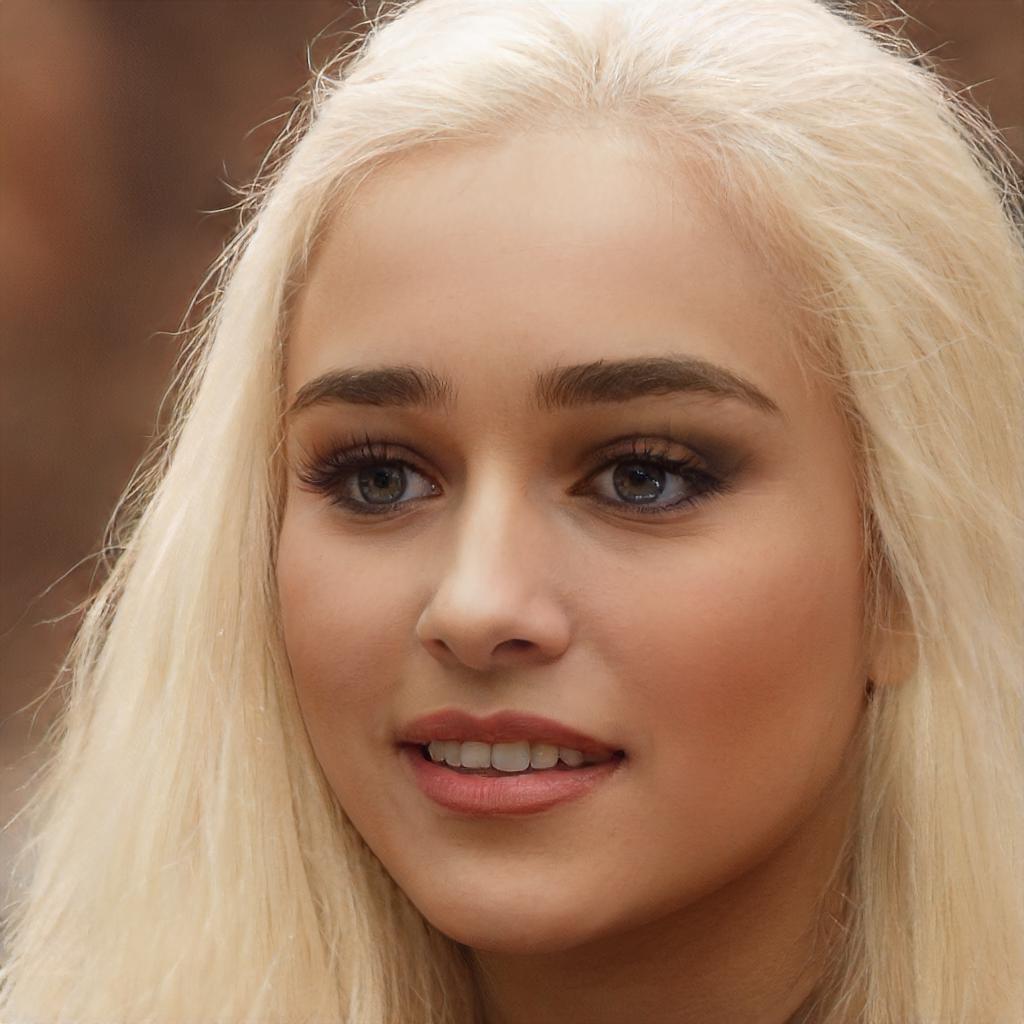} \\
            & Source & Inversion & \multicolumn{4}{c}{\ruleline{0.52\linewidth}{Edits}} \\
            \end{tabular}
    \caption{Additional comparison of configurations \texttt{A} and \texttt{D} trained on FFHQ. Here, results are displayed on real images of celebrities collected from the internet.
    To display the versatile edit capability of configuration \texttt{D}, each pair of rows depicts randomly selected edits performed using StyleFlow \cite{abdal2020styleflow}.}
    \label{fig:internet_celebs_1}
\end{figure*}

\begin{figure*}

    \setlength{\tabcolsep}{1pt}
    \centering
        \centering
            \begin{tabular}{c c c c c c c}

            & Source & Inversion & \multicolumn{4}{c}{\ruleline{0.52\linewidth}{Edits}} \\
            
            \raisebox{0.06\textwidth}{\texttt{A}} & \includegraphics[width=0.135\textwidth]{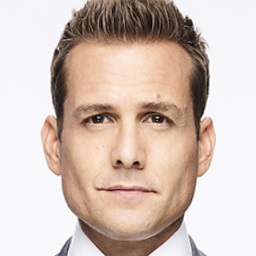} & 
            \includegraphics[width=0.135\textwidth]{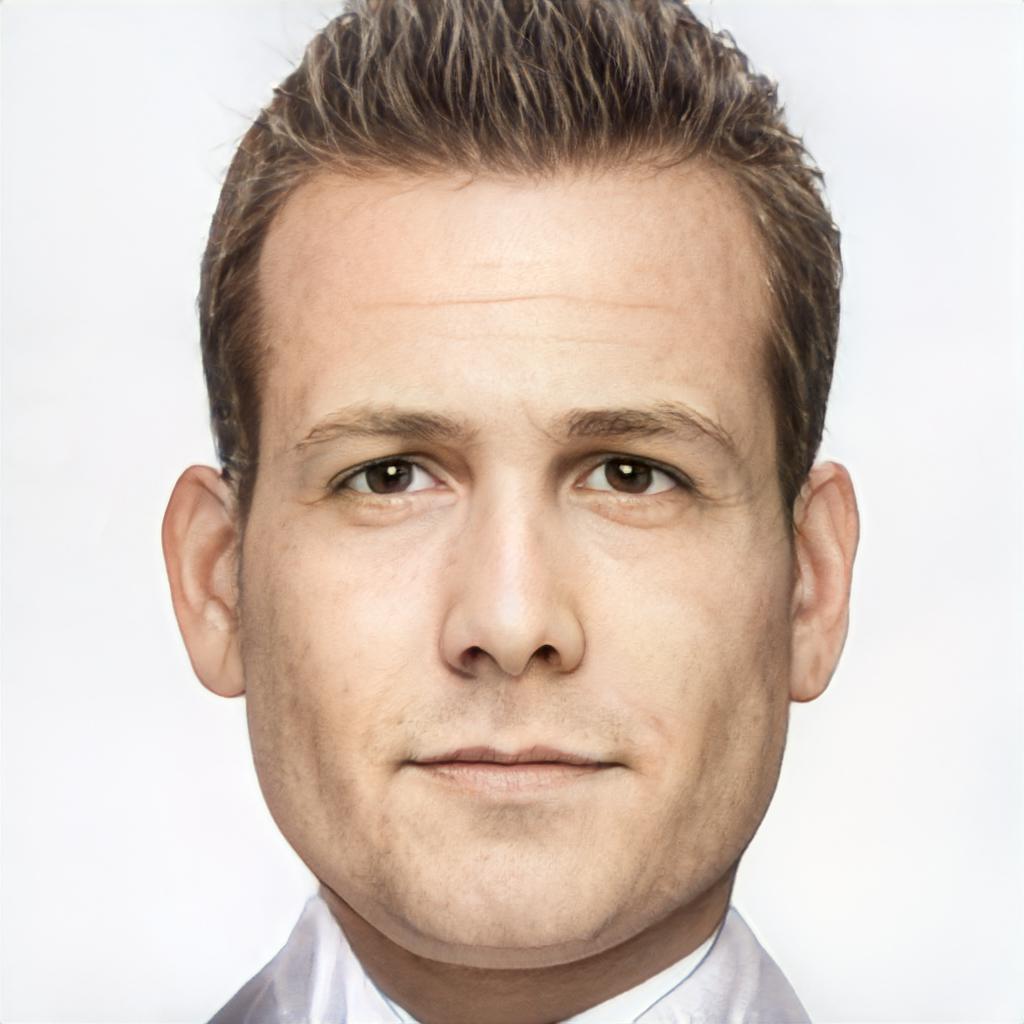} &
            \includegraphics[width=0.135\textwidth]{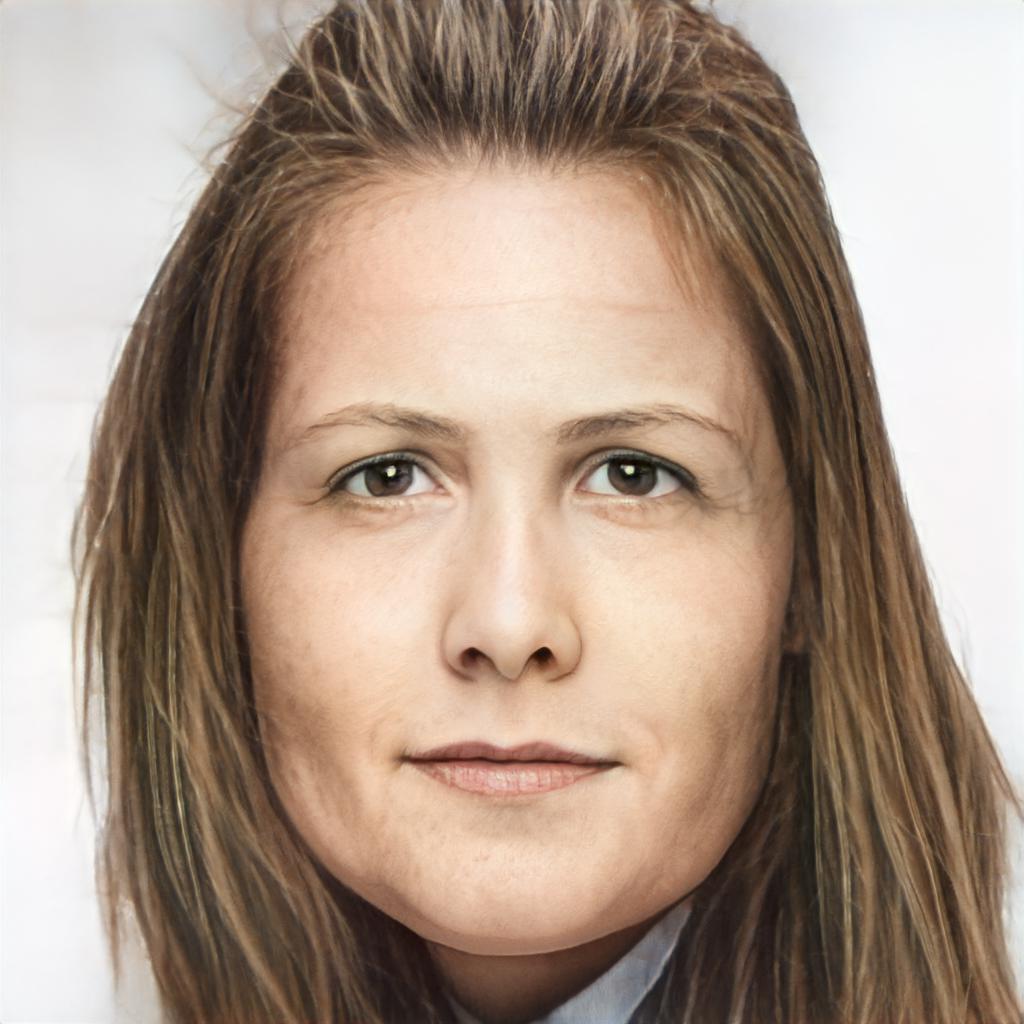} &
            \includegraphics[width=0.135\textwidth]{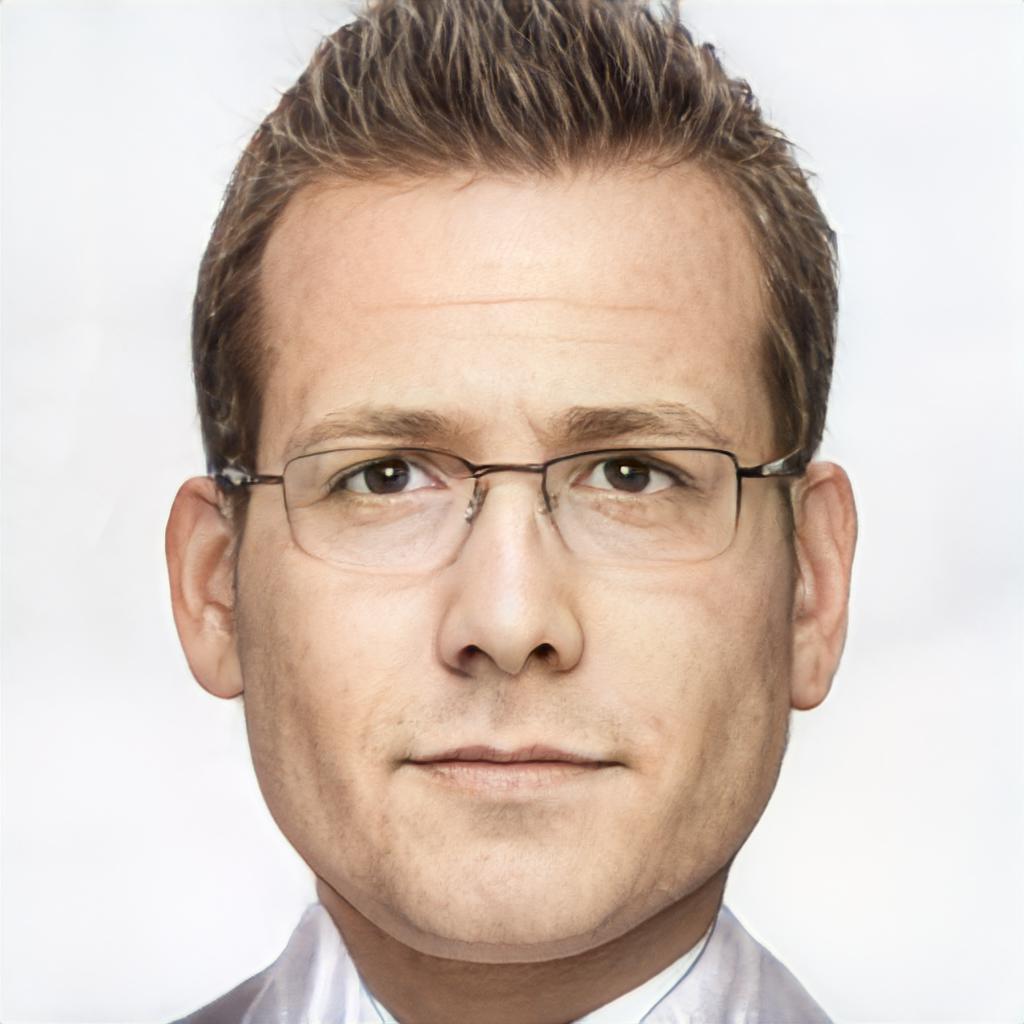} &
            \includegraphics[width=0.135\textwidth]{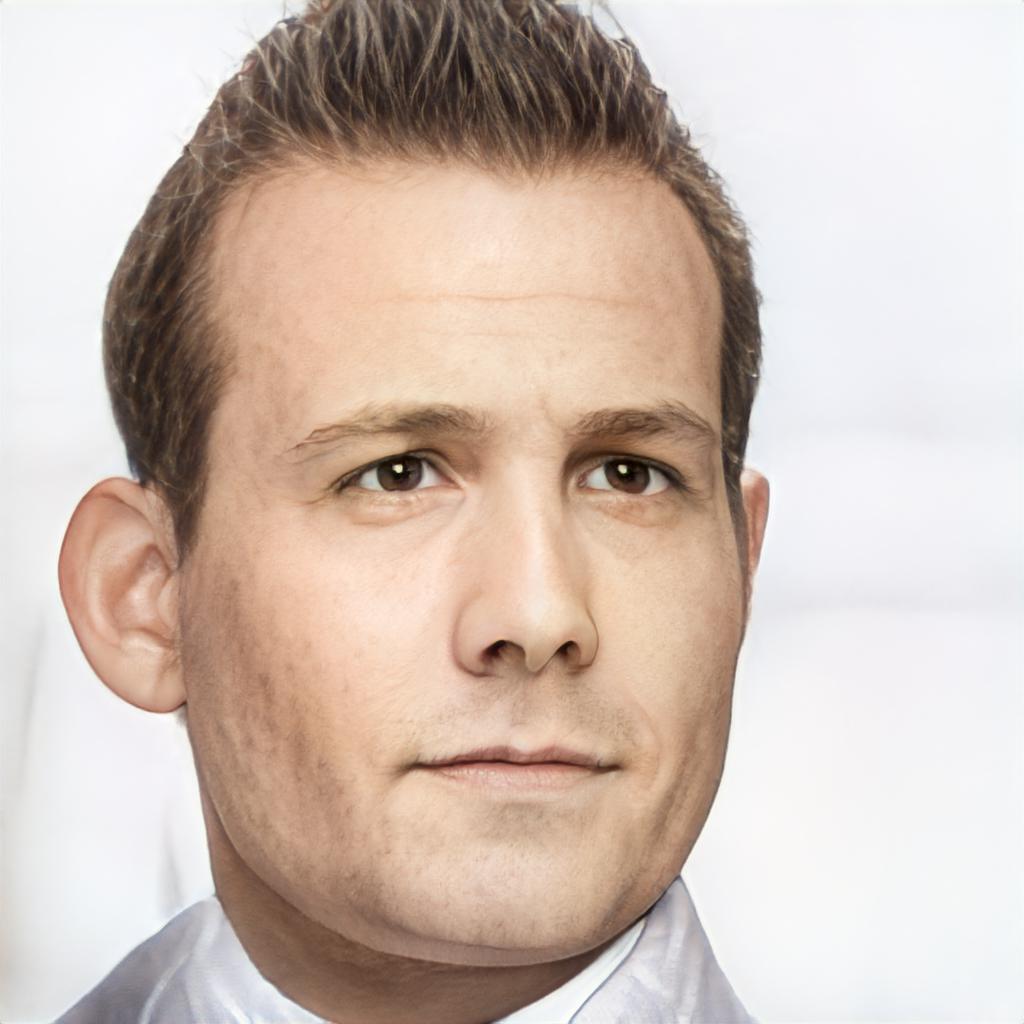} &
            \includegraphics[width=0.135\textwidth]{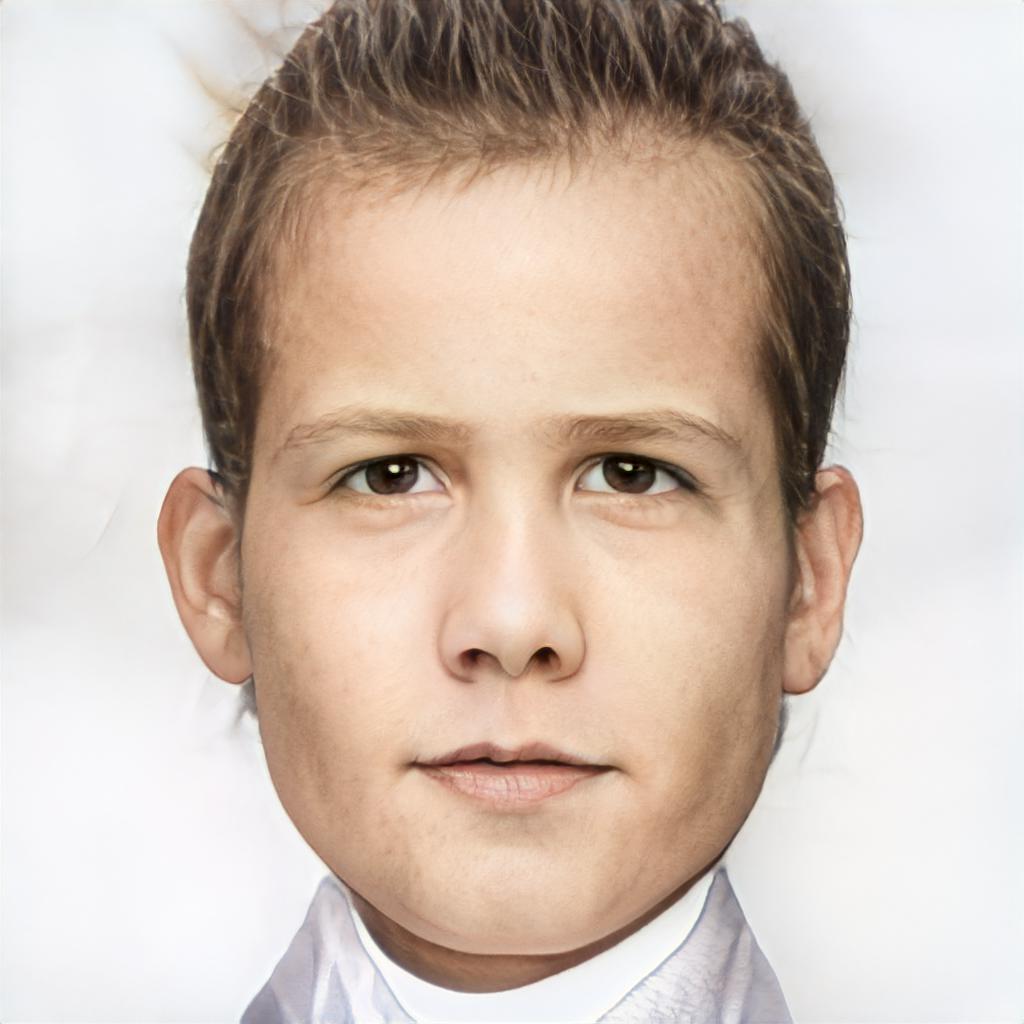} \\
            \raisebox{0.06\textwidth}{\texttt{D}} & \includegraphics[width=0.135\textwidth]{images/appendix/styleflow_edit_celebs_ours/harvey_src.jpg} &
            \includegraphics[width=0.135\textwidth]{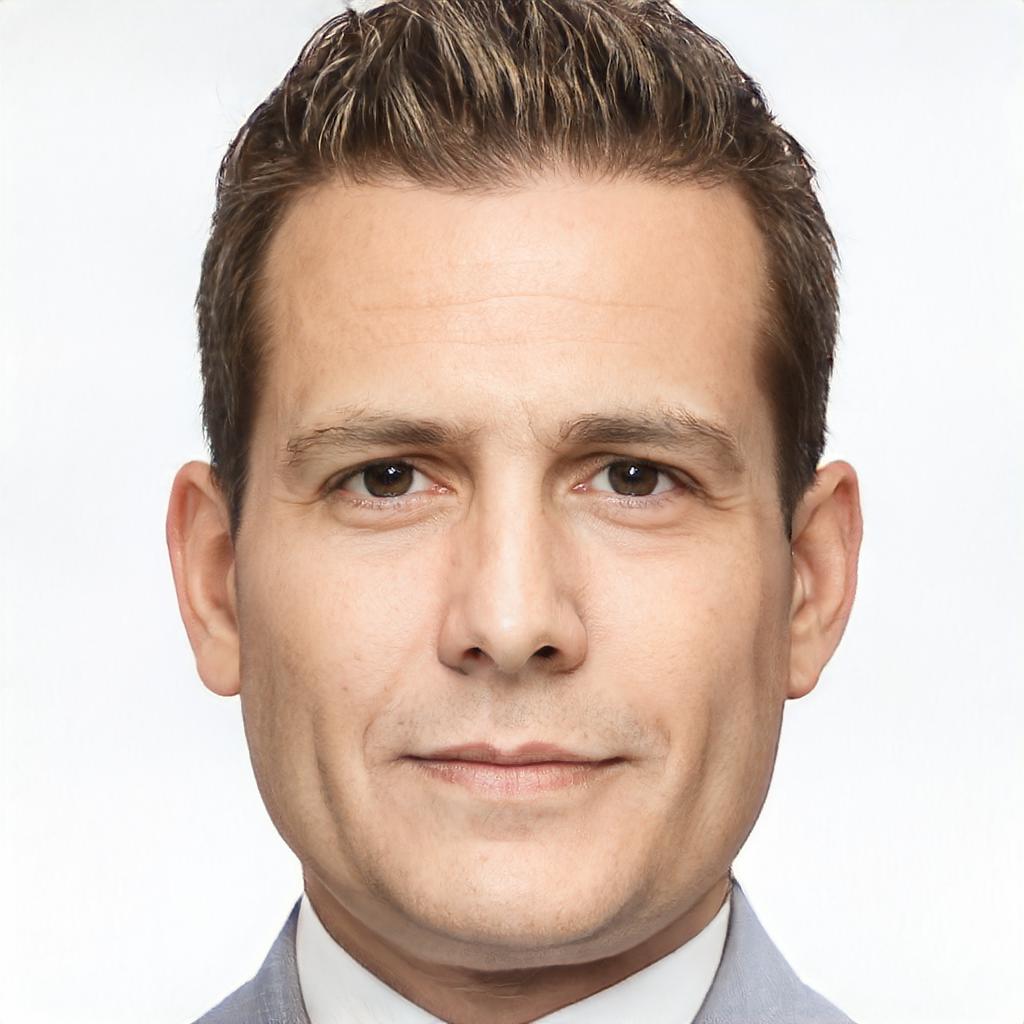} &
            \includegraphics[width=0.135\textwidth]{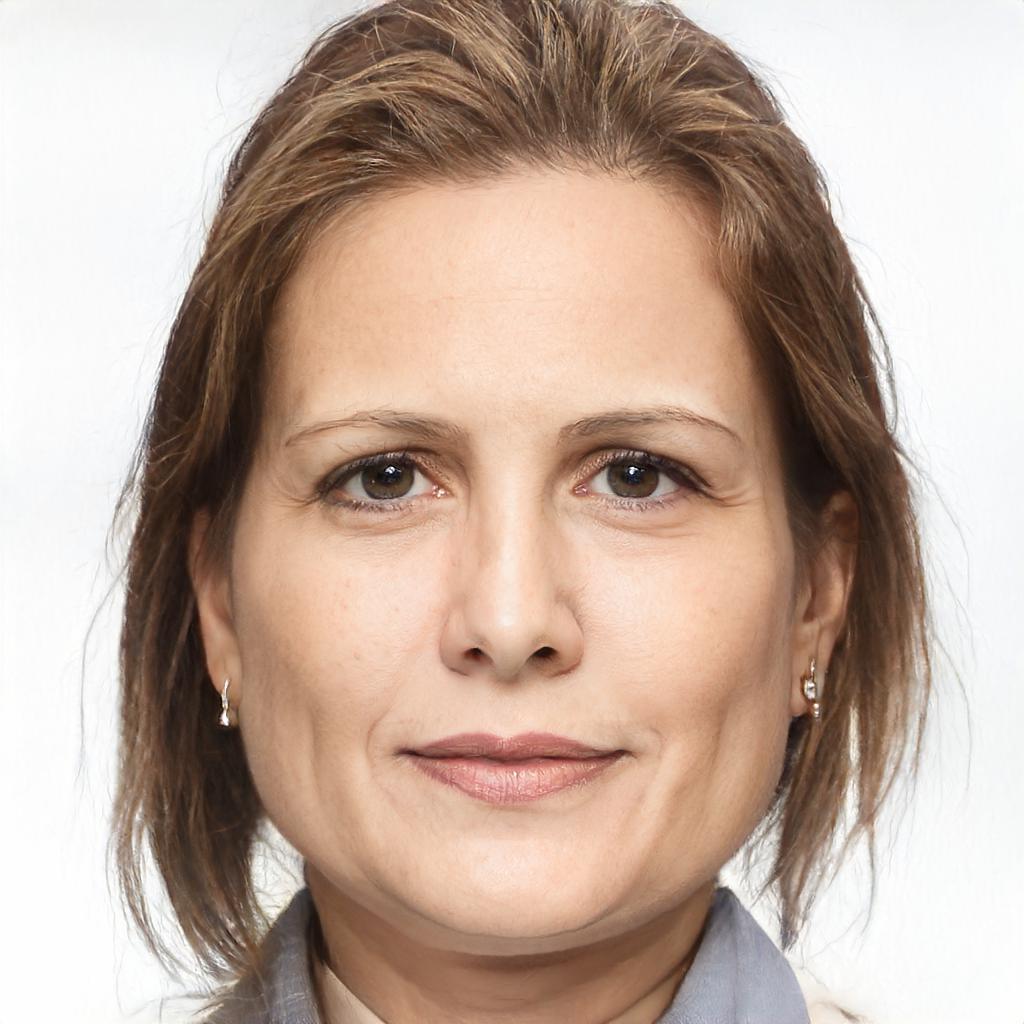} &
            \includegraphics[width=0.135\textwidth]{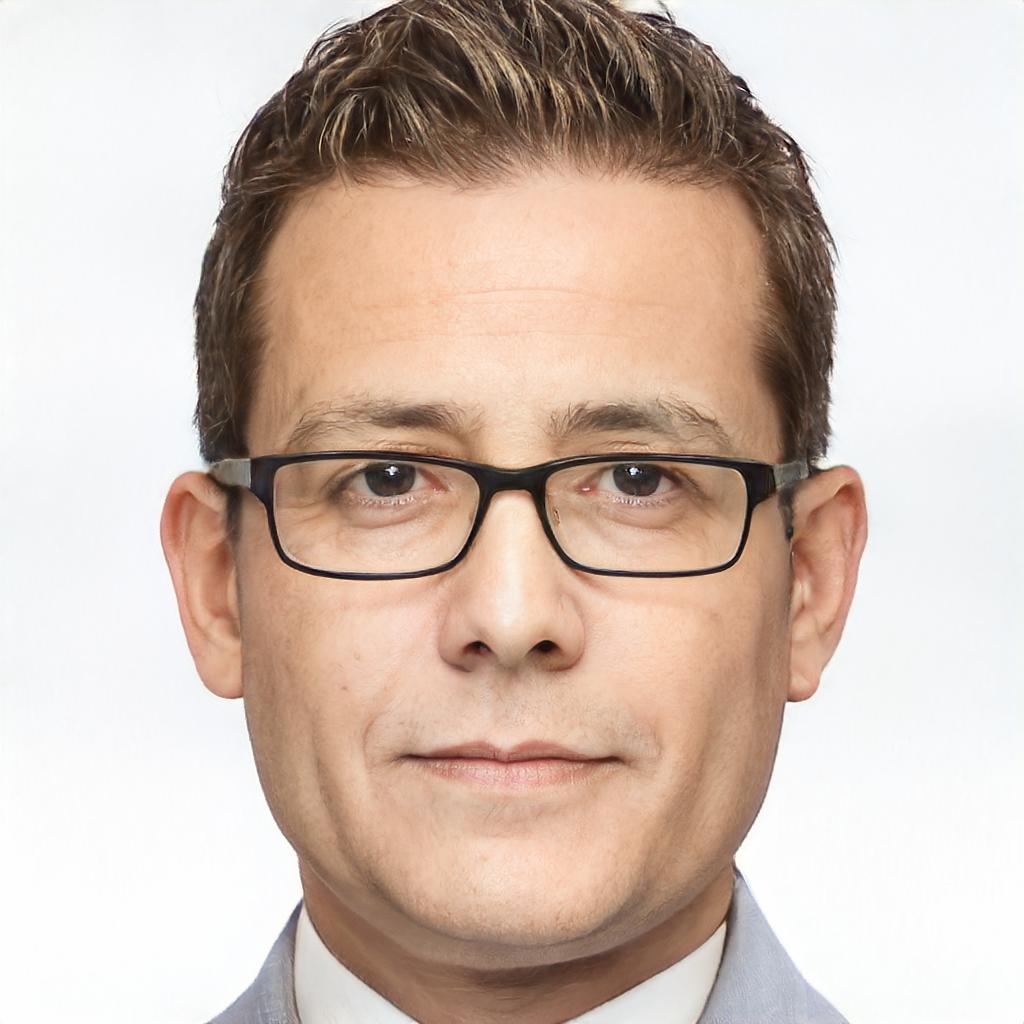} &
            \includegraphics[width=0.135\textwidth]{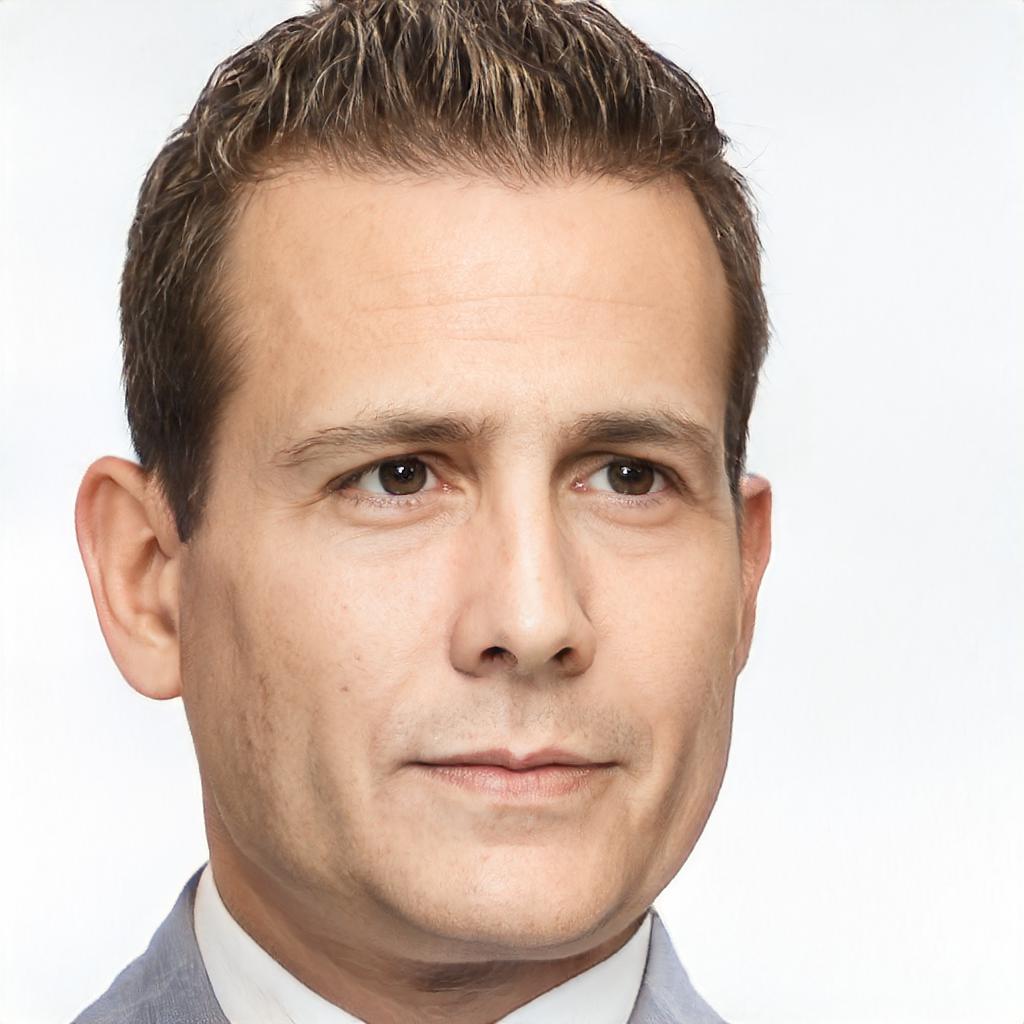} &
            \includegraphics[width=0.135\textwidth]{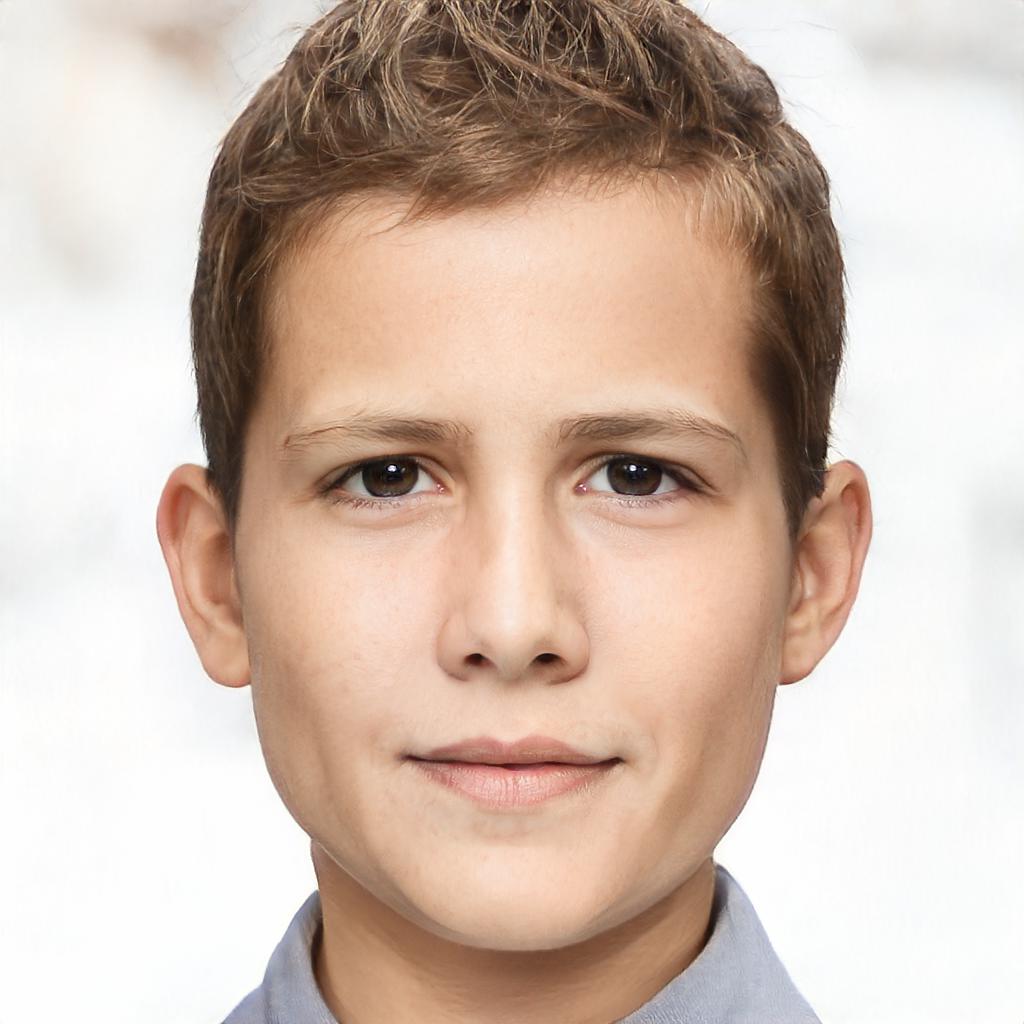} \\
            \raisebox{0.06\textwidth}{\texttt{A}} & \includegraphics[width=0.135\textwidth]{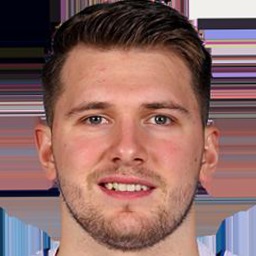} & 
            \includegraphics[width=0.135\textwidth]{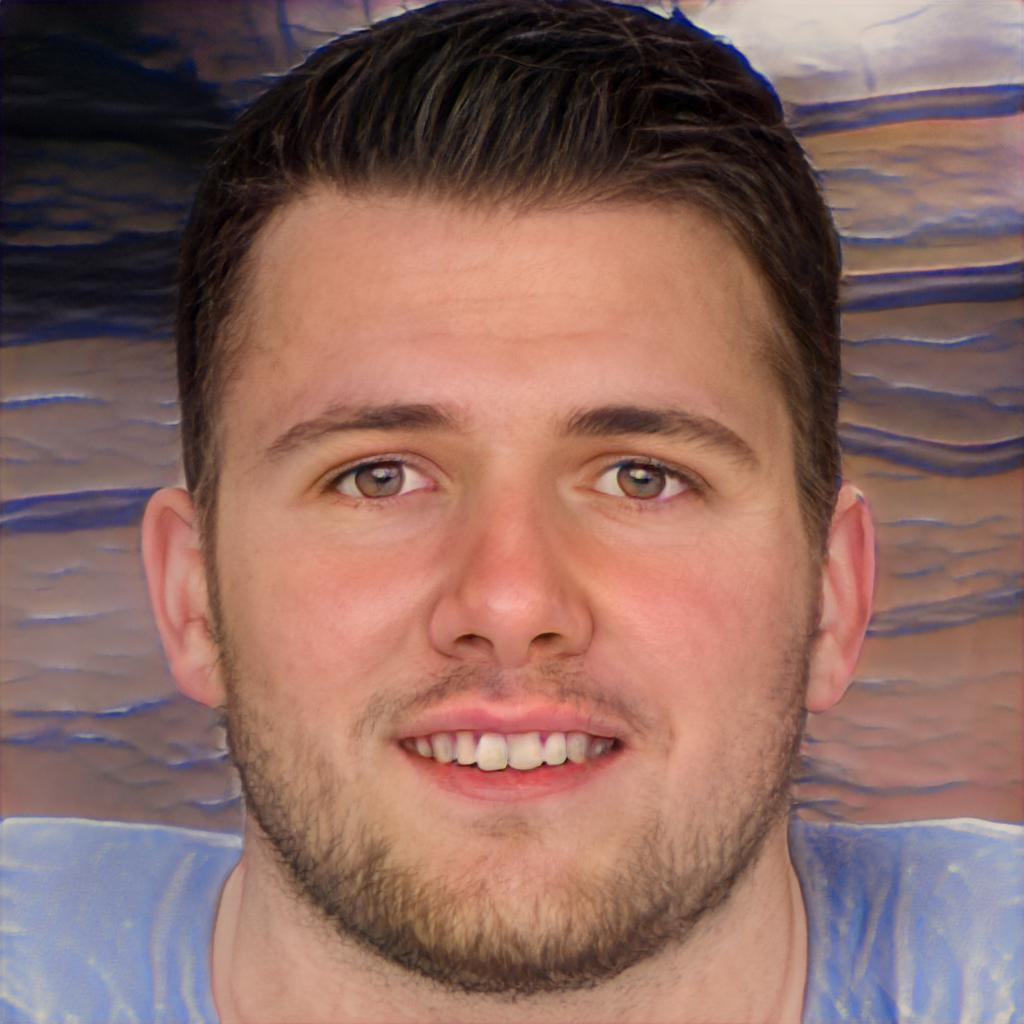} &
            \includegraphics[width=0.135\textwidth]{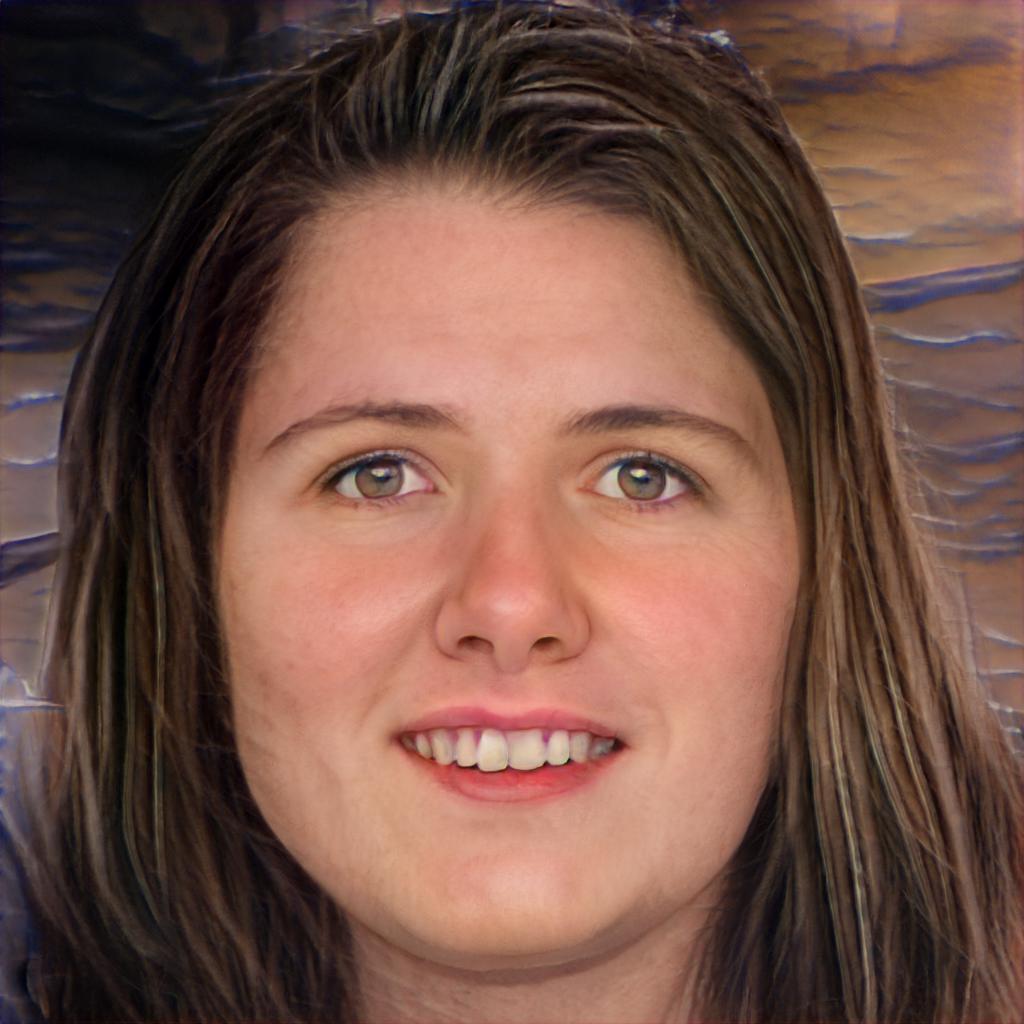} &
            \includegraphics[width=0.135\textwidth]{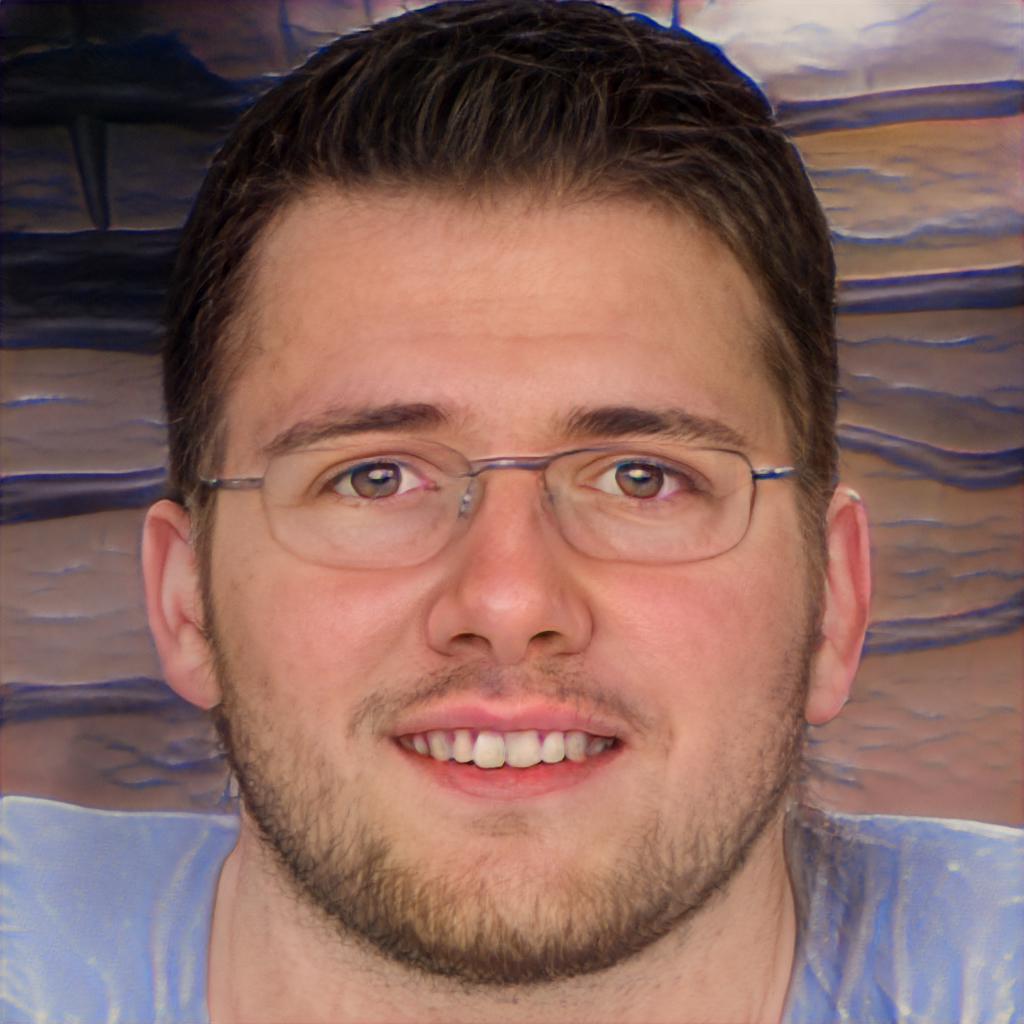} &
            \includegraphics[width=0.135\textwidth]{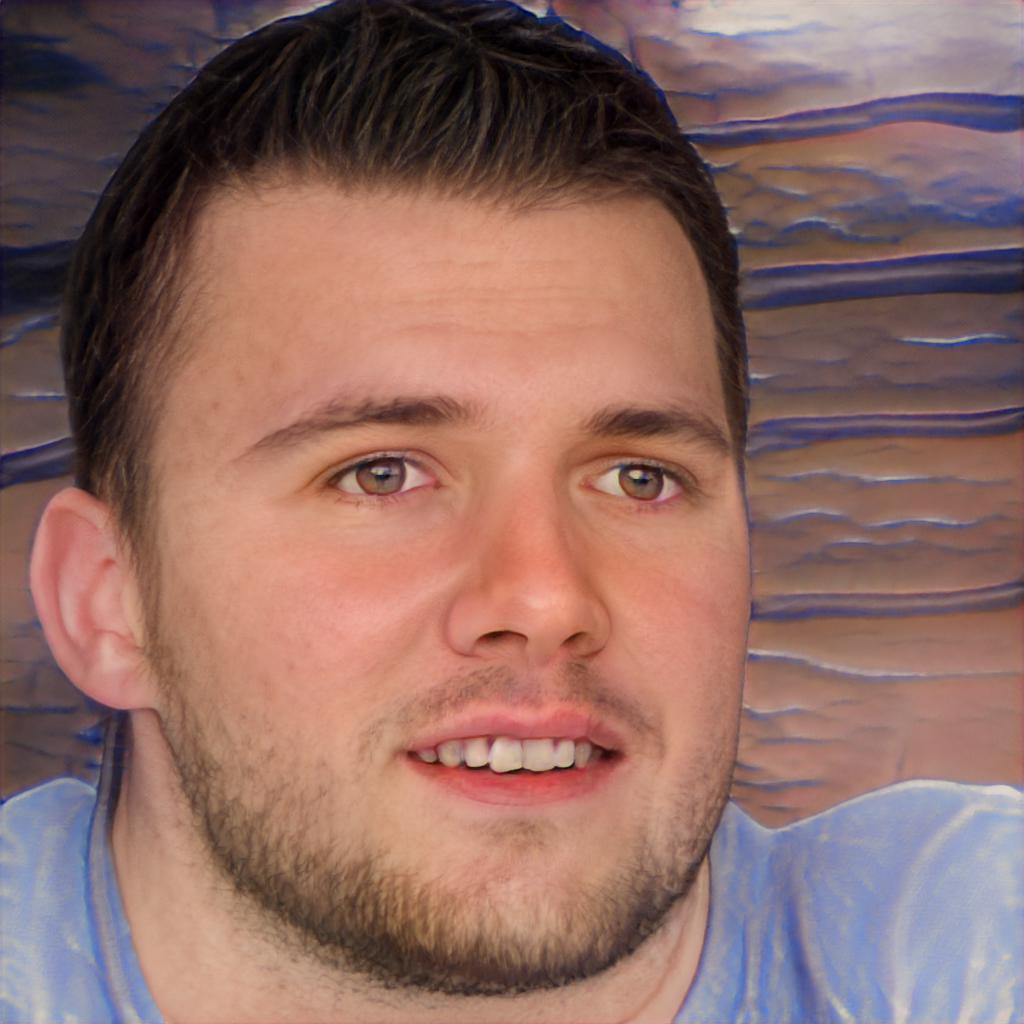} &
            \includegraphics[width=0.135\textwidth]{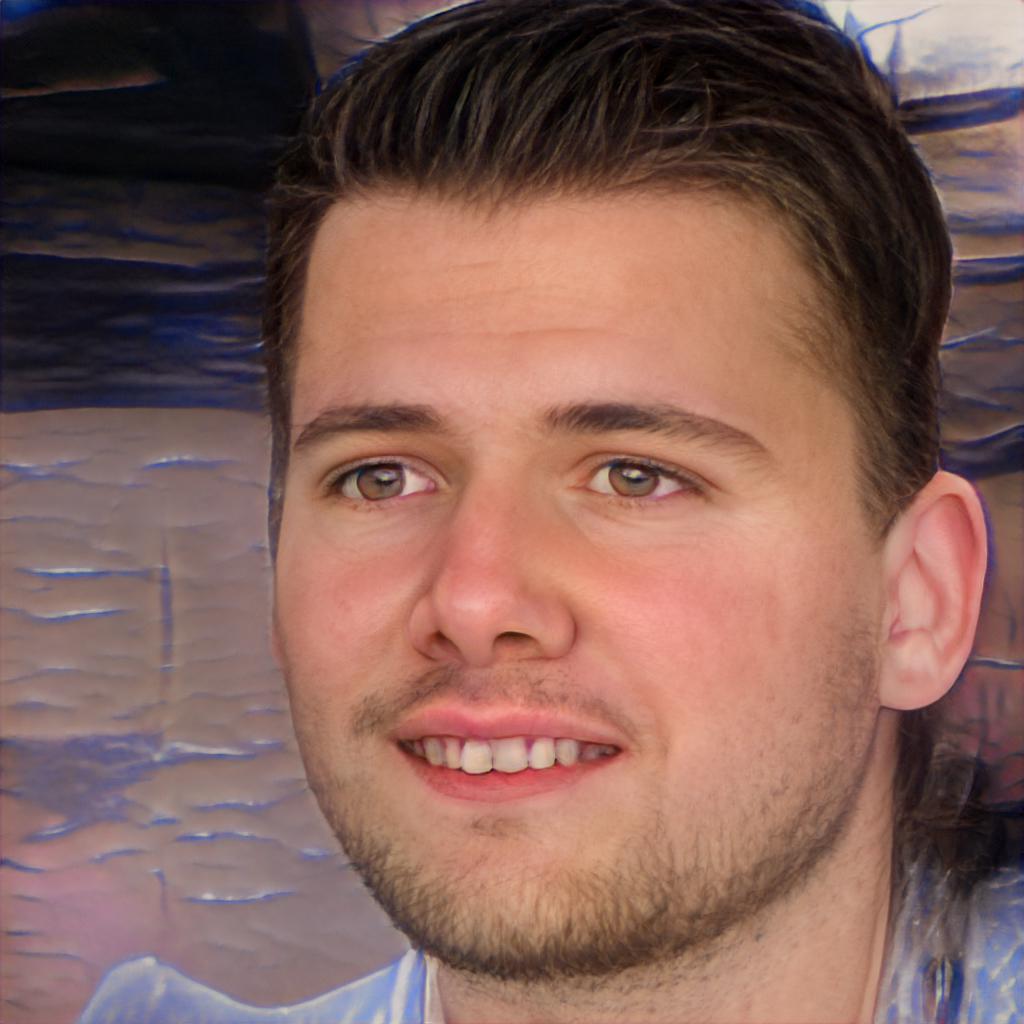} \\
            \raisebox{0.06\textwidth}{\texttt{D}} & \includegraphics[width=0.135\textwidth]{images/appendix/styleflow_edit_celebs_ours/luka_src.jpg} &
            \includegraphics[width=0.135\textwidth]{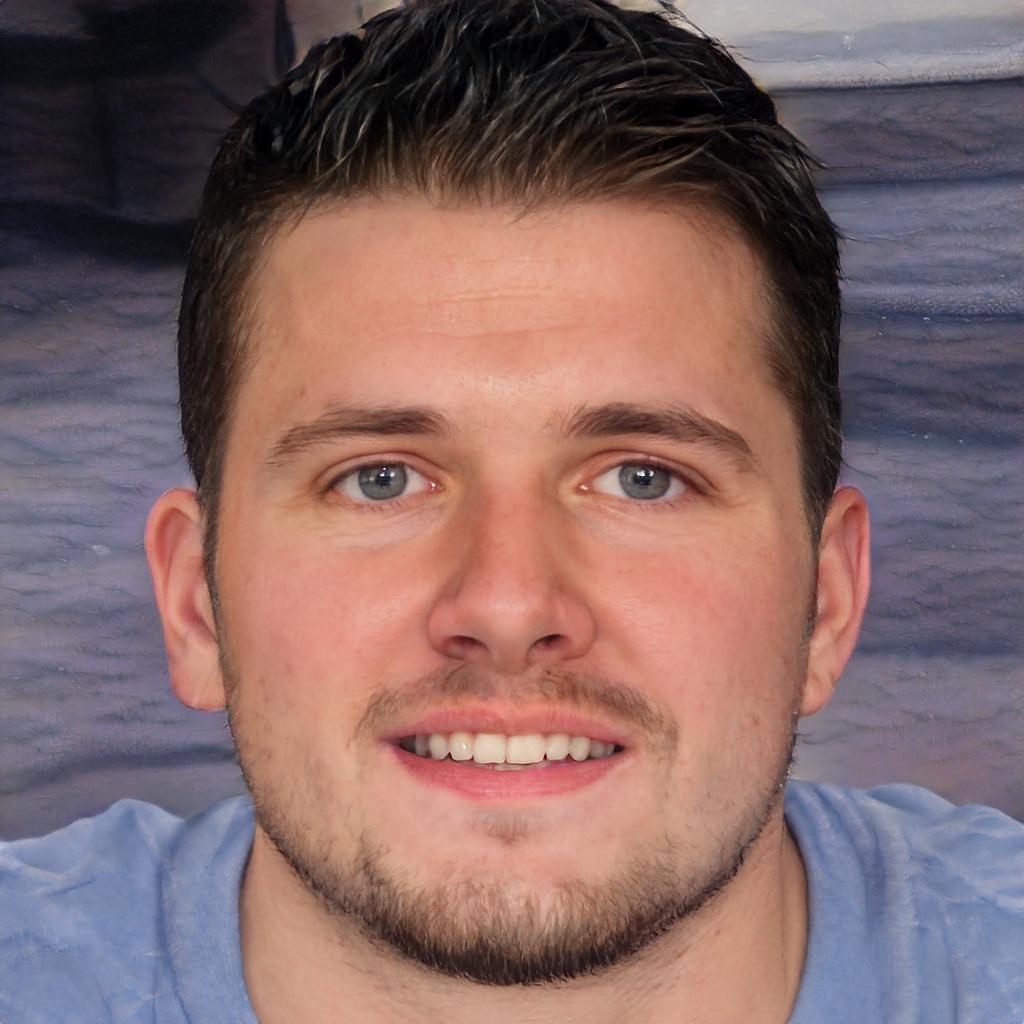} &
            \includegraphics[width=0.135\textwidth]{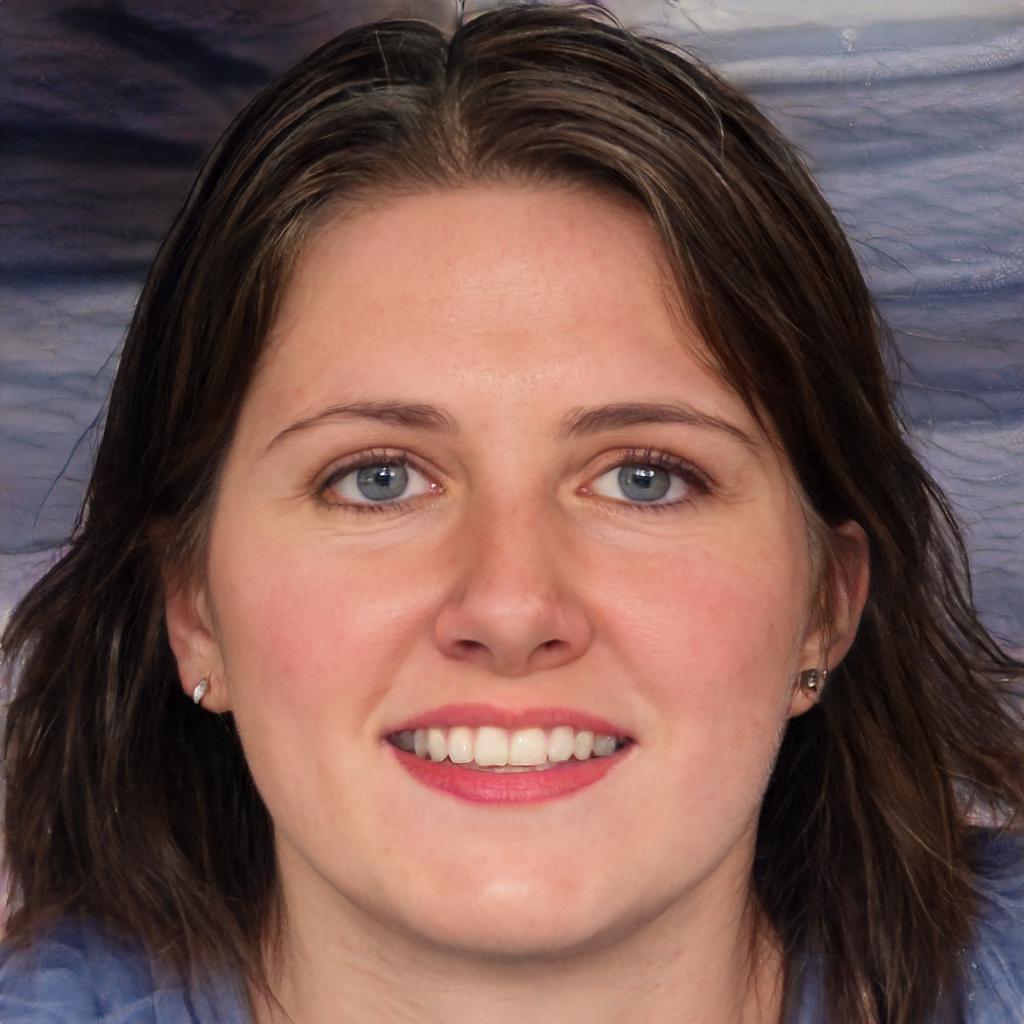} &
            \includegraphics[width=0.135\textwidth]{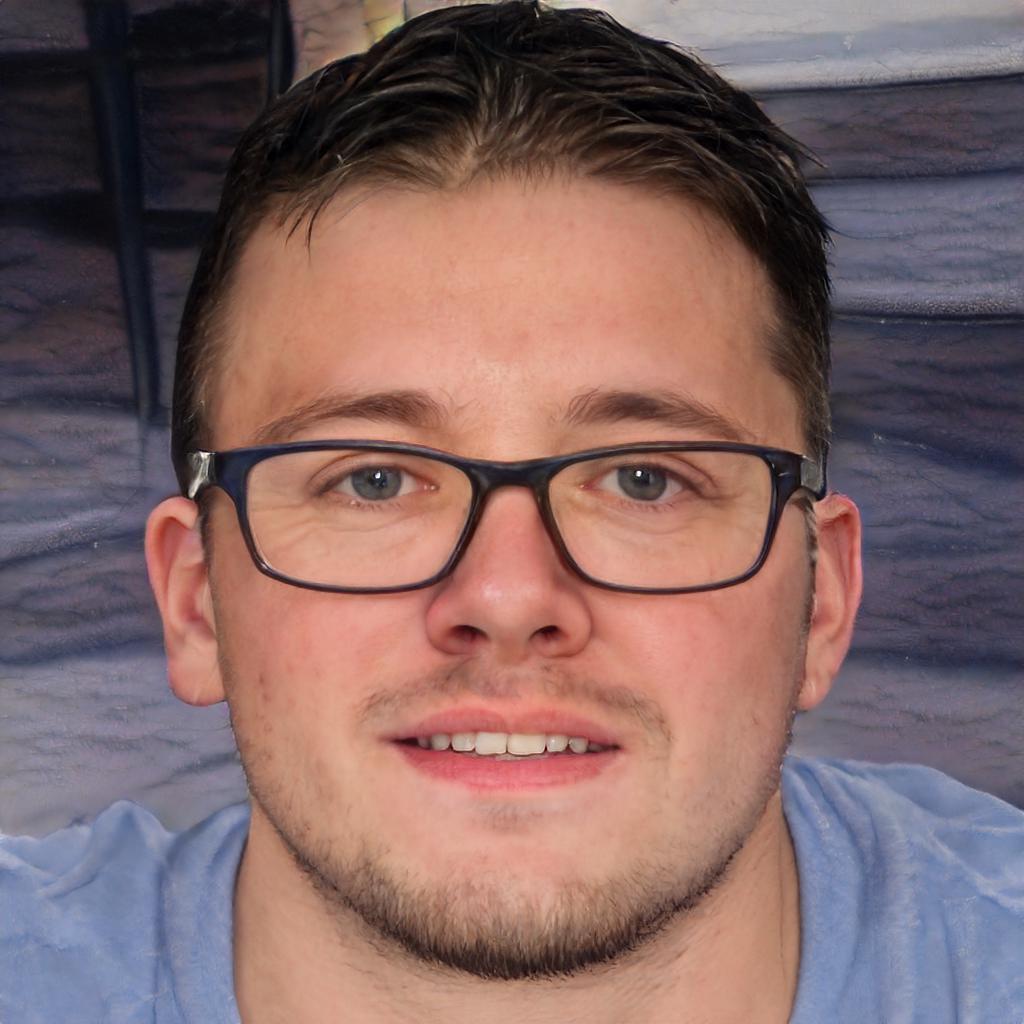} &
            \includegraphics[width=0.135\textwidth]{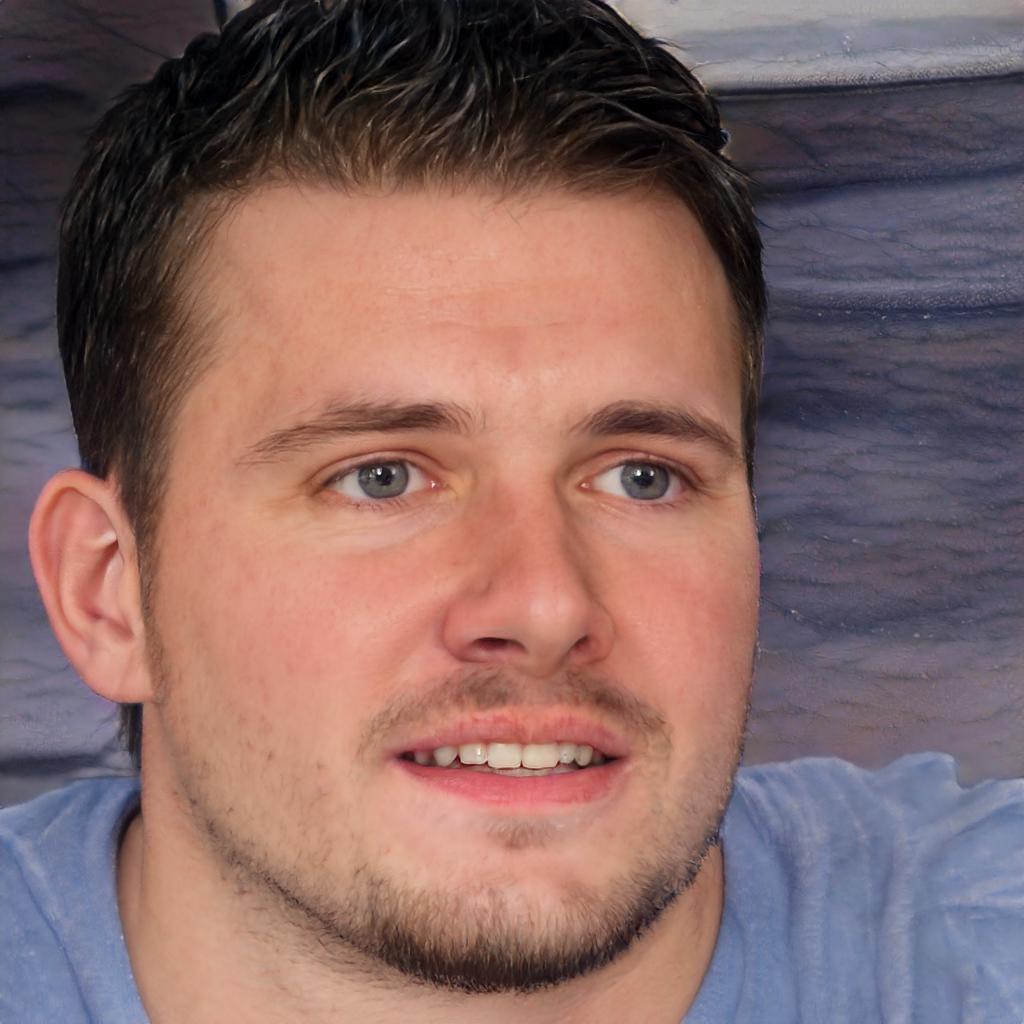} &
            \includegraphics[width=0.135\textwidth]{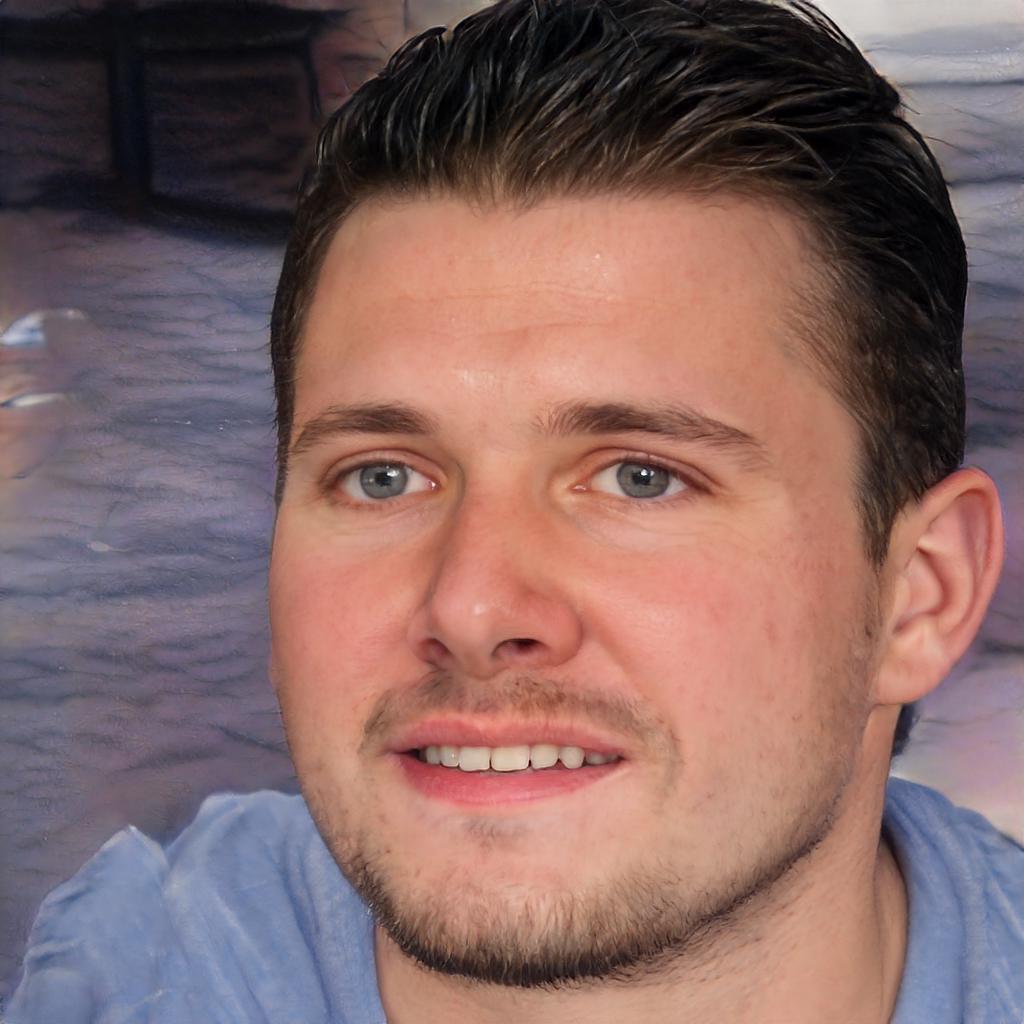} \\
            \raisebox{0.06\textwidth}{\texttt{A}} & \includegraphics[width=0.135\textwidth]{images/optimization/messi/messi5_src.jpg} & 
            \includegraphics[width=0.135\textwidth]{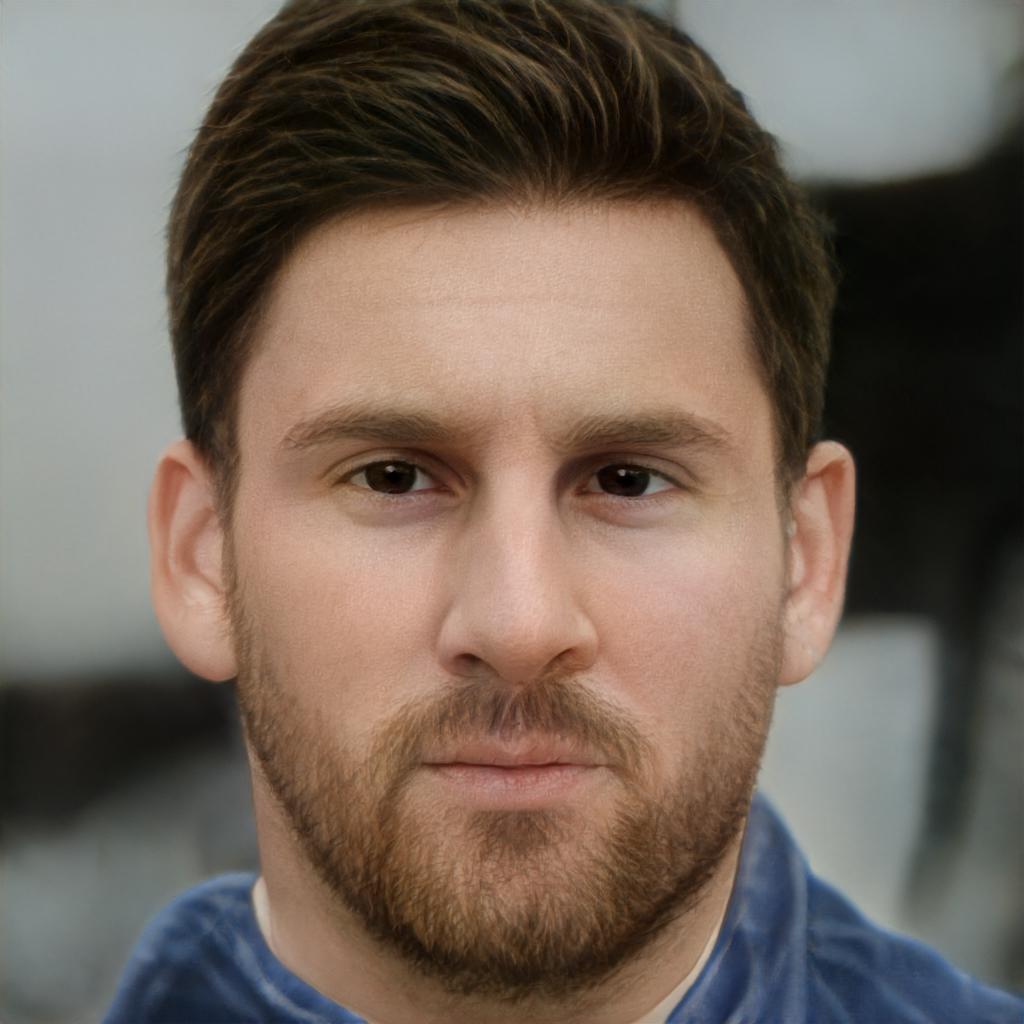} &
            \includegraphics[width=0.135\textwidth]{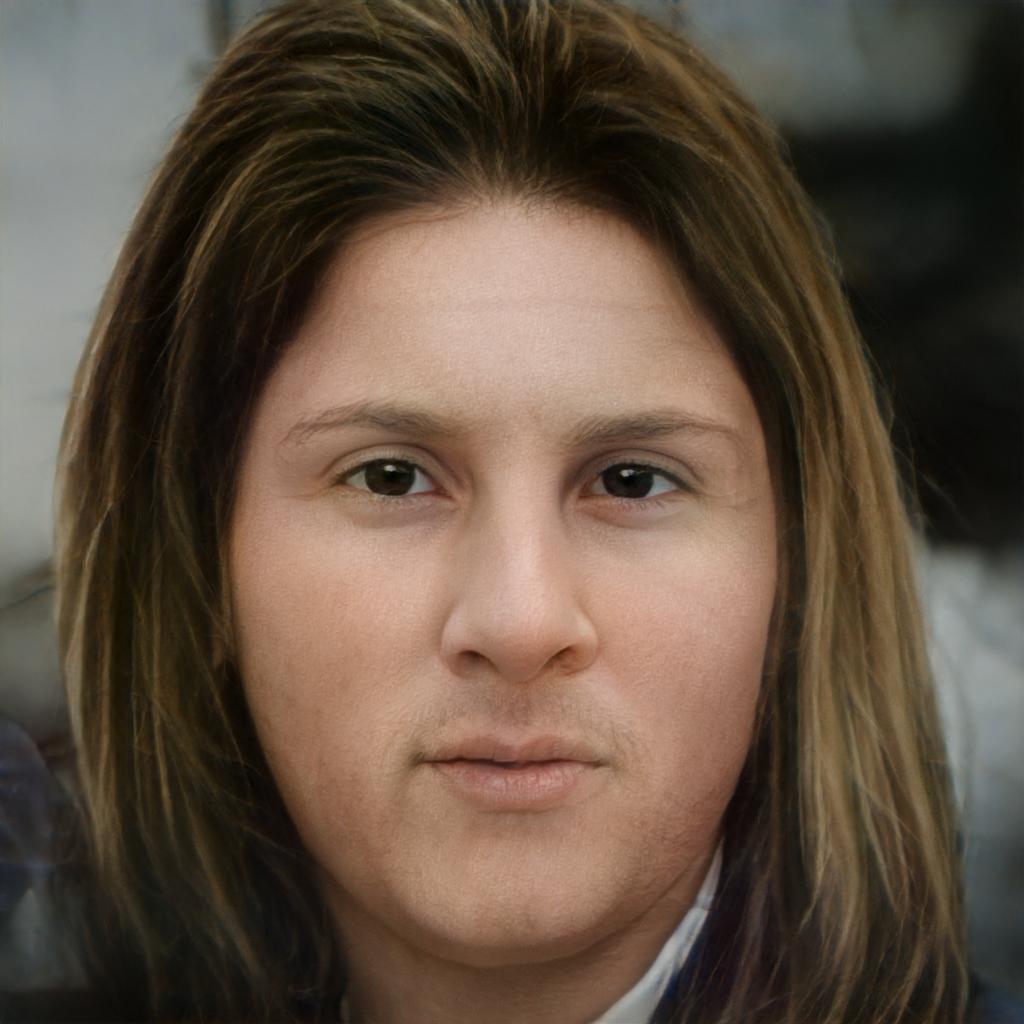} &
            \includegraphics[width=0.135\textwidth]{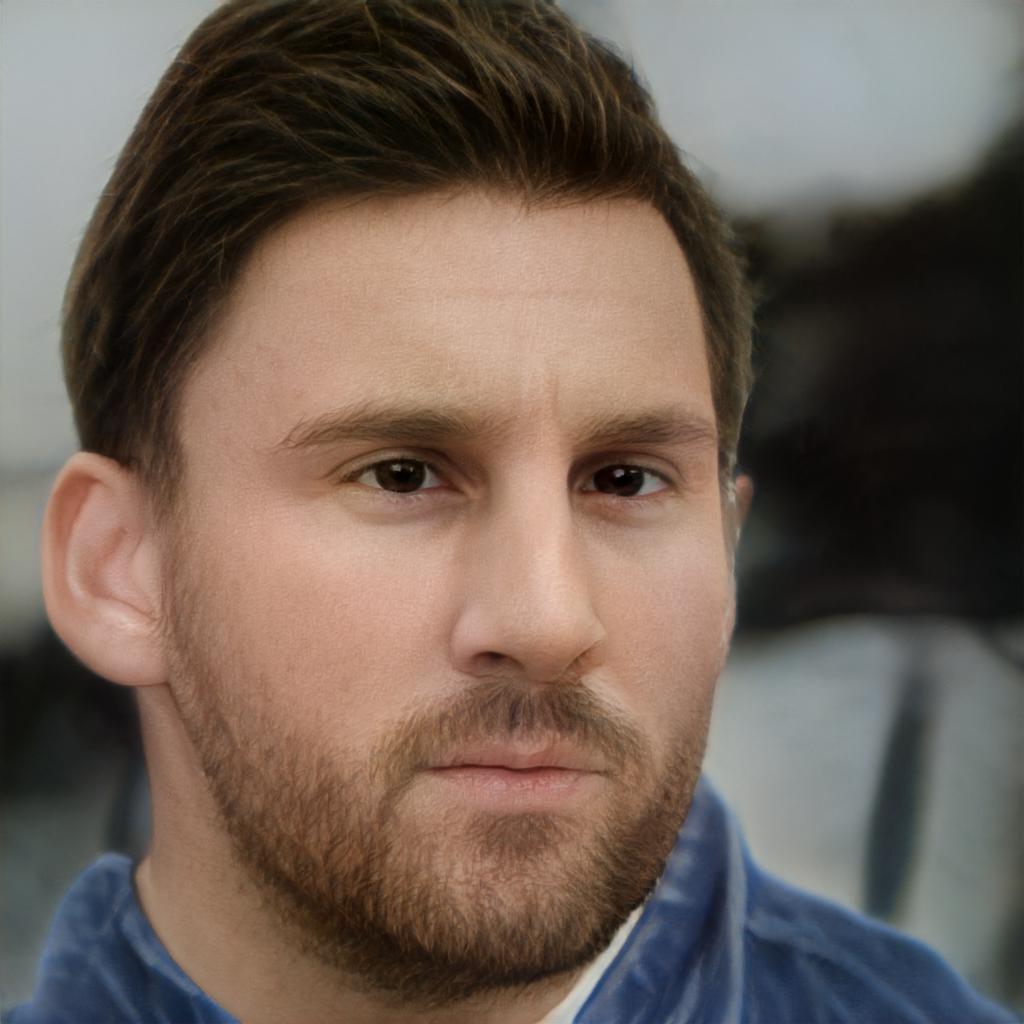} &
            \includegraphics[width=0.135\textwidth]{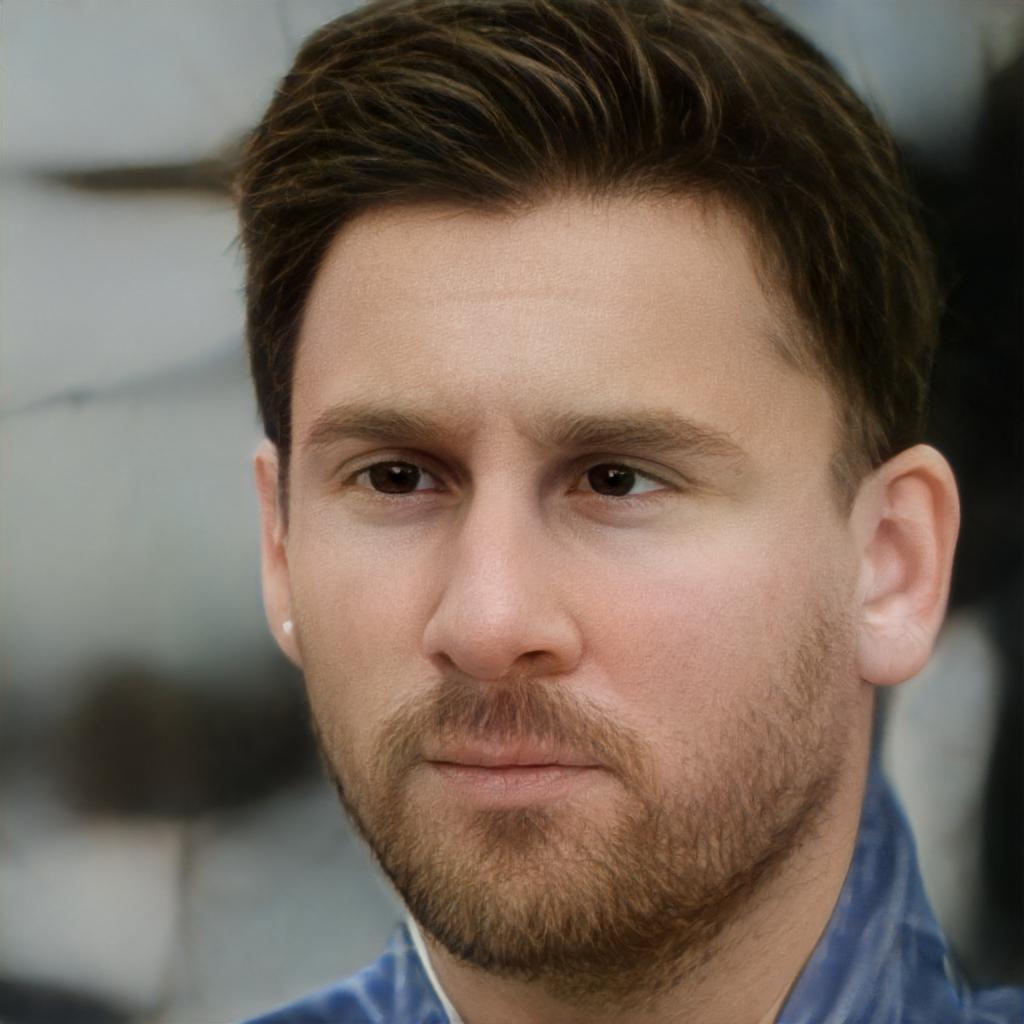} &
            \includegraphics[width=0.135\textwidth]{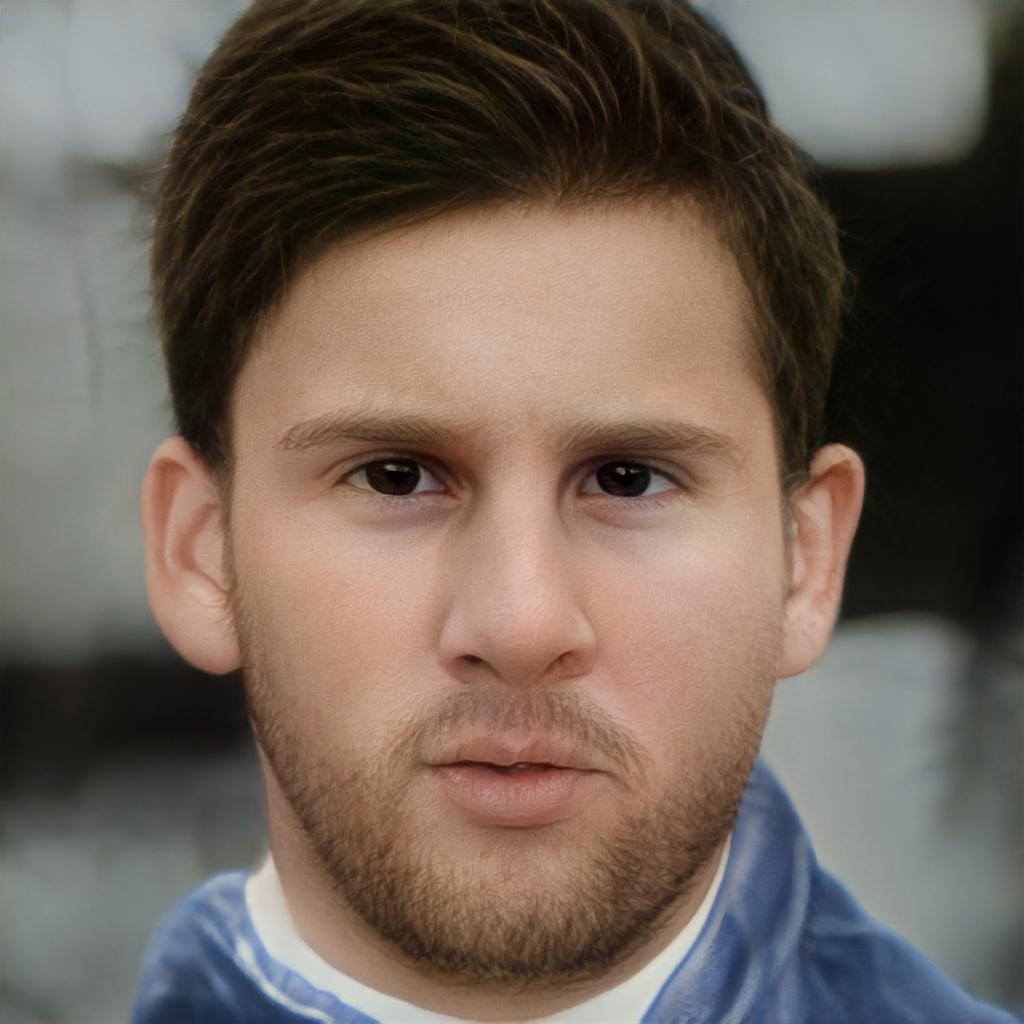} \\
            \raisebox{0.06\textwidth}{\texttt{D}} & \includegraphics[width=0.135\textwidth]{images/optimization/messi/messi5_src.jpg} & 
            \includegraphics[width=0.135\textwidth]{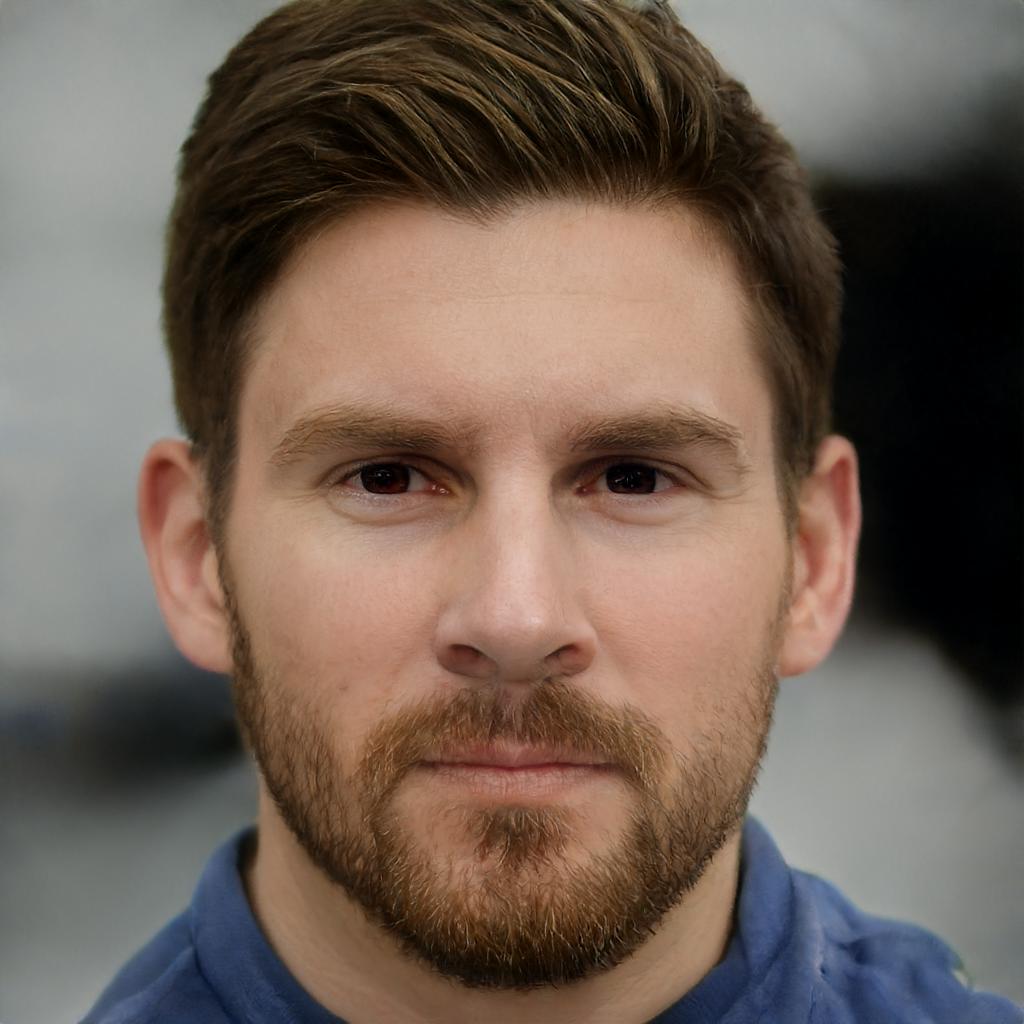} &
            \includegraphics[width=0.135\textwidth]{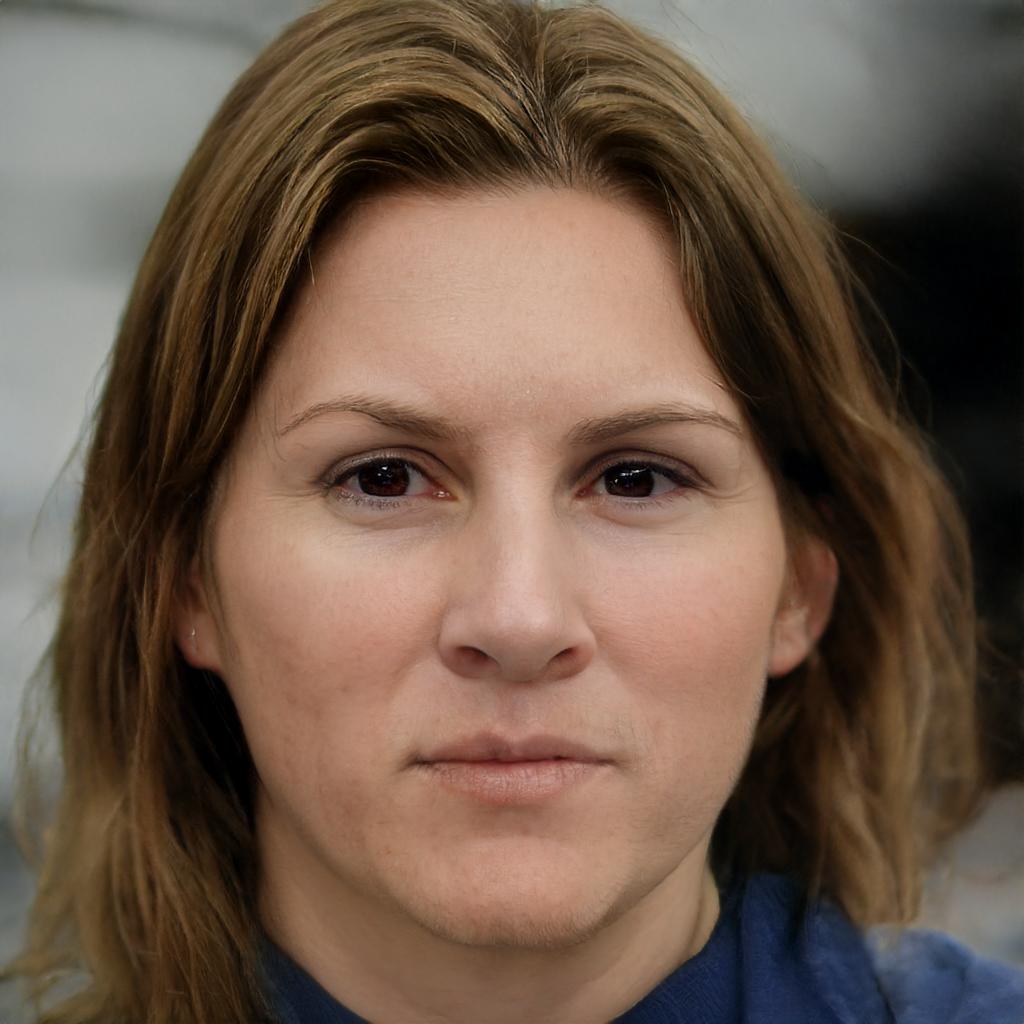} &
            \includegraphics[width=0.135\textwidth]{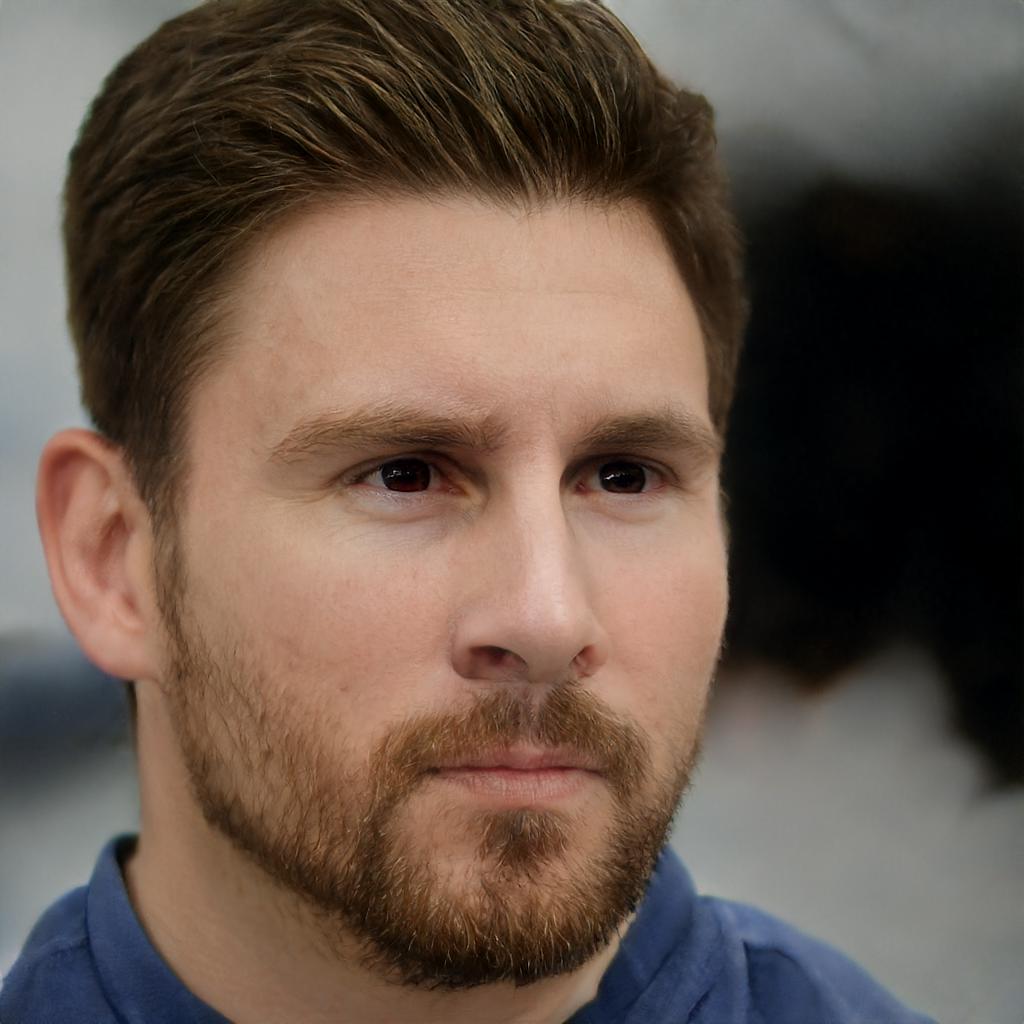} &
            \includegraphics[width=0.135\textwidth]{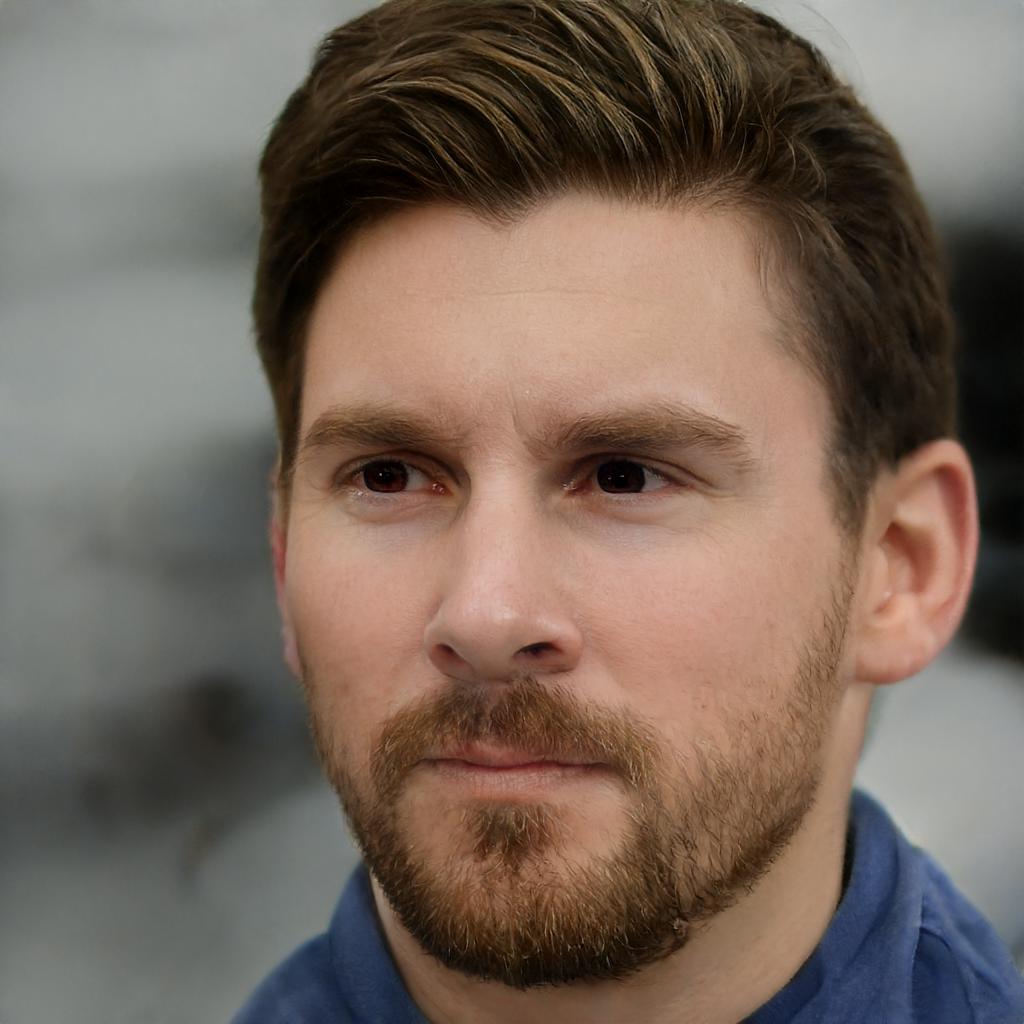} &
            \includegraphics[width=0.135\textwidth]{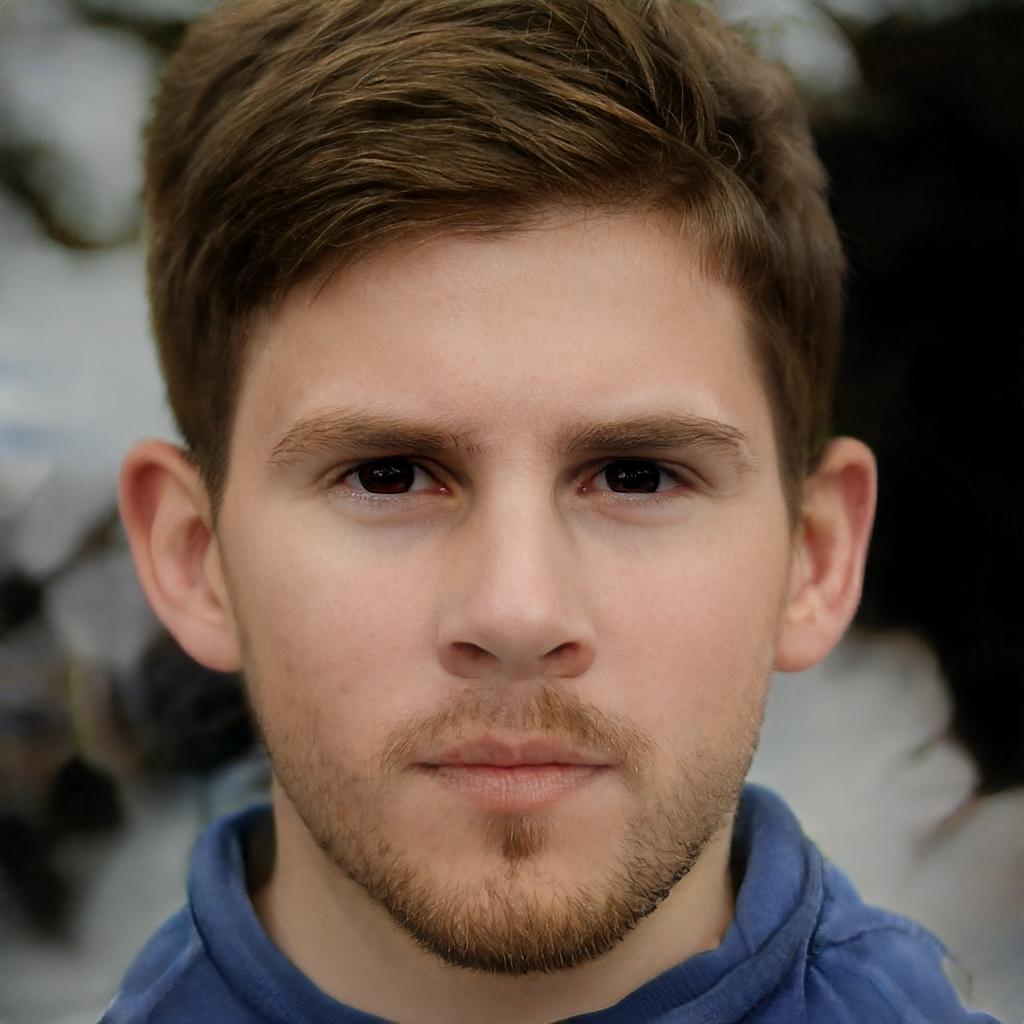} \\
            \raisebox{0.06\textwidth}{\texttt{A}} & \includegraphics[width=0.135\textwidth]{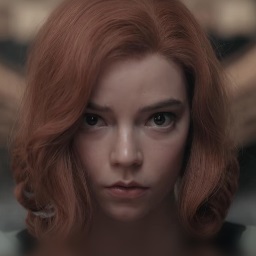} & 
            \includegraphics[width=0.135\textwidth]{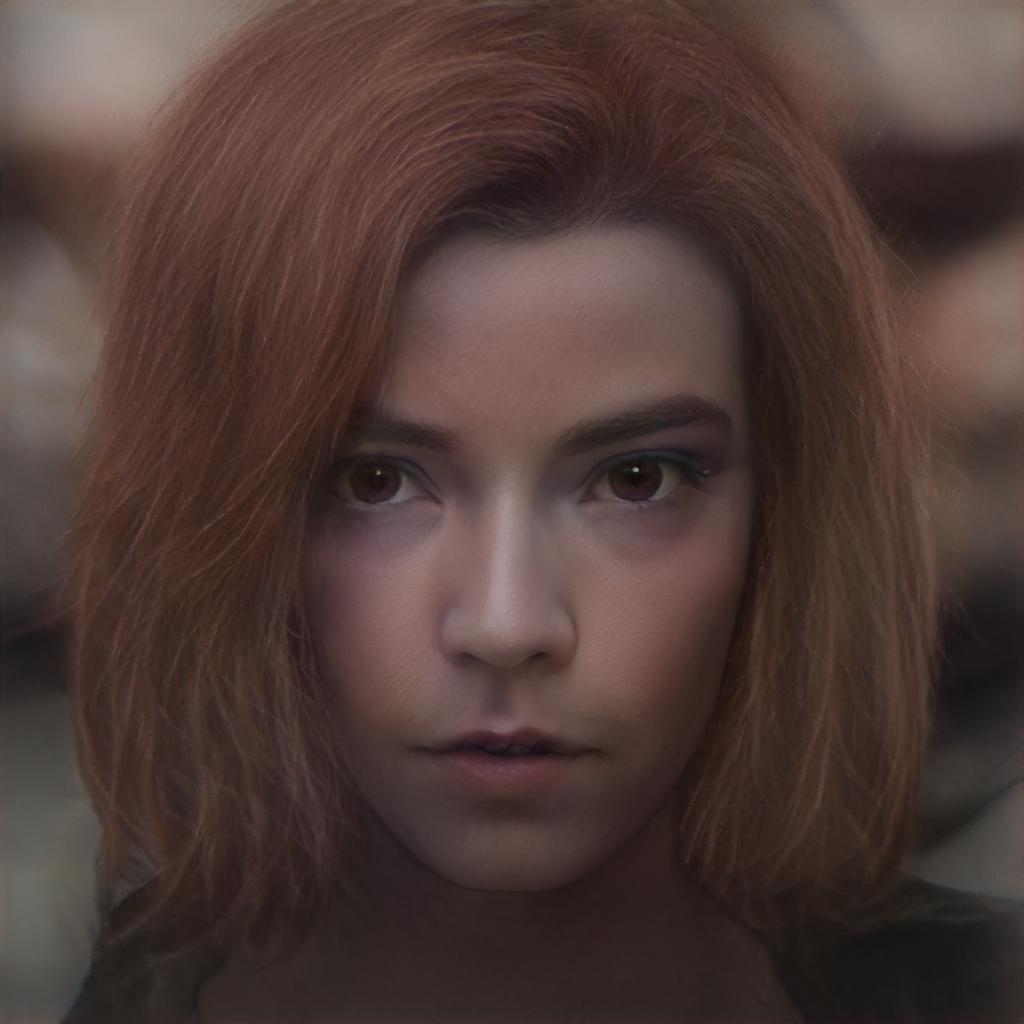} &
            \includegraphics[width=0.135\textwidth]{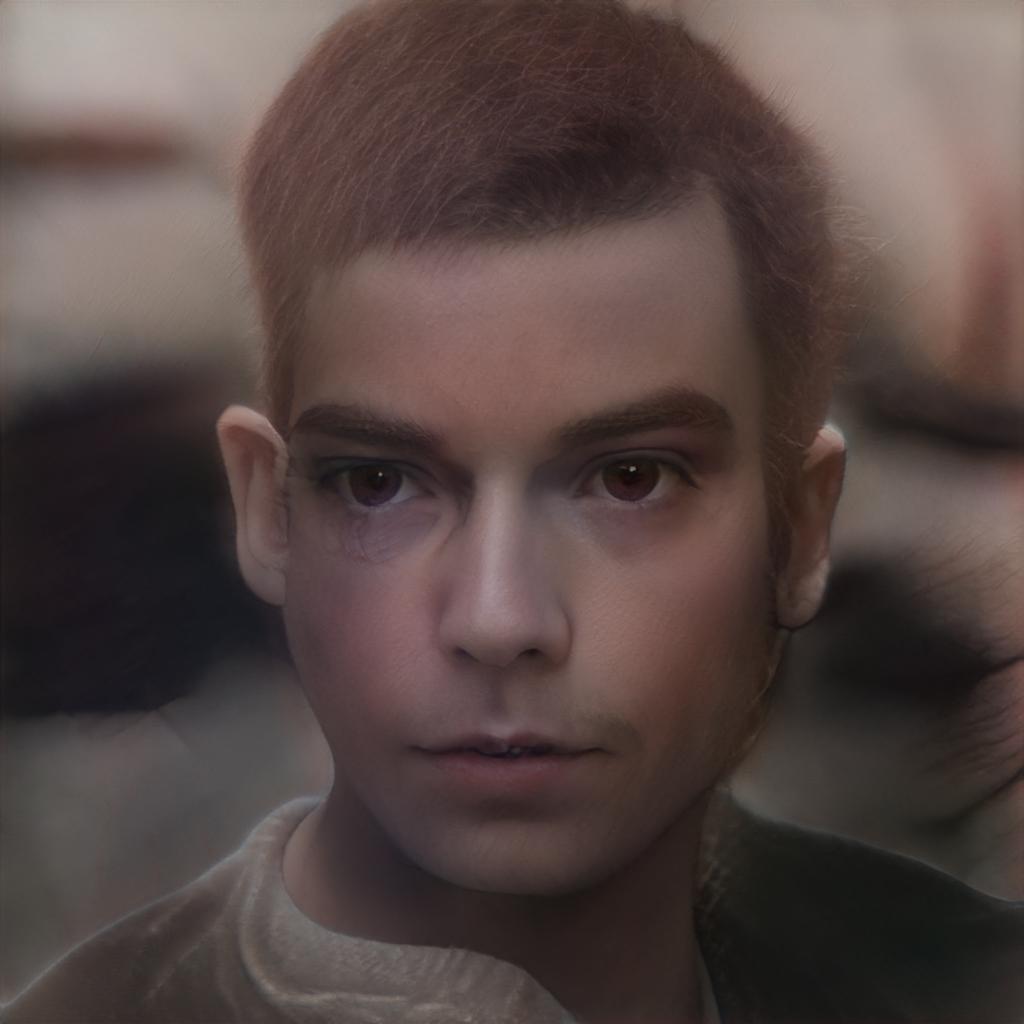} &
            \includegraphics[width=0.135\textwidth]{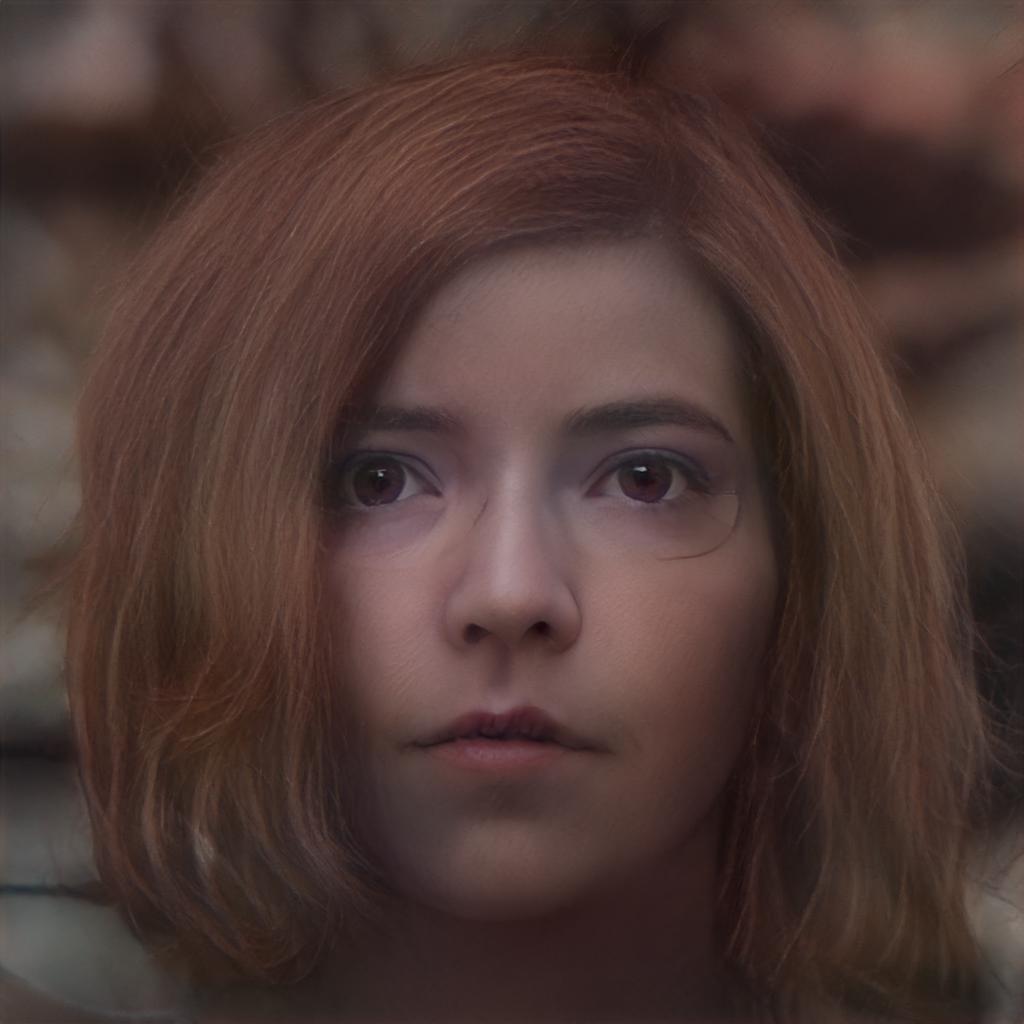} &
            \includegraphics[width=0.135\textwidth]{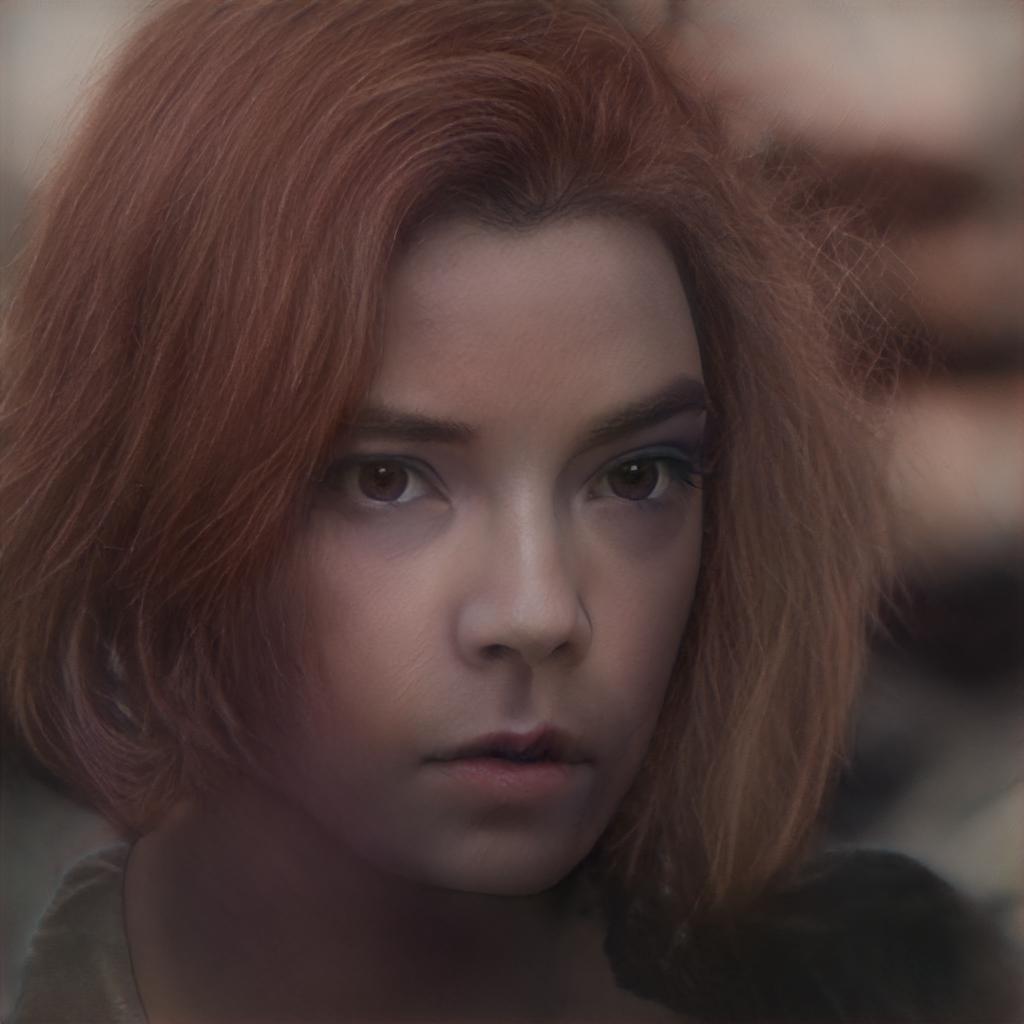} &
            \includegraphics[width=0.135\textwidth]{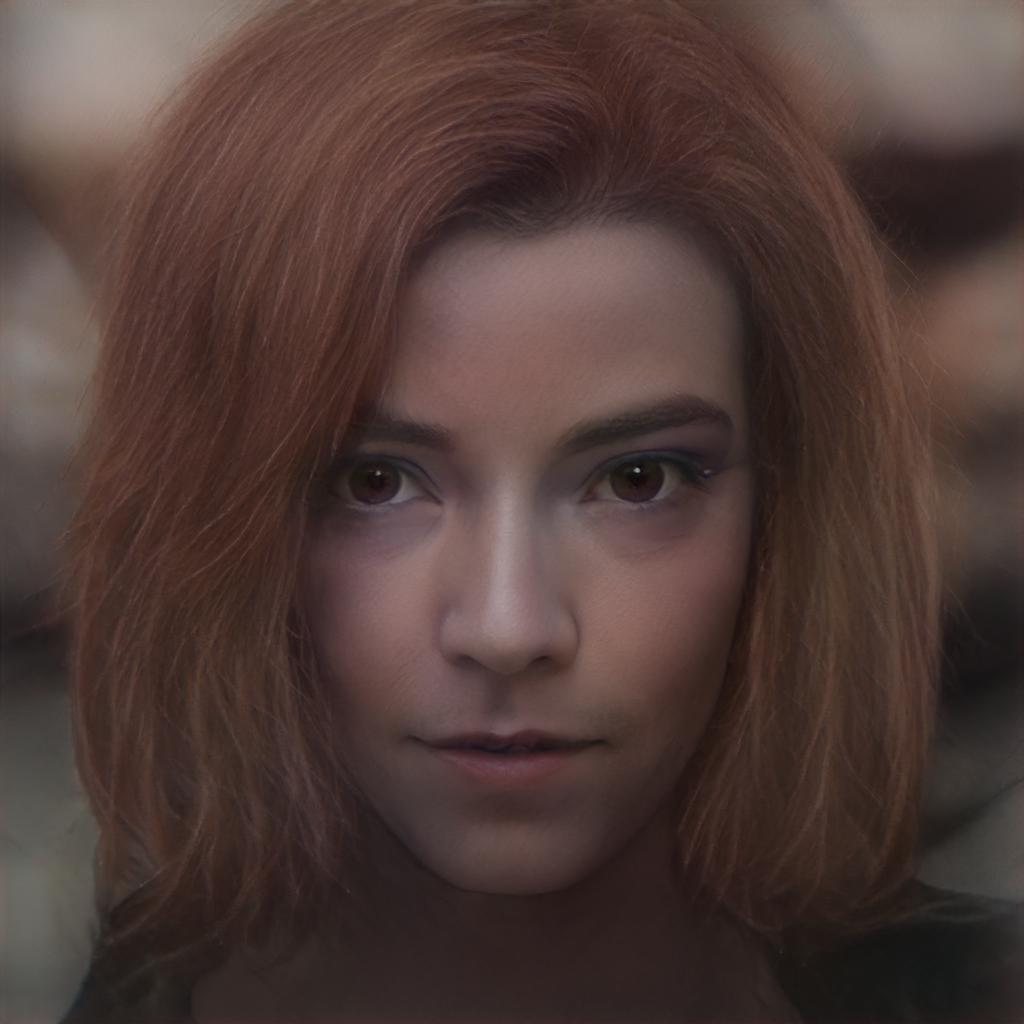}  \\
            \raisebox{0.06\textwidth}{\texttt{D}} & \includegraphics[width=0.135\textwidth]{images/appendix/styleflow_edit_celebs_ours/queens-gambit_src.jpg} &
            \includegraphics[width=0.135\textwidth]{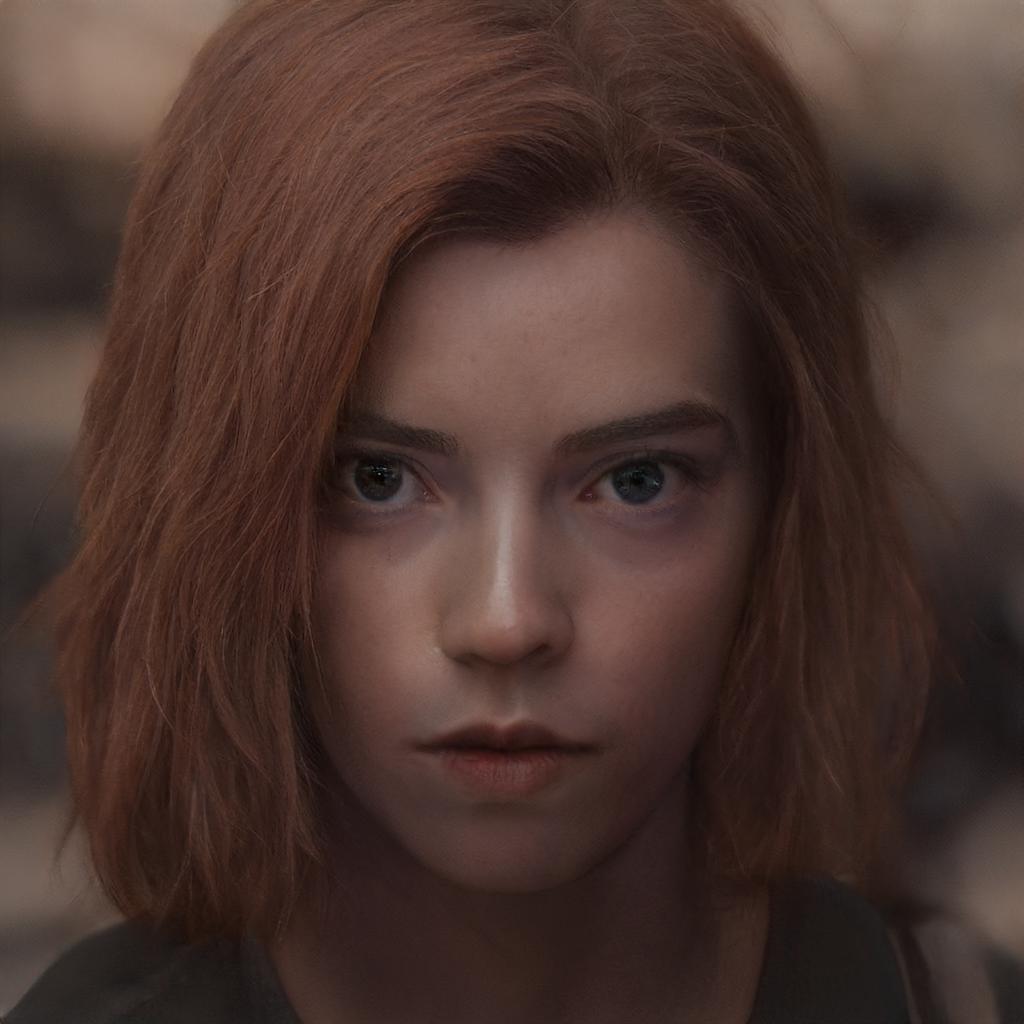} &
            \includegraphics[width=0.135\textwidth]{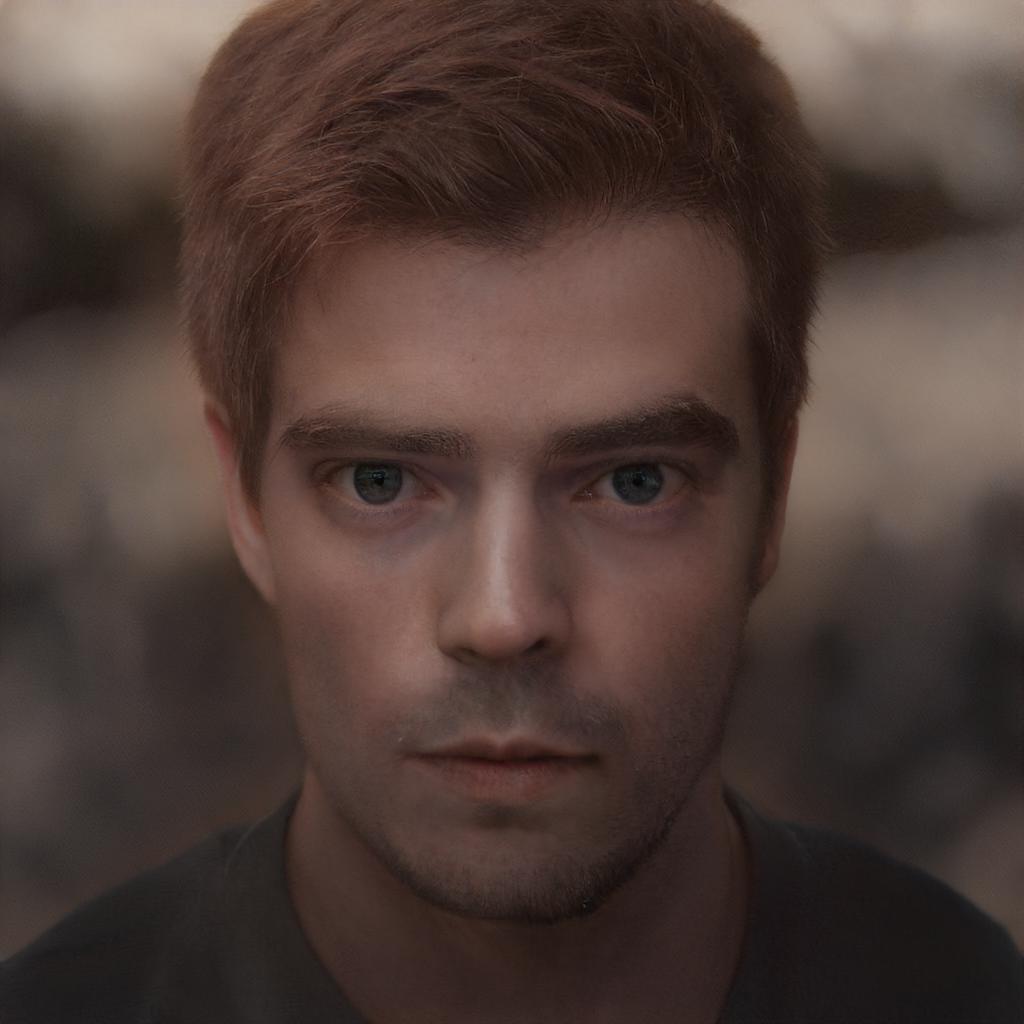} &
            \includegraphics[width=0.135\textwidth]{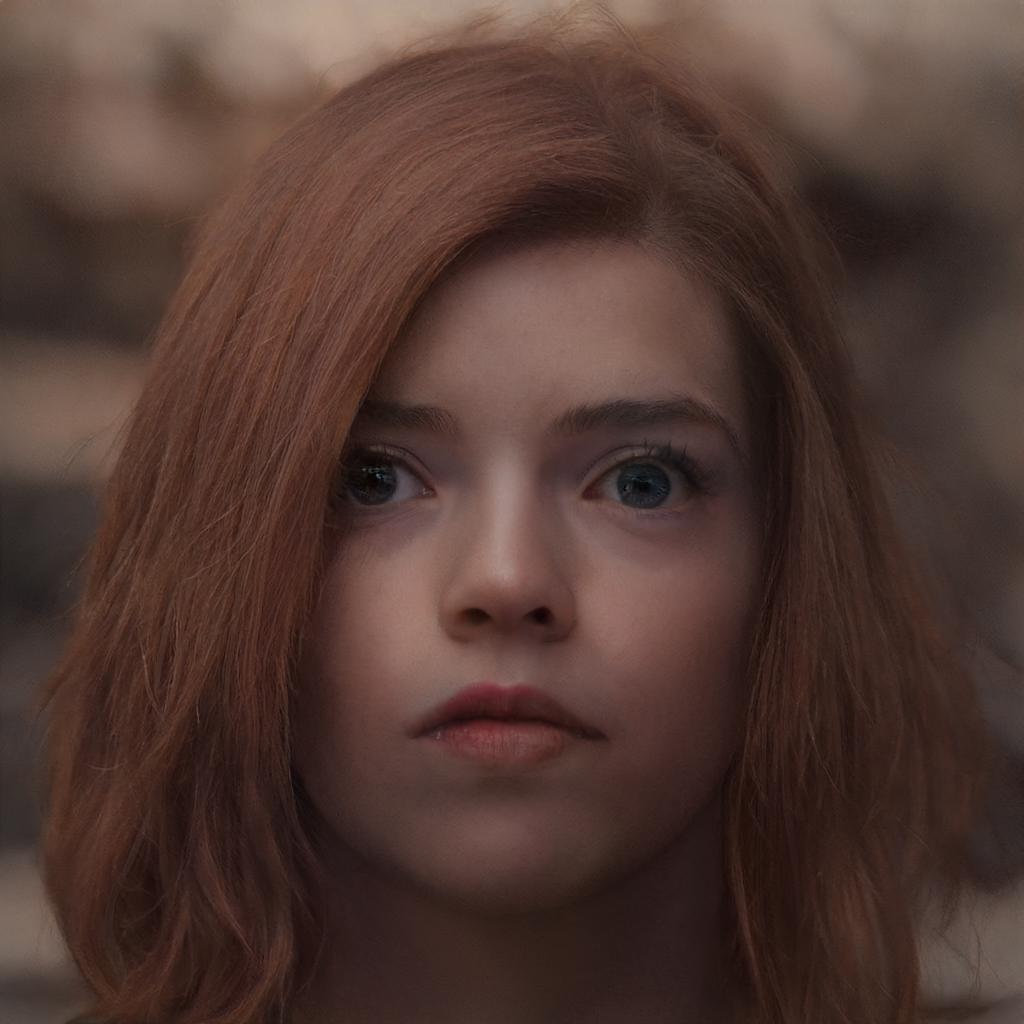} &
            \includegraphics[width=0.135\textwidth]{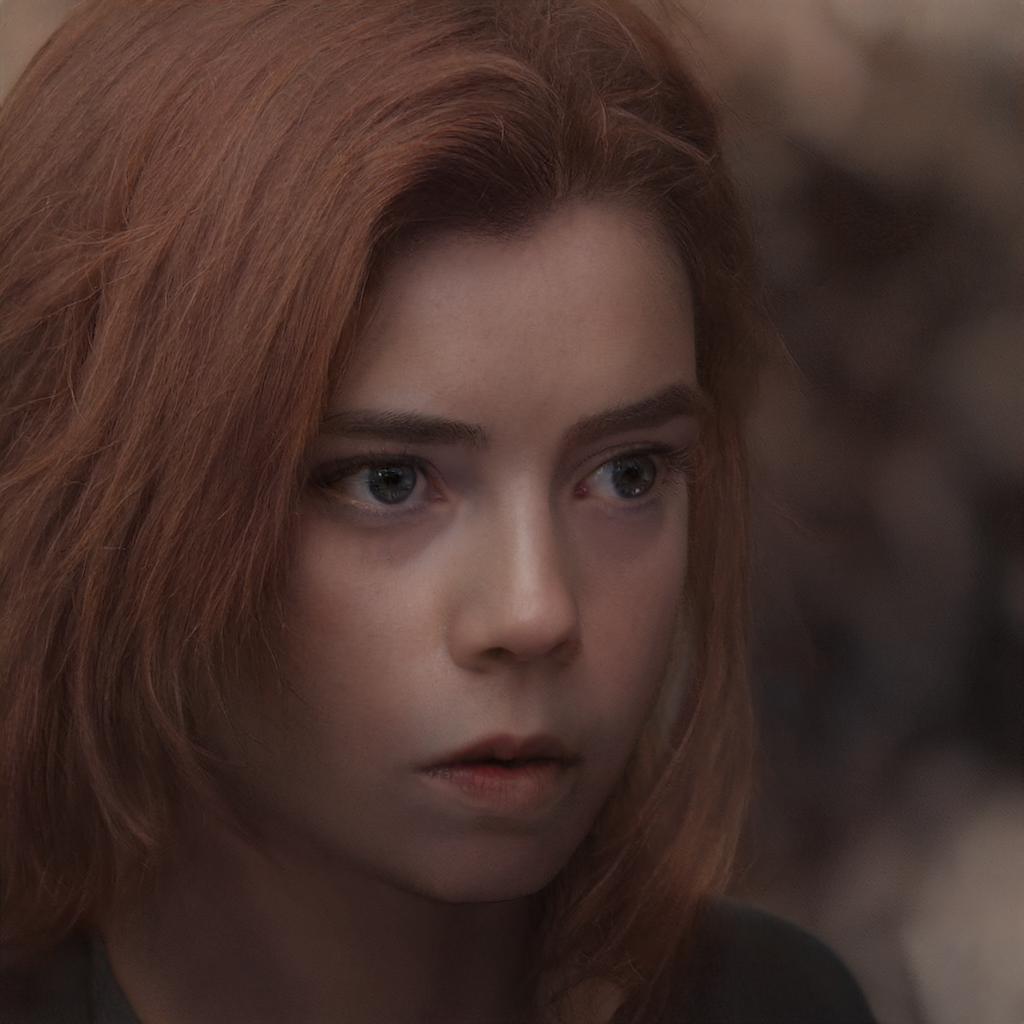} &
            \includegraphics[width=0.135\textwidth]{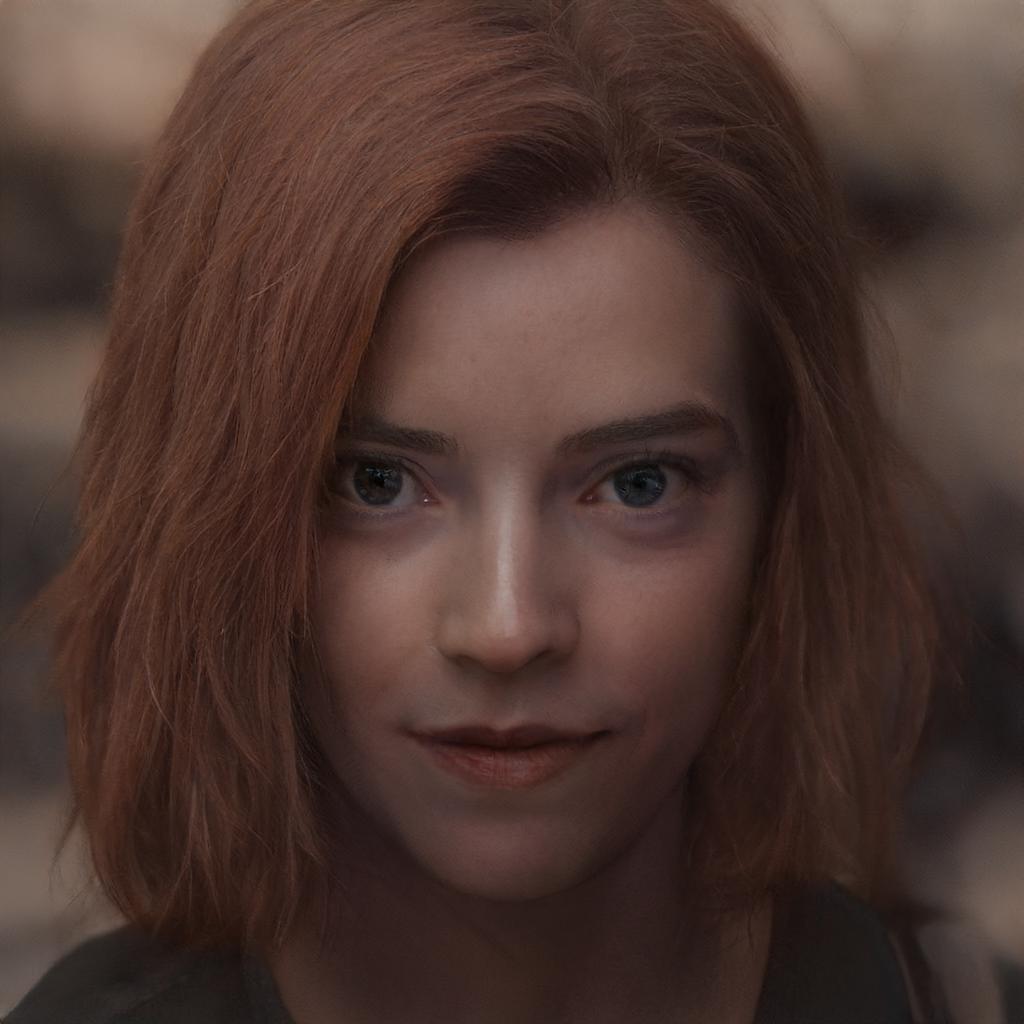} \\
            
            & Source & Inversion & \multicolumn{4}{c}{\ruleline{0.52\linewidth}{Edits}} \\
            
            \end{tabular}
    \caption{Additional comparison of configurations \texttt{A} and \texttt{D}, following the same format as Figure \ref{fig:internet_celebs_1}.}
    \label{fig:internet_celebs_2}
\end{figure*}

\begin{figure*}

    \setlength{\tabcolsep}{1pt}
    \centering
        \centering
            \begin{tabular}{c c c c c c c c c c c c c}
             & Source & Inversion & Viewpoint I & Viewpoint II & Cube & Color & Grass \\
            \raisebox{0.04\textwidth}{\texttt{A}} & 
            \includegraphics[height=0.1\textwidth]{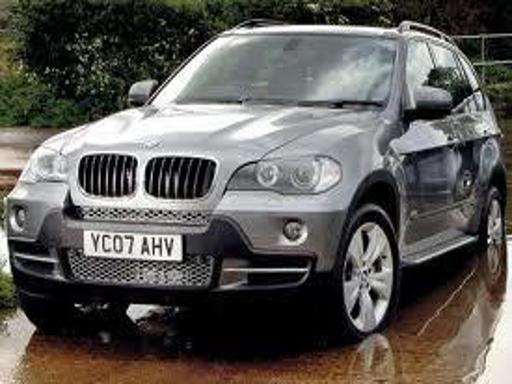} &
            \includegraphics[height=0.1\textwidth]{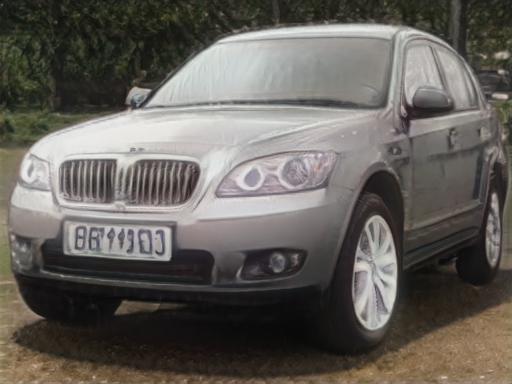} &
            \includegraphics[height=0.1\textwidth]{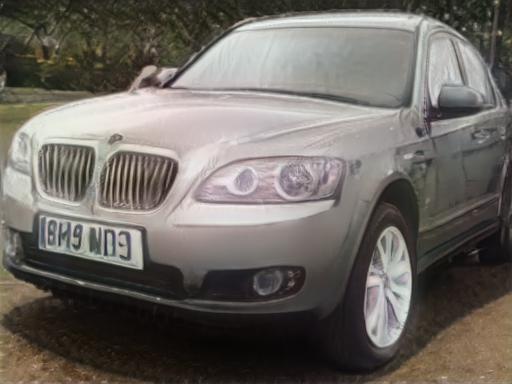} &
            \includegraphics[height=0.1\textwidth]{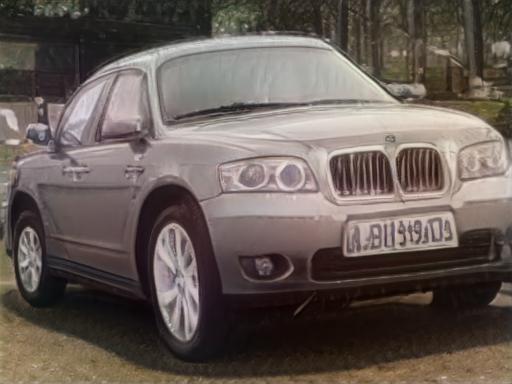} &
            \includegraphics[height=0.1\textwidth]{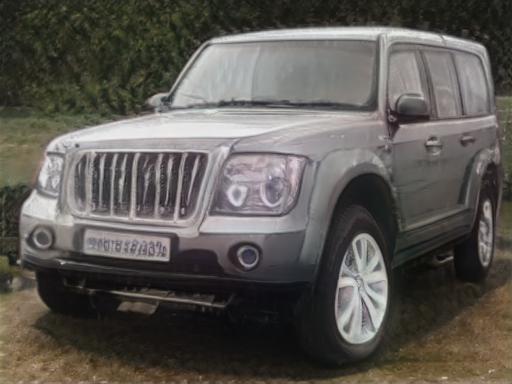} &
            \includegraphics[height=0.1\textwidth]{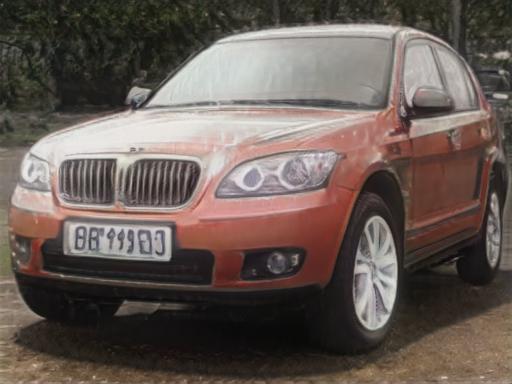} &
            \includegraphics[height=0.1\textwidth]{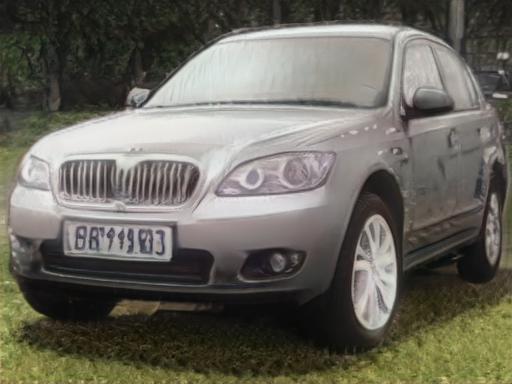} \\
            \raisebox{0.04\textwidth}{\texttt{D}} &
            \includegraphics[height=0.1\textwidth]{images/appendix/cars/00009_src.jpg} &
            \includegraphics[height=0.1\textwidth]{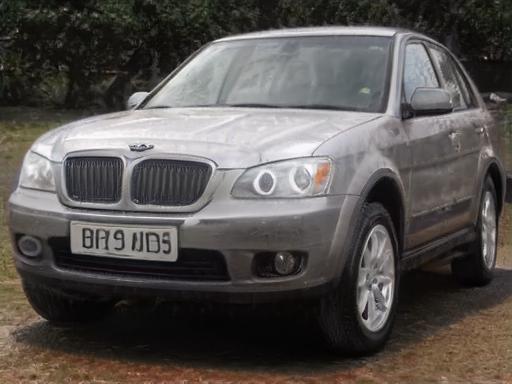} &
            \includegraphics[height=0.1\textwidth]{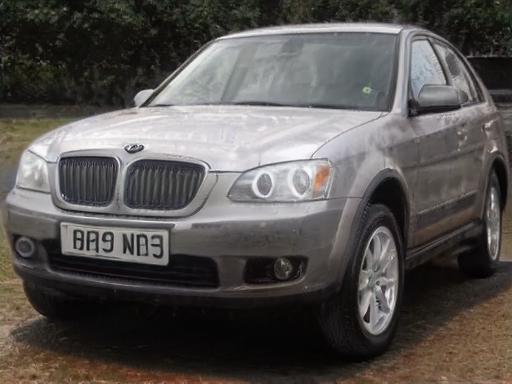} &
            \includegraphics[height=0.1\textwidth]{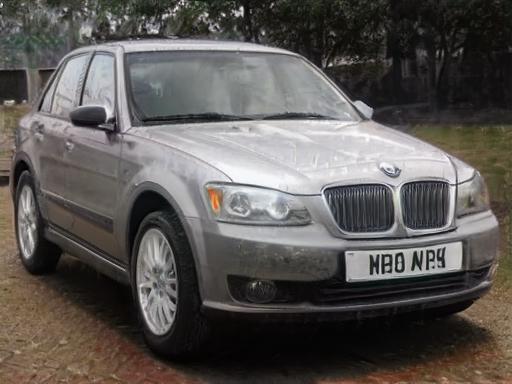} &
            \includegraphics[height=0.1\textwidth]{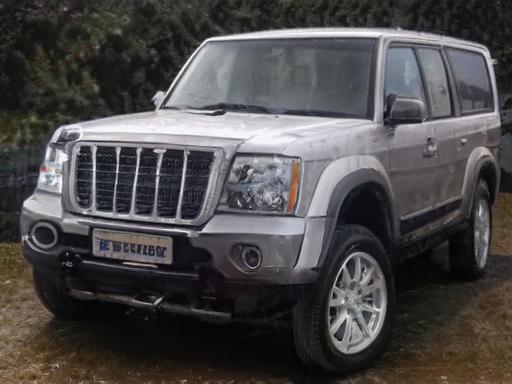} &
            \includegraphics[height=0.1\textwidth]{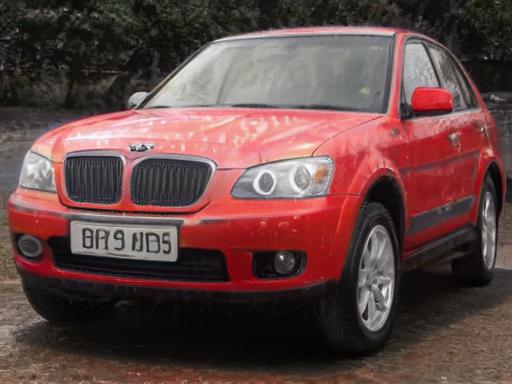} &
            \includegraphics[height=0.1\textwidth]{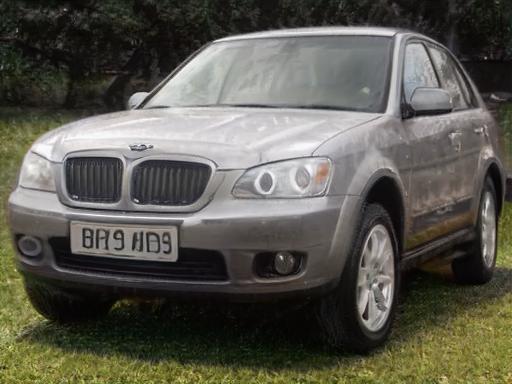} \\
            
            \raisebox{0.04\textwidth}{\texttt{A}} & 
            \includegraphics[height=0.1\textwidth]{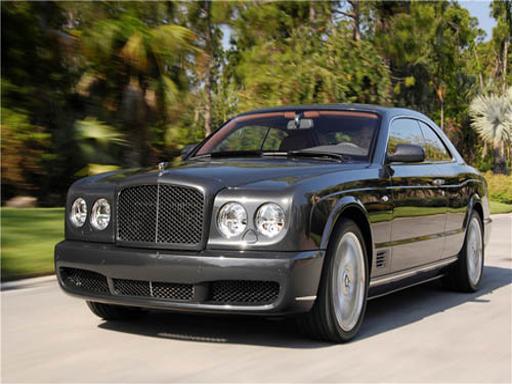} &
            \includegraphics[height=0.1\textwidth]{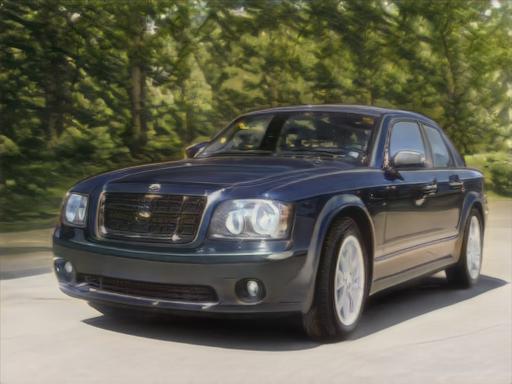} &
            \includegraphics[height=0.1\textwidth]{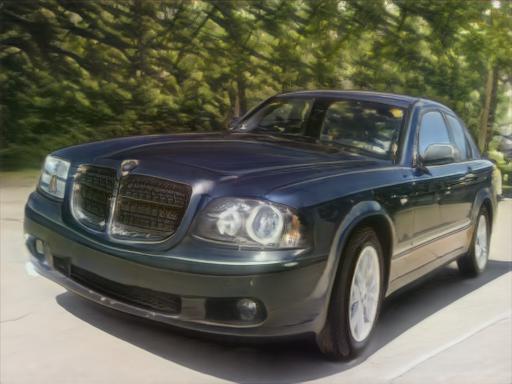} &
            \includegraphics[height=0.1\textwidth]{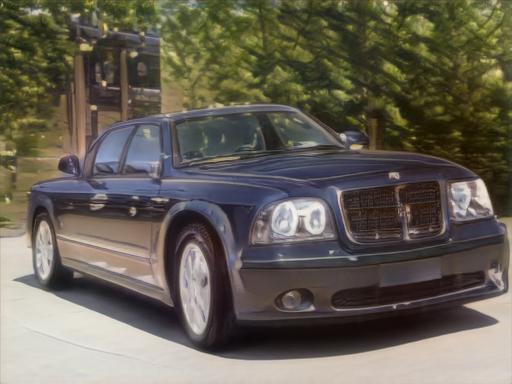} &
            \includegraphics[height=0.1\textwidth]{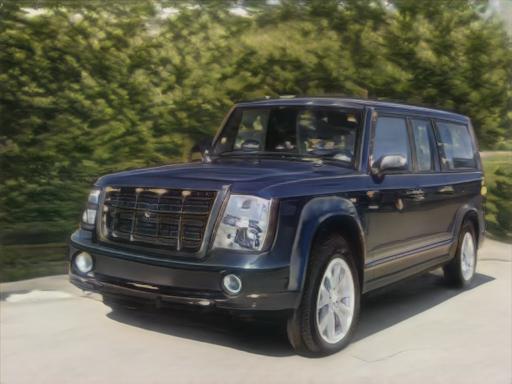} &
            \includegraphics[height=0.1\textwidth]{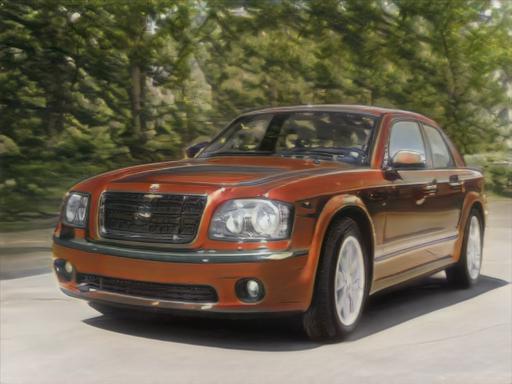} &
            \includegraphics[height=0.1\textwidth]{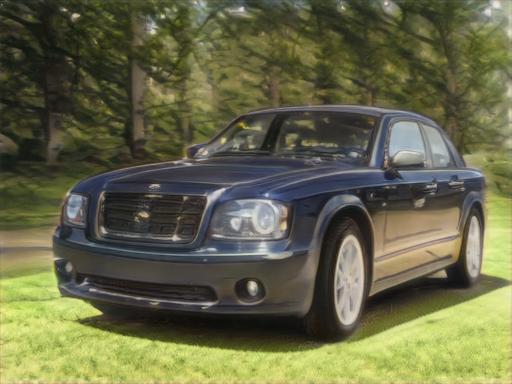} \\
            \raisebox{0.04\textwidth}{\texttt{D}} &
            \includegraphics[height=0.1\textwidth]{images/appendix/cars/00277_src.jpg} &
            \includegraphics[height=0.1\textwidth]{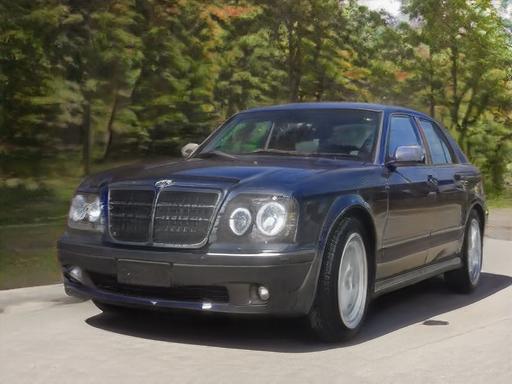} &
            \includegraphics[height=0.1\textwidth]{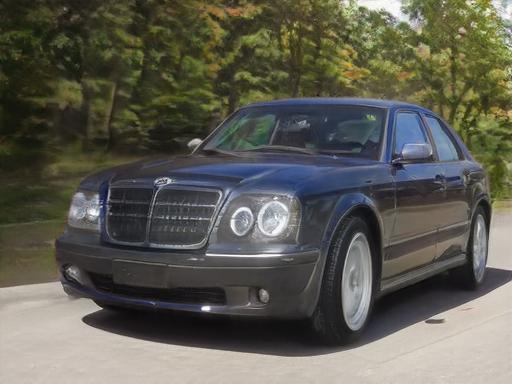} &
            \includegraphics[height=0.1\textwidth]{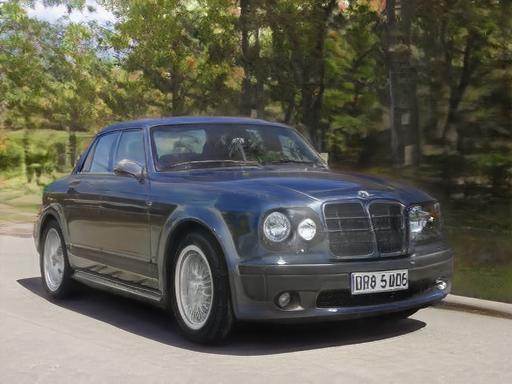} &
            \includegraphics[height=0.1\textwidth]{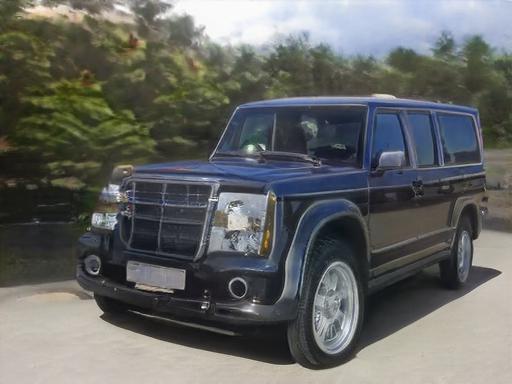} &
            \includegraphics[height=0.1\textwidth]{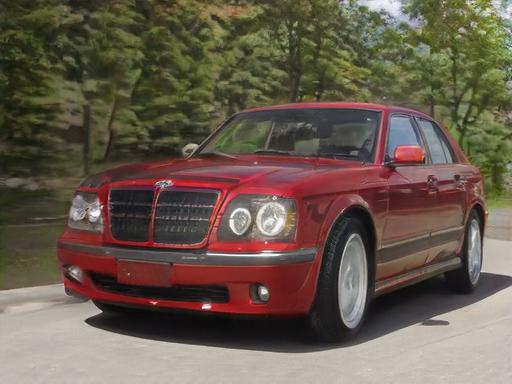} &
            \includegraphics[height=0.1\textwidth]{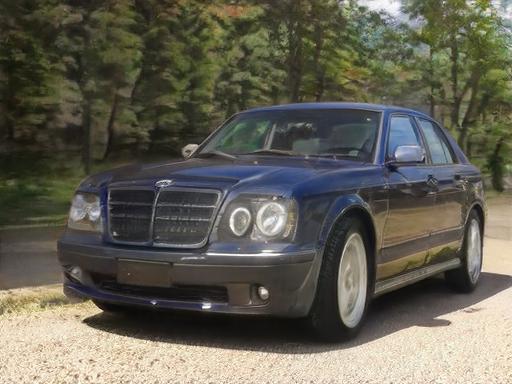} \\
            
            \raisebox{0.04\textwidth}{\texttt{A}} & 
            \includegraphics[height=0.1\textwidth]{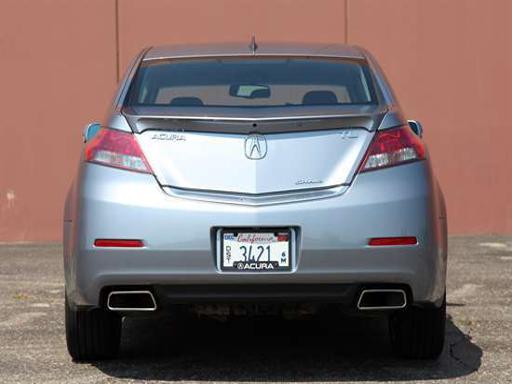} &
            \includegraphics[height=0.1\textwidth]{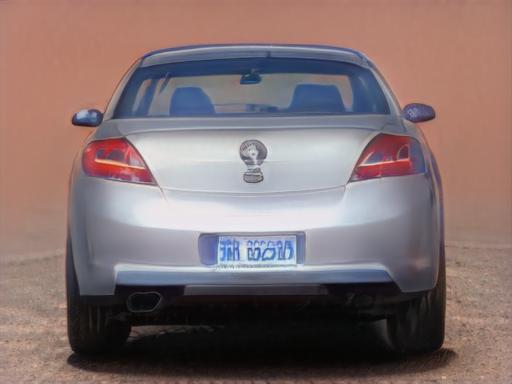} &
            \includegraphics[height=0.1\textwidth]{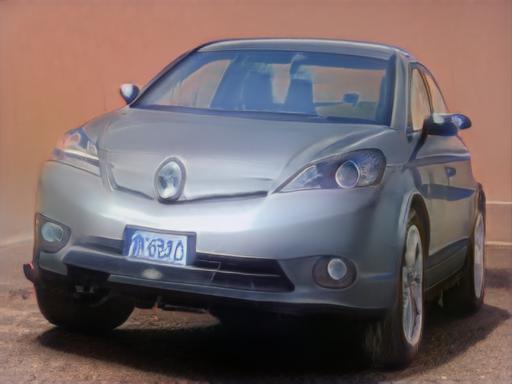} &
            \includegraphics[height=0.1\textwidth]{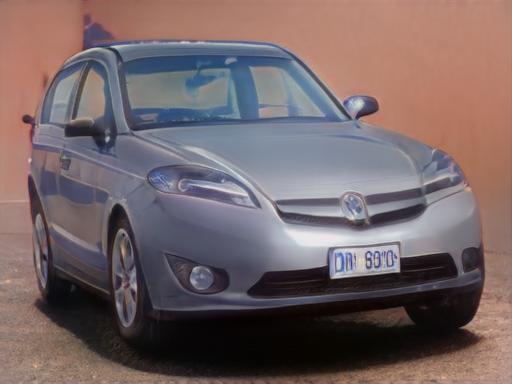} &
            \includegraphics[height=0.1\textwidth]{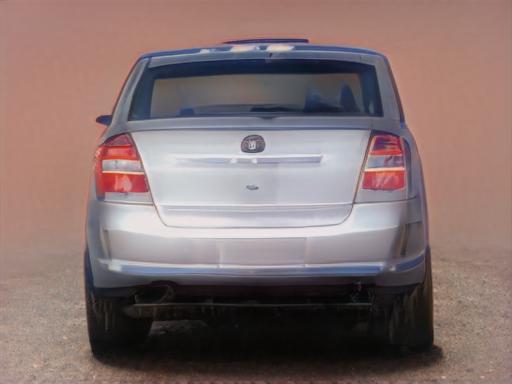} &
            \includegraphics[height=0.1\textwidth]{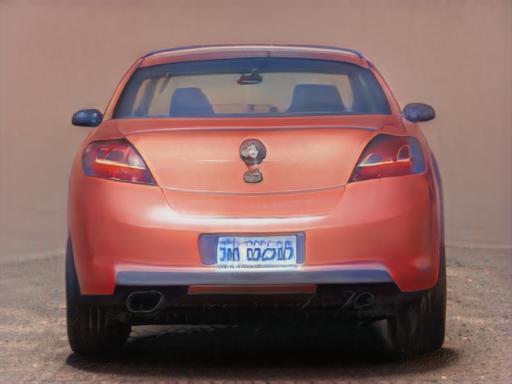} &
            \includegraphics[height=0.1\textwidth]{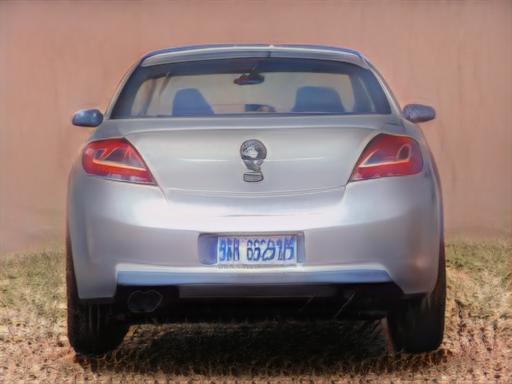} \\
            \raisebox{0.04\textwidth}{\texttt{D}} &
            \includegraphics[height=0.1\textwidth]{images/appendix/cars/00397_src.jpg} &
            \includegraphics[height=0.1\textwidth]{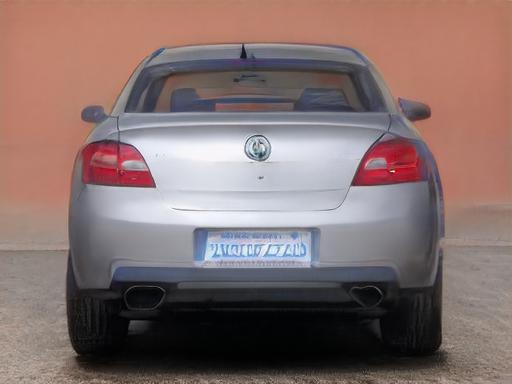} &
            \includegraphics[height=0.1\textwidth]{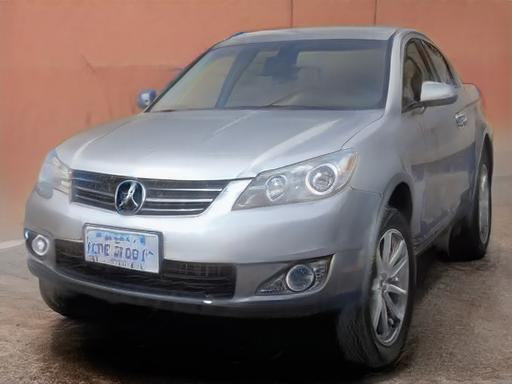} &
            \includegraphics[height=0.1\textwidth]{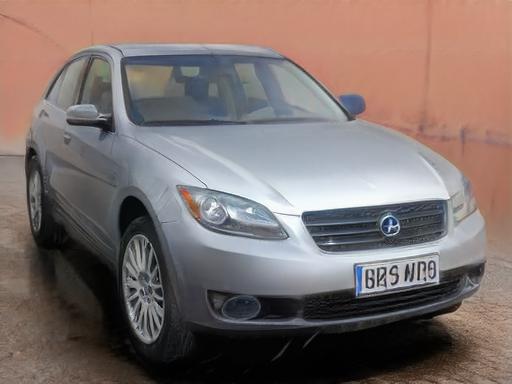} &
            \includegraphics[height=0.1\textwidth]{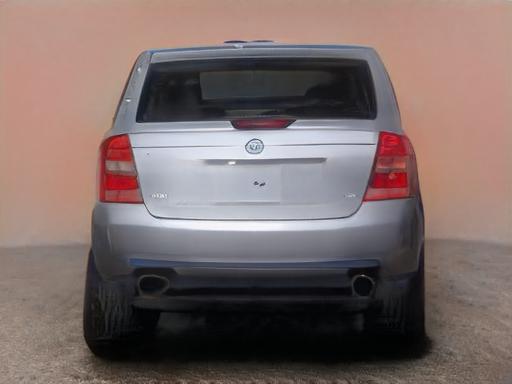} &
            \includegraphics[height=0.1\textwidth]{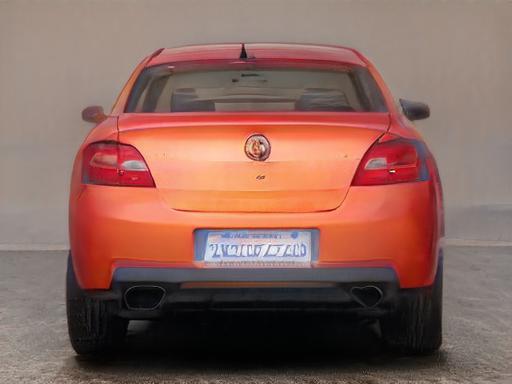} &
            \includegraphics[height=0.1\textwidth]{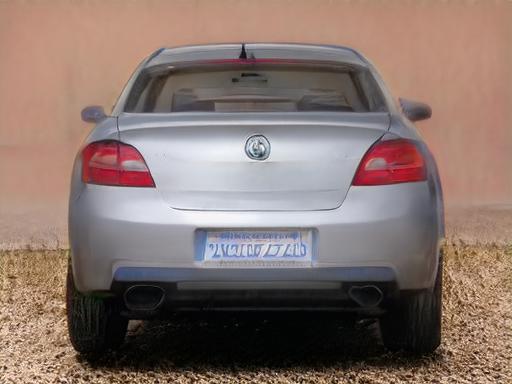} \\
            
            \raisebox{0.04\textwidth}{\texttt{A}} & 
            \includegraphics[height=0.1\textwidth]{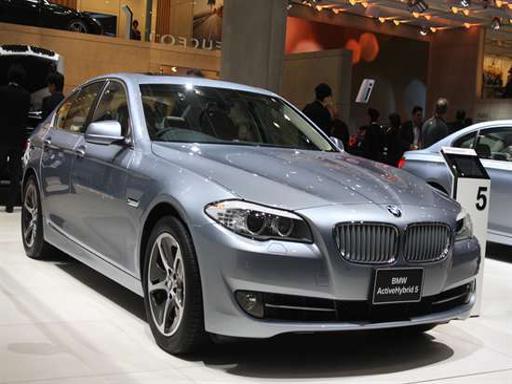} &
            \includegraphics[height=0.1\textwidth]{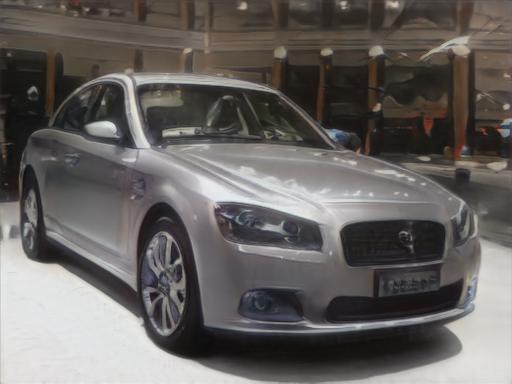} &
            \includegraphics[height=0.1\textwidth]{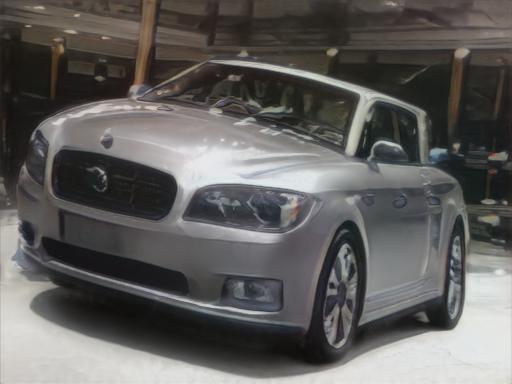} &
            \includegraphics[height=0.1\textwidth]{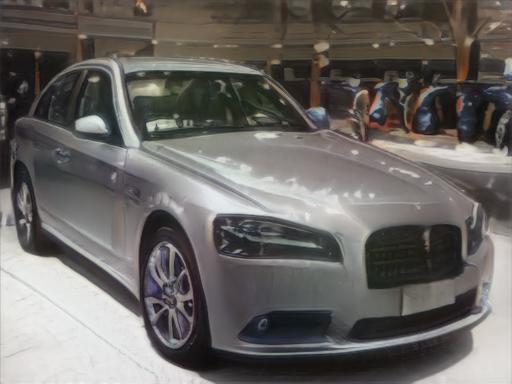} &
            \includegraphics[height=0.1\textwidth]{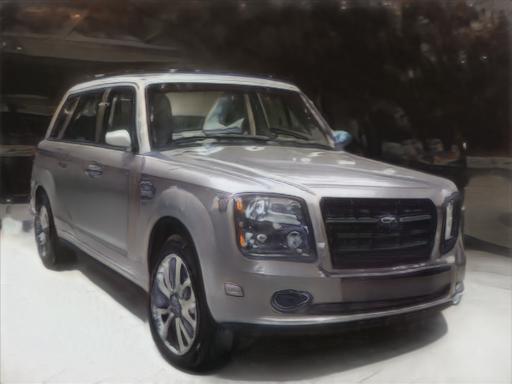} &
            \includegraphics[height=0.1\textwidth]{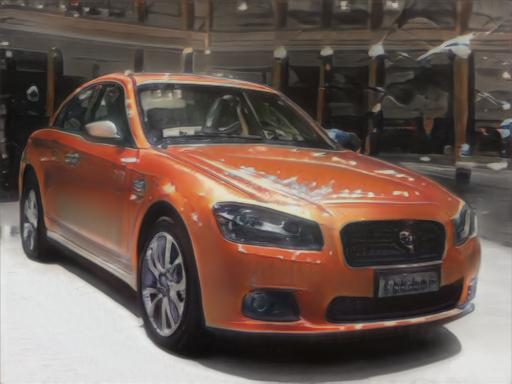} &
            \includegraphics[height=0.1\textwidth]{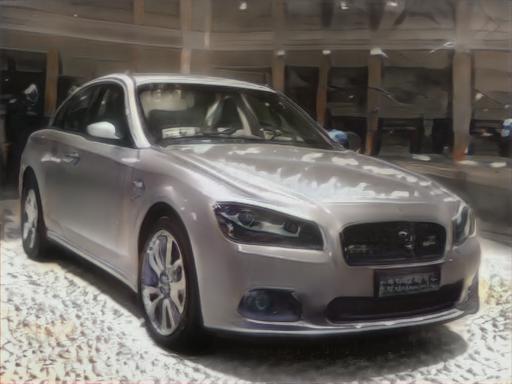} \\
            \raisebox{0.04\textwidth}{\texttt{D}} &
            \includegraphics[height=0.1\textwidth]{images/appendix/cars/00439_src.jpg} &
            \includegraphics[height=0.1\textwidth]{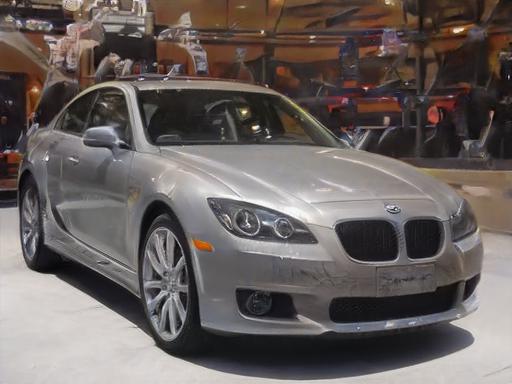} &
            \includegraphics[height=0.1\textwidth]{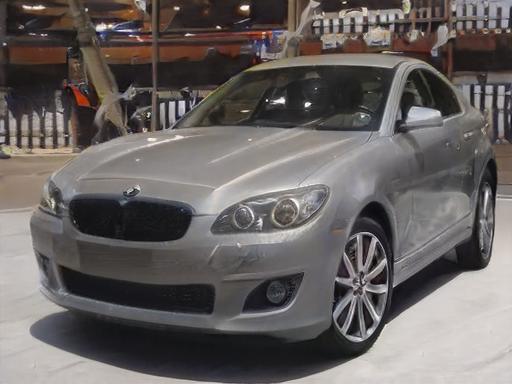} &
            \includegraphics[height=0.1\textwidth]{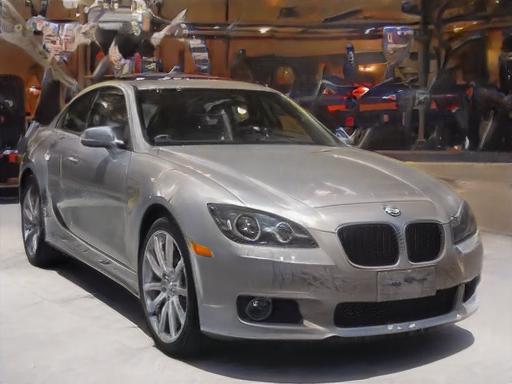} &
            \includegraphics[height=0.1\textwidth]{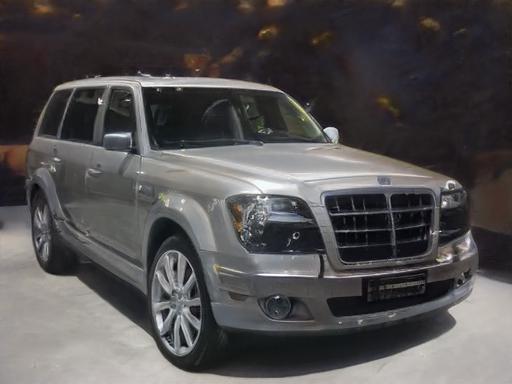} &
            \includegraphics[height=0.1\textwidth]{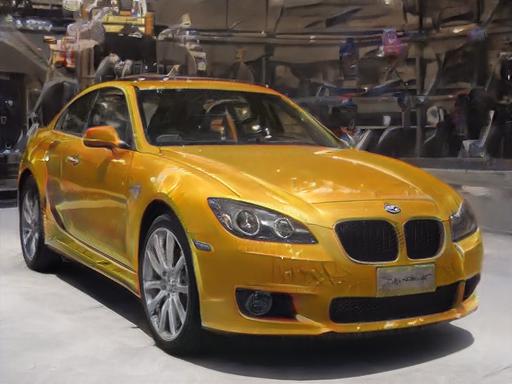} &
            \includegraphics[height=0.1\textwidth]{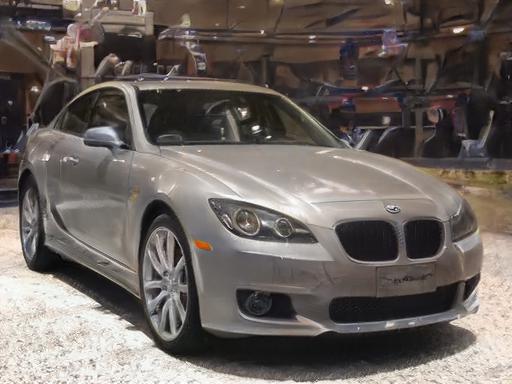} \\
            
            \raisebox{0.04\textwidth}{\texttt{A}} & 
            \includegraphics[height=0.1\textwidth]{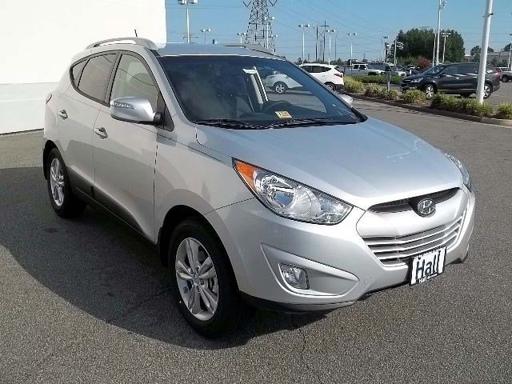} &
            \includegraphics[height=0.1\textwidth]{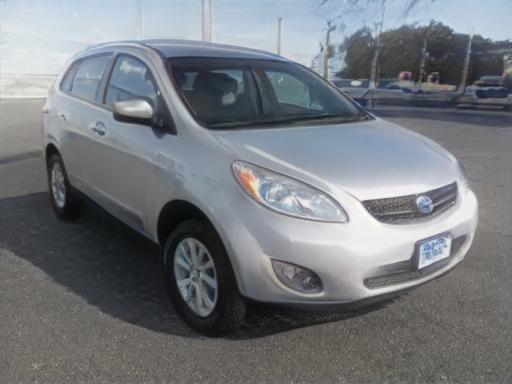} &
            \includegraphics[height=0.1\textwidth]{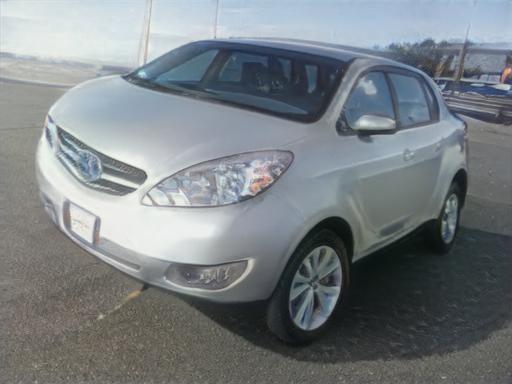} &
            \includegraphics[height=0.1\textwidth]{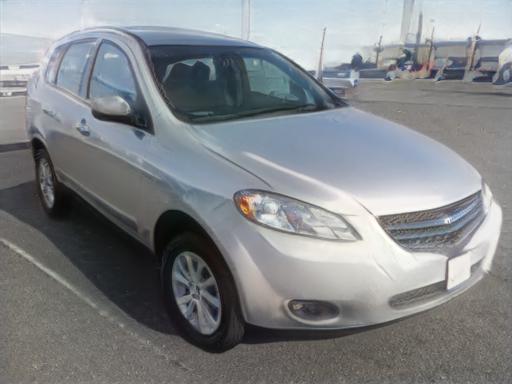} &
            \includegraphics[height=0.1\textwidth]{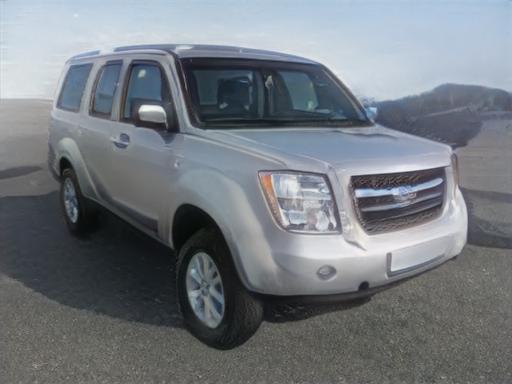} &
            \includegraphics[height=0.1\textwidth]{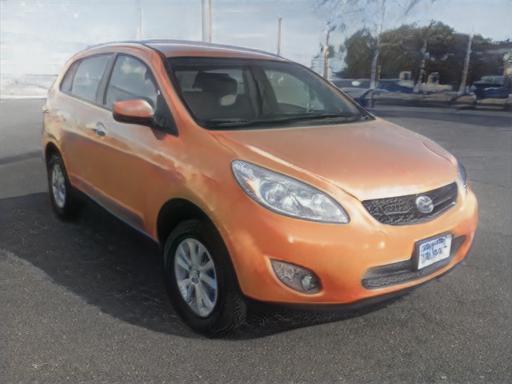} &
            \includegraphics[height=0.1\textwidth]{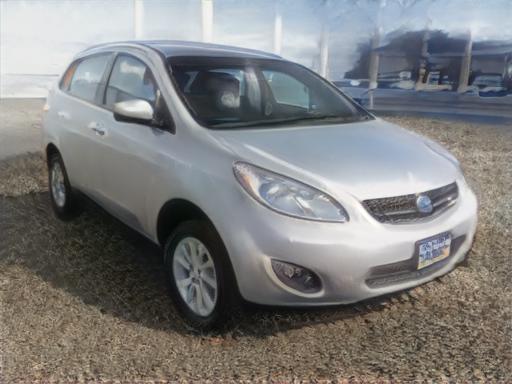} \\
            \raisebox{0.04\textwidth}{\texttt{D}} &
            \includegraphics[height=0.1\textwidth]{images/appendix/cars/00270_src.jpg} &
            \includegraphics[height=0.1\textwidth]{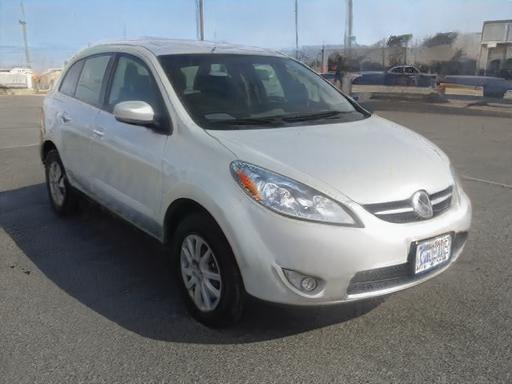} &
            \includegraphics[height=0.1\textwidth]{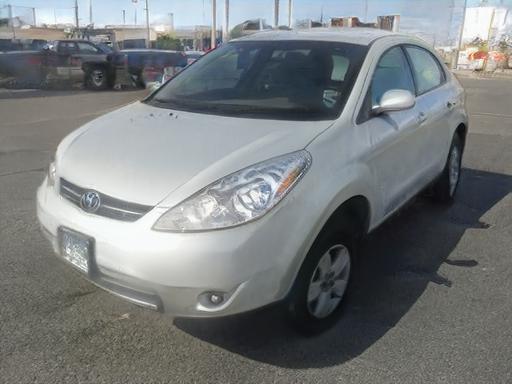} &
            \includegraphics[height=0.1\textwidth]{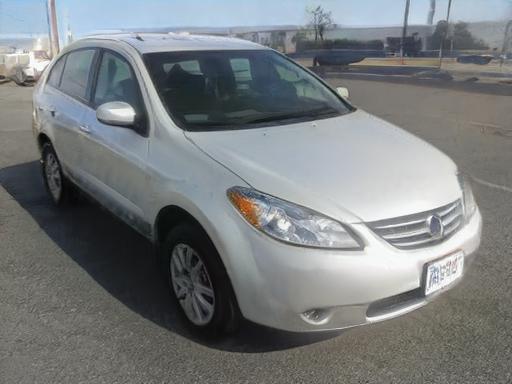} &
            \includegraphics[height=0.1\textwidth]{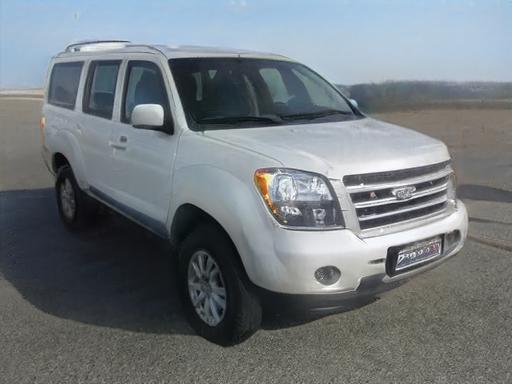} &
            \includegraphics[height=0.1\textwidth]{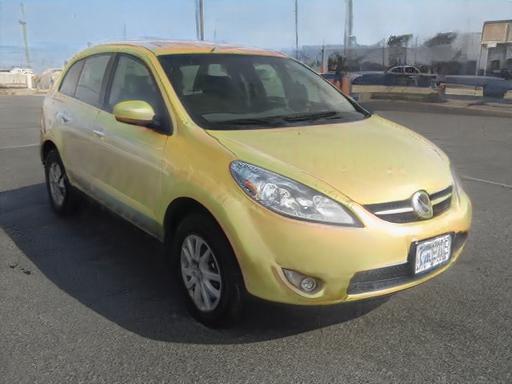} &
            \includegraphics[height=0.1\textwidth]{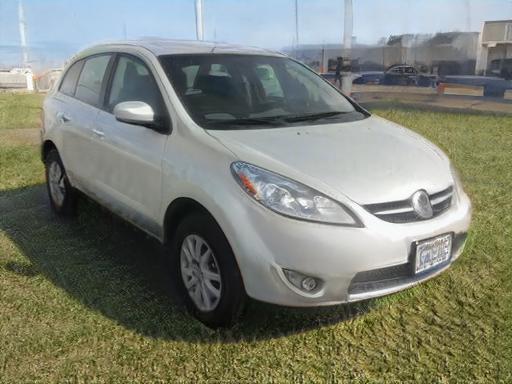} \\
            
             & Source & Inversion & Viewpoint I & Viewpoint II & Cube & Color & Grass \\
            \end{tabular}
    \caption{Comparing inversion and editing results of configurations \texttt{A} and \texttt{D} on the Stanford Cars test set~\cite{KrauseStarkDengFei-Fei_3DRR2013}. The leftmost column is the real source image, to its right is the reconstruction through StyleGAN2~\cite{karras2020analyzing}. Remaining columns are edits along the directions obtained by GANSpace \cite{harkonen2020ganspace}. The column header specifies the edit performed. Note that while configuration \texttt{A} often presents less, i.e. better, distortion, configuration \texttt{D} yields reconstruction and editings with better perceptual quality. 
    }
    \label{fig:tradeoff-cars}
\end{figure*}

\begin{figure*}
    \setlength{\tabcolsep}{1pt}
    \centering
        \centering
            \begin{tabular}{c c c c c c c c c c c c c}
            & Source & Inversion & Viewpoint I & Viewpoint II & Cube & Color & Grass \\
           \raisebox{0.04\textwidth}{\texttt{A}} & 
            \includegraphics[height=0.1\textwidth]{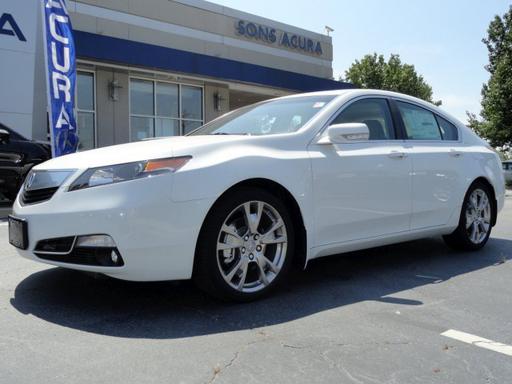} &
            \includegraphics[height=0.1\textwidth]{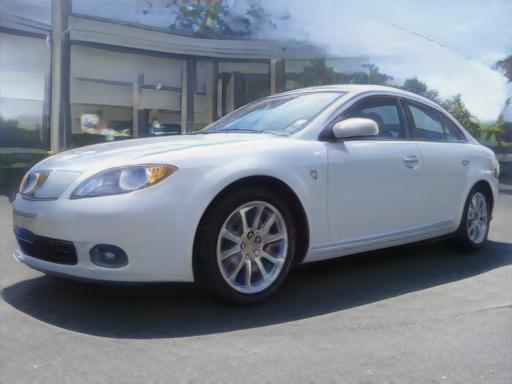} &
            \includegraphics[height=0.1\textwidth]{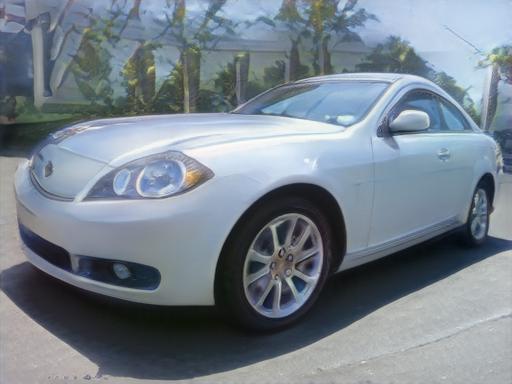} &
            \includegraphics[height=0.1\textwidth]{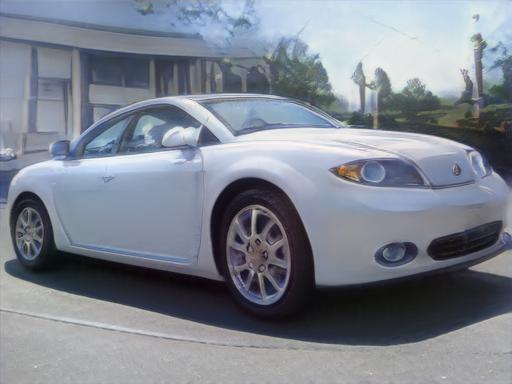} &
            \includegraphics[height=0.1\textwidth]{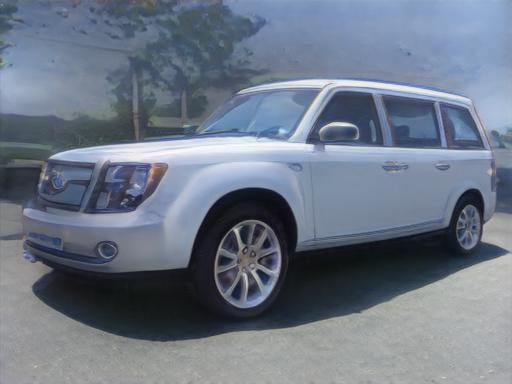} &
            \includegraphics[height=0.1\textwidth]{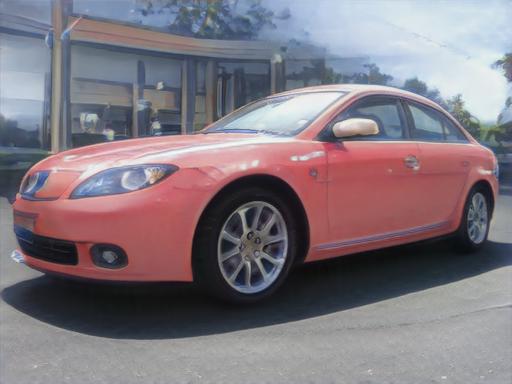} &
            \includegraphics[height=0.1\textwidth]{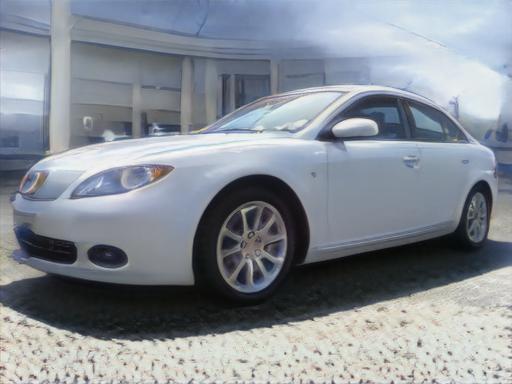} \\
           \raisebox{0.04\textwidth}{\texttt{D}} &
            \includegraphics[height=0.1\textwidth]{images/appendix/cars/00282_src.jpg} &
            \includegraphics[height=0.1\textwidth]{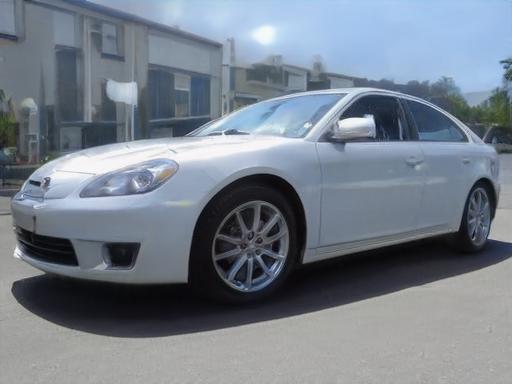} &
            \includegraphics[height=0.1\textwidth]{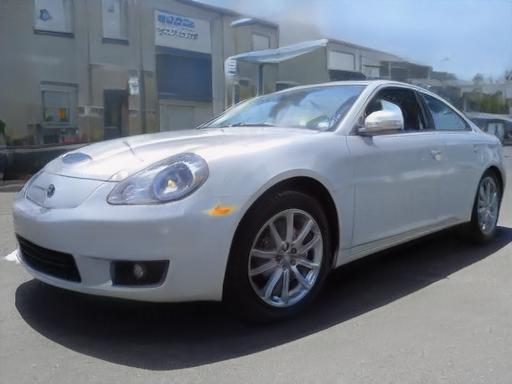} &
            \includegraphics[height=0.1\textwidth]{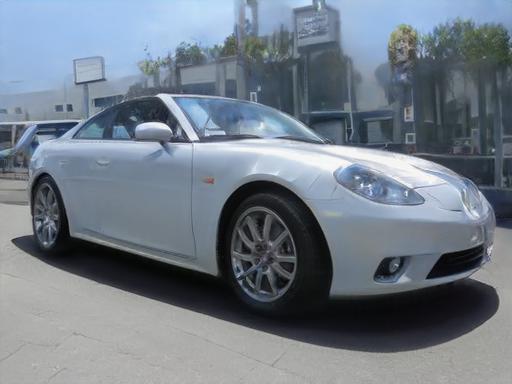} &
            \includegraphics[height=0.1\textwidth]{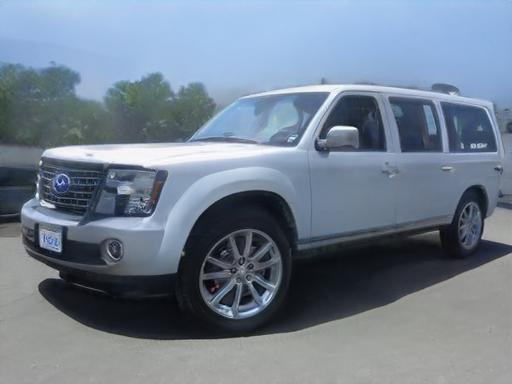} &
            \includegraphics[height=0.1\textwidth]{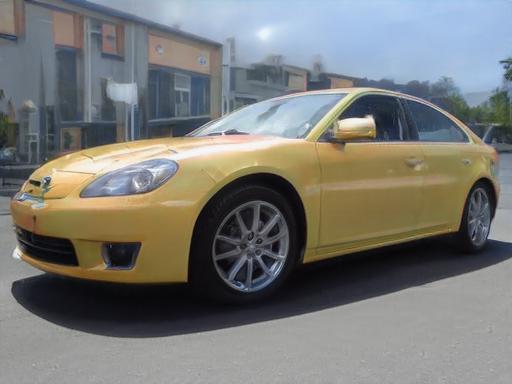} &
            \includegraphics[height=0.1\textwidth]{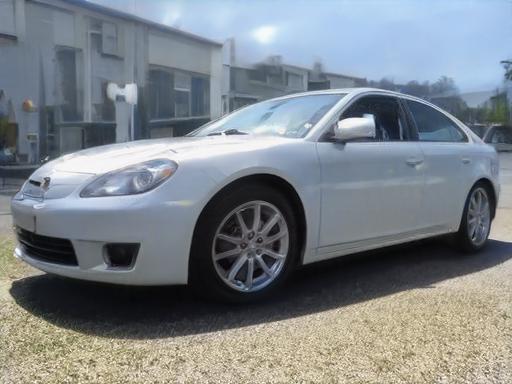} \\
            
           \raisebox{0.04\textwidth}{\texttt{A}} & 
            \includegraphics[height=0.1\textwidth]{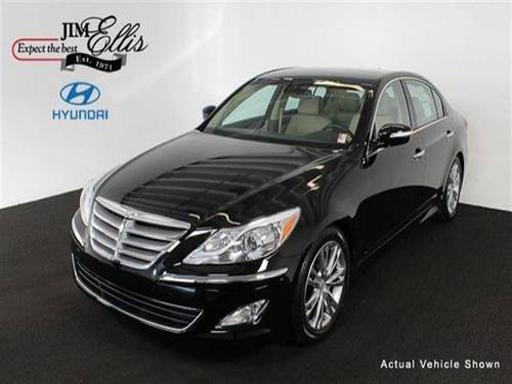} &
            \includegraphics[height=0.1\textwidth]{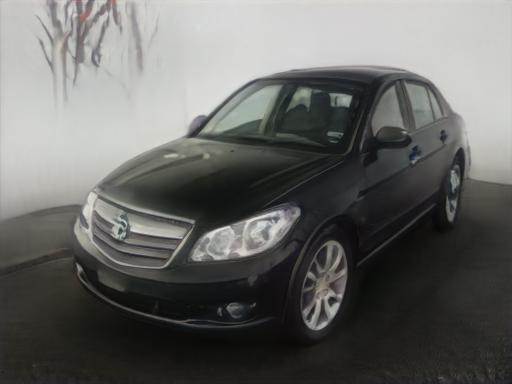} &
            \includegraphics[height=0.1\textwidth]{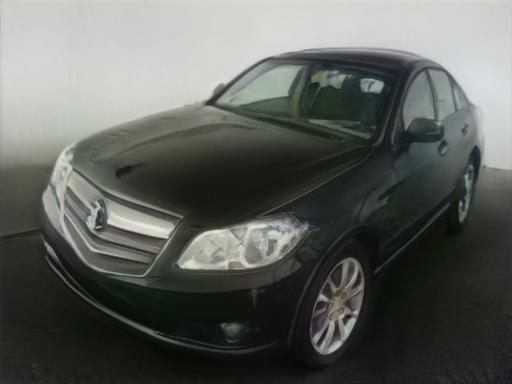} &
            \includegraphics[height=0.1\textwidth]{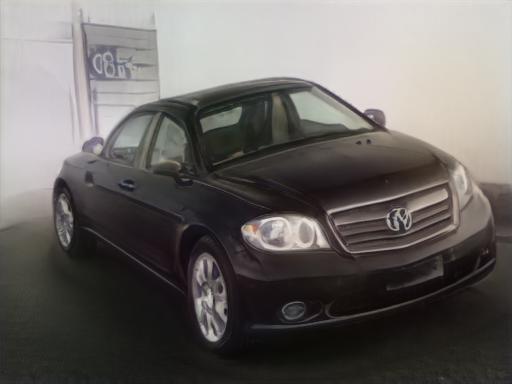} &
            \includegraphics[height=0.1\textwidth]{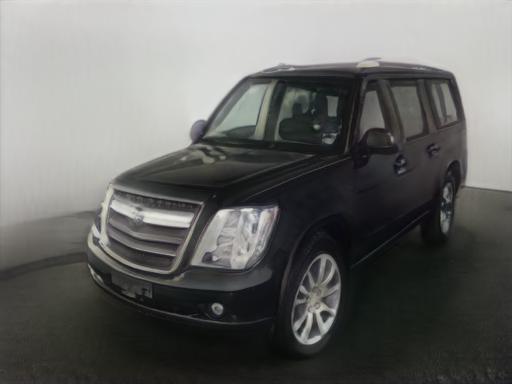} &
            \includegraphics[height=0.1\textwidth]{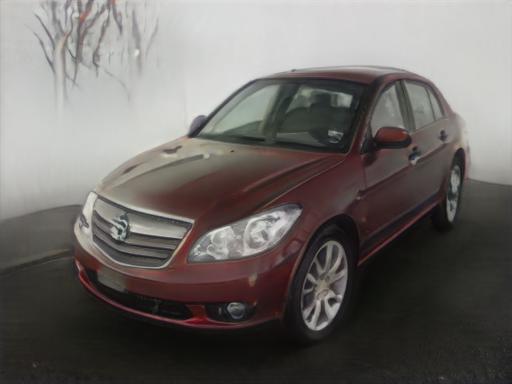} &
            \includegraphics[height=0.1\textwidth]{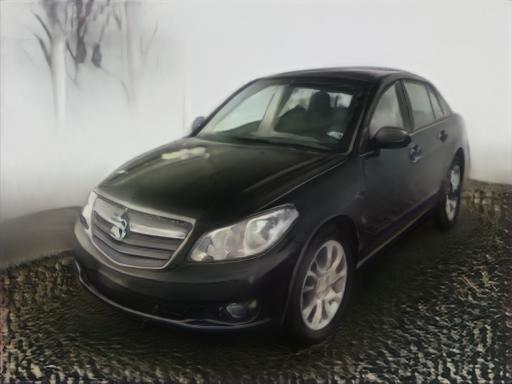} \\
           \raisebox{0.04\textwidth}{\texttt{D}} &
            \includegraphics[height=0.1\textwidth]{images/appendix/cars/00284_src.jpg} &
            \includegraphics[height=0.1\textwidth]{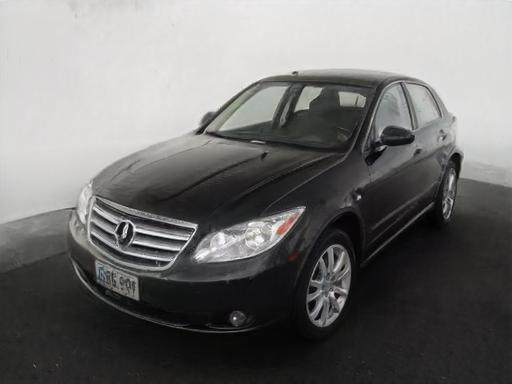} &
            \includegraphics[height=0.1\textwidth]{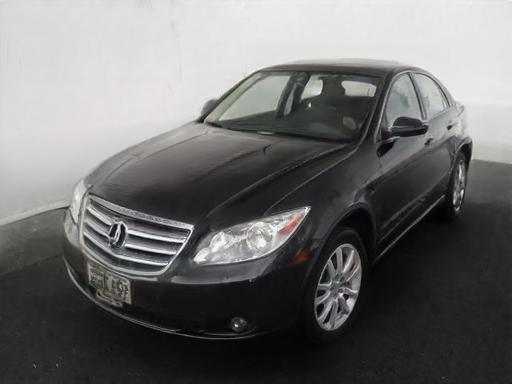} &
            \includegraphics[height=0.1\textwidth]{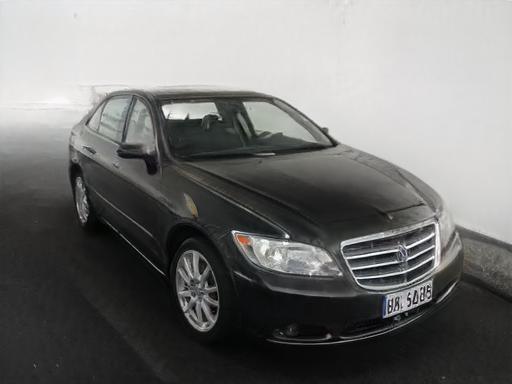} &
            \includegraphics[height=0.1\textwidth]{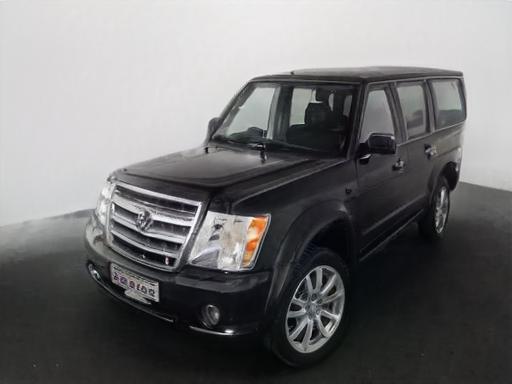} &
            \includegraphics[height=0.1\textwidth]{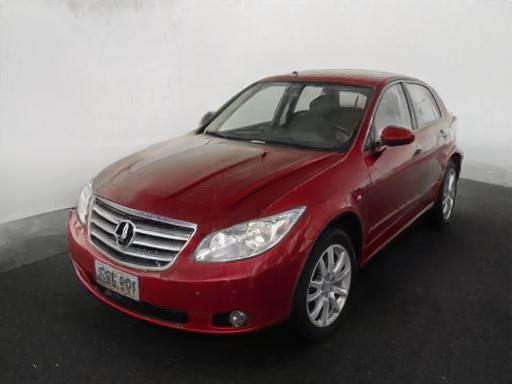} &
            \includegraphics[height=0.1\textwidth]{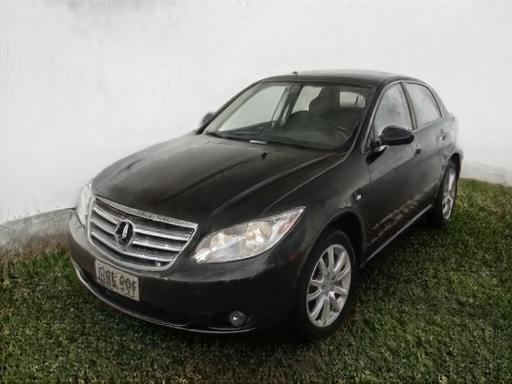} \\
            
           \raisebox{0.04\textwidth}{\texttt{A}} & 
            \includegraphics[height=0.1\textwidth]{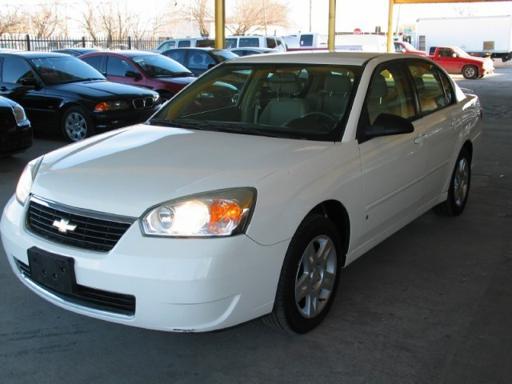} &
            \includegraphics[height=0.1\textwidth]{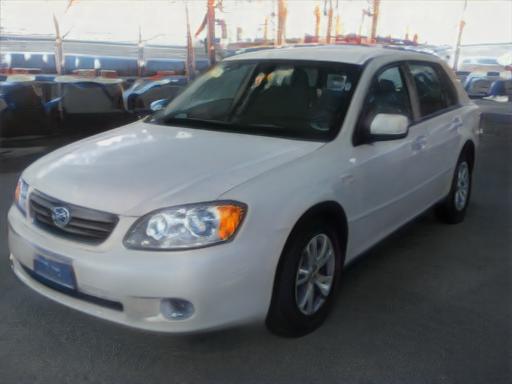} &
            \includegraphics[height=0.1\textwidth]{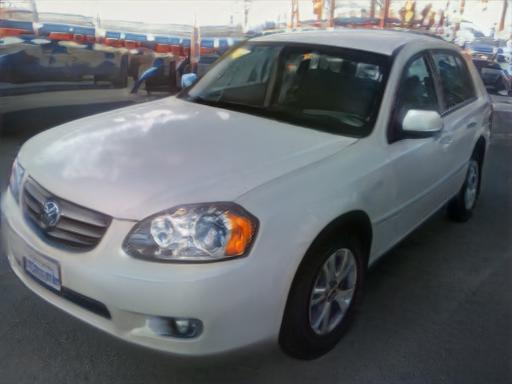} &
            \includegraphics[height=0.1\textwidth]{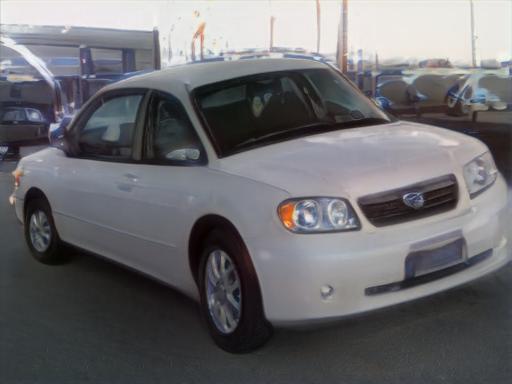} &
            \includegraphics[height=0.1\textwidth]{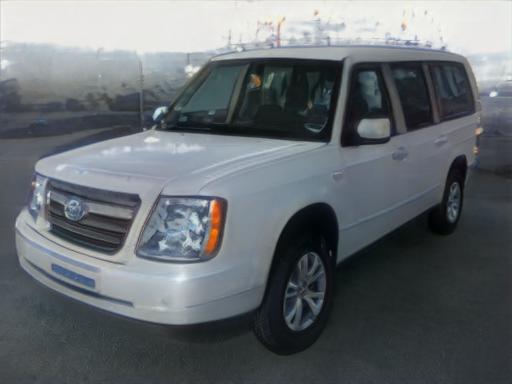} &
            \includegraphics[height=0.1\textwidth]{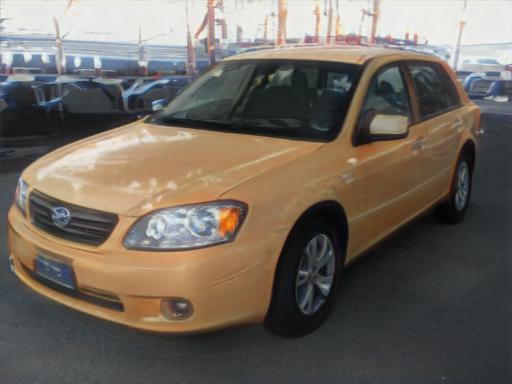} &
            \includegraphics[height=0.1\textwidth]{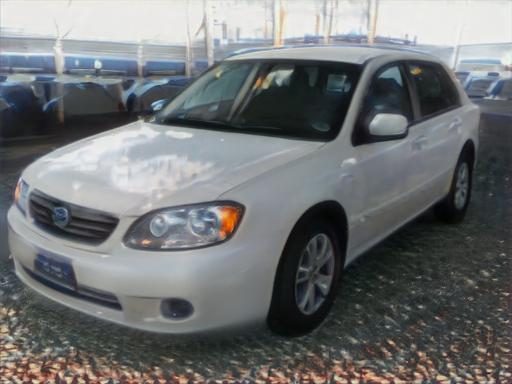} \\
           \raisebox{0.04\textwidth}{\texttt{D}} &
            \includegraphics[height=0.1\textwidth]{images/appendix/cars/00290_src.jpg} &
            \includegraphics[height=0.1\textwidth]{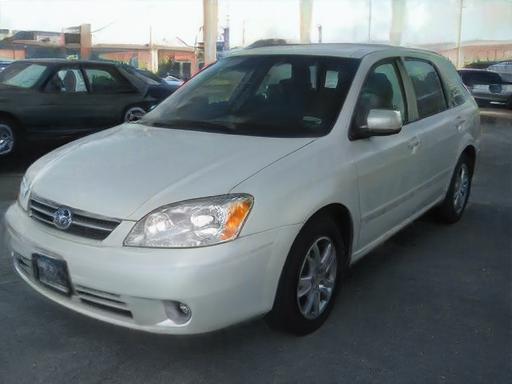} &
            \includegraphics[height=0.1\textwidth]{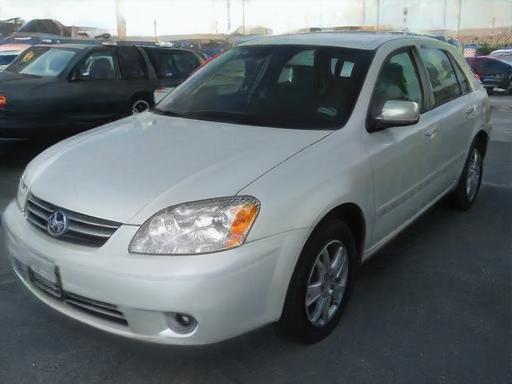} &
            \includegraphics[height=0.1\textwidth]{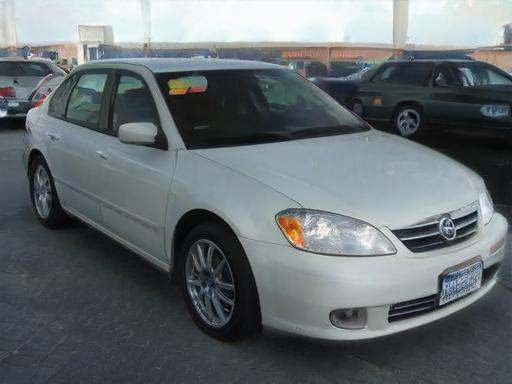} &
            \includegraphics[height=0.1\textwidth]{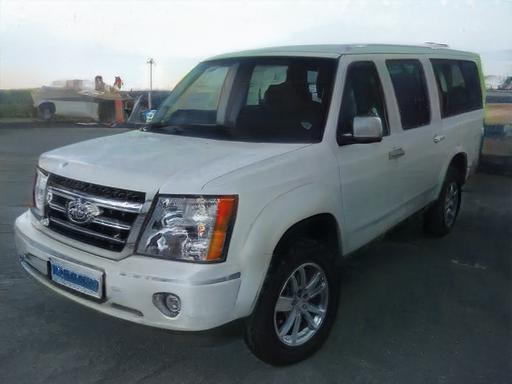} &
            \includegraphics[height=0.1\textwidth]{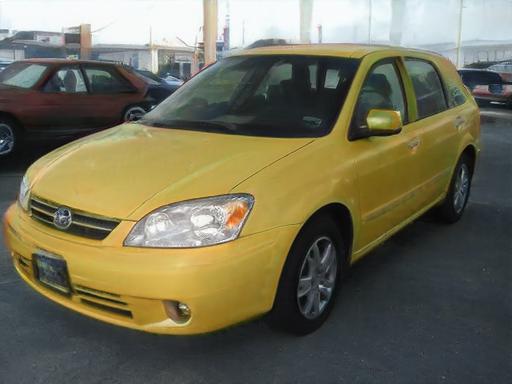} &
            \includegraphics[height=0.1\textwidth]{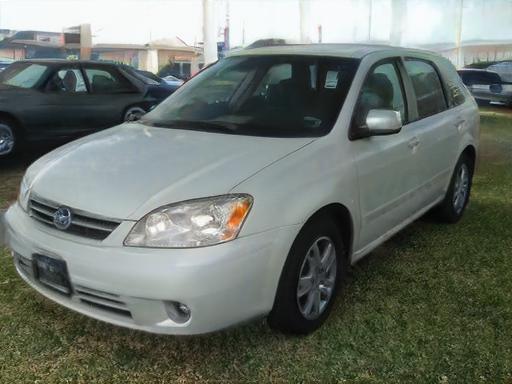} \\
            
           \raisebox{0.04\textwidth}{\texttt{A}} & 
            \includegraphics[height=0.1\textwidth]{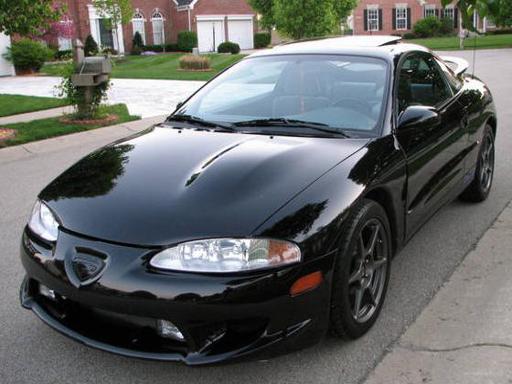} &
            \includegraphics[height=0.1\textwidth]{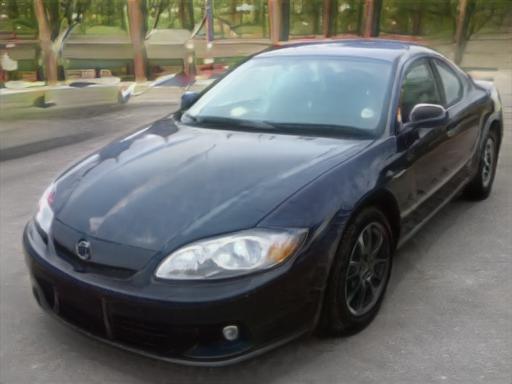} &
            \includegraphics[height=0.1\textwidth]{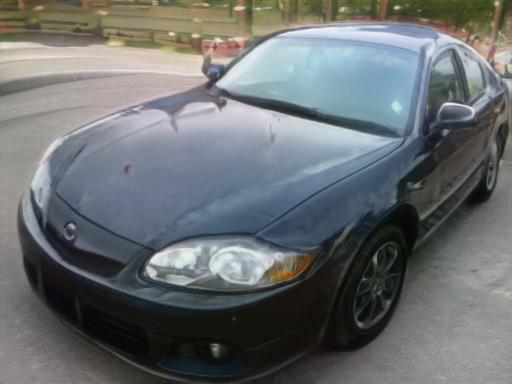} &
            \includegraphics[height=0.1\textwidth]{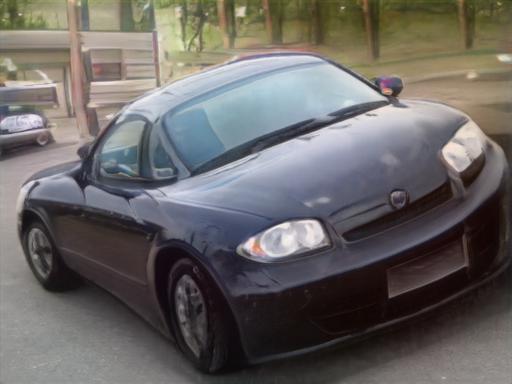} &
            \includegraphics[height=0.1\textwidth]{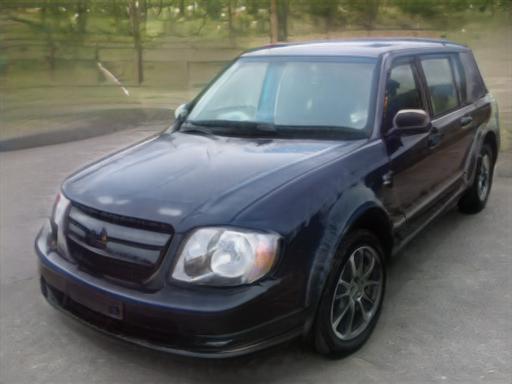} &
            \includegraphics[height=0.1\textwidth]{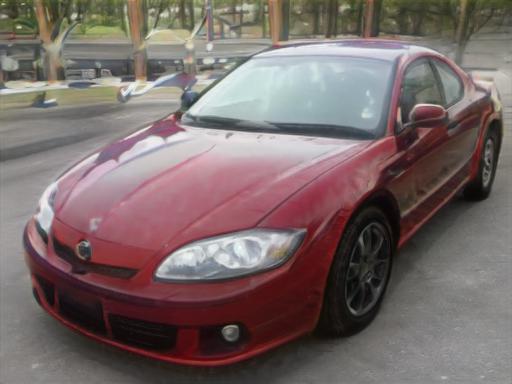} &
            \includegraphics[height=0.1\textwidth]{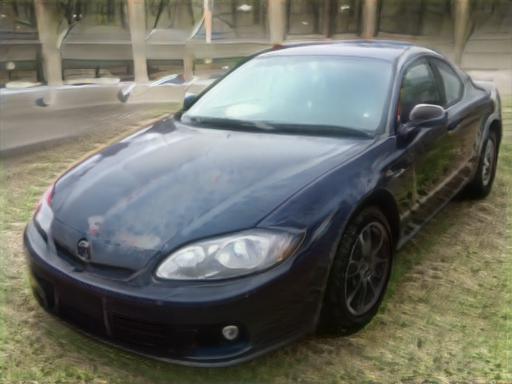} \\
           \raisebox{0.04\textwidth}{\texttt{D}} &
            \includegraphics[height=0.1\textwidth]{images/appendix/cars/00320_src.jpg} &
            \includegraphics[height=0.1\textwidth]{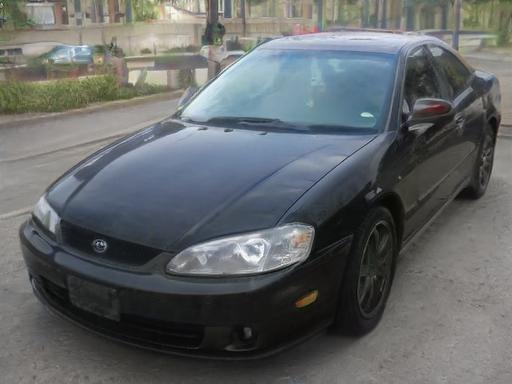} &
            \includegraphics[height=0.1\textwidth]{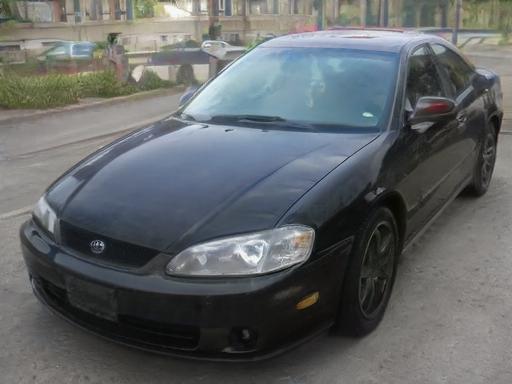} &
            \includegraphics[height=0.1\textwidth]{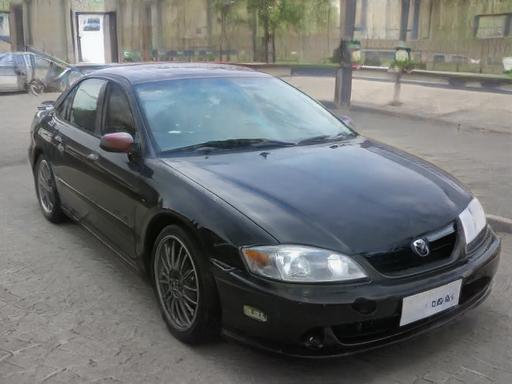} &
            \includegraphics[height=0.1\textwidth]{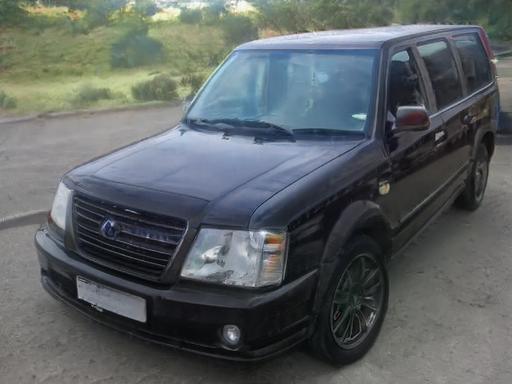} &
            \includegraphics[height=0.1\textwidth]{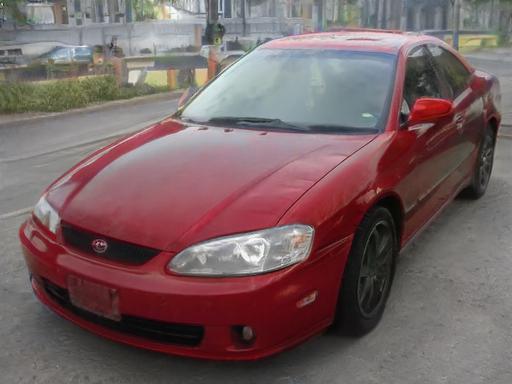} &
            \includegraphics[height=0.1\textwidth]{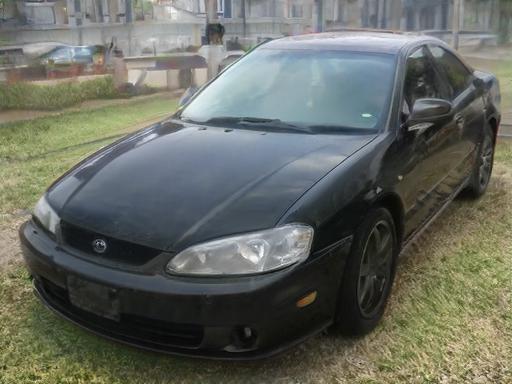} \\
            
           \raisebox{0.04\textwidth}{\texttt{A}} & 
            \includegraphics[height=0.1\textwidth]{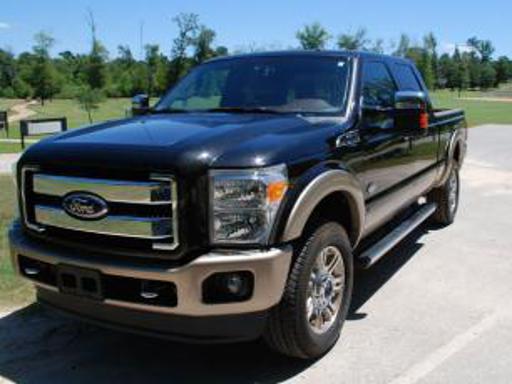} &
            \includegraphics[height=0.1\textwidth]{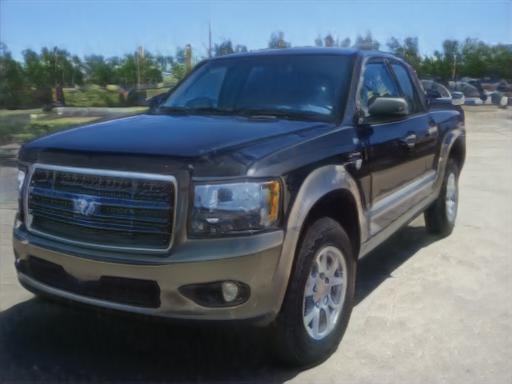} &
            \includegraphics[height=0.1\textwidth]{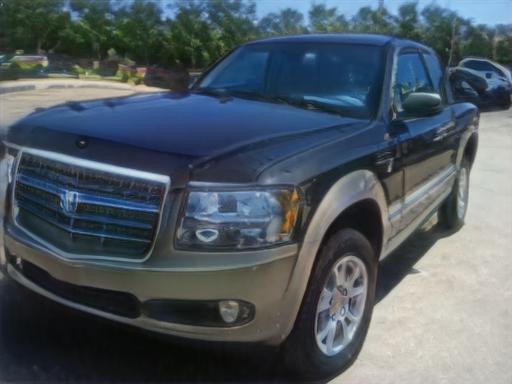} &
            \includegraphics[height=0.1\textwidth]{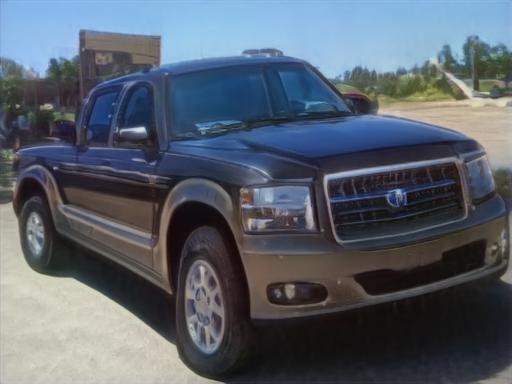} &
            \includegraphics[height=0.1\textwidth]{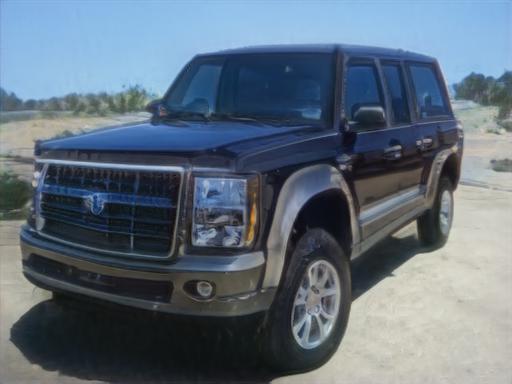} &
            \includegraphics[height=0.1\textwidth]{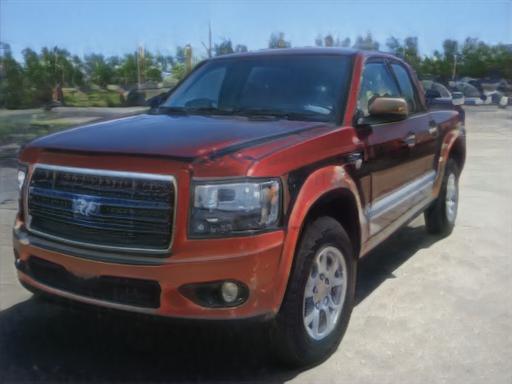} &
            \includegraphics[height=0.1\textwidth]{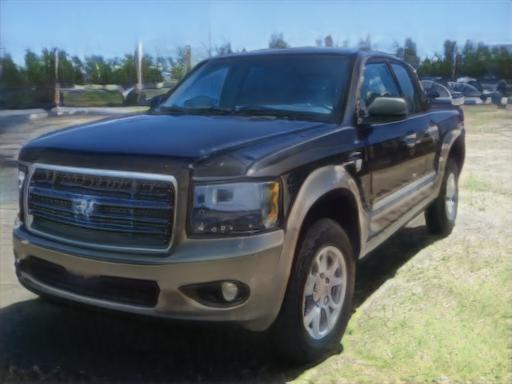} \\
           \raisebox{0.04\textwidth}{\texttt{D}} &
            \includegraphics[height=0.1\textwidth]{images/appendix/cars/00511_src.jpg} &
            \includegraphics[height=0.1\textwidth]{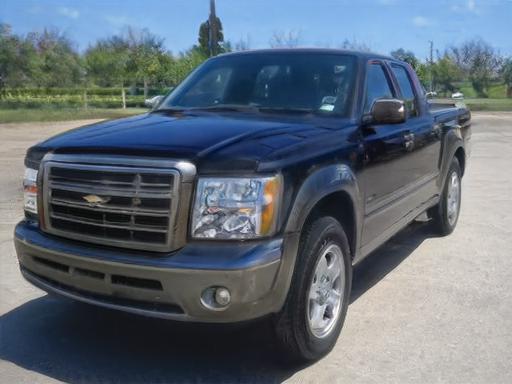} &
            \includegraphics[height=0.1\textwidth]{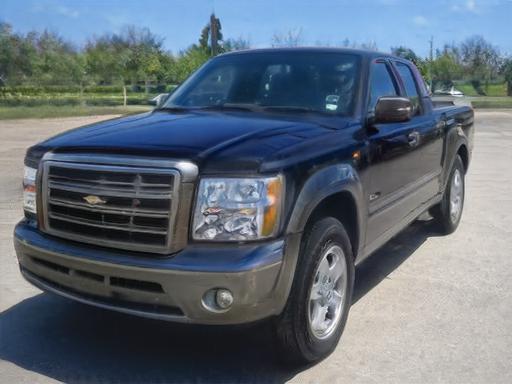} &
            \includegraphics[height=0.1\textwidth]{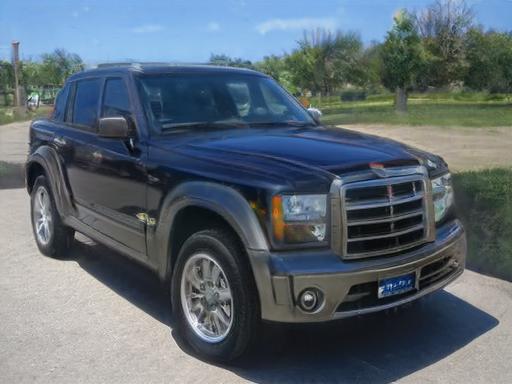} &
            \includegraphics[height=0.1\textwidth]{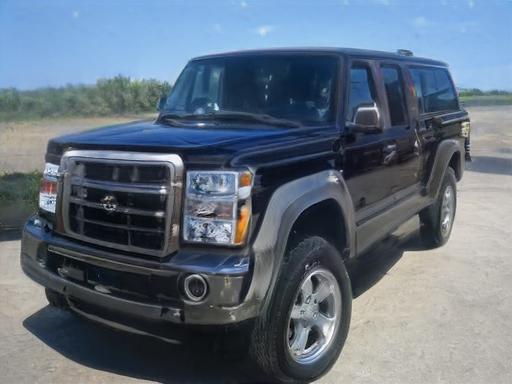} &
            \includegraphics[height=0.1\textwidth]{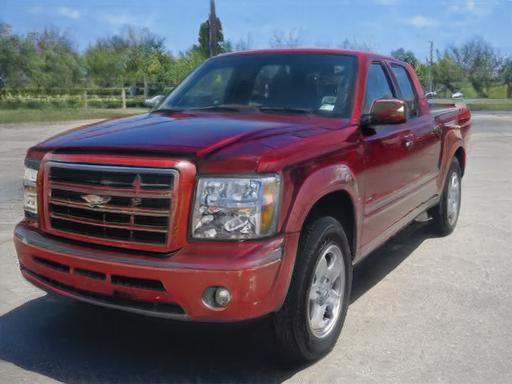} &
            \includegraphics[height=0.1\textwidth]{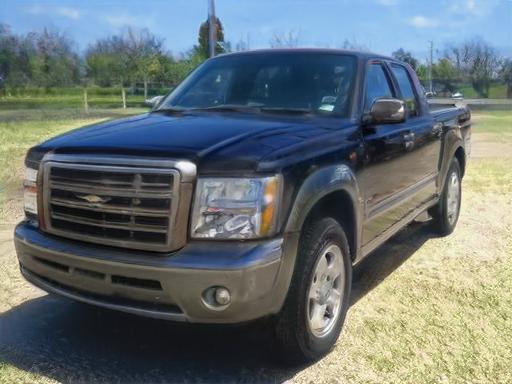} \\
            
             & Source & Inversion & Viewpoint I & Viewpoint II & Cube & Color & Grass \\
            \end{tabular}
    \caption{Additional comparison of configurations \texttt{A} and \texttt{D}, following the same format as Figure \ref{fig:tradeoff-cars}.}
    \label{fig:tradeoff-cars-2}
\end{figure*}

We provide additional figures in this document. See \Cref{fig:internet_celebs_1,fig:internet_celebs_2,fig:tradeoff-cars,fig:tradeoff-cars-2}. Observe, that our method achieves significantly superior editing in terms of perceptual quality and expressiveness. 

\section{The Continuous Distortion-Editability Tradeoff}
\begin{figure*}
    \setlength{\tabcolsep}{1pt}
    \centering
        \centering
            \begin{tabular}{c c c c c c c c}
            
            \raisebox{0.05\textwidth}{\texttt{1}} &
            \includegraphics[width=0.13\textwidth]{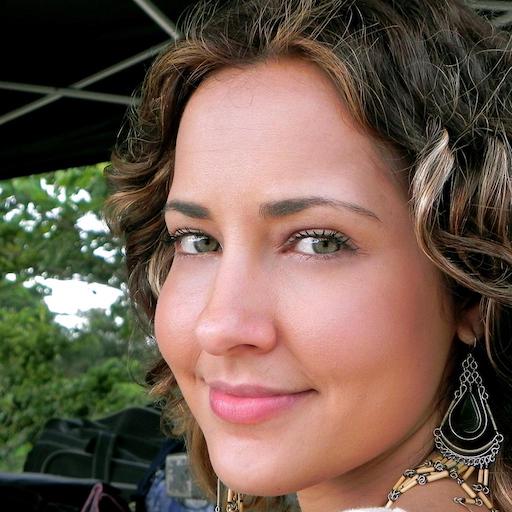} &
            \includegraphics[width=0.13\textwidth]{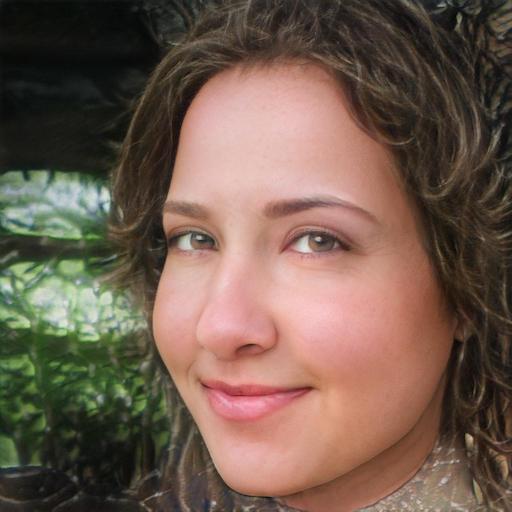} &
            \includegraphics[width=0.13\textwidth]{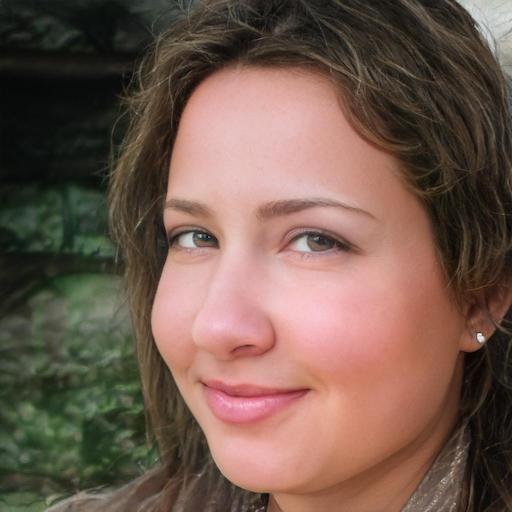} &
            \includegraphics[width=0.13\textwidth]{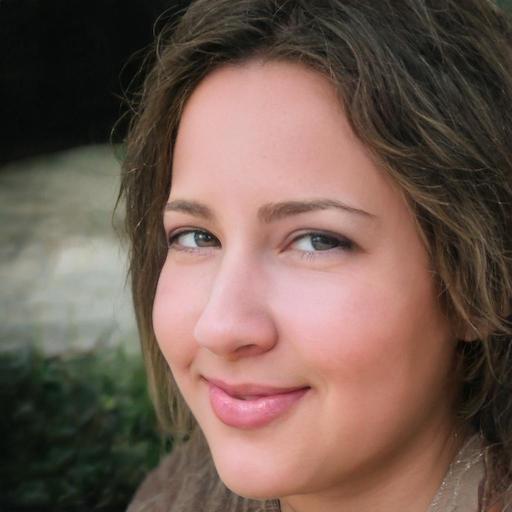} &
            \includegraphics[width=0.13\textwidth]{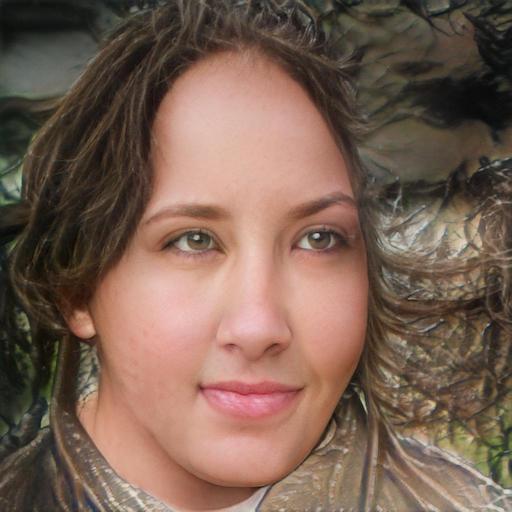} &
            \includegraphics[width=0.13\textwidth]{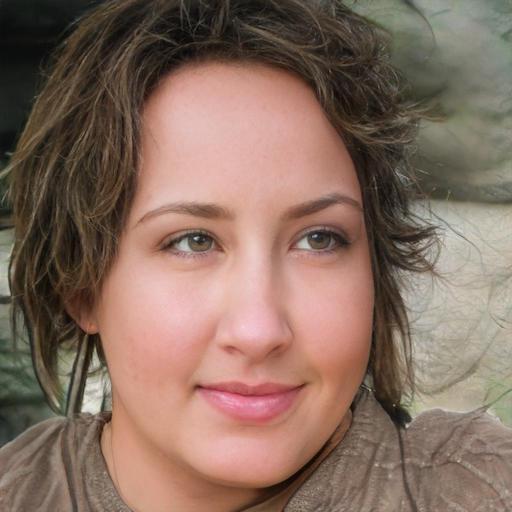} &
            \includegraphics[width=0.13\textwidth]{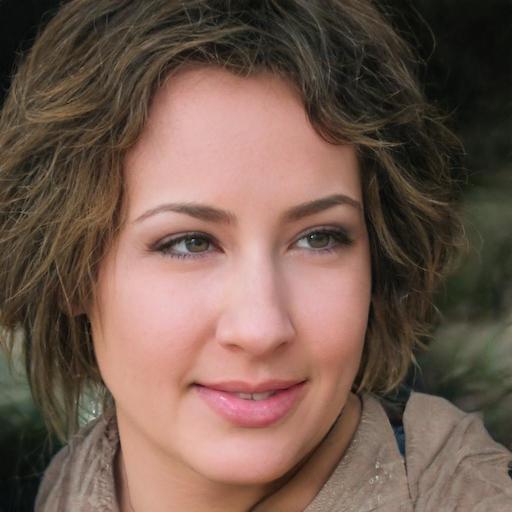} \\
            \raisebox{0.05\textwidth}{\texttt{2}} &
            \includegraphics[width=0.13\textwidth]{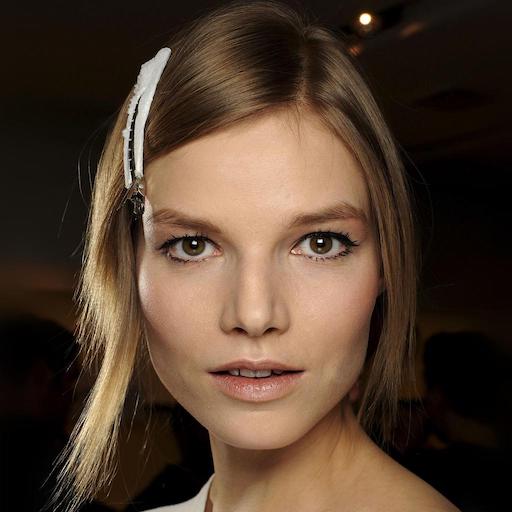} &
            \includegraphics[width=0.13\textwidth]{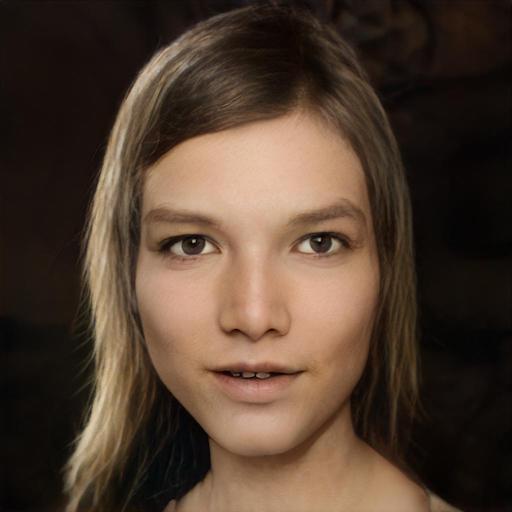} &
            \includegraphics[width=0.13\textwidth]{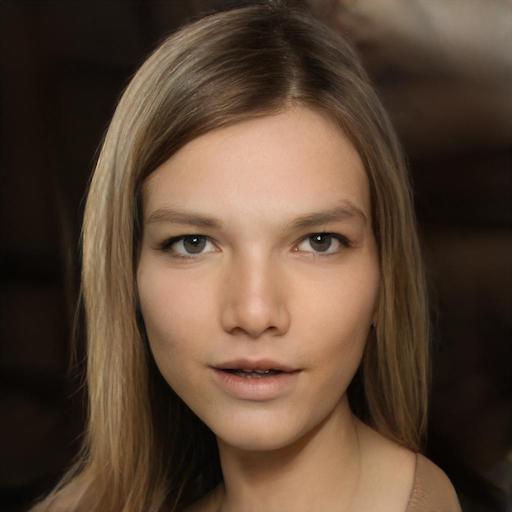} &
            \includegraphics[width=0.13\textwidth]{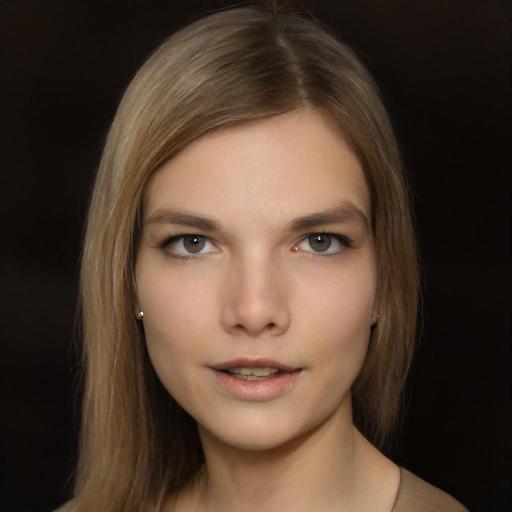} &
            \includegraphics[width=0.13\textwidth]{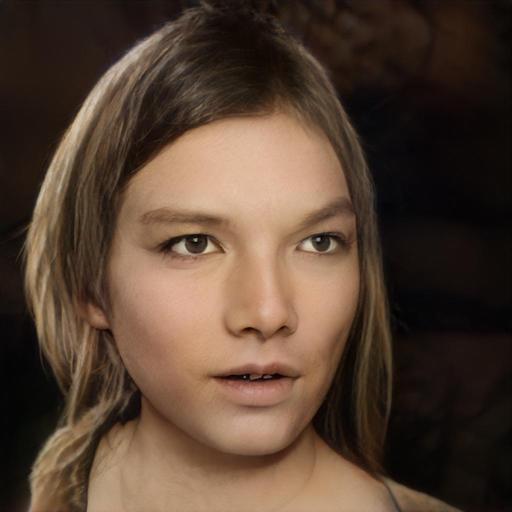} &
            \includegraphics[width=0.13\textwidth]{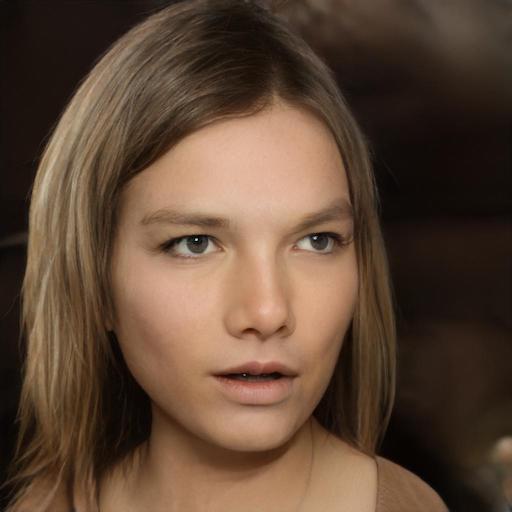} &
            \includegraphics[width=0.13\textwidth]{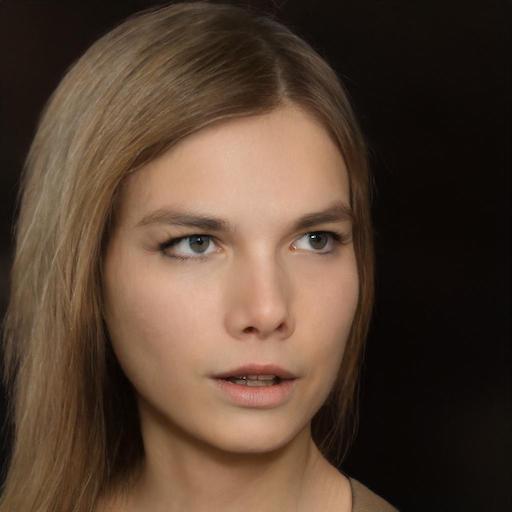} \\
            \raisebox{0.05\textwidth}{\texttt{3}} &
            \includegraphics[width=0.13\textwidth]{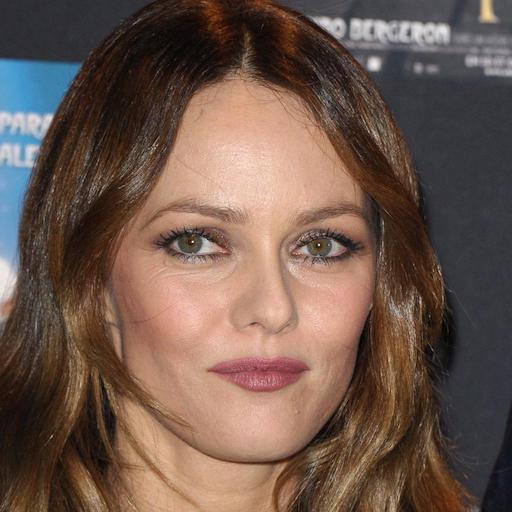} &
            \includegraphics[width=0.13\textwidth]{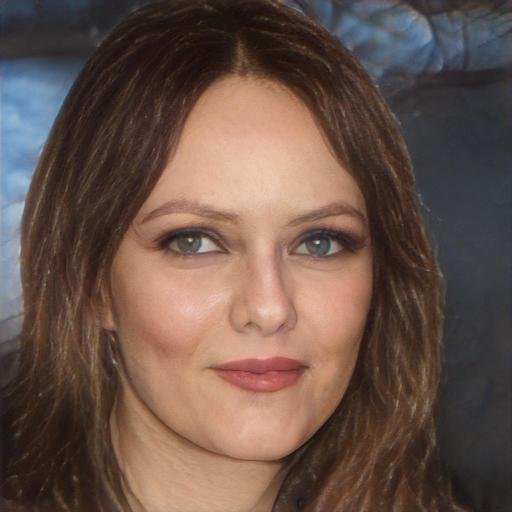} &
            \includegraphics[width=0.13\textwidth]{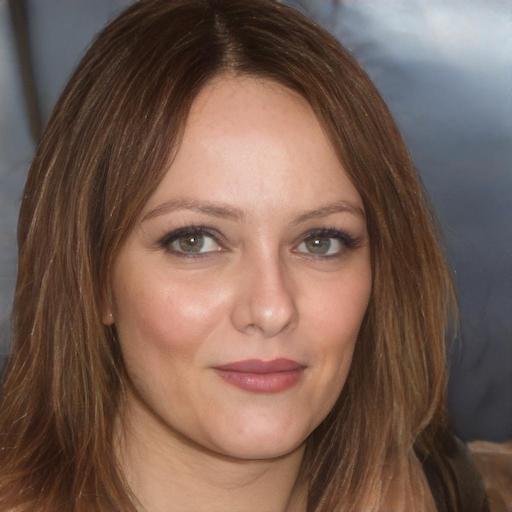} &
            \includegraphics[width=0.13\textwidth]{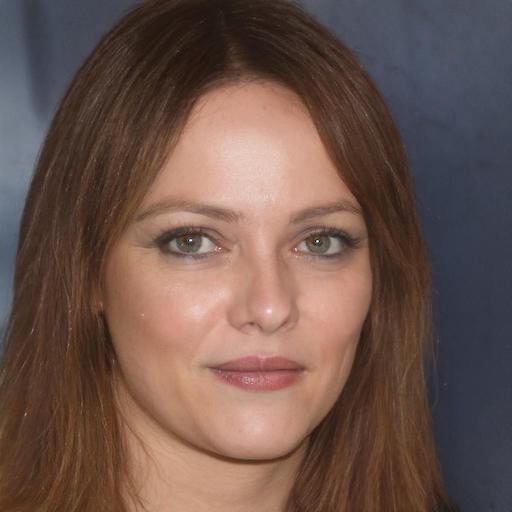} &
            \includegraphics[width=0.13\textwidth]{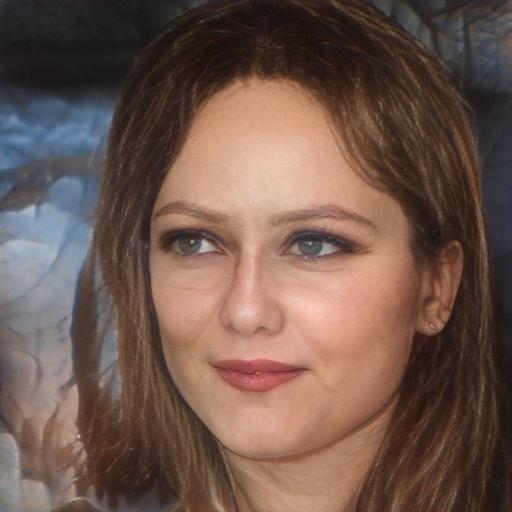} &
            \includegraphics[width=0.13\textwidth]{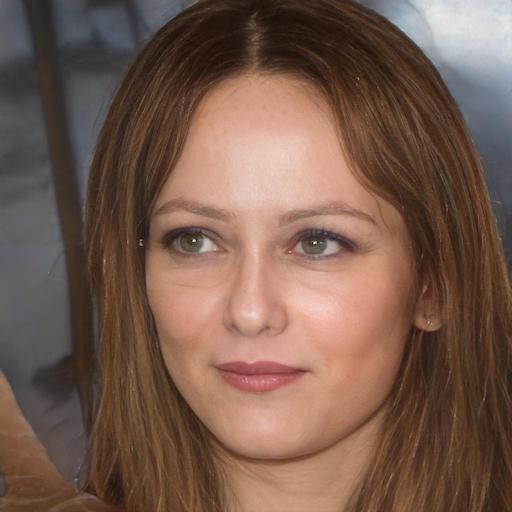} &
            \includegraphics[width=0.13\textwidth]{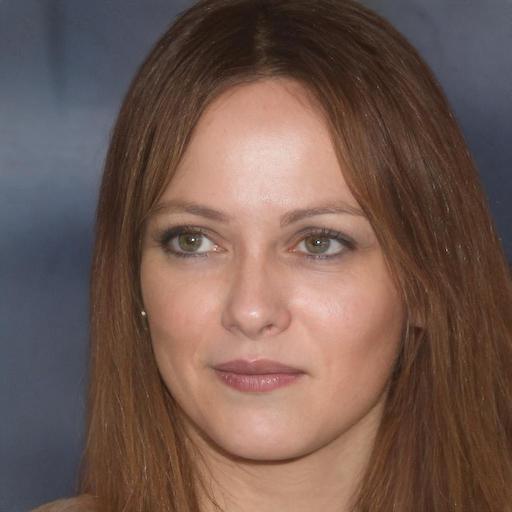} \\
            \raisebox{0.05\textwidth}{\texttt{4}} &
            \includegraphics[width=0.13\textwidth]{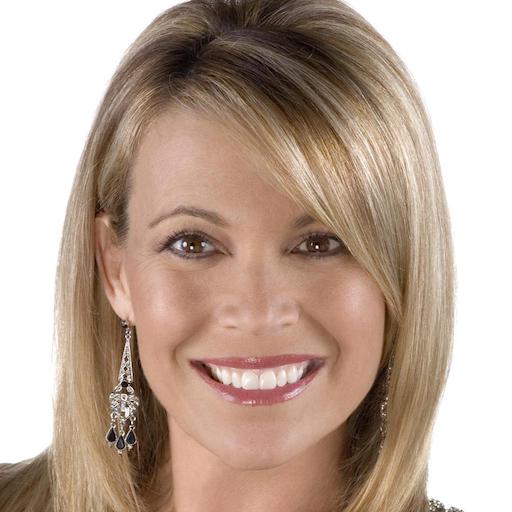} &
            \includegraphics[width=0.13\textwidth]{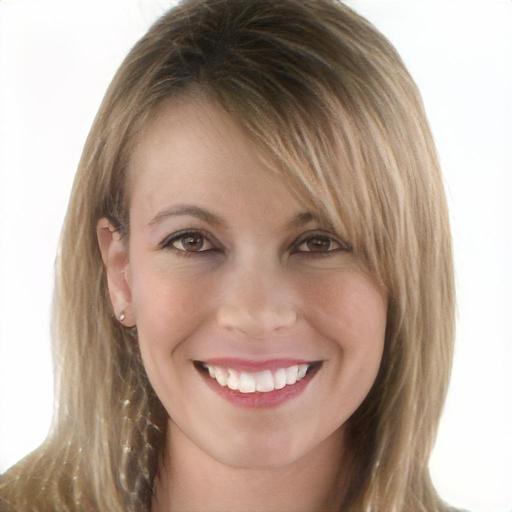} &
            \includegraphics[width=0.13\textwidth]{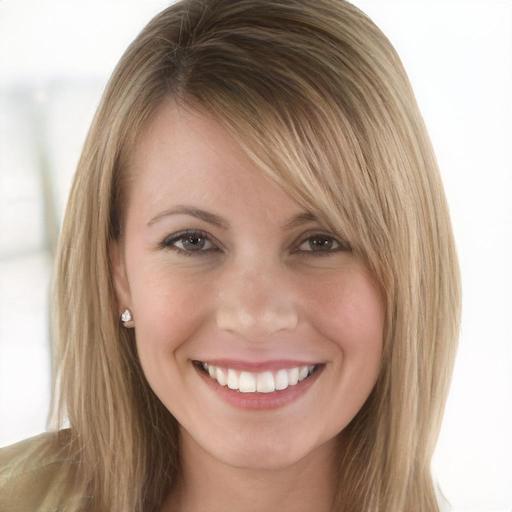} &
            \includegraphics[width=0.13\textwidth]{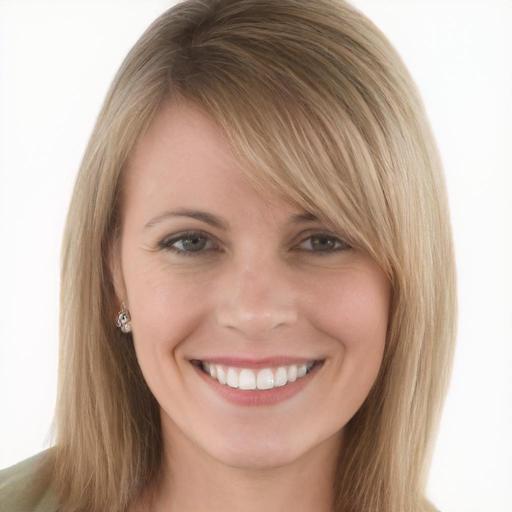} &
            \includegraphics[width=0.13\textwidth]{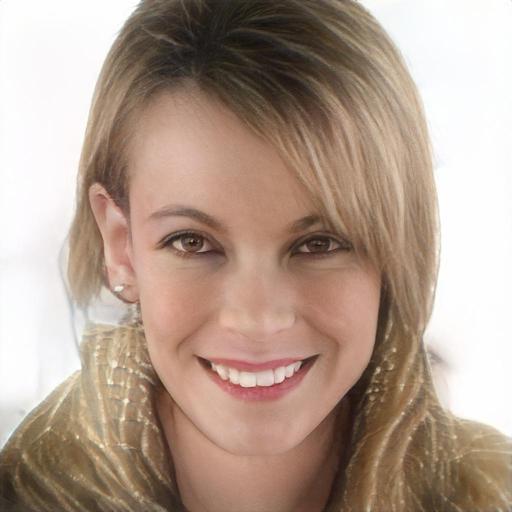} &
            \includegraphics[width=0.13\textwidth]{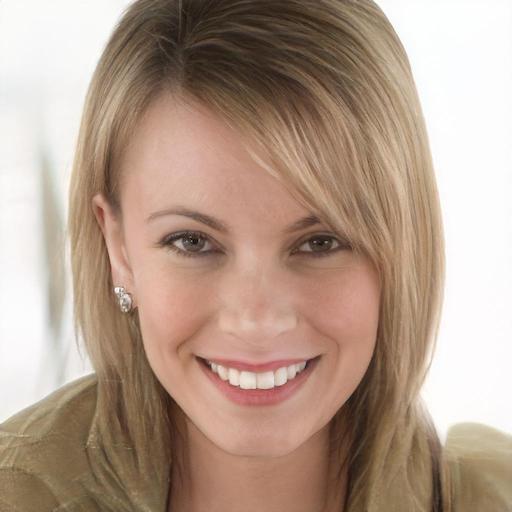} &
            \includegraphics[width=0.13\textwidth]{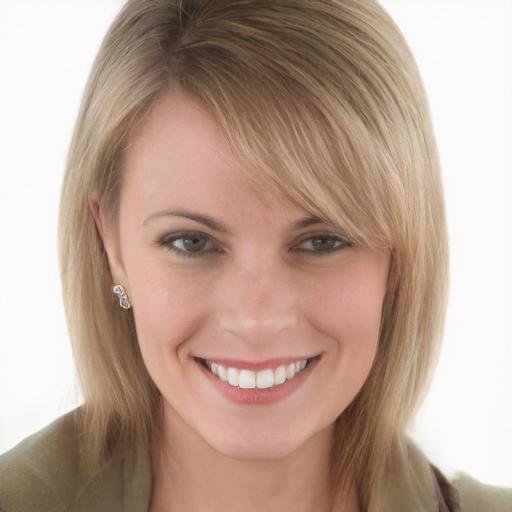} \\
            \raisebox{0.05\textwidth}{\texttt{5}} &
            \includegraphics[width=0.13\textwidth]{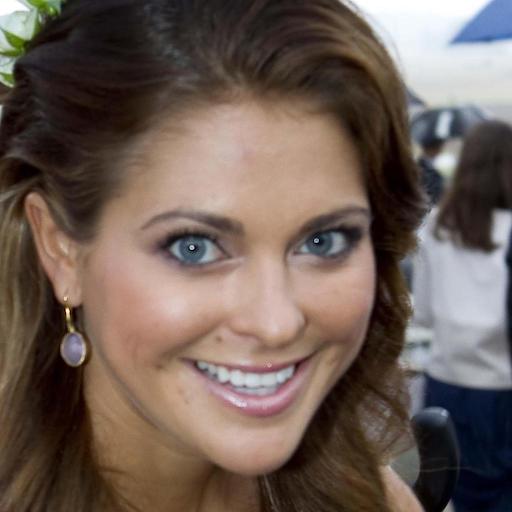} &
            \includegraphics[width=0.13\textwidth]{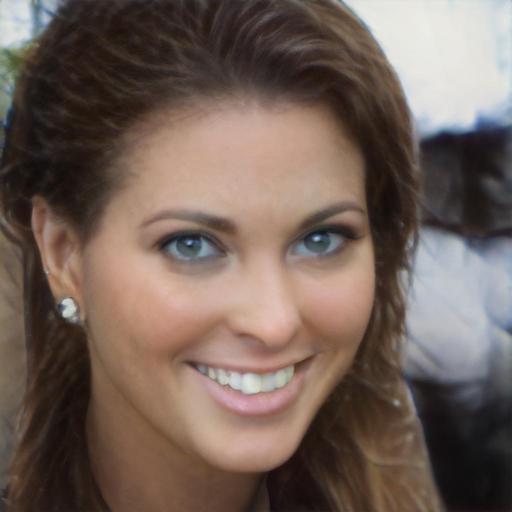} &
            \includegraphics[width=0.13\textwidth]{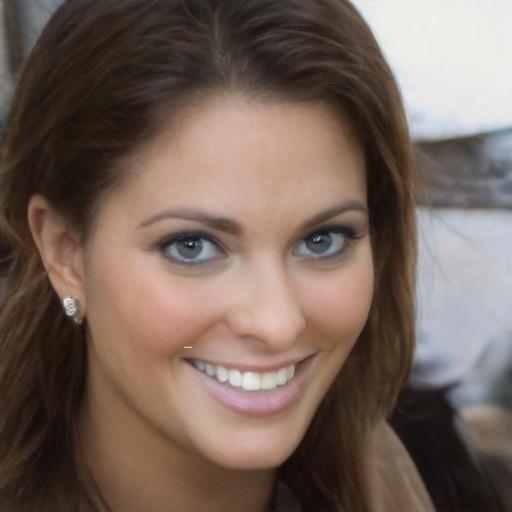} &
            \includegraphics[width=0.13\textwidth]{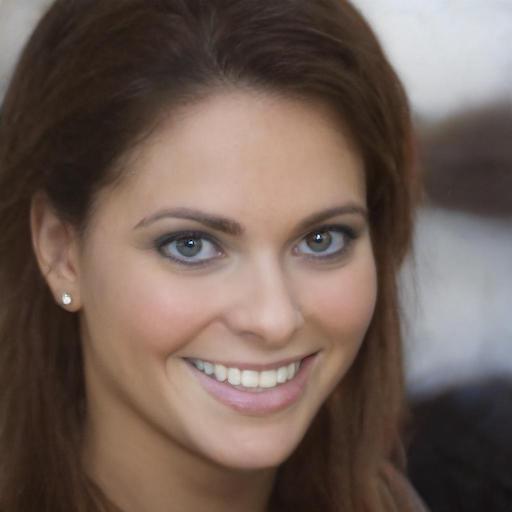} &
            \includegraphics[width=0.13\textwidth]{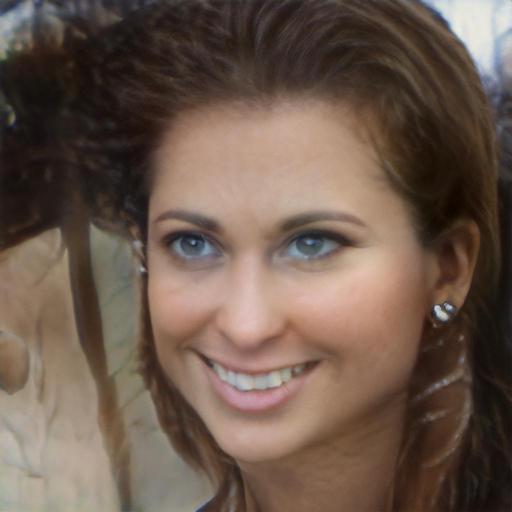} &
            \includegraphics[width=0.13\textwidth]{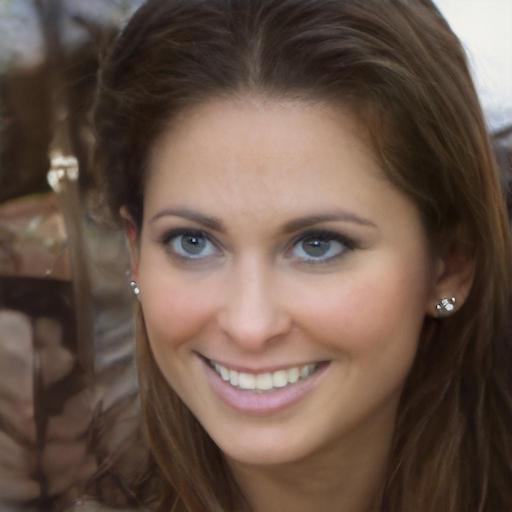} &
            \includegraphics[width=0.13\textwidth]{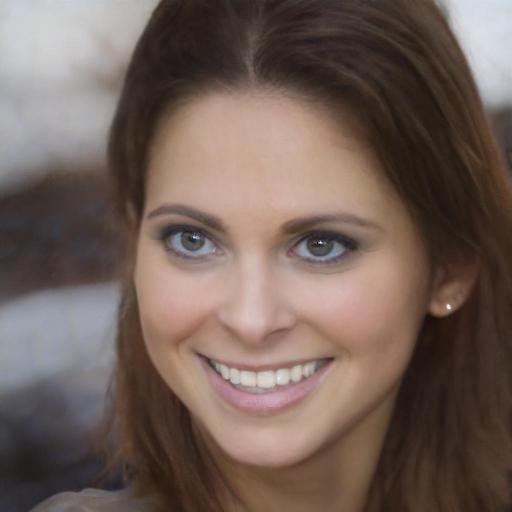} \\
            \raisebox{0.05\textwidth}{\texttt{6}} &
            \includegraphics[width=0.13\textwidth]{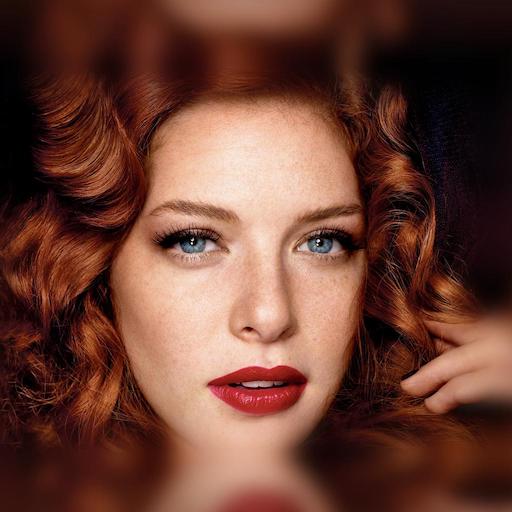} &
            \includegraphics[width=0.13\textwidth]{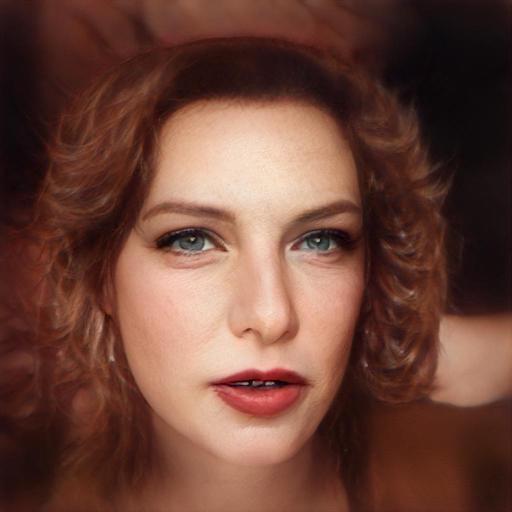} &
            \includegraphics[width=0.13\textwidth]{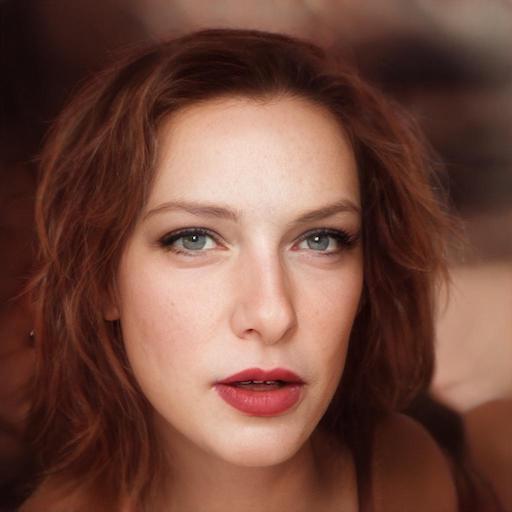} &
            \includegraphics[width=0.13\textwidth]{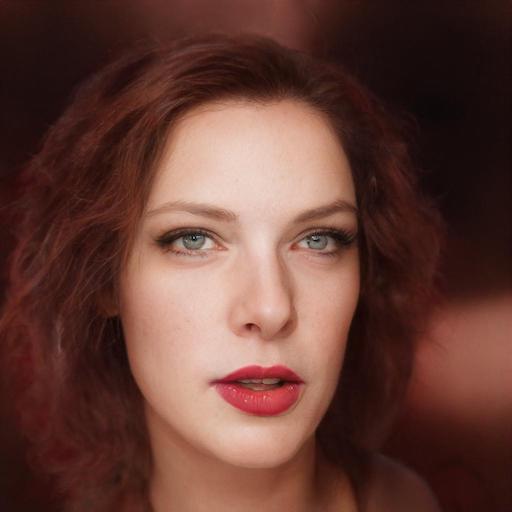} &
            \includegraphics[width=0.13\textwidth]{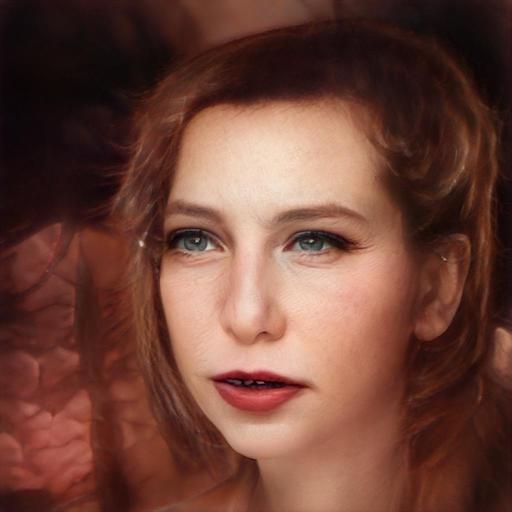} &
            \includegraphics[width=0.13\textwidth]{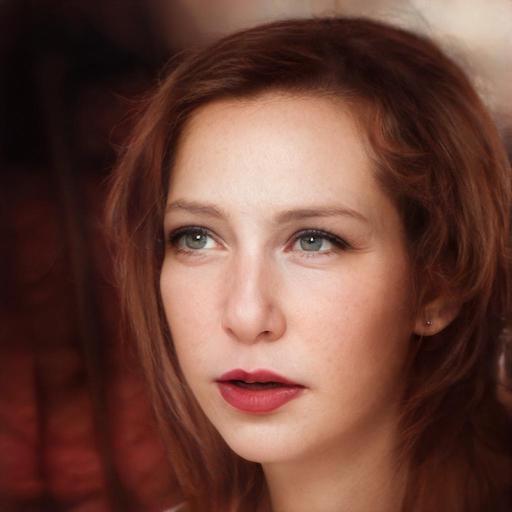} &
            \includegraphics[width=0.13\textwidth]{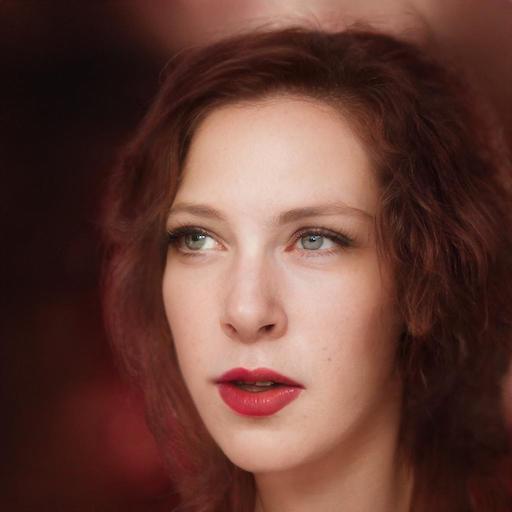} \\
            \raisebox{0.05\textwidth}{\texttt{7}} &
            \includegraphics[width=0.13\textwidth]{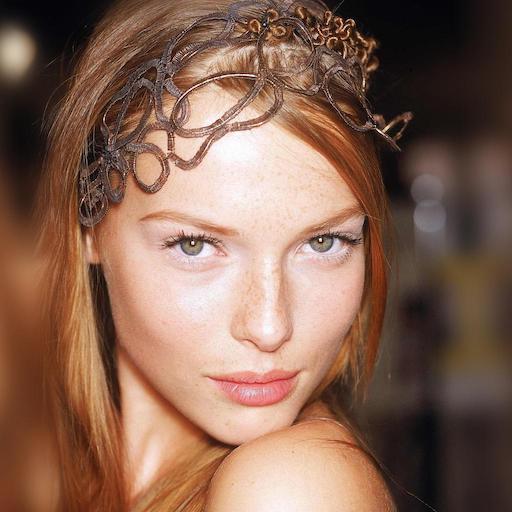} &
            \includegraphics[width=0.13\textwidth]{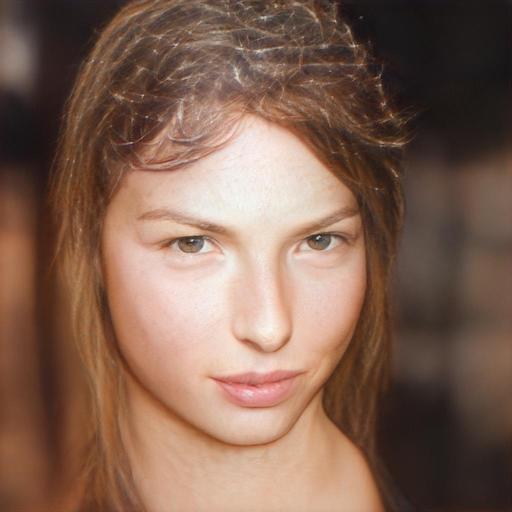} &
            \includegraphics[width=0.13\textwidth]{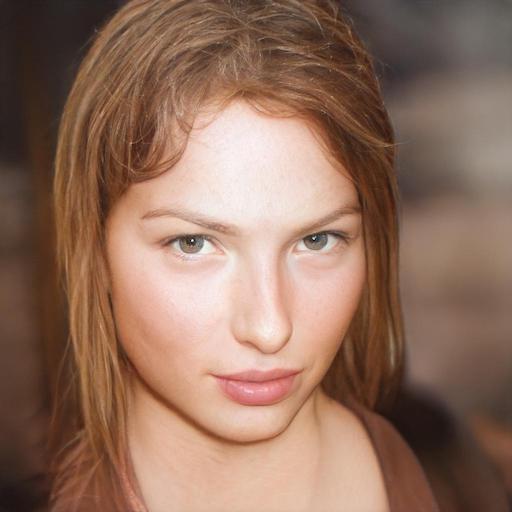} &
            \includegraphics[width=0.13\textwidth]{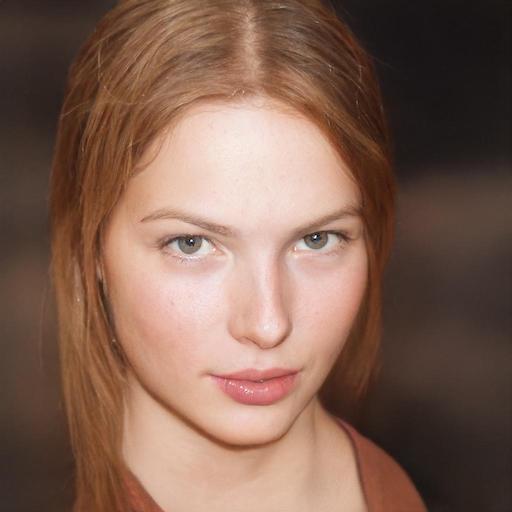} &
            \includegraphics[width=0.13\textwidth]{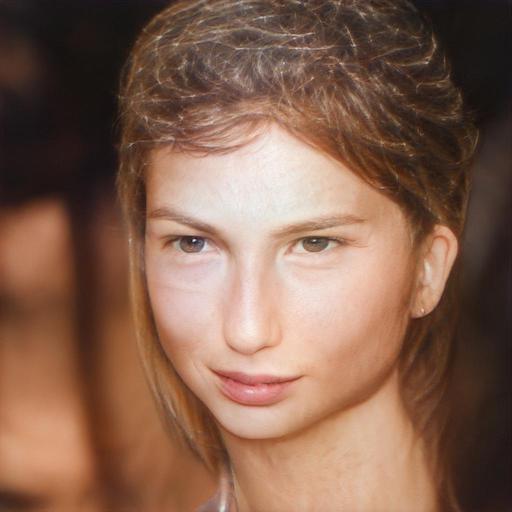} &
            \includegraphics[width=0.13\textwidth]{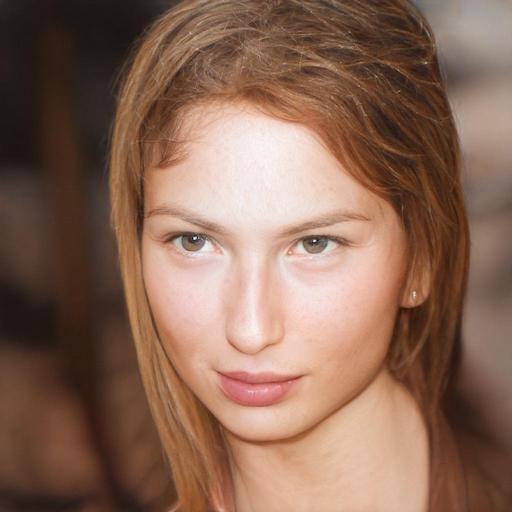} &
            \includegraphics[width=0.13\textwidth]{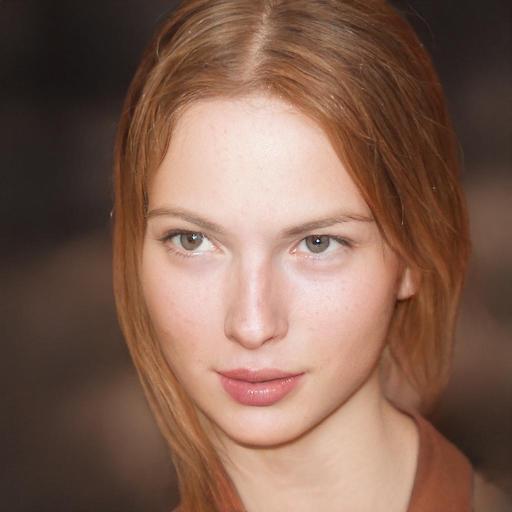} \\
            \raisebox{0.05\textwidth}{\texttt{8}} &
            \includegraphics[width=0.13\textwidth]{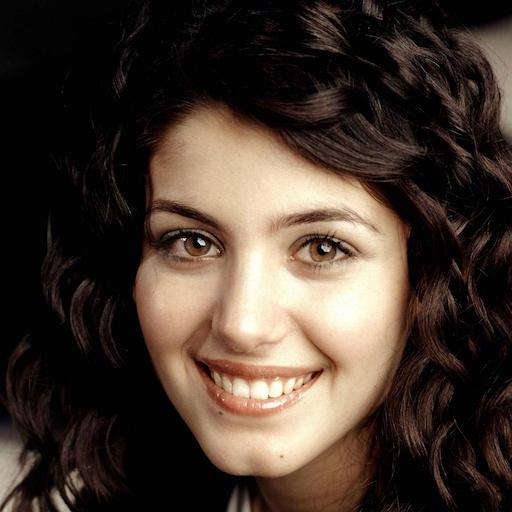} &
            \includegraphics[width=0.13\textwidth]{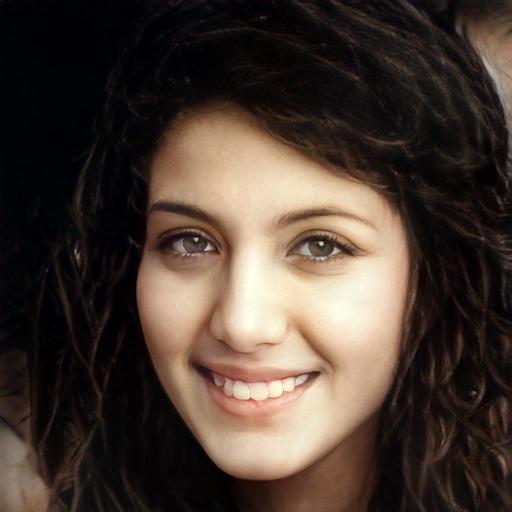} &
            \includegraphics[width=0.13\textwidth]{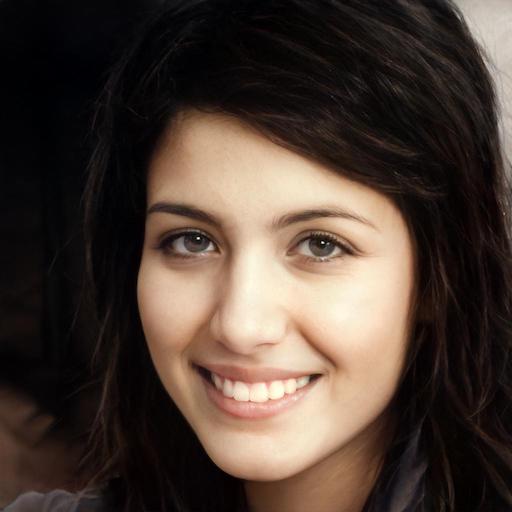} &
            \includegraphics[width=0.13\textwidth]{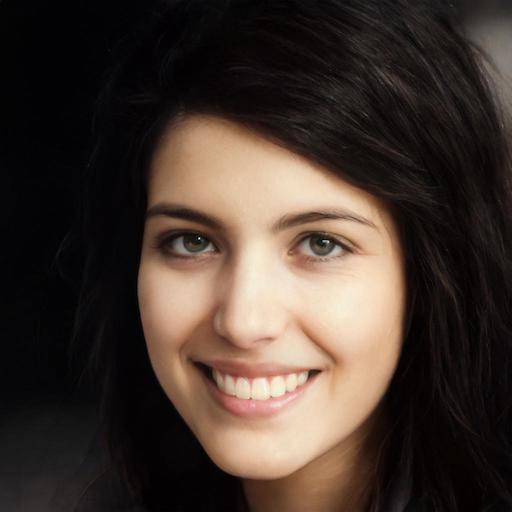} &
            \includegraphics[width=0.13\textwidth]{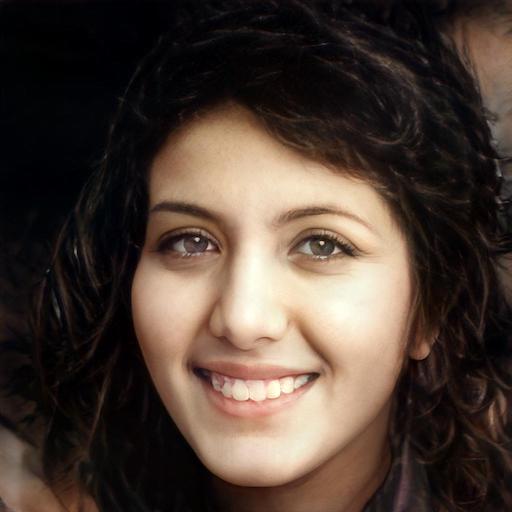} &
            \includegraphics[width=0.13\textwidth]{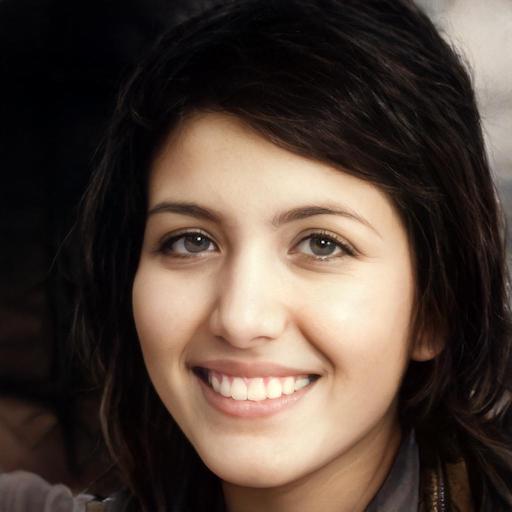} &
            \includegraphics[width=0.13\textwidth]{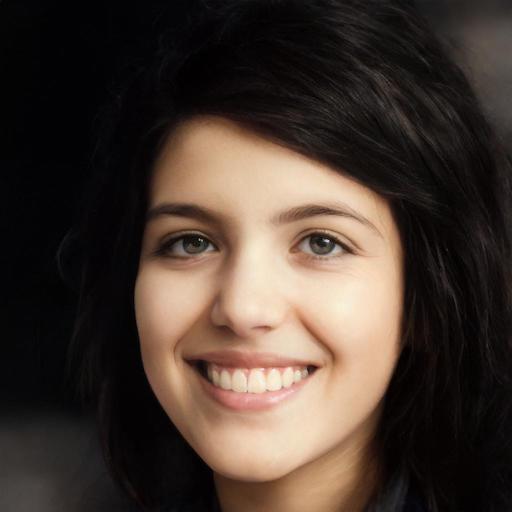} \\
            
            & Source & 
            \begin{tabular}{@{}c@{}}pSp \\ Inversion\end{tabular} &
            \begin{tabular}{@{}c@{}}Inversion \\ Interpolation \end{tabular} &
            \begin{tabular}{@{}c@{}}e4e \\ Inversion\end{tabular} &
            \begin{tabular}{@{}c@{}}pSp \\ Pose Edit\end{tabular} &
            \begin{tabular}{@{}c@{}}Interpolation \\ Pose Edit\end{tabular} &
            \begin{tabular}{@{}c@{}}e4e \\ Pose Edit\end{tabular} 

            \end{tabular}
    \caption{Displaying the continuous nature of the distortion-editability tradeoff on the first $8$ images of CelebA-HQ~\cite{karras2017progressive} test set. The leftmost column is the source, the second and fourth columns are inversions obtained by pSp and e4e, respectively. The third column is the mid-point interpolation between the inversions. The three rightmost columns are pose edits of the inversions and interpolated inversion performed using StyleFlow~\cite{abdal2020styleflow}. Note that, the proximity to the pSp or e4e inversion controls where the latent code lies on the tradeoff curve.}
    \label{fig:continuous_celeba}
\end{figure*}

\begin{figure*}
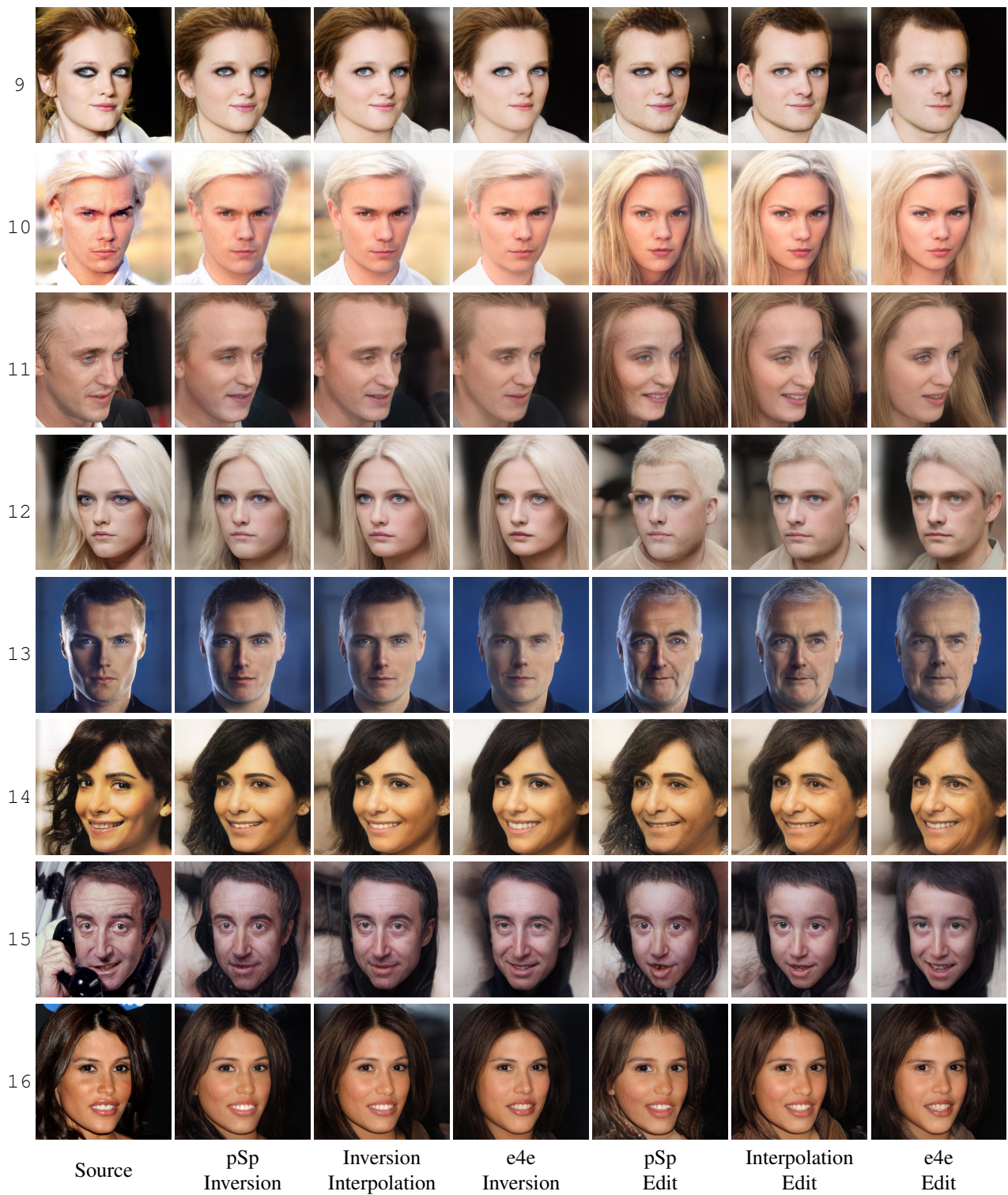


    \setlength{\tabcolsep}{1pt}
    \centering
        \centering

    \caption{Similar to Figure~\ref{fig:continuous_celeba}, displaying the next $8$ images from the CelebA-HQ~\cite{karras2017progressive} test set. The first four pairs of rows perform a gender edit while the last four pairs of rows perform an age edit.}
    \label{fig:next_ten_celeba}
\end{figure*}

\begin{figure*}
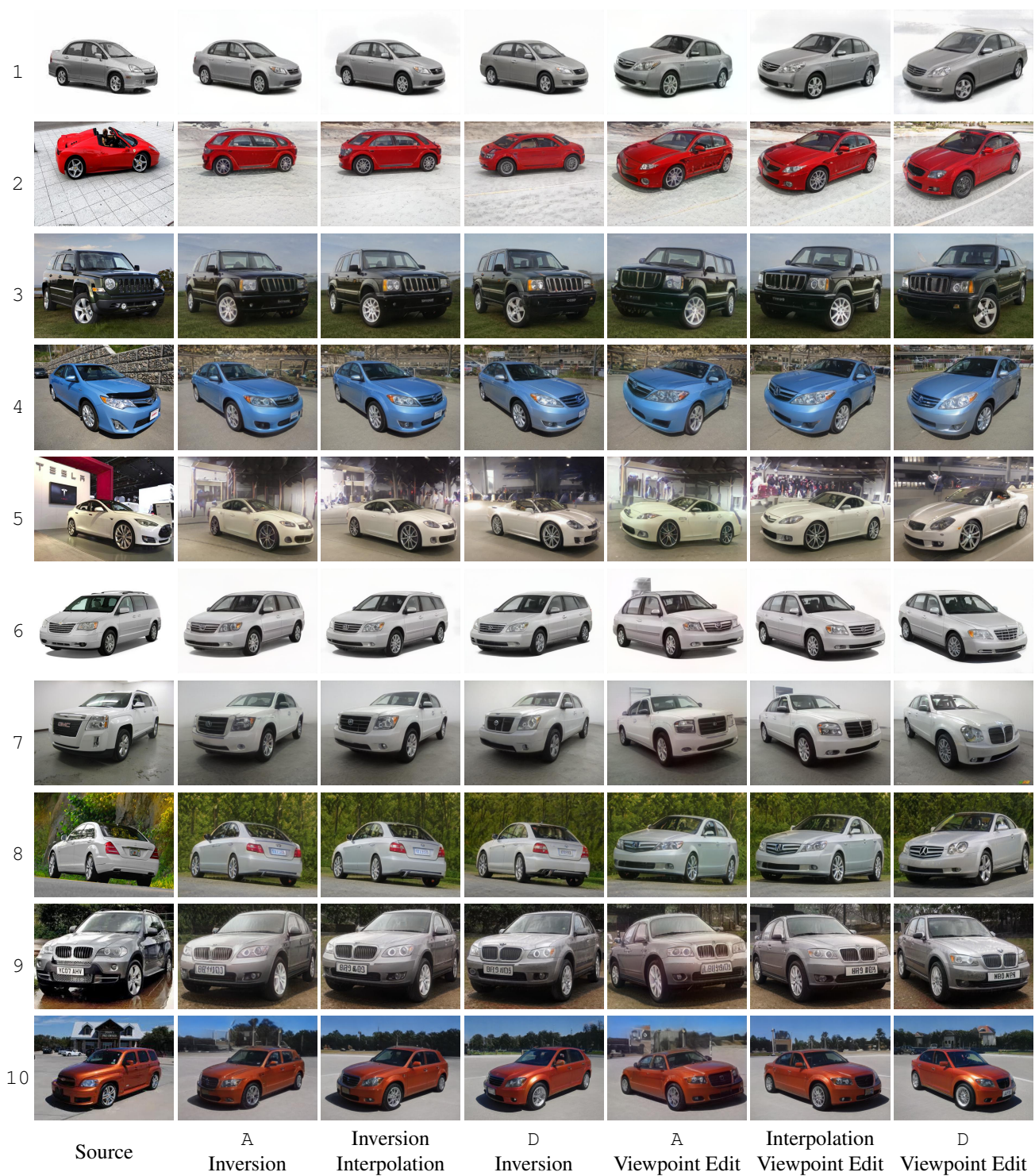

\label{fig:car_top_10}
    \setlength{\tabcolsep}{1pt}
    \centering
        \centering
            %
    \caption{Similar to Figure \ref{fig:continuous_celeba}, displaying the first $10$ images from the Stanford Cars~\cite{KrauseStarkDengFei-Fei_3DRR2013} test set. Here, we perform viewpoint edits using GANSpace~\cite{harkonen2020ganspace}.}
    \label{fig:appendix_first_cars_1}
\end{figure*}

\begin{figure*}
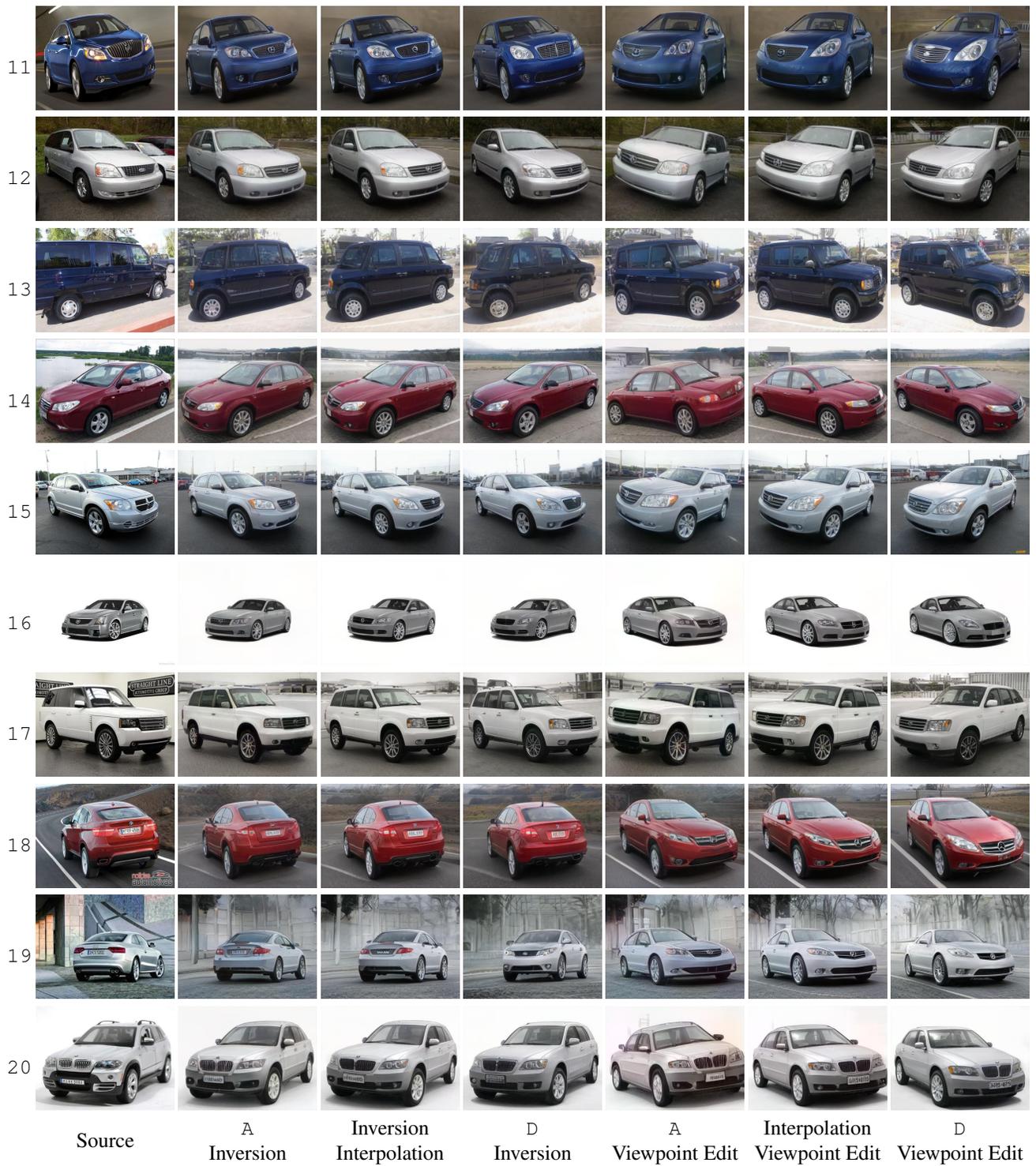


    \setlength{\tabcolsep}{1pt}
    \centering
        \centering
            %
    \caption{Similar to Figure \ref{fig:continuous_celeba}, displaying the next ten images from the Stanford Cars~\cite{KrauseStarkDengFei-Fei_3DRR2013} test set.}
    \label{fig:appendix_first_cars_2_}
\end{figure*}

We provide additional results, similar to Figure 10 in the main paper, where we demonstrate that the distortion-editability tradeoff and the fact that the location on the tradeoff curve can easily be controlled by interpolation in the latent space. To prevent the suspicion of cherry-picking, we provide uncurated results of the first samples in the CelebA-HQ~\cite{karras2017progressive} and Stanford Cars~\cite{KrauseStarkDengFei-Fei_3DRR2013} datasets. See \Cref{fig:continuous_celeba,fig:next_ten_celeba,fig:appendix_first_cars_1,fig:appendix_first_cars_2_}.

\section{LEC}
\begin{figure*}

    \setlength{\tabcolsep}{1pt}
    \centering
        \centering
            \begin{tabular}{c c c c c c}

            \qquad\qquad Source \qquad\qquad\qquad Inversion \qquad\qquad\qquad\quad Edit \qquad\qquad\qquad Edit's Inverse \qquad\qquad Inversion Editing \\

            \multicolumn{5}{c}{\includegraphics[width=\linewidth]{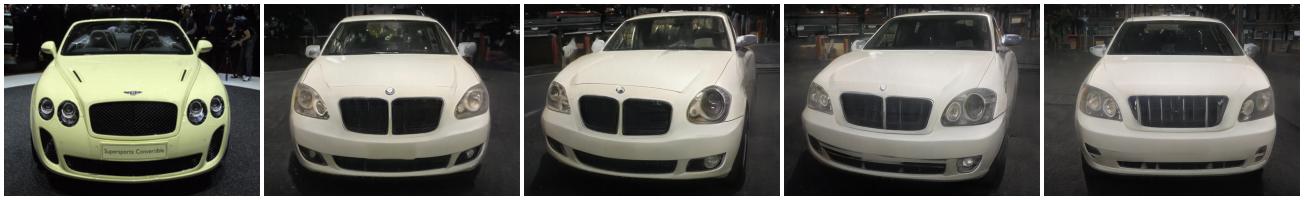} } \\
            \multicolumn{5}{c}{\includegraphics[width=\linewidth]{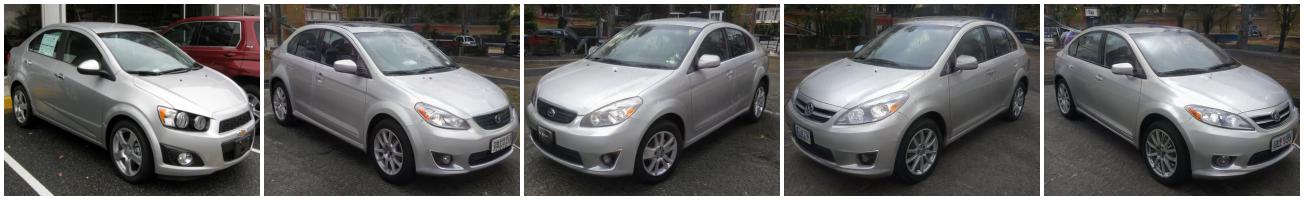} } \\
            \multicolumn{5}{c}{\includegraphics[width=\linewidth]{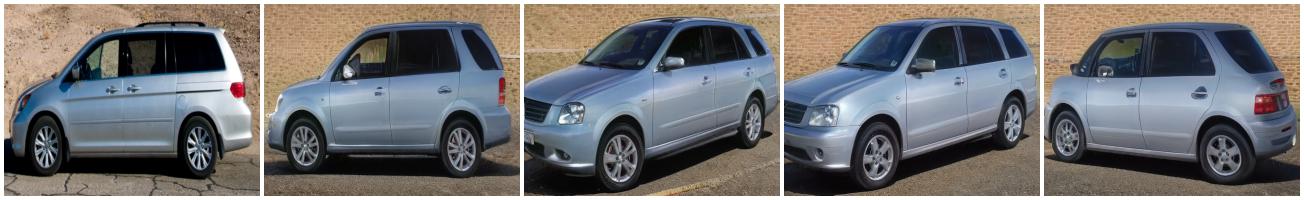} } \\
            \multicolumn{5}{c}{\includegraphics[width=\linewidth]{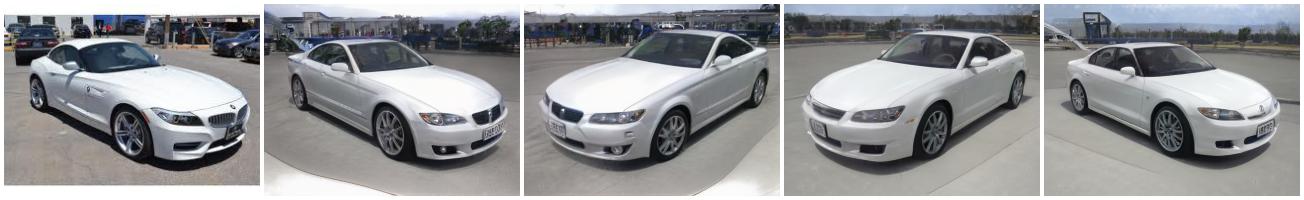} } \\
            \multicolumn{5}{c}{\includegraphics[width=\linewidth]{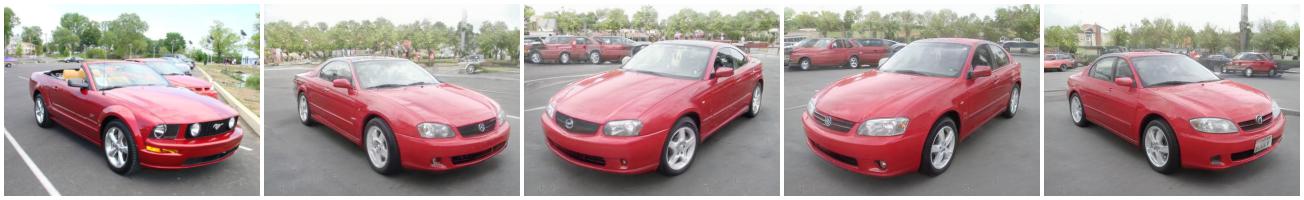} } \\
            \multicolumn{5}{c}{\includegraphics[width=\linewidth]{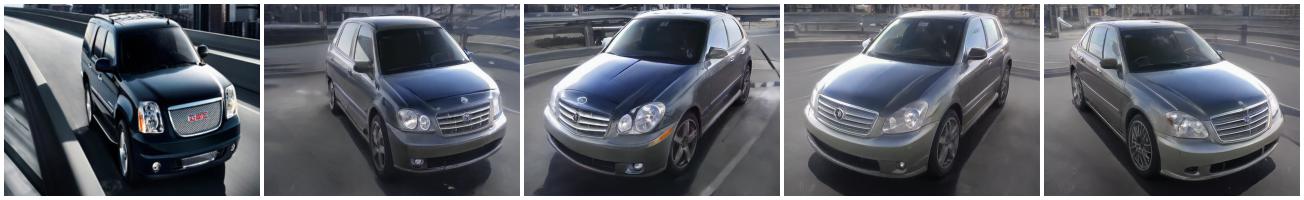} } \\
            \multicolumn{5}{c}{\includegraphics[width=\linewidth]{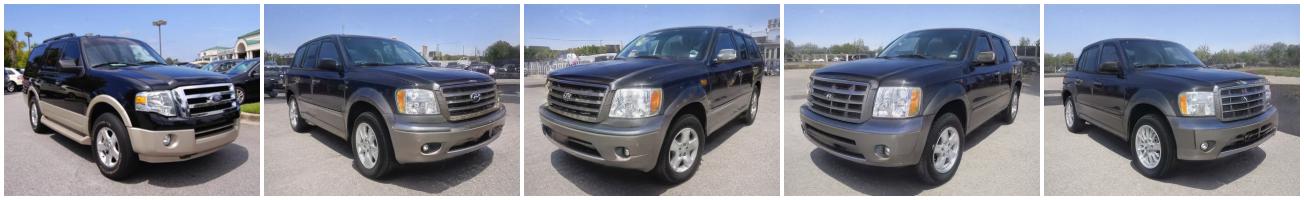} } \\
            
            \qquad\qquad Source \qquad\qquad\qquad Inversion \qquad\qquad\qquad\quad Edit \qquad\qquad\qquad Edit's Inverse \qquad\qquad Inversion Editing
        
        \end{tabular}
    \caption{Examples of the LEC protocol over the cars domain using GANSpace~\cite{harkonen2020ganspace} to edit the car's viewpoint. Note that the "Inversion Editing" almost perfectly reconstructs the "Inversion" which serves as evidence that our e4e encoder is well-behaved.}
    \label{fig:appendix_lec_cars}
\end{figure*}

We provide additional results of the LEC measure, similar to Figure 11 in the main paper. See \Cref{fig:appendix_lec_cars}. Note that our method successfully reconstructs the inversion image after the LEC protocol. This serves as additional evidence that our encoder is well-behaved and suitable  for successive semantic latent editing. 

\section{Additional Results}
\begin{figure*}
    \setlength{\tabcolsep}{1pt}
    \centering
        \centering
            \begin{tabular}{c c c c c c}
            
            Source & Inversion & Edit & Source & Inversion & Edit \\
            
            \includegraphics[width=0.15\textwidth]{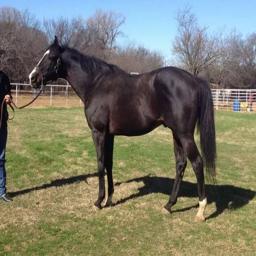} &
            \includegraphics[width=0.15\textwidth]{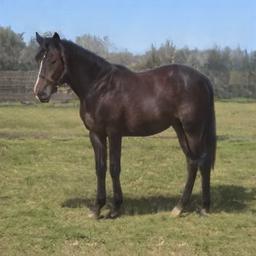} &
            \includegraphics[width=0.15\textwidth]{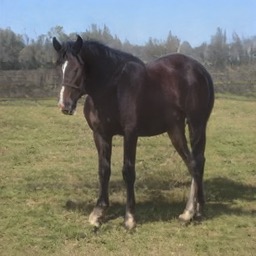} &
            \includegraphics[width=0.15\textwidth]{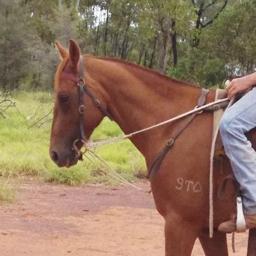} &
            \includegraphics[width=0.15\textwidth]{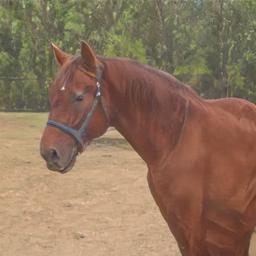} &
            \includegraphics[width=0.15\textwidth]{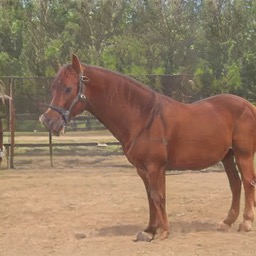} \\
            \includegraphics[width=0.15\textwidth]{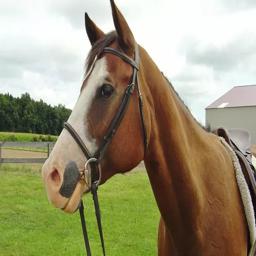} &
            \includegraphics[width=0.15\textwidth]{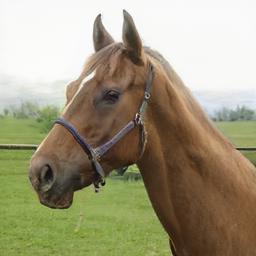} &
            \includegraphics[width=0.15\textwidth]{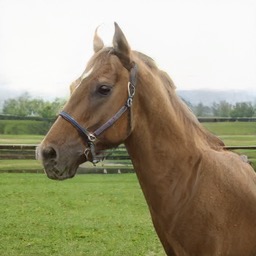} &
            \includegraphics[width=0.15\textwidth]{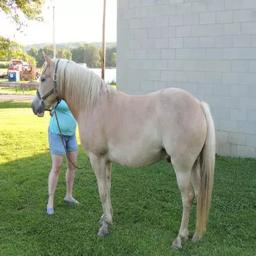} &
            \includegraphics[width=0.15\textwidth]{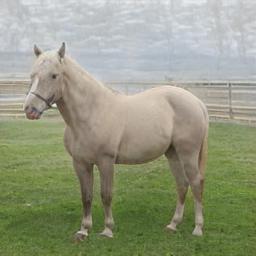} &
            \includegraphics[width=0.15\textwidth]{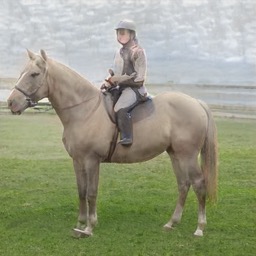} \\
            \includegraphics[width=0.15\textwidth]{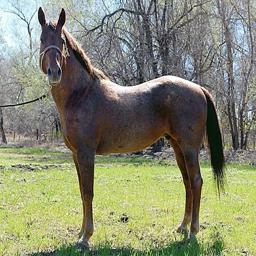} &
            \includegraphics[width=0.15\textwidth]{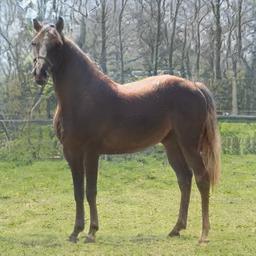} &
            \includegraphics[width=0.15\textwidth]{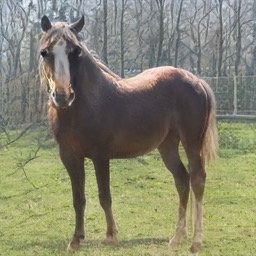} &
            \includegraphics[width=0.15\textwidth]{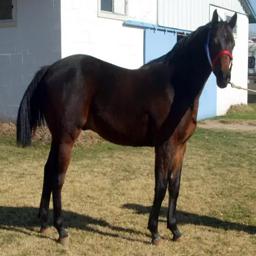} &
            \includegraphics[width=0.15\textwidth]{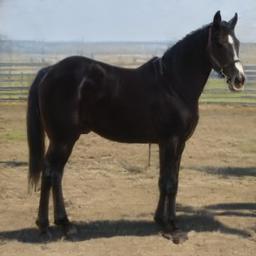} &
            \includegraphics[width=0.15\textwidth]{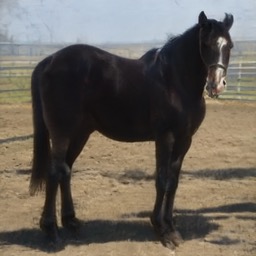} \\
\end{tabular}
    \caption{Additional results obtained by e4e over the LSUN horses domain~\cite{yu2016lsun}. Various edits, such as pose, head change, and horse rider, are performed using SeFa ~\cite{shen2020closedform}.}
    \label{fig:appendix_horses}
\end{figure*}

\begin{figure*}
    \setlength{\tabcolsep}{1pt}
    \centering
        \centering
            \begin{tabular}{c c c c c c}
            
            Source & Inversion & Edit & Source & Inversion & Edit \\
            
            \includegraphics[width=0.15\textwidth]{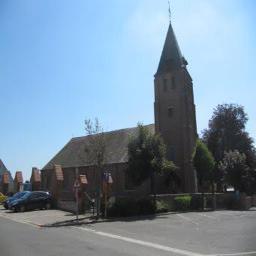} &
            \includegraphics[width=0.15\textwidth]{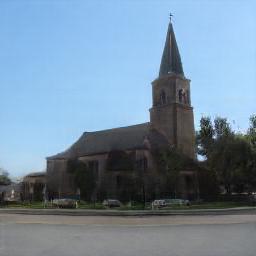} &
            \includegraphics[width=0.15\textwidth]{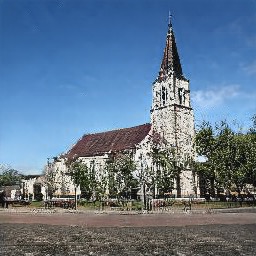} &
            \includegraphics[width=0.15\textwidth]{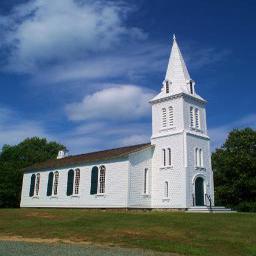} &
            \includegraphics[width=0.15\textwidth]{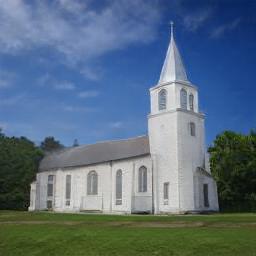} &
            \includegraphics[width=0.15\textwidth]{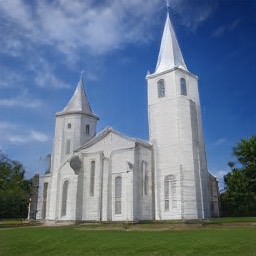} \\
            \includegraphics[width=0.15\textwidth]{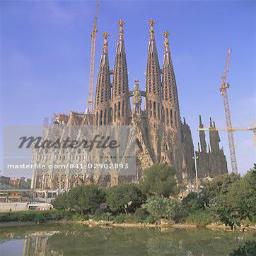} &
            \includegraphics[width=0.15\textwidth]{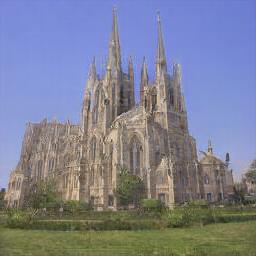} &
            \includegraphics[width=0.15\textwidth]{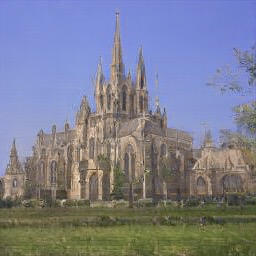} &
            \includegraphics[width=0.15\textwidth]{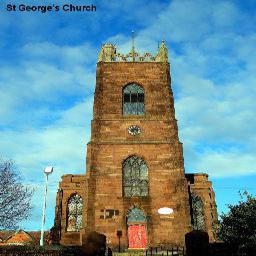} &
            \includegraphics[width=0.15\textwidth]{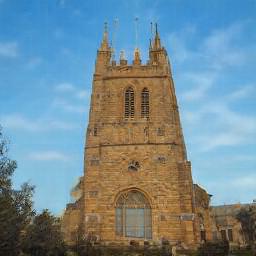} &
            \includegraphics[width=0.15\textwidth]{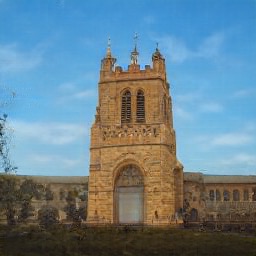} \\
            \includegraphics[width=0.15\textwidth]{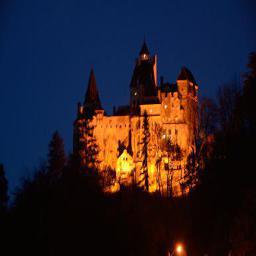} &
            \includegraphics[width=0.15\textwidth]{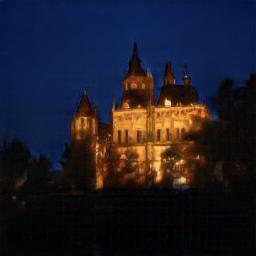} &
            \includegraphics[width=0.15\textwidth]{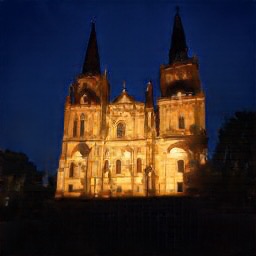} &
            \includegraphics[width=0.15\textwidth]{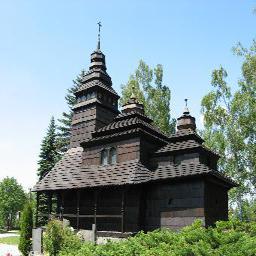} &
            \includegraphics[width=0.15\textwidth]{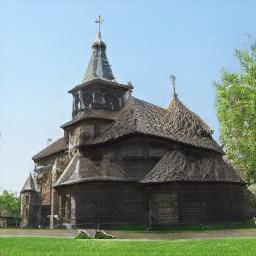} &
            \includegraphics[width=0.15\textwidth]{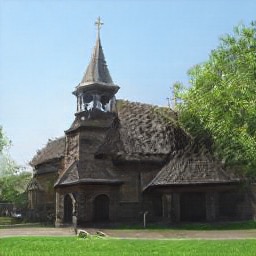} \\
\end{tabular}
    \caption{Additional results obtained by e4e over the LSUN churches domain~\cite{yu2016lsun}. Various edits, such as viewpoint, geometric change, and daylight, are performed using SeFa ~\cite{shen2020closedform}.}
    \label{fig:appendix_church}
\end{figure*}

\begin{figure*}
    \setlength{\tabcolsep}{1pt}
    \centering
        \centering
            \begin{tabular}{c c c c c c c c}
            
            Source & Inversion & Young & Old & Source & Inversion & Young & Old \\
            
            \includegraphics[width=0.12\textwidth]{images/first_celebs/0.jpg} &
            \includegraphics[width=0.12\textwidth]{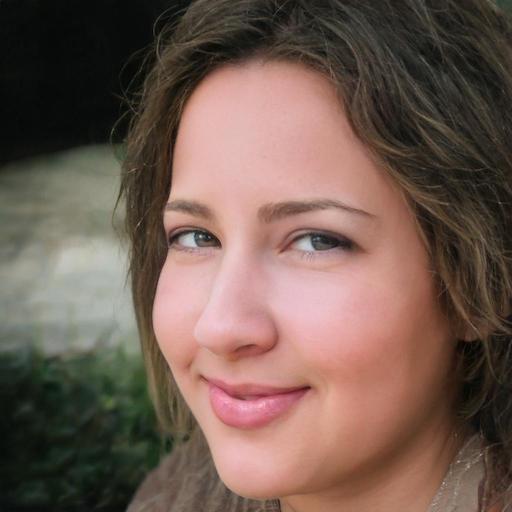} &
            \includegraphics[width=0.12\textwidth]{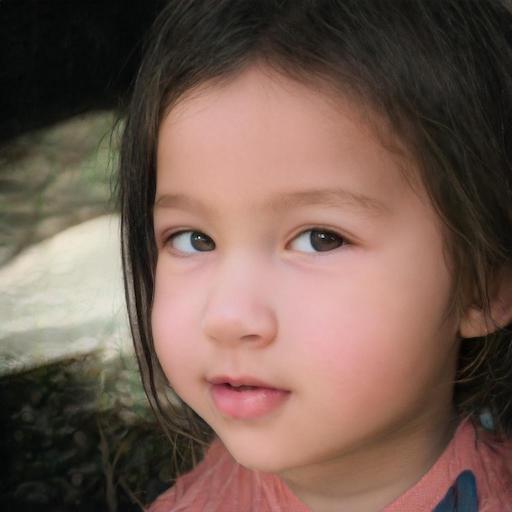} &
            \includegraphics[width=0.12\textwidth]{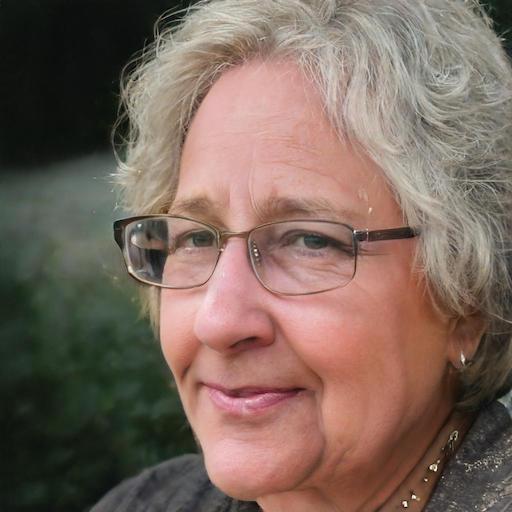} &
            \includegraphics[width=0.12\textwidth]{images/first_celebs/1.jpg} &
            \includegraphics[width=0.12\textwidth]{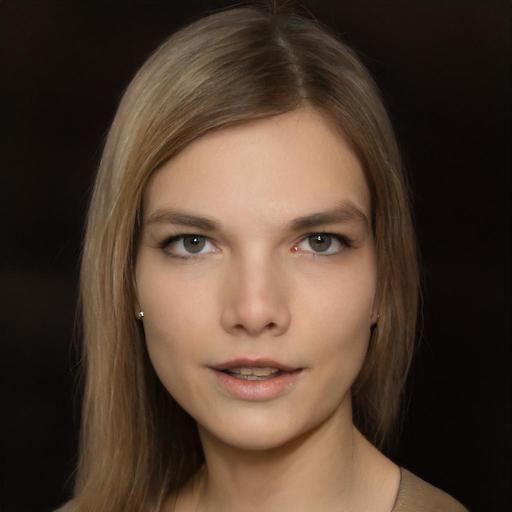} &
            \includegraphics[width=0.12\textwidth]{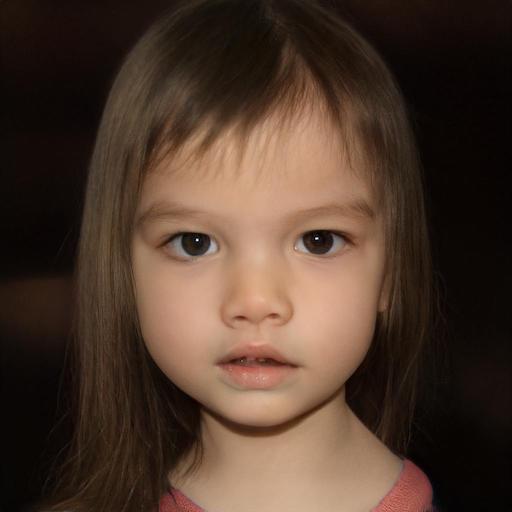} &
            \includegraphics[width=0.12\textwidth]{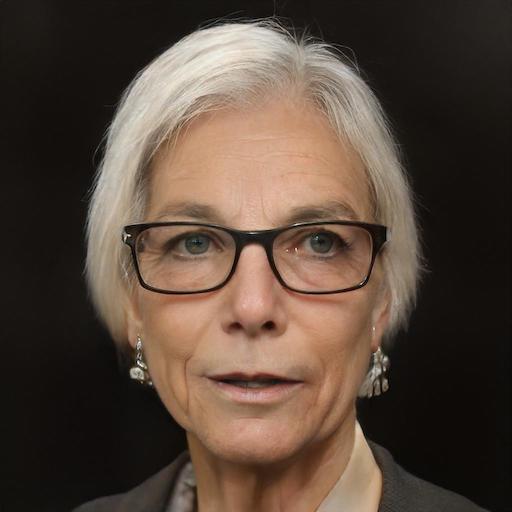}  \\
            \includegraphics[width=0.12\textwidth]{images/first_celebs/2.jpg} &
            \includegraphics[width=0.12\textwidth]{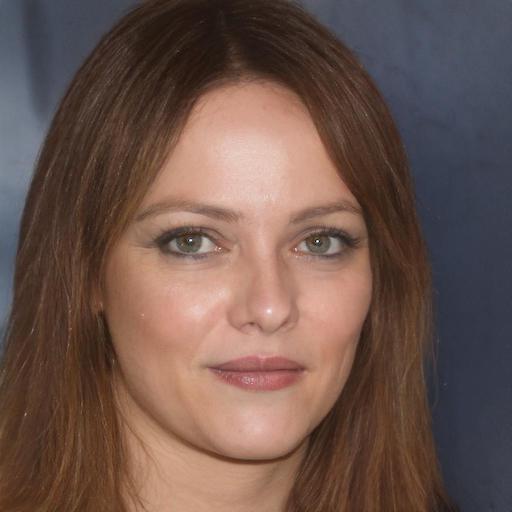} &
            \includegraphics[width=0.12\textwidth]{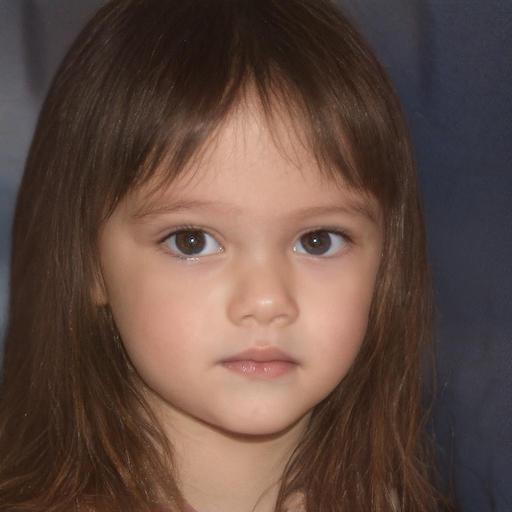} &
            \includegraphics[width=0.12\textwidth]{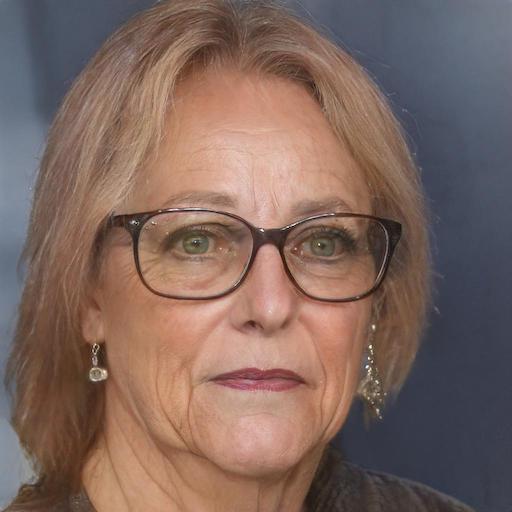} &
            \includegraphics[width=0.12\textwidth]{images/first_celebs/3.jpg} &
            \includegraphics[width=0.12\textwidth]{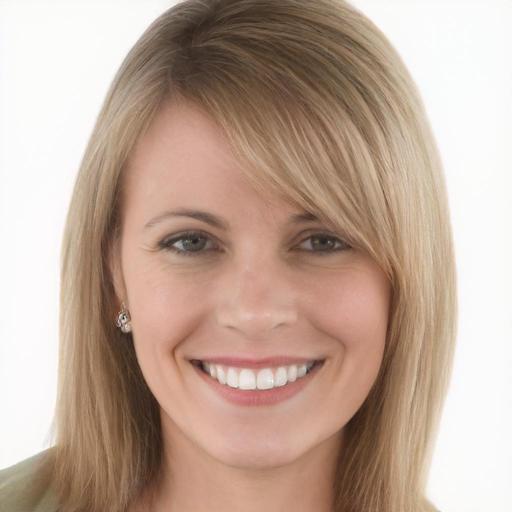} &
            \includegraphics[width=0.12\textwidth]{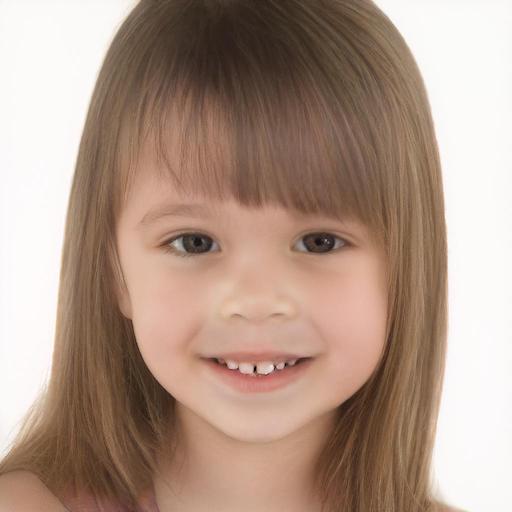} &
            \includegraphics[width=0.12\textwidth]{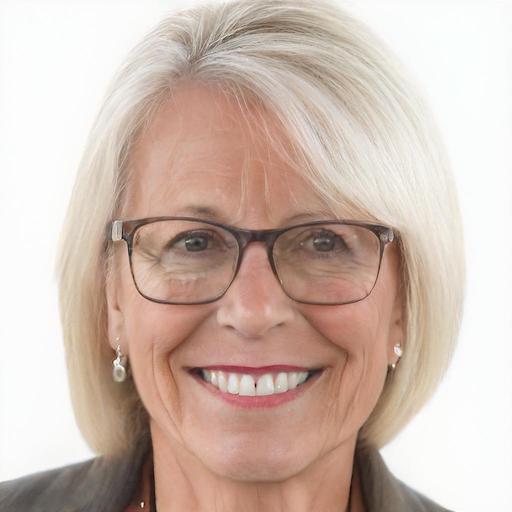}  \\
\end{tabular}
    \caption{Additional results of our e4e encoder over the first 48 images in the CelebA-HQ~\cite{karras2017progressive} test set using InterFaceGAN~\cite{shen2020interpreting} age editing. In this figure, images $1-16$ are presented.}
    \label{fig:first_celebs_interfacegan_1}
\end{figure*}

\begin{figure*}
    \setlength{\tabcolsep}{1pt}
    \centering
        \centering
            \begin{tabular}{c c c c c c c c}
            
            Source & Inversion & Young & Old & Source & Inversion & Young & Old \\
            
            \includegraphics[width=0.12\textwidth]{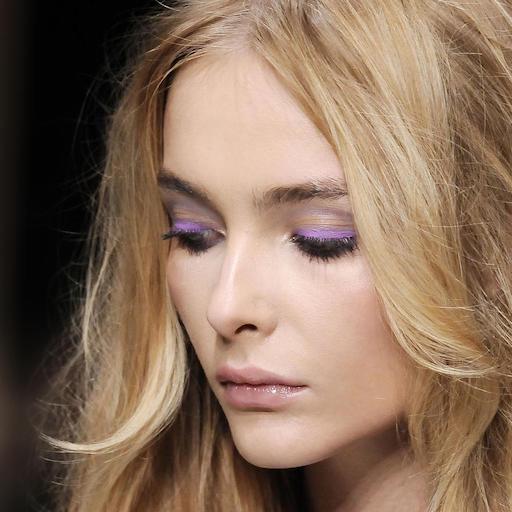} &
            \includegraphics[width=0.12\textwidth]{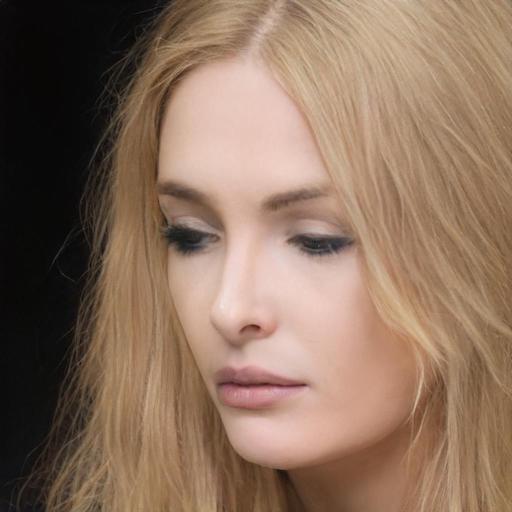} &
            \includegraphics[width=0.12\textwidth]{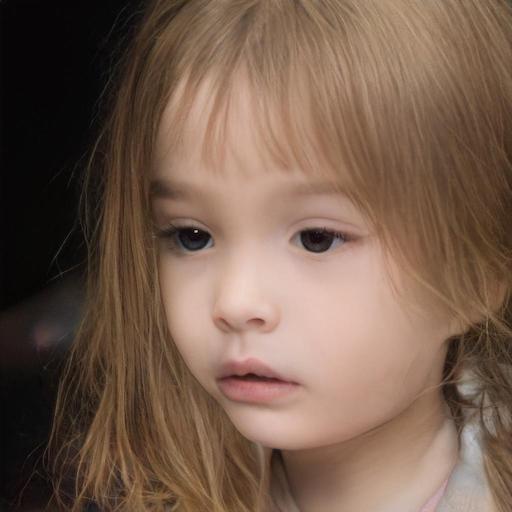} &
            \includegraphics[width=0.12\textwidth]{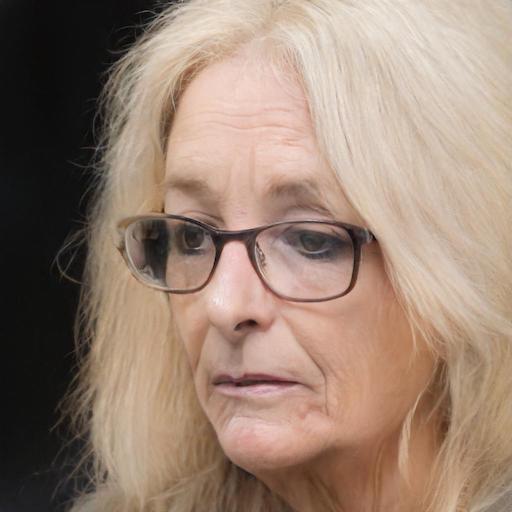} &
            \includegraphics[width=0.12\textwidth]{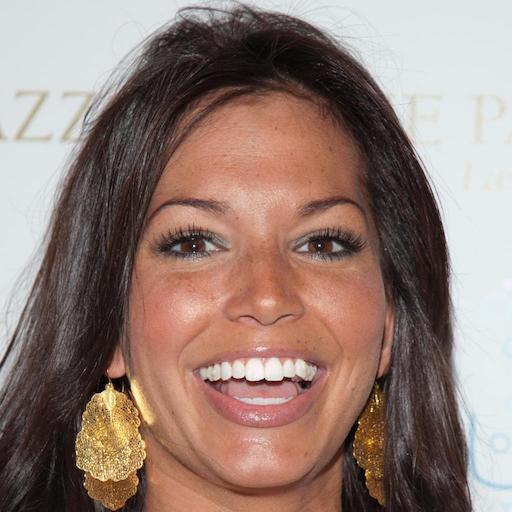} &
            \includegraphics[width=0.12\textwidth]{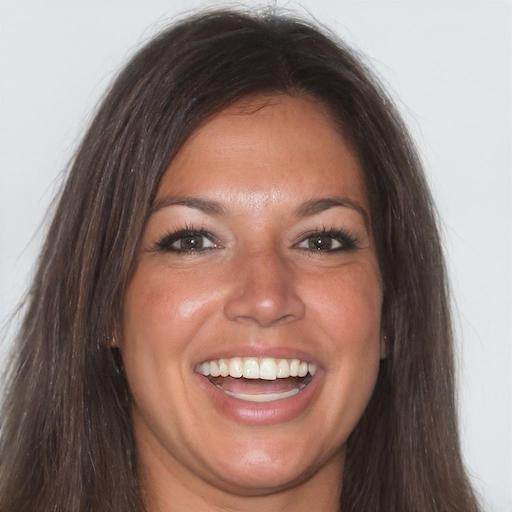} &
            \includegraphics[width=0.12\textwidth]{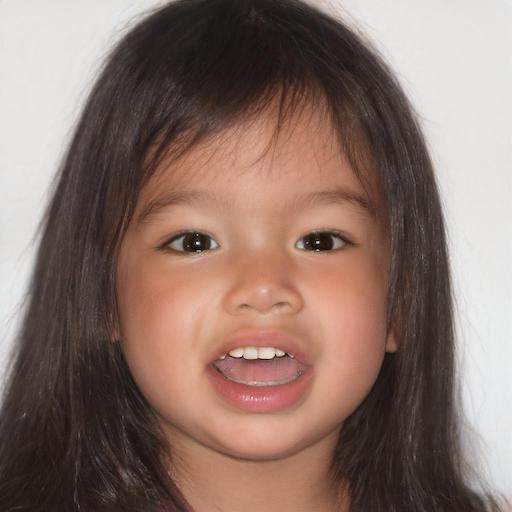} &
            \includegraphics[width=0.12\textwidth]{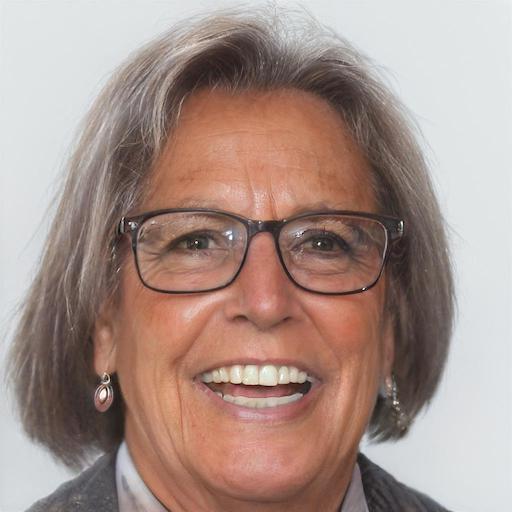}  \\
            \includegraphics[width=0.12\textwidth]{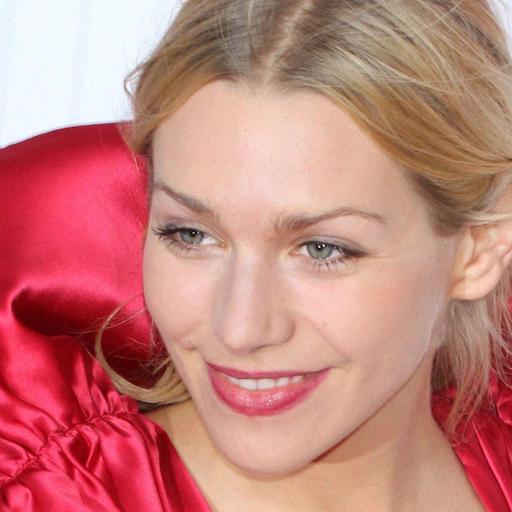} &
            \includegraphics[width=0.12\textwidth]{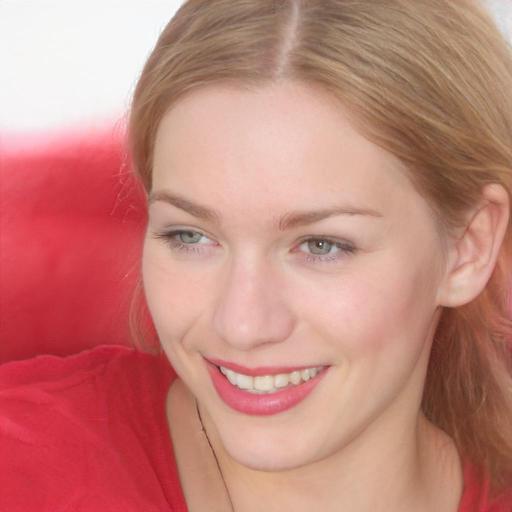} &
            \includegraphics[width=0.12\textwidth]{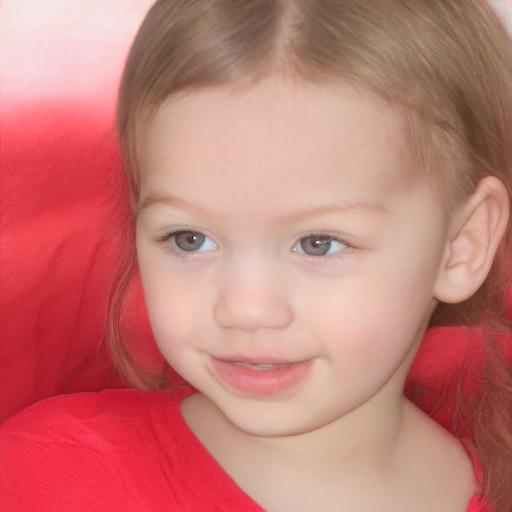} &
            \includegraphics[width=0.12\textwidth]{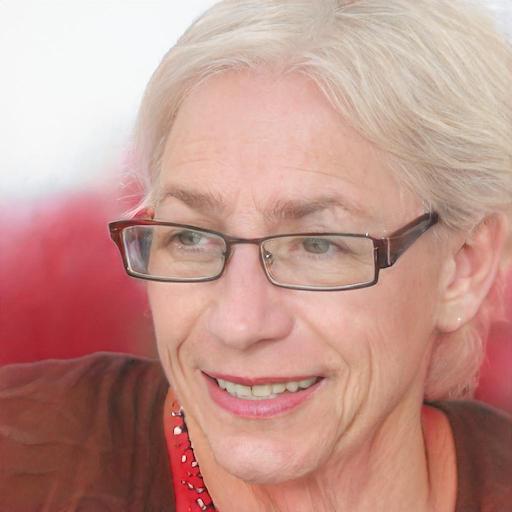} &
            \includegraphics[width=0.12\textwidth]{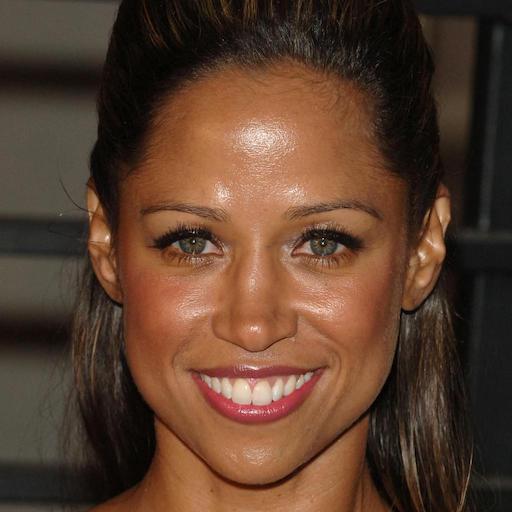} &
            \includegraphics[width=0.12\textwidth]{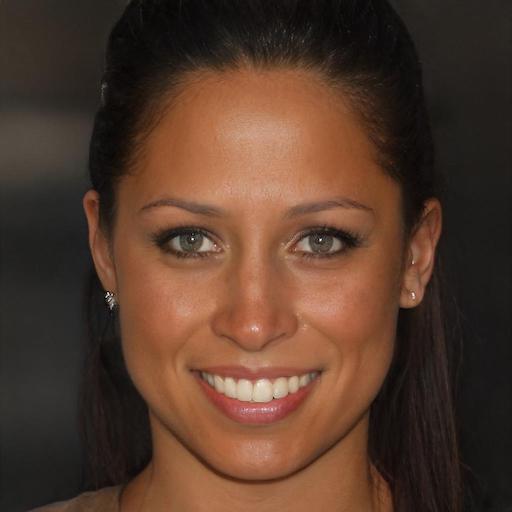} &
            \includegraphics[width=0.12\textwidth]{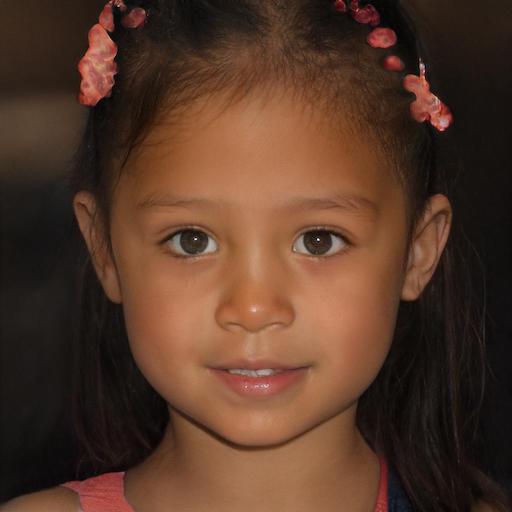} &
            \includegraphics[width=0.12\textwidth]{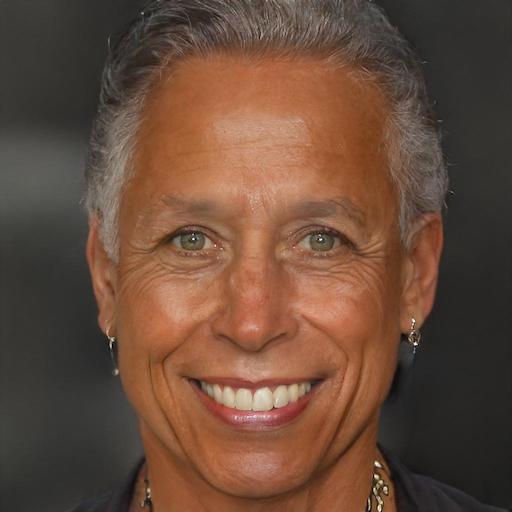}  \\
\end{tabular}
    \caption{Additional results of our e4e encoder over the first 48 images in the CelebA-HQ~\cite{karras2017progressive} test set using InterFaceGAN~\cite{shen2020interpreting} age editing. In this figure, images $17-32$ are presented.}
    \label{fig:first_celebs_interfacegan_2}
\end{figure*}

\begin{figure*}
\label{fig:first_ten_celeba}
    \setlength{\tabcolsep}{1pt}
    \centering
        \centering
            \begin{tabular}{c c c c c c c c}
            
            Source & Inversion & Young & Old & Source & Inversion & Young & Old \\
            
            \includegraphics[width=0.12\textwidth]{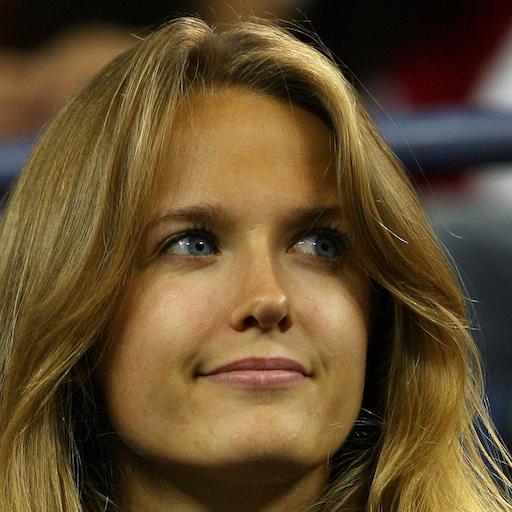} &
            \includegraphics[width=0.12\textwidth]{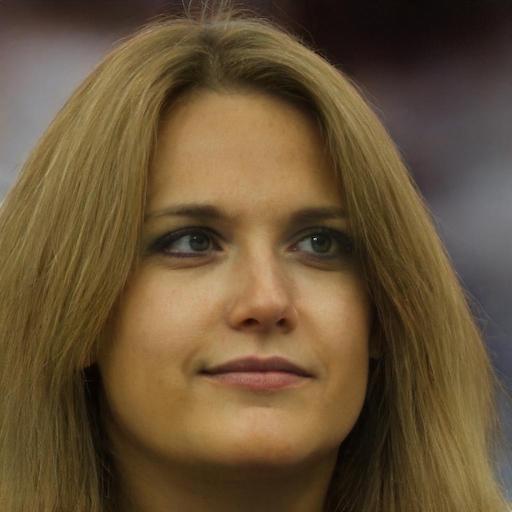} &
            \includegraphics[width=0.12\textwidth]{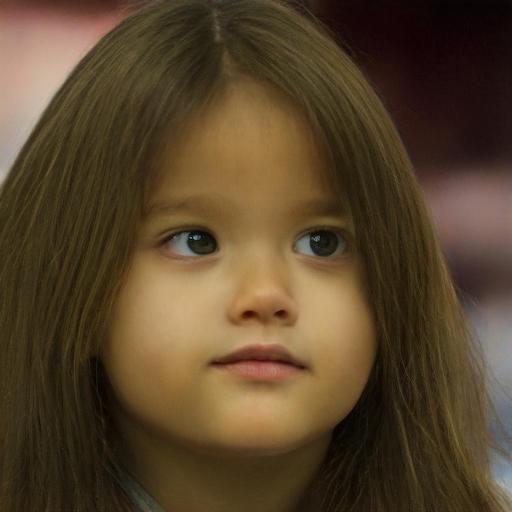} &
            \includegraphics[width=0.12\textwidth]{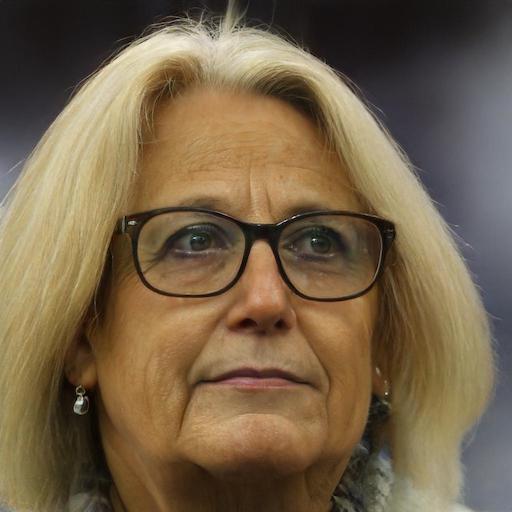} &
            \includegraphics[width=0.12\textwidth]{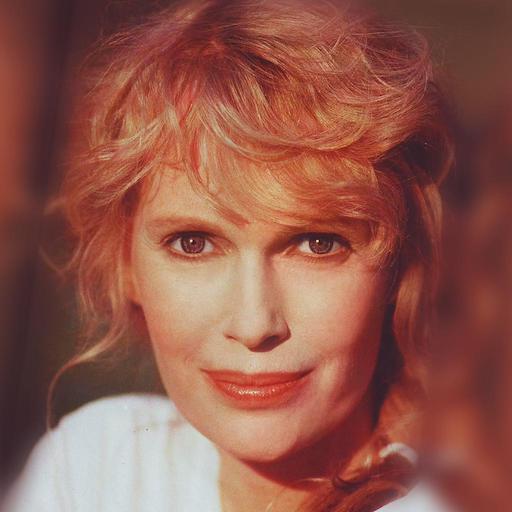} &
            \includegraphics[width=0.12\textwidth]{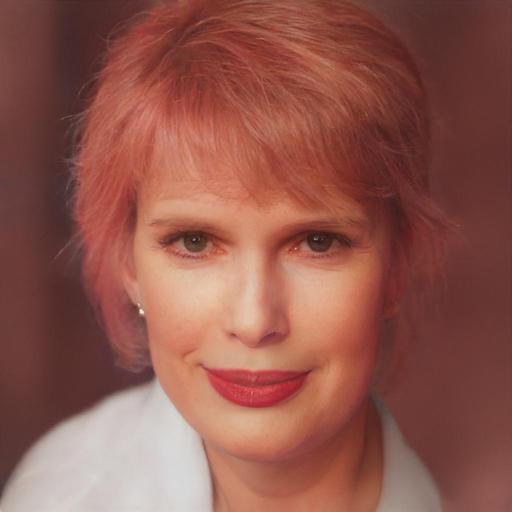} &
            \includegraphics[width=0.12\textwidth]{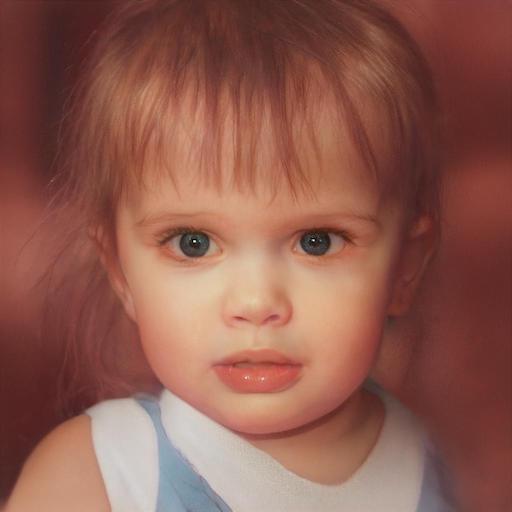} &
            \includegraphics[width=0.12\textwidth]{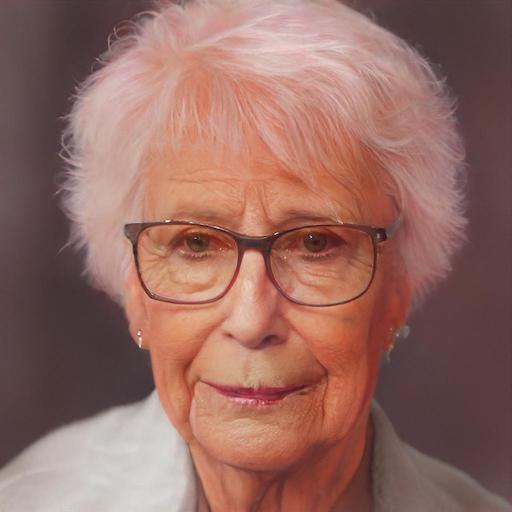}  \\
            \includegraphics[width=0.12\textwidth]{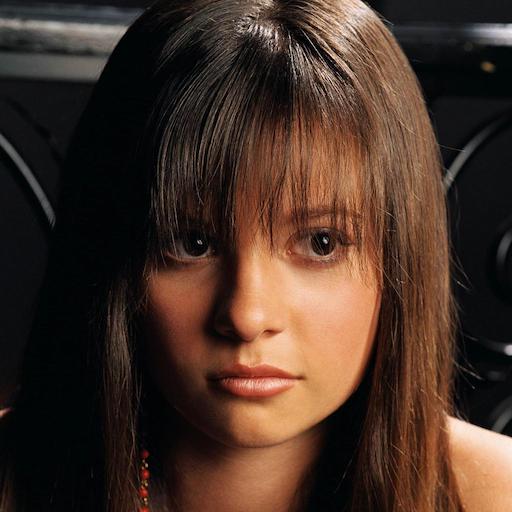} &
            \includegraphics[width=0.12\textwidth]{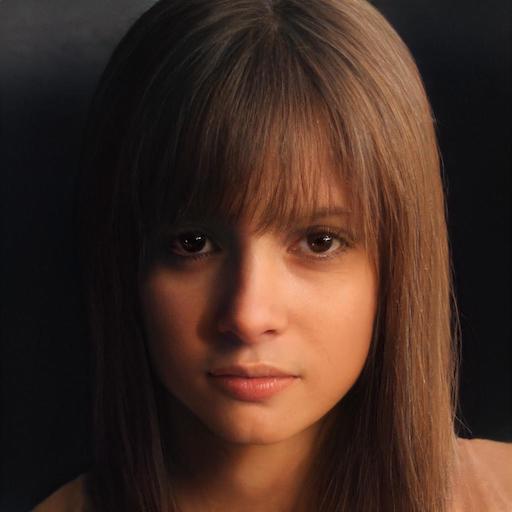} &
            \includegraphics[width=0.12\textwidth]{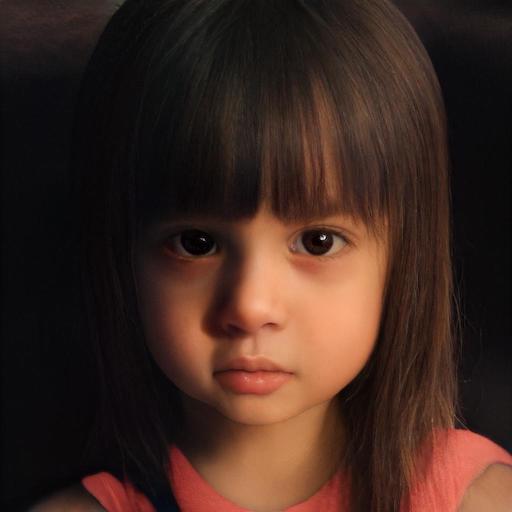} &
            \includegraphics[width=0.12\textwidth]{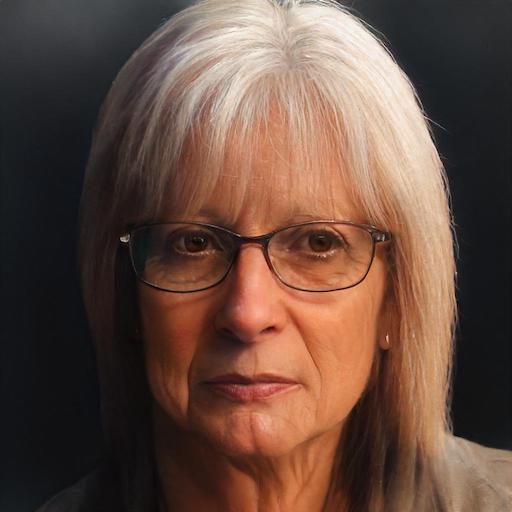} &
            \includegraphics[width=0.12\textwidth]{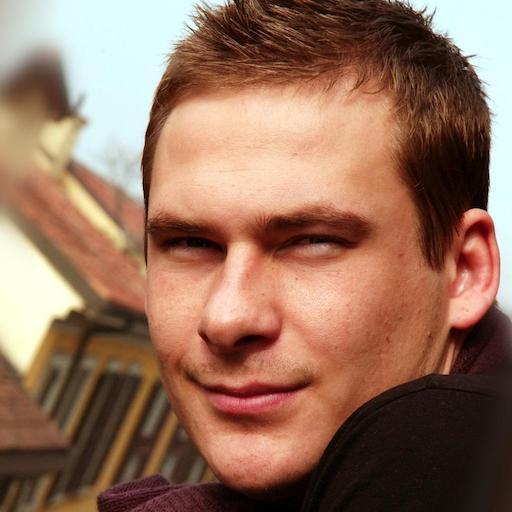} &
            \includegraphics[width=0.12\textwidth]{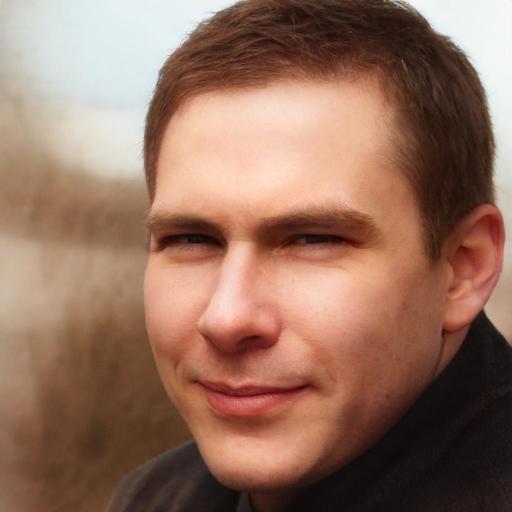} &
            \includegraphics[width=0.12\textwidth]{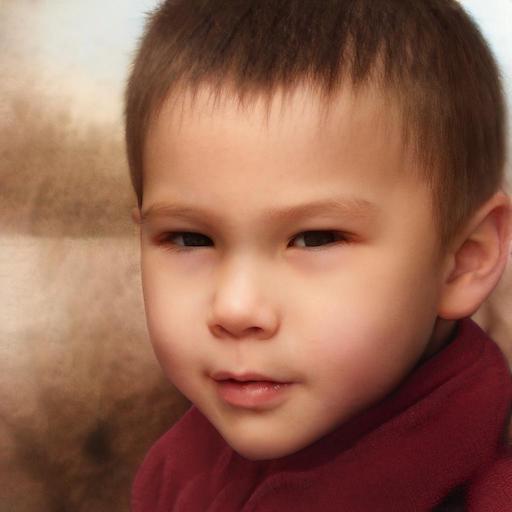} &
            \includegraphics[width=0.12\textwidth]{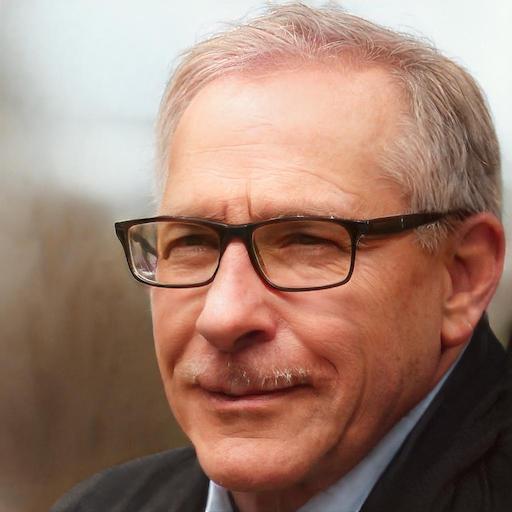}  \\
\end{tabular}
    \caption{Additional results of our e4e encoder over the first 48 images in the CelebA-HQ~\cite{karras2017progressive} test set using InterFaceGAN~\cite{shen2020interpreting} age editing. In this figure, images $33-48$ are presented}.
    \label{fig:first_celebs_interfacegan_3}
\end{figure*}

\begin{figure*}

    \setlength{\tabcolsep}{1pt}
    \centering
        \centering
            \begin{tabular}{c c c c c c c c c c c c c}

             Source & Inversion & Viewpoint I & Viewpoint II & Cube & Color & Grass \\
             
            \includegraphics[height=0.1\textwidth]{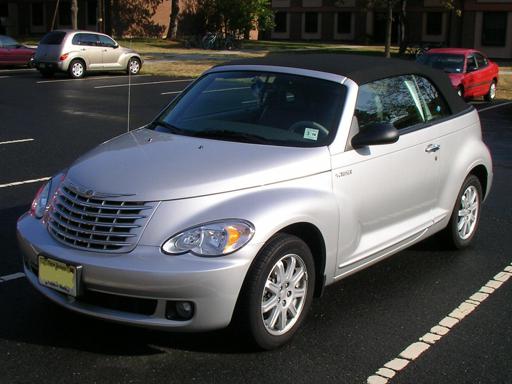} &
            \includegraphics[height=0.1\textwidth]{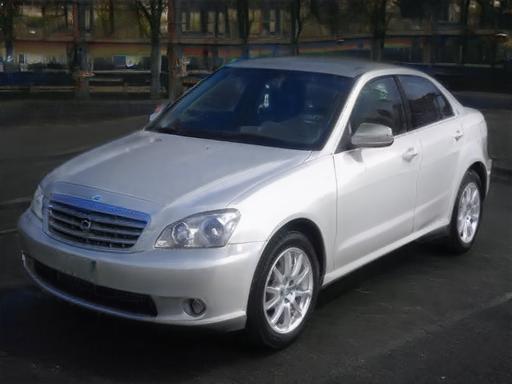} &
            \includegraphics[height=0.1\textwidth]{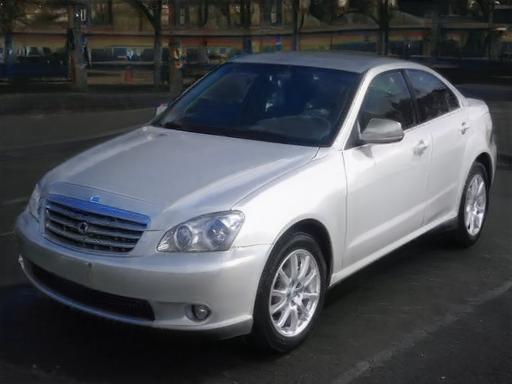} &
            \includegraphics[height=0.1\textwidth]{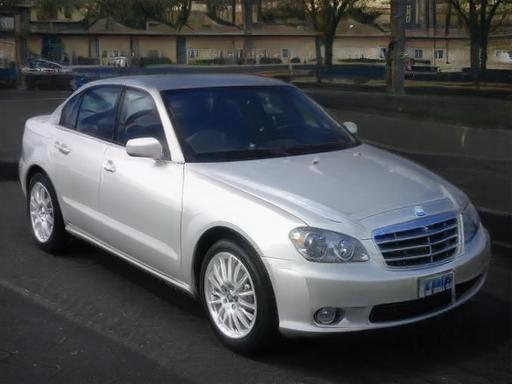} &
            \includegraphics[height=0.1\textwidth]{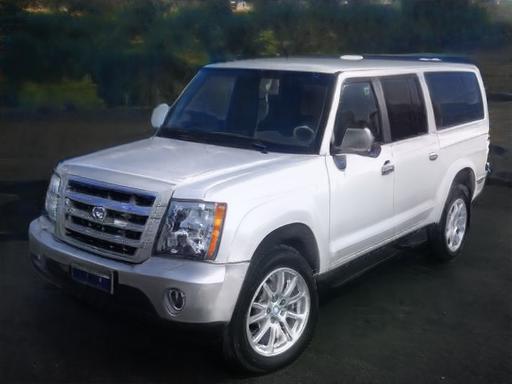} &
            \includegraphics[height=0.1\textwidth]{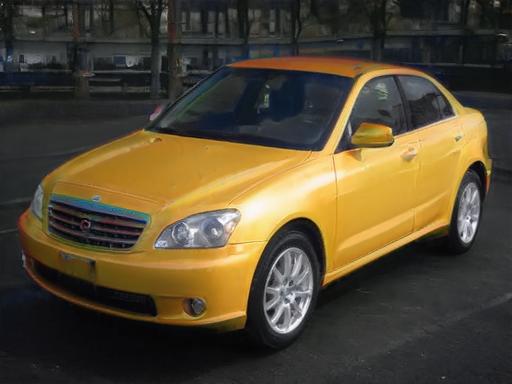} &
            \includegraphics[height=0.1\textwidth]{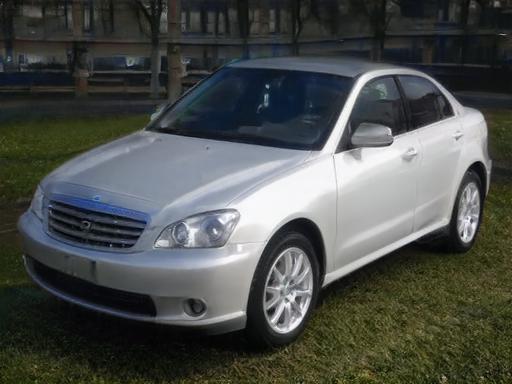} \\
            
            \includegraphics[height=0.1\textwidth]{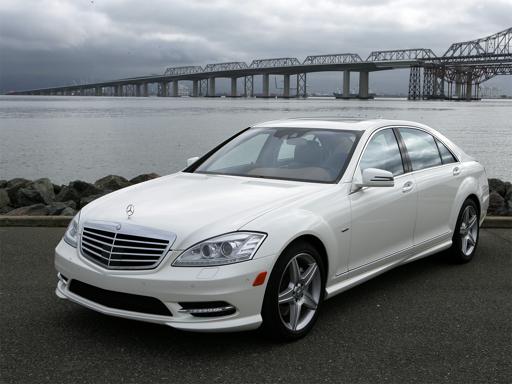} &
            \includegraphics[height=0.1\textwidth]{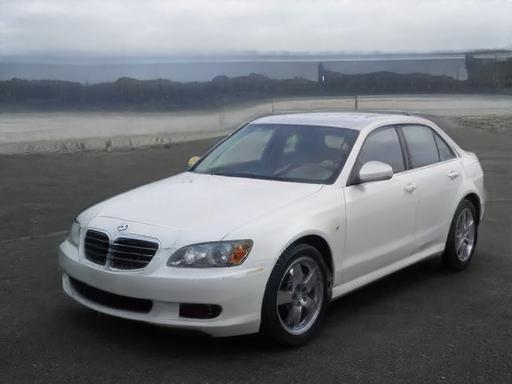} &
            \includegraphics[height=0.1\textwidth]{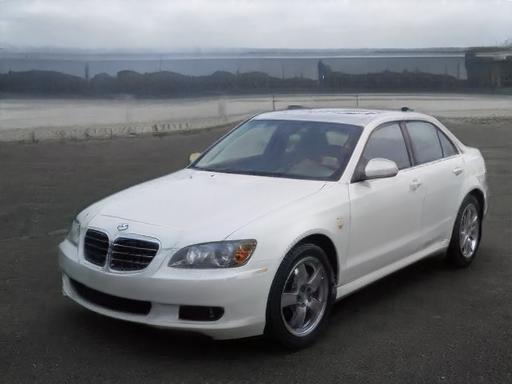} &
            \includegraphics[height=0.1\textwidth]{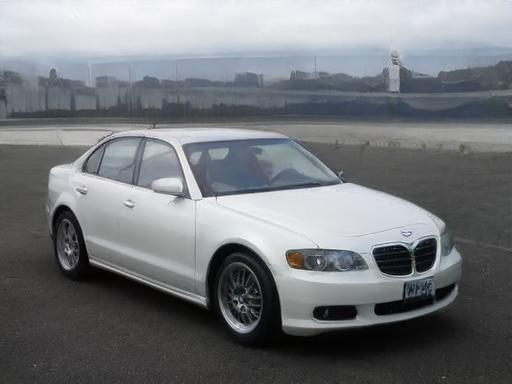} &
            \includegraphics[height=0.1\textwidth]{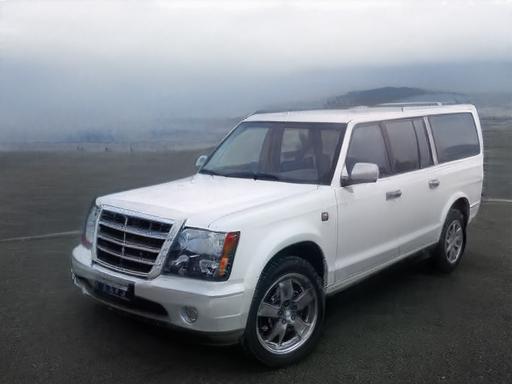} &
            \includegraphics[height=0.1\textwidth]{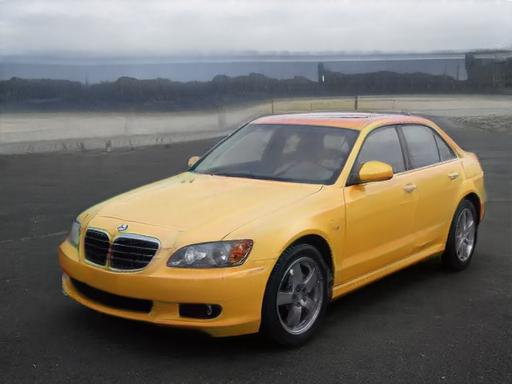} &
            \includegraphics[height=0.1\textwidth]{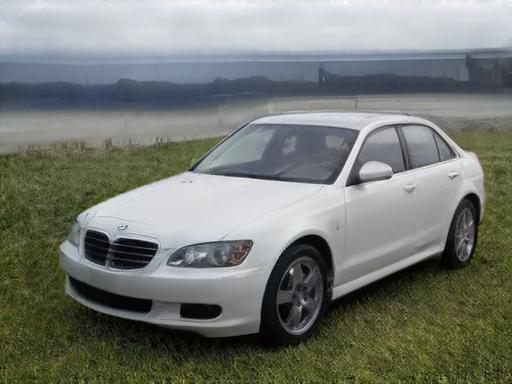} \\
\end{tabular}
    \caption{Additional results of our e4e encoder on the cars domain, similar to Figure \ref{fig:internet_celebs_1} on the Stanford Cars test set ~\cite{KrauseStarkDengFei-Fei_3DRR2013}.}
    \label{fig:appendix_first_cars_3}
\end{figure*}

\begin{figure*}

    \setlength{\tabcolsep}{1pt}
    \centering
        \centering
            \begin{tabular}{c c c c c c c c c c c c c}

             Source & Inversion & Viewpoint I & Viewpoint II & Cube & Color & Grass \\
             
            \includegraphics[height=0.1\textwidth]{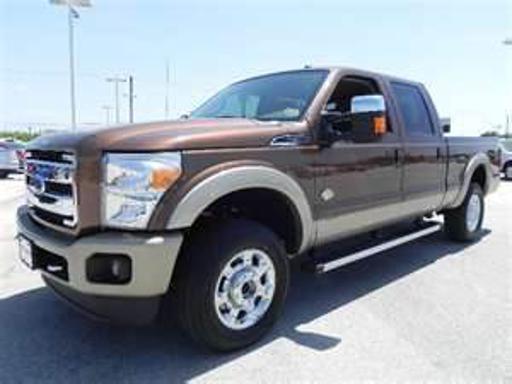} &
            \includegraphics[height=0.1\textwidth]{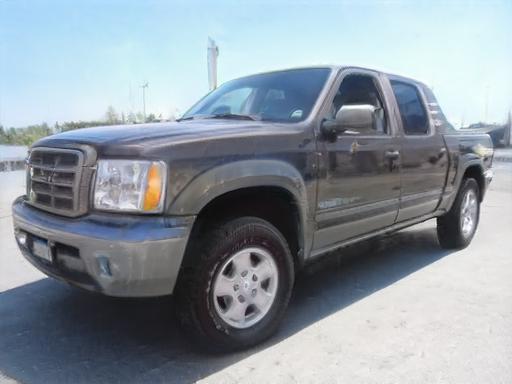} &
            \includegraphics[height=0.1\textwidth]{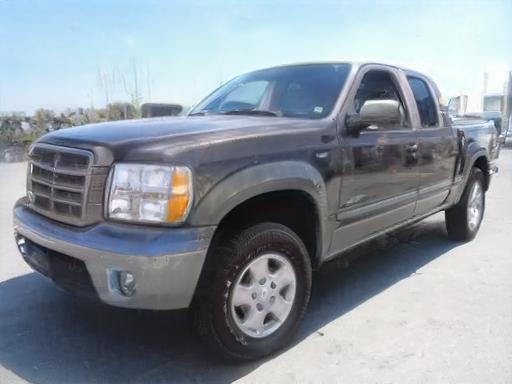} &
            \includegraphics[height=0.1\textwidth]{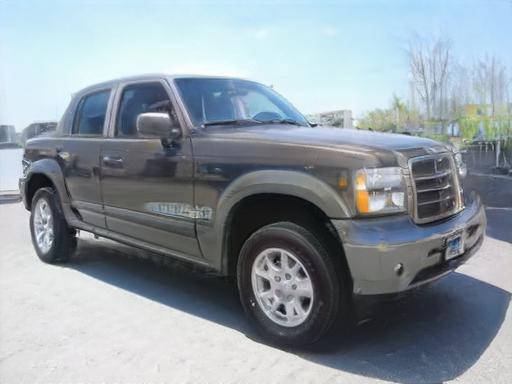} &
            \includegraphics[height=0.1\textwidth]{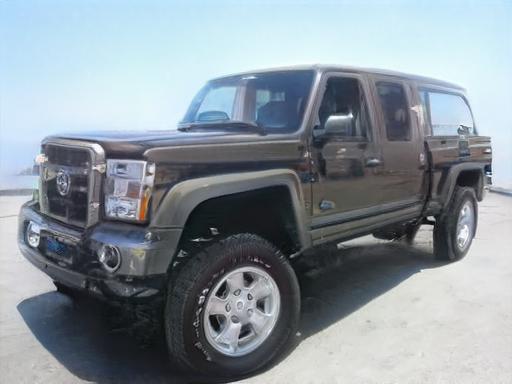} &
            \includegraphics[height=0.1\textwidth]{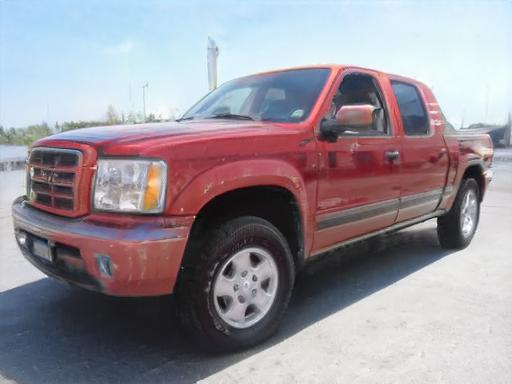} &
            \includegraphics[height=0.1\textwidth]{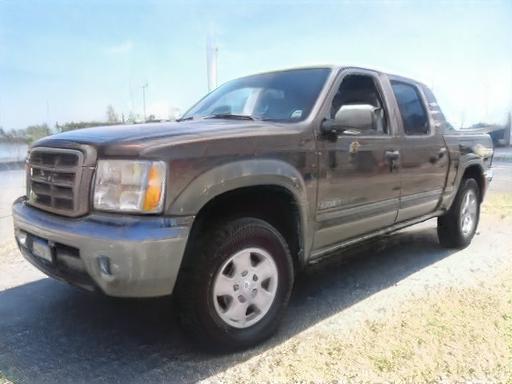}  \\
            
            \includegraphics[height=0.1\textwidth]{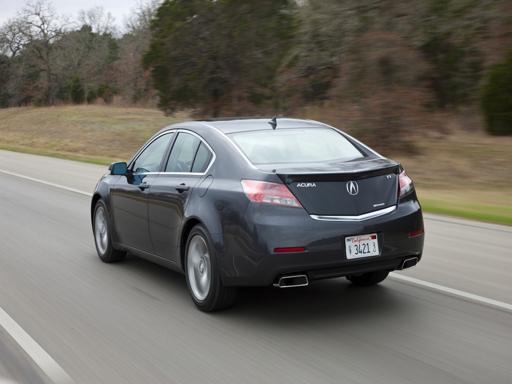} &
            \includegraphics[height=0.1\textwidth]{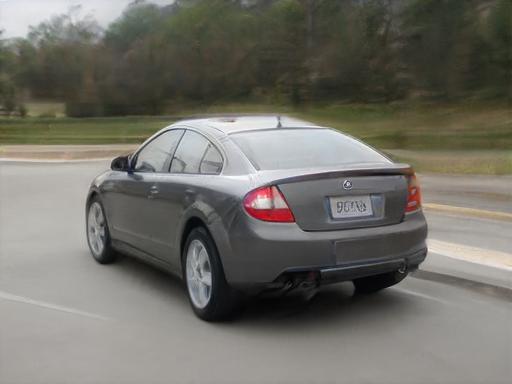} &
            \includegraphics[height=0.1\textwidth]{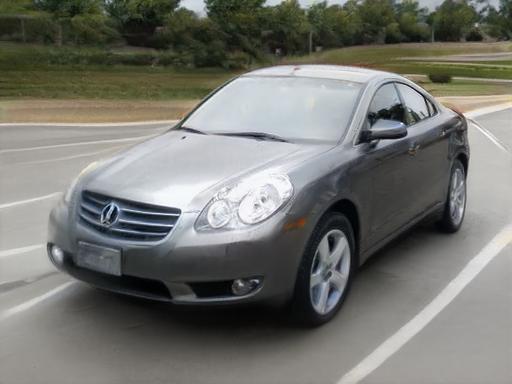} &
            \includegraphics[height=0.1\textwidth]{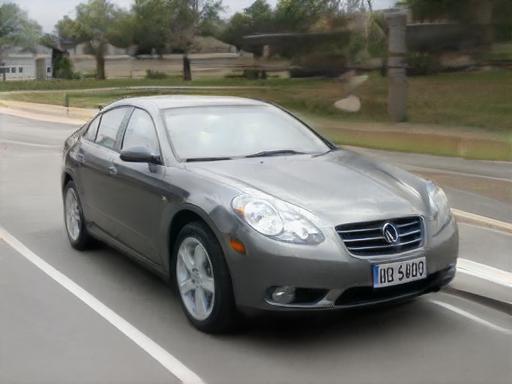} &
            \includegraphics[height=0.1\textwidth]{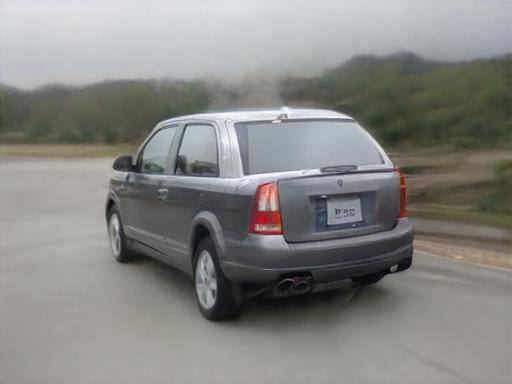} &
            \includegraphics[height=0.1\textwidth]{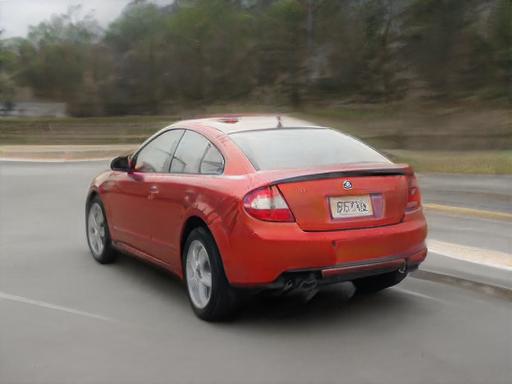} &
            \includegraphics[height=0.1\textwidth]{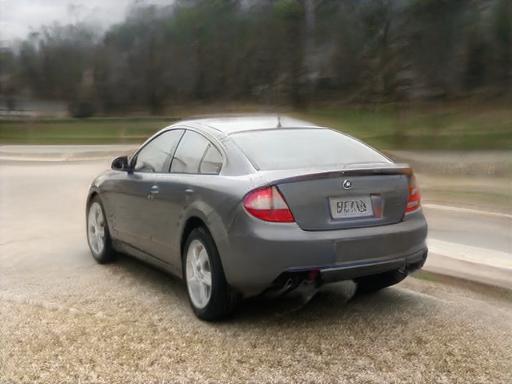}  \\
\end{tabular}
    \caption{Additional results of our e4e encoder on the cars domain, similar to Figure \ref{fig:internet_celebs_1} on the Stanford Cars test set ~\cite{KrauseStarkDengFei-Fei_3DRR2013}.}
    \label{fig:appendix_first_cars_4}
\end{figure*}

\begin{figure*}

    \setlength{\tabcolsep}{1pt}
    \centering
        \centering
            \begin{tabular}{c c c c c c c c c c c c c}

             Source & Inversion & Viewpoint I & Viewpoint II & Cube & Color & Grass \\
             
            \includegraphics[height=0.1\textwidth]{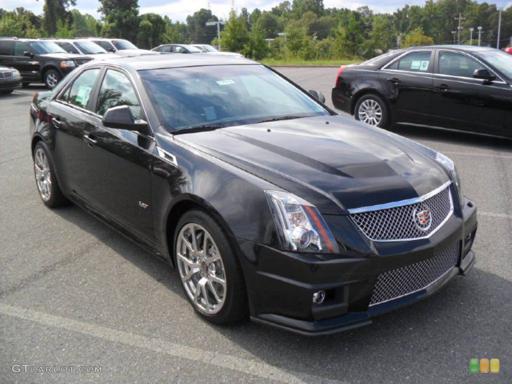} &
            \includegraphics[height=0.1\textwidth]{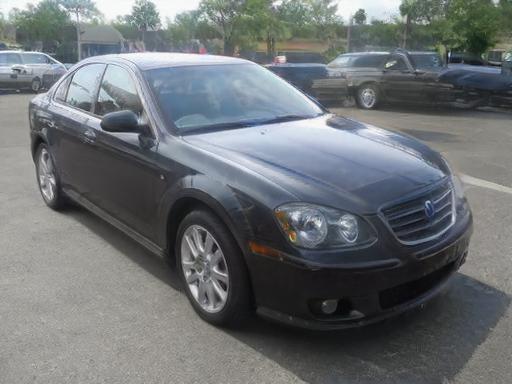} &
            \includegraphics[height=0.1\textwidth]{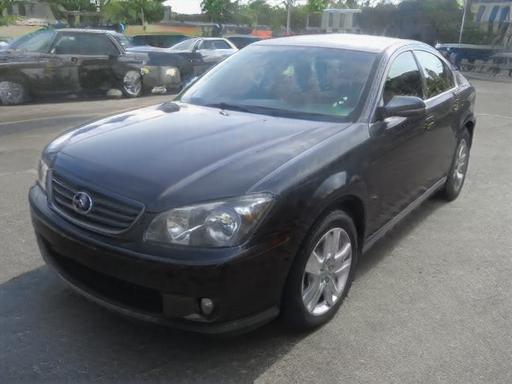} &
            \includegraphics[height=0.1\textwidth]{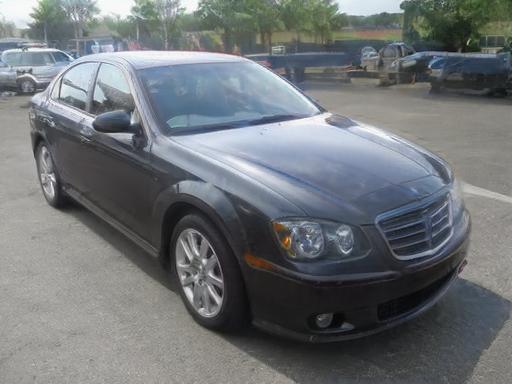} &
            \includegraphics[height=0.1\textwidth]{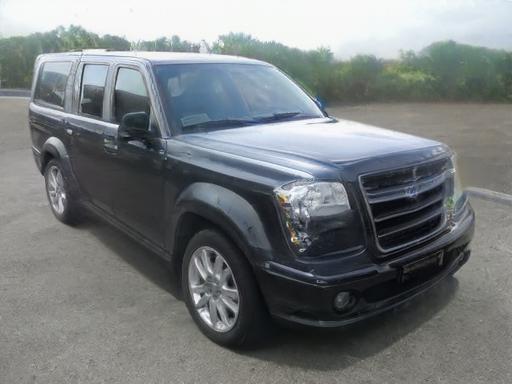} &
            \includegraphics[height=0.1\textwidth]{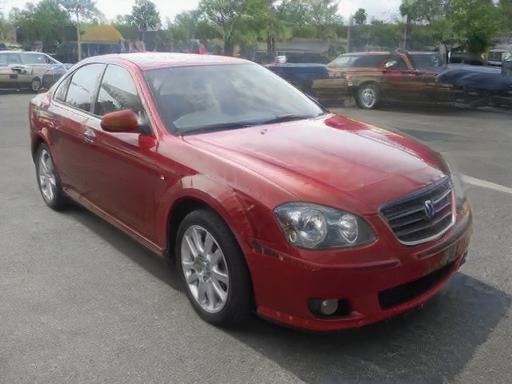} &
            \includegraphics[height=0.1\textwidth]{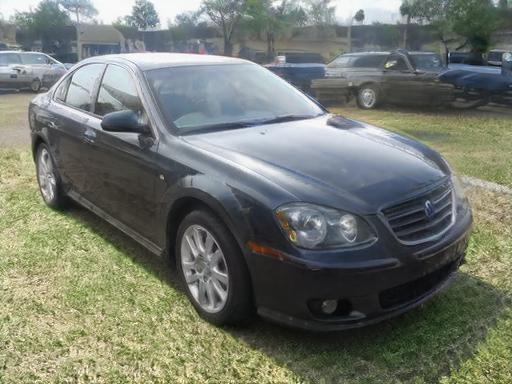}  \\
            
            \includegraphics[height=0.1\textwidth]{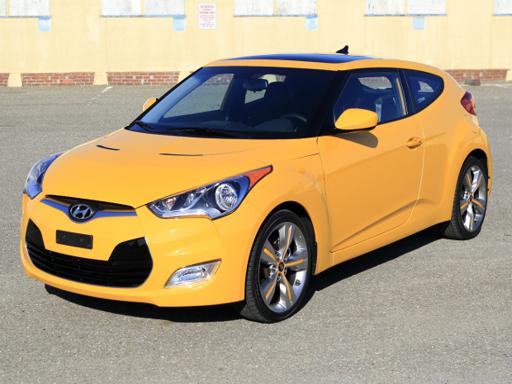} &
            \includegraphics[height=0.1\textwidth]{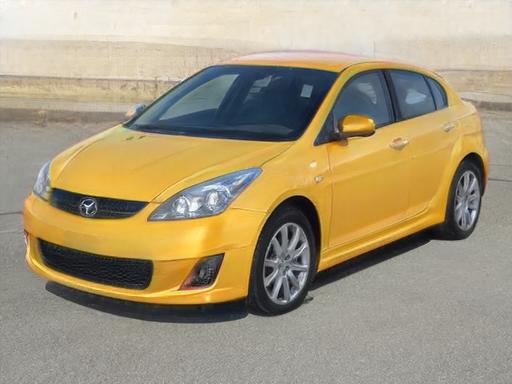} &
            \includegraphics[height=0.1\textwidth]{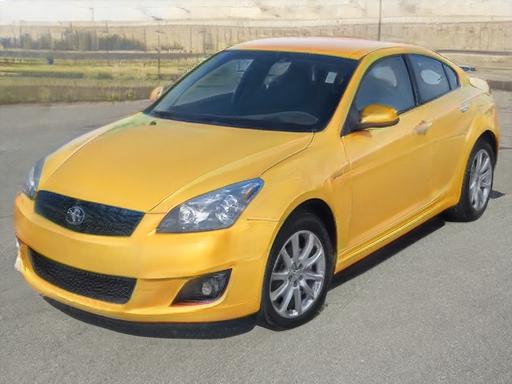} &
            \includegraphics[height=0.1\textwidth]{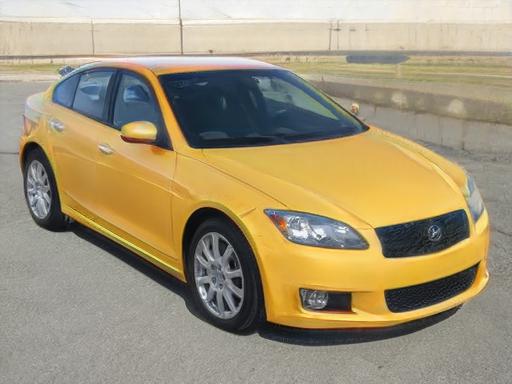} &
            \includegraphics[height=0.1\textwidth]{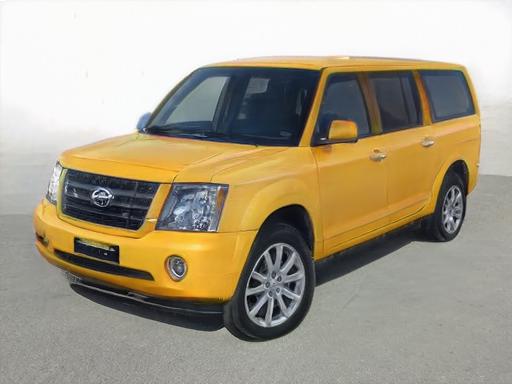} &
            \includegraphics[height=0.1\textwidth]{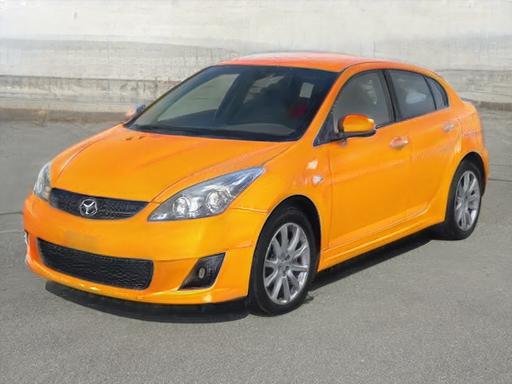} &
            \includegraphics[height=0.1\textwidth]{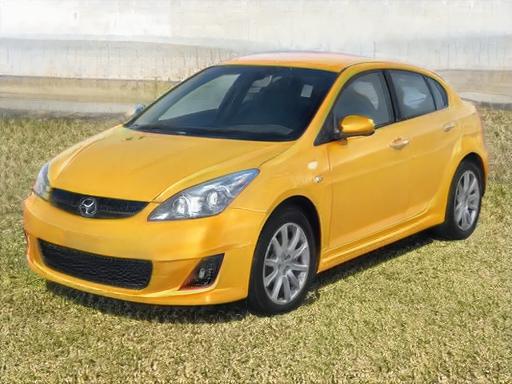}  \\
\end{tabular}
    \caption{Additional results of our e4e encoder on the cars domain, similar to Figure \ref{fig:internet_celebs_1} on the Stanford Cars test set ~\cite{KrauseStarkDengFei-Fei_3DRR2013}.}
    \label{fig:appendix_first_cars_5}
\end{figure*}

Last, we provide numerous inversion and editing results obtained using our e4e approach in \Cref{fig:first_celebs_interfacegan_1,fig:first_celebs_interfacegan_2,fig:first_celebs_interfacegan_3,fig:appendix_first_cars_3,fig:appendix_first_cars_4,fig:appendix_first_cars_5,fig:appendix_horses,fig:appendix_church}.




\end{document}